%% file: main.tex
\RequirePackage{fix-cm}
\documentclass[11pt]{article}

\usepackage[preprint]{acl}

\usepackage{times}
\usepackage{latexsym}
\usepackage[T1]{fontenc}
\usepackage[utf8]{inputenc}
\usepackage{microtype}
\usepackage{inconsolata}

\usepackage{graphicx}
\usepackage{float} %
\usepackage{booktabs}
\usepackage{multirow}
\usepackage{amsfonts}
\usepackage{amssymb}   %
\usepackage{amsmath}
\usepackage{nicefrac}
\usepackage{enumitem}   %
\usepackage{url}
\usepackage{hyperref}
\usepackage{fontawesome5}
\newcommand{\mRep}{\faIcon{file-alt}}          %
\newcommand{\mMeas}{\faIcon{ruler}}            %
\newcommand{\mRetr}{\faIcon{fire}\hspace{0.4pt}\faIcon{ruler}} %
\usepackage{colortbl}
\definecolor{avgbg}{RGB}{232,240,250}
\newcolumntype{A}{>{\columncolor{avgbg}}c}
\graphicspath{{media/}{figure/}}

\usepackage{pgfplots}
\pgfplotsset{compat=1.18}
\usepgfplotslibrary{groupplots}
\usetikzlibrary{shapes.geometric}
\definecolor{cEagle}{HTML}{0072B2}
\definecolor{cDFlash}{HTML}{E69F00}
\definecolor{cDFlare}{HTML}{56B4E9}
\definecolor{cDomino}{HTML}{CC79A7}
\definecolor{cDPace}{HTML}{8C62B5}
\definecolor{cDSpark}{HTML}{009E73}
\definecolor{cDCore}{HTML}{000000}
\definecolor{taupos}{HTML}{56B4E9}   %
\definecolor{tauneg}{HTML}{E69F00}   %
\input{figure/Figure-colors}

\title{When Does Better Acceptance Pay?\\A Controlled Study of Speculative Decoding}

\input{authors}

\begin{document}
\maketitle
\baiduinternshipnote

\input{section/abstract}
\input{section/introduction}
\input{section/related-work}
\input{section/controlled-protocol}
\input{section/results}
\input{section/design-costs}
\input{section/analysis}
\input{section/conclusion}

\input{section/limitations}

\input{section/acknowledgments}

\bibliography{references}

\appendix

\input{section/Appendix-training}

\input{section/Appendix-measurement}
\input{section/Appendix-serving}
\input{section/Appendix-source-protocols}
\input{section/Appendix-alignment}
\input{section/Appendix-task-entropy}
\input{section/Appendix-dcore}
\input{section/Appendix-design-notes}
\input{section/Appendix-eagle-concurrency}
\input{section/Appendix-benchmark-results}
\input{section/Appendix-concurrency}
\input{section/Appendix-sampled-drafts}
\clearpage
\input{section/Appendix-loss-ablation}
\input{figure/Figure-path-entropy}
\input{section/Appendix-training-curves}
\input{section/Appendix-accuracy}

\end{document}

%% file: figure/Figure-colors.tex
\definecolor{vir0}{RGB}{68,1,84}
\definecolor{vir1}{RGB}{69,4,87}
\definecolor{vir2}{RGB}{70,8,92}
\definecolor{vir3}{RGB}{70,11,94}
\definecolor{vir4}{RGB}{71,16,99}
\definecolor{vir5}{RGB}{71,19,101}
\definecolor{vir6}{RGB}{72,23,105}
\definecolor{vir7}{RGB}{72,26,108}
\definecolor{vir8}{RGB}{72,29,111}
\definecolor{vir9}{RGB}{72,33,115}
\definecolor{vir10}{RGB}{72,36,117}
\definecolor{vir11}{RGB}{72,40,120}
\definecolor{vir12}{RGB}{71,42,122}
\definecolor{vir13}{RGB}{71,46,124}
\definecolor{vir14}{RGB}{70,48,126}
\definecolor{vir15}{RGB}{70,52,128}
\definecolor{vir16}{RGB}{69,55,129}
\definecolor{vir17}{RGB}{68,58,131}
\definecolor{vir18}{RGB}{67,62,133}
\definecolor{vir19}{RGB}{66,64,134}
\definecolor{vir20}{RGB}{65,68,135}
\definecolor{vir21}{RGB}{64,70,136}
\definecolor{vir22}{RGB}{62,73,137}
\definecolor{vir23}{RGB}{62,76,138}
\definecolor{vir24}{RGB}{60,79,138}
\definecolor{vir25}{RGB}{59,82,139}
\definecolor{vir26}{RGB}{58,84,140}
\definecolor{vir27}{RGB}{56,88,140}
\definecolor{vir28}{RGB}{55,90,140}
\definecolor{vir29}{RGB}{54,93,141}
\definecolor{vir30}{RGB}{53,95,141}
\definecolor{vir31}{RGB}{51,98,141}
\definecolor{vir32}{RGB}{50,100,142}
\definecolor{vir33}{RGB}{49,103,142}
\definecolor{vir34}{RGB}{48,106,142}
\definecolor{vir35}{RGB}{47,108,142}
\definecolor{vir36}{RGB}{46,111,142}
\definecolor{vir37}{RGB}{45,113,142}
\definecolor{vir38}{RGB}{44,115,142}
\definecolor{vir39}{RGB}{43,117,142}
\definecolor{vir40}{RGB}{42,120,142}
\definecolor{vir41}{RGB}{41,122,142}
\definecolor{vir42}{RGB}{40,125,142}
\definecolor{vir43}{RGB}{39,128,142}
\definecolor{vir44}{RGB}{38,130,142}
\definecolor{vir45}{RGB}{37,132,142}
\definecolor{vir46}{RGB}{36,134,142}
\definecolor{vir47}{RGB}{35,137,142}
\definecolor{vir48}{RGB}{34,139,141}
\definecolor{vir49}{RGB}{33,142,141}
\definecolor{vir50}{RGB}{33,145,140}
\definecolor{vir51}{RGB}{32,146,140}
\definecolor{vir52}{RGB}{31,149,139}
\definecolor{vir53}{RGB}{31,151,139}
\definecolor{vir54}{RGB}{31,154,138}
\definecolor{vir55}{RGB}{30,156,137}
\definecolor{vir56}{RGB}{31,159,136}
\definecolor{vir57}{RGB}{31,161,136}
\definecolor{vir58}{RGB}{32,163,134}
\definecolor{vir59}{RGB}{33,166,133}
\definecolor{vir60}{RGB}{34,168,132}
\definecolor{vir61}{RGB}{37,171,130}
\definecolor{vir62}{RGB}{38,173,129}
\definecolor{vir63}{RGB}{41,175,127}
\definecolor{vir64}{RGB}{44,177,126}
\definecolor{vir65}{RGB}{47,180,124}
\definecolor{vir66}{RGB}{50,182,122}
\definecolor{vir67}{RGB}{55,184,120}
\definecolor{vir68}{RGB}{59,187,117}
\definecolor{vir69}{RGB}{63,188,115}
\definecolor{vir70}{RGB}{68,191,112}
\definecolor{vir71}{RGB}{72,193,110}
\definecolor{vir72}{RGB}{78,195,107}
\definecolor{vir73}{RGB}{82,197,105}
\definecolor{vir74}{RGB}{88,199,101}
\definecolor{vir75}{RGB}{94,201,98}
\definecolor{vir76}{RGB}{99,203,95}
\definecolor{vir77}{RGB}{105,205,91}
\definecolor{vir78}{RGB}{110,206,88}
\definecolor{vir79}{RGB}{117,208,84}
\definecolor{vir80}{RGB}{122,209,81}
\definecolor{vir81}{RGB}{129,211,77}
\definecolor{vir82}{RGB}{134,213,73}
\definecolor{vir83}{RGB}{142,214,69}
\definecolor{vir84}{RGB}{149,216,64}
\definecolor{vir85}{RGB}{155,217,60}
\definecolor{vir86}{RGB}{162,218,55}
\definecolor{vir87}{RGB}{168,219,52}
\definecolor{vir88}{RGB}{176,221,47}
\definecolor{vir89}{RGB}{181,222,43}
\definecolor{vir90}{RGB}{189,223,38}
\definecolor{vir91}{RGB}{194,223,35}
\definecolor{vir92}{RGB}{202,225,31}
\definecolor{vir93}{RGB}{210,226,27}
\definecolor{vir94}{RGB}{216,226,25}
\definecolor{vir95}{RGB}{223,227,24}
\definecolor{vir96}{RGB}{229,228,25}
\definecolor{vir97}{RGB}{236,229,27}
\definecolor{vir98}{RGB}{241,229,29}
\definecolor{vir99}{RGB}{248,230,33}
\definecolor{vir100}{RGB}{253,231,37}

%% file: authors.tex
\author{%
\normalsize\bfseries Yongchao Long\textsuperscript{1,2,}\thanks{Equal contribution.}\quad
Chunrui Zheng\textsuperscript{2,4,*}\quad
Yuxuan Zhou\textsuperscript{2}\quad Guangkun Nie\textsuperscript{2,5}\quad
Tian-Chi Ma\textsuperscript{2,6}\\
\normalsize\bfseries Chao Pang\textsuperscript{2}\quad
Shuohuan Wang\textsuperscript{2}\quad Yuchen Ding\textsuperscript{2}\quad
Xianbin Wen\textsuperscript{1}\quad Dianhai Yu\textsuperscript{2}\\
\normalsize\bfseries Junyuan Shang\textsuperscript{2,}\thanks{Corresponding authors:
\mbox{\href{mailto:shangjunyuan@baidu.com}{\nolinkurl{shangjunyuan@baidu.com}}}
(Junyuan Shang) and
\mbox{\href{mailto:joy_yuxi@pku.edu.cn}{\nolinkurl{joy_yuxi@pku.edu.cn}}}
(Yuxi Zhou).}\quad
Yuxi Zhou\textsuperscript{1,\textdagger}\quad
Shenda Hong\textsuperscript{3}\\[4pt]
\small\normalfont\textsuperscript{1}Department of Computer Science, Tianjin University of Technology
\quad\textsuperscript{2}Baidu Inc.\\[1pt]
\small\normalfont\textsuperscript{3}National Institute of Health Data Science, Peking University\\[1pt]
\small\normalfont\textsuperscript{4}School of Information Science and Engineering, Shandong University\\[1pt]
\small\normalfont\textsuperscript{5}School of Intelligence Science and Technology, Peking University\\[1pt]
\small\normalfont\textsuperscript{6}National Key Laboratory for Novel Software Technology, Nanjing University\\[5pt]
\small\normalfont\href{https://github.com/ernie-research/DCore}{%
\faIcon{github}\enspace\nolinkurl{github.com/ernie-research/DCore}}\quad
\href{https://huggingface.co/collections/ernie-research/dcore}{%
\raisebox{-0.2em}{\includegraphics[height=1.2em]{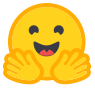}}%
\enspace Hugging Face}
}

\newcommand{\baiduinternshipnote}{%
\begingroup
\renewcommand{\thefootnote}{}%
\begin{NoHyper}
\footnotetext{This work was completed while Yongchao Long, Chunrui Zheng,
Guangkun Nie, and Tian-Chi Ma were interns at Baidu.}%
\end{NoHyper}
\endgroup
}

\hypersetup{pdfauthor={Yongchao Long, Chunrui Zheng, Yuxuan Zhou, Guangkun Nie, Tian-Chi Ma, Chao Pang, Shuohuan Wang, Yuchen Ding, Xianbin Wen, Dianhai Yu, Junyuan Shang, Yuxi Zhou, Shenda Hong},
pdftitle={When Does Better Acceptance Pay? A Controlled Study of Speculative Decoding}}

%% file: section/abstract.tex
\begin{abstract}
The throughput gains of speculative decoding depend on the trade-off between
acceptance length and execution latency. We ask when additional acceptance
is worth the cost of obtaining it. We establish a comparison
protocol spanning training, inference, and evaluation, remeasuring released
checkpoints and retraining six drafter recipes on shared target-generated
data across ten benchmarks and two target models. Alignment changes the
picture: gaps that appear large in published comparisons shrink substantially
under shared training and inference conditions. This narrowing motivates two
deeper questions: how is acceptance improved at different costs, and when do
these gains improve throughput? We find that acceptance can be improved
through nearly free objective and context changes or more expensive
token-to-token conditioning, which pays off most clearly under sampled
drafting. DCore illustrates the former route as a strong
baseline with little additional measured drafting overhead. The payoff is
conditional: task uncertainty and serving conditions affect acceptance
returns, execution costs, and the preferred drafter and budget.
These findings inform future drafter design and deployment choices.
\end{abstract}

%% file: section/introduction.tex
\section{Introduction}
\label{sec:intro}

\input{figure/Figure-alignment.tex}

Speculative decoding uses a lightweight drafter to propose tokens that a
target model verifies in parallel, preserving the target's output
distribution through lossless verification
\citep{leviathan2023fast,chen2023accelerating}. A memory-bound target pass
can then commit several tokens instead of one
\citep{pope2023efficient,shazeer2019fast}, making longer accepted drafts
appealing. \textbf{The ultimate goal, however, is higher throughput;
acceptance length is an intermediate measure.} Acceptance length is
relatively straightforward to measure, whereas throughput depends on the
serving implementation and operating conditions, making comparisons more
demanding. The accepted tokens must repay the computation spent proposing
and verifying them. We therefore ask: \textbf{when is additional acceptance
worth the cost of obtaining it?}

We begin by examining how acceptance is improved. Training data and
supervision shape proposal quality, while architectural components change
how proposals are generated. Recent recipes pursue these improvements
across three families. Autoregressive (AR) Eagle-3 maintains token-to-token
conditioning through sequential backbone calls
\citep{NEURIPS2025_c7b5a35e}. Block-parallel DFlash predicts multiple positions
in one pass \citep{chen2026dflash}; DFlare changes target-state conditioning
\citep{zhang2026dflare}, and D-PACE changes the training objective
\citep{wu2026dpacedynamicpositionawarecrossentropy}. Semi-autoregressive
(semi-AR) Domino and DSpark add causal correction to parallel predictions
\citep{huang2026dominodecouplingcausalmodeling,cheng2026dspark}. Published
comparisons often change training data, components, and inference protocols
together, obscuring which routes deliver the observed gains.

To make these trade-offs interpretable, we establish a comparison protocol
spanning training, inference, and evaluation. It defines shared experimental
conditions, documents recipe-specific differences, and measures acceptance
alongside throughput, providing a common basis for assessing what each
design gains and what it costs. Within this framework, we first remeasure
released checkpoints under common inference and evaluation conditions,
then retrain the recipes on shared target-generated data. The observed
method gaps narrow substantially: on GSM8K, the fastest--slowest spread
falls from $4.60$ as reported to $2.84$ after remeasurement and $1.53$ after
retraining, in units of speedup over each stage's target-only reference
(Fig.~\ref{fig:published_vs_controlled}). This narrowing reflects the
combined effect of aligning training and evaluation conditions and provides
a common reference for examining the remaining design trade-offs.

With this reference established, we next ask which components improve
acceptance and at what execution cost. Our analysis distinguishes two
routes: improving proposal quality with little additional inference
overhead, or spending more computation on token-to-token conditioning.
To explore the nearly free route, we start from DFlash's low-latency
block-parallel backbone and ablate candidate components
(Fig.~\ref{fig:roadmap}), retaining improvements that raise acceptance
without a measurable increase in drafting latency. This yields DCore, a
strong baseline combining existing context and objective components, which
achieves the highest mean throughput in $15$ of the $24$ single-request
settings spanning target, budget, temperature pair $T_t|T_d$, and entropy
group, including every greedy-draft $1|0$ setting
(Tab.~\ref{tab:main_results_temps}). More expensive conditioning can also
be worthwhile: for comparable measurements, throughput improves when the
proportional acceptance gain exceeds the proportional increase in complete
step cost (Section~\ref{sec:evaluation}). Sampled drafting is one such
case: semi-AR DSpark leads all eight $1|1$ settings, while block-parallel
drafters, which propose each position without seeing the tokens sampled
before it, lose far more acceptance than semi-AR recipes after a non-argmax
predecessor, an accepted previous token that is not the target's argmax
(Fig.~\ref{fig:sampling_gap}).

Having identified how acceptance can be improved and at what cost, we
examine when those gains translate into higher throughput. Task uncertainty,
characterized by the target's next-token entropy, helps describe these
returns. Across three drafter families, accepted prefixes tend to have
lower target entropy than the rejected positions ending their paths
(Fig.~\ref{fig:draftability}). Complementary benchmark-level analysis
associates higher measured entropy with shorter acceptance and smaller
budget returns for block-parallel and semi-AR recipes; tree-budget returns
show a weaker association. These observations help characterize which
workloads reward more speculation. Whether those rewards repay their cost
also depends on execution conditions: concurrency and target scale change
relative costs and the measured preferences among methods and budgets.
Together, these analyses connect workload-dependent acceptance gains to
their throughput value under particular operating conditions.

Our study covers six drafter recipes and ten benchmarks using target models
supported by most of the evaluated methods
(App.~\ref{app:published-protocols}).

\paragraph{Findings and their implications.}
\begin{itemize}[leftmargin=*,nosep]
\item \textbf{Alignment narrows the observed method gaps.}
Report released-checkpoint remeasurement and controlled retraining
separately, since they answer different questions (Section~\ref{sec:results}).
\item \textbf{Acceptance can be improved at different execution costs.}
Test improvements with little additional drafting overhead, exemplified by
DCore, and assess more expensive conditioning against the complete step-cost
increase; under sampled drafting, prefix-conditioned proposals avoid a
mismatch that loss reweighting alone does not remove
(Sections~\ref{sec:design} and~\ref{sec:analysis}).
\item \textbf{Workload and serving conditions determine whether extra
acceptance pays.}
Use task uncertainty to characterize acceptance and budget returns, and
evaluate the drafter and budget jointly under the intended operating
conditions (Section~\ref{sec:analysis}).
\end{itemize}

%% file: figure/Figure-alignment.tex
\newcommand{\pvcSwatch}[1]{\tikz[baseline=-0.55ex]{%
    \fill[#1] (0,-0.9mm) rectangle (2.6mm,0.9mm);}}
\newcommand{\pvcIbeam}{\tikz[baseline=-0.55ex]{%
    \draw[black!55, line width=0.5pt] (0.9mm,-1.1mm) -- (0.9mm,1.1mm);
    \draw[black!55, line width=0.5pt] (0,-1.1mm) -- (1.8mm,-1.1mm);
    \draw[black!55, line width=0.5pt] (0,1.1mm) -- (1.8mm,1.1mm);}}

\begin{figure}[t]
    \centering
    \mbox{\tiny
     \pvcSwatch{cEagle}\,Eagle-3\hspace{4pt}
     \pvcSwatch{cDFlash}\,DFlash\hspace{4pt}
     \pvcSwatch{cDFlare}\,DFlare\hspace{4pt}
     \pvcSwatch{cDomino}\,Domino\hspace{8pt}
     \pvcIbeam\,max$-$min speedup}\\[3pt]
    \begin{tikzpicture}
    \pgfplotsset{
        ideal/.style={ybar, bar shift=0pt, bar width=10.35pt,
            fill=#1!22, draw=#1!55},
        idealpaper/.style={ybar, bar shift=0pt, bar width=10.35pt,
            fill=#1!22, draw=#1, dashed, semithick},
        real/.style={ybar, bar shift=0pt, bar width=10.35pt,
            fill=#1, draw=none,
            nodes near coords, point meta=explicit symbolic,
            every node near coord/.append style={
                font=\fontsize{4.4}{5}\selectfont,
                inner sep=1.2pt, text=white, anchor=north}},
        panel/.style={
            width=0.755\columnwidth, scale only axis,
            xmin=0.3, xmax=15.6, ymin=0,
            yticklabel style={font=\scriptsize},
            ylabel style={font=\scriptsize},
            xtick={2.5,7.5,12.5},
            xtick style={draw=none}, axis x line*=bottom,
            ymajorgrids=true, grid style={gray!25},
        },
    }
    \tikzset{tag/.style={font=\fontsize{5.5}{6}\selectfont,
        anchor=north west, inner sep=1pt,
        fill=white, fill opacity=0.75, text opacity=1,
        rounded corners=0.5pt}}
    \newcommand{\spreadmark}[4]{%
        \draw[black!55, line width=0.5pt] (axis cs:#1,#2) -- (axis cs:#1,#3);
        \draw[black!55, line width=0.5pt]
            ([xshift=-2.6pt]axis cs:#1,#2) -- ([xshift=2.6pt]axis cs:#1,#2);
        \draw[black!55, line width=0.5pt]
            ([xshift=-2.6pt]axis cs:#1,#3) -- ([xshift=2.6pt]axis cs:#1,#3);
        \node[font=\fontsize{4.4}{5}\selectfont, inner sep=1pt,
            anchor=south, text=black] at ([yshift=1.5pt]axis cs:#1,#3) {#4};}
    \begin{axis}[panel, name=L, height=2.600cm, ymax=9.4, ytick={0,2,4,6,8},
        axis y line*=left, xticklabels={},
        ylabel={$\tau$},
    ]
    \addplot[idealpaper=cEagle,  forget plot] coordinates {(1,3.12)};
    \addplot[idealpaper=cDFlash, forget plot] coordinates {(2,5.98)};
    \addplot[idealpaper=cDFlare, forget plot] coordinates {(3,7.07)};
    \addplot[idealpaper=cDomino, forget plot] coordinates {(4,8.34)};

    \addplot[ideal=cEagle,  forget plot] coordinates {(6,3.14)};
    \addplot[ideal=cDFlash, forget plot] coordinates {(7,5.56)};
    \addplot[ideal=cDFlare, forget plot] coordinates {(8,6.49)};
    \addplot[ideal=cDomino, forget plot] coordinates {(9,7.39)};

    \addplot[ideal=cEagle,  forget plot] coordinates {(11,5.94)};
    \addplot[ideal=cDFlash, forget plot] coordinates {(12,6.71)};
    \addplot[ideal=cDFlare, forget plot] coordinates {(13,6.88)};
    \addplot[ideal=cDomino, forget plot] coordinates {(14,7.67)};

    \addplot[real=cEagle,  forget plot] coordinates {(1,1.87) [1.87]};
    \addplot[real=cDFlash, forget plot] coordinates {(2,4.67) [4.67]};
    \addplot[real=cDFlare, forget plot] coordinates {(3,5.34) [5.34]};
    \addplot[real=cDomino, forget plot] coordinates {(4,6.47) [6.47]};

    \addplot[real=cEagle,  forget plot] coordinates {(6,1.73) [1.73]};
    \addplot[real=cDFlash, forget plot] coordinates {(7,3.79) [3.79]};
    \addplot[real=cDFlare, forget plot] coordinates {(8,4.24) [4.24]};
    \addplot[real=cDomino, forget plot] coordinates {(9,4.57) [4.57]};

    \addplot[real=cEagle,  forget plot] coordinates {(11,3.20) [3.20]};
    \addplot[real=cDFlash, forget plot] coordinates {(12,4.54) [4.54]};
    \addplot[real=cDFlare, forget plot] coordinates {(13,4.67) [4.67]};
    \addplot[real=cDomino, forget plot] coordinates {(14,4.73) [4.73]};

    \spreadmark{5}{1.87}{6.47}{4.60}
    \spreadmark{10}{1.73}{4.57}{2.84}
    \spreadmark{15}{3.20}{4.73}{1.53}

    \node[tag] at (rel axis cs:0.012,0.985) {GSM8K \ (Low Entropy)};
    \end{axis}

    \begin{axis}[panel, at={(L.south west)}, anchor=south west,
        height=2.600cm, ymax=9.4, ytick={0,2,4,6,8},
        axis x line=none, axis y line*=right,
        ylabel={Speedup},
        yticklabels={0$\times$,2$\times$,4$\times$,6$\times$,8$\times$}]
    \end{axis}
    \begin{axis}[panel, name=B, at={(L.south)}, yshift=-9pt, anchor=north,
        height=1.383cm, ymax=5.0, ytick={0,2,4},
        axis y line*=left, ylabel={$\tau$},
        xticklabels={{As reported\\{\tiny their protocol}},%
                     {Re-measured\\{\tiny our protocol}},%
                     {Re-trained\\{\tiny + our data}}},
        xticklabel style={font=\scriptsize, align=center, yshift=-1pt},
    ]
    \addplot[idealpaper=cEagle,  forget plot] coordinates {(1,2.70)};
    \addplot[idealpaper=cDFlash, forget plot] coordinates {(2,3.80)};
    \addplot[idealpaper=cDFlare, forget plot] coordinates {(3,4.37)};
    \addplot[idealpaper=cDomino, forget plot] coordinates {(4,4.61)};

    \addplot[ideal=cEagle,  forget plot] coordinates {(6,2.53)};
    \addplot[ideal=cDFlash, forget plot] coordinates {(7,2.89)};
    \addplot[ideal=cDFlare, forget plot] coordinates {(8,3.37)};
    \addplot[ideal=cDomino, forget plot] coordinates {(9,3.28)};

    \addplot[ideal=cEagle,  forget plot] coordinates {(11,3.59)};
    \addplot[ideal=cDFlash, forget plot] coordinates {(12,3.00)};
    \addplot[ideal=cDFlare, forget plot] coordinates {(13,3.13)};
    \addplot[ideal=cDomino, forget plot] coordinates {(14,3.43)};

    \addplot[real=cEagle,  forget plot] coordinates {(1,1.58) [1.58]};
    \addplot[real=cDFlash, forget plot] coordinates {(2,2.47) [2.47]};
    \addplot[real=cDFlare, forget plot] coordinates {(3,2.81) [2.81]};
    \addplot[real=cDomino, forget plot] coordinates {(4,2.94) [2.94]};

    \addplot[real=cEagle,  forget plot] coordinates {(6,1.41) [1.41]};
    \addplot[real=cDFlash, forget plot] coordinates {(7,2.04) [2.04]};
    \addplot[real=cDFlare, forget plot] coordinates {(8,2.26) [2.26]};
    \addplot[real=cDomino, forget plot] coordinates {(9,2.12) [2.12]};

    \addplot[real=cEagle,  forget plot] coordinates {(11,1.99) [1.99]};
    \addplot[real=cDFlash, forget plot] coordinates {(12,2.10) [2.10]};
    \addplot[real=cDFlare, forget plot] coordinates {(13,2.19) [2.19]};
    \addplot[real=cDomino, forget plot] coordinates {(14,2.22) [2.22]};

    \spreadmark{5}{1.58}{2.94}{1.36}
    \spreadmark{10}{1.41}{2.26}{0.85}
    \spreadmark{15}{1.99}{2.22}{0.23}

    \node[tag] at (rel axis cs:0.40,0.99) {MT-Bench \ (High Entropy)};
    \end{axis}

    \begin{axis}[panel, at={(B.south west)}, anchor=south west,
        height=1.383cm, ymax=5.0, ytick={0,2,4},
        axis x line=none, axis y line*=right,
        ylabel={Speedup},
        yticklabels={0$\times$,2$\times$,4$\times$}]
    \end{axis}
    \end{tikzpicture}
    \caption{\textbf{Observed gaps narrow after inference and training alignment.} Four drafters on
    GSM8K, representing the low-entropy half of our benchmark suite, and
    MT-Bench, representing the high-entropy half, on Qwen3-8B. Pale
    bars: acceptance length $\tau$ (left axis); solid bars: speedup over the
    target-only baseline (right axis). \emph{As reported}: each paper's own
    published numbers. \emph{Re-measured}: its released checkpoint re-run in our
    engine. \emph{Re-trained}: a drafter we train on the same training data.
    }
    \label{fig:published_vs_controlled}
\end{figure}
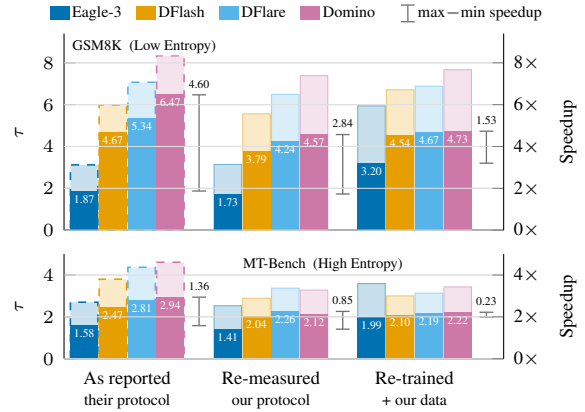

%% file: section/related-work.tex
\section{Related Work}
\label{sec:related}

\paragraph{Obtaining acceptance at different costs.}
Blockwise parallel decoding introduced the draft-and-verify approach
\citep{stern2018blockwise}, and speculative sampling established lossless
decoding that preserves the target distribution
\citep{leviathan2023fast,chen2023accelerating}. Retrieval and parallel
prediction reduce proposal cost \citep{he2024rest,fu2024lookahead}, while
candidate trees spend additional computation to increase the chance of
acceptance \citep{miao2024specinfer,li2024eagle2}.
Learned drafting has practical relevance across design families.
DeepSeek-V3 explicitly relates its causal MTP design to EAGLE
\citep{li2024eagle,deepseekai2024deepseekv3}; Qwen3.8 and Tencent Hy4 preview
also support native autoregressive MTP drafting
\citep{qwen2026qwen38mtp,tencent2026hy4preview}.
MiMo-V2.5-Pro's UltraSpeed mode adopts block-parallel DFlash
\citep{xiaomi2026mimoultraspeed}, and NVIDIA Nemotron 3.5 Lightning provides
both DFlash and semi-AR DSpark drafters
\citep{nvidia2026nemotron35lightning,nvidia2026nemotron35dflash};
DeepSeek-V4-Pro also provides a DSpark module
\citep{deepseekai2026v4prodspark}.
Motivated by this coverage, we select six target-hidden-state-conditioned
recipes spanning AR, block-parallel, and semi-AR drafting
\citep{NEURIPS2025_c7b5a35e,chen2026dflash,zhang2026dflare,
wu2026dpacedynamicpositionawarecrossentropy,
huang2026dominodecouplingcausalmodeling,cheng2026dspark}.
Their shared conditioning basis supports a common comparison protocol;
variations in context, supervision, and causal modeling expose different
routes to acceptance and their execution costs.
Tab.~\ref{tab:drafter-summary} summarizes the evaluated recipes;
full method definitions appear in the cited papers.

\paragraph{From controlled comparison to deployment decisions.}
Benchmarks and system studies characterize speculative decoding across
tasks, model scales, and serving conditions
\citep{xia-etal-2024-unlocking,abramovich2026speedbench,sun2025benchmark,
kim2026runtime,yan-etal-2025-decoding,liu2026illusion,
mainardi-etal-2026-empirical}.
Shared-data evaluations and component ablations examine drafter design
\citep{hui2026peagle,chen2026dflash,zhang2026dflare,
huang2026dominodecouplingcausalmodeling,cheng2026dspark,liu2026angelspec},
while entropy analyses characterize token rejection and task-dependent
acceptance \citep{cheng-etal-2026-edsd,abramovich2026speedbench,
rusanovsky2026lilicorr}.
However, a shared-data ablation can establish a component's benefit in its
own setting without showing how much of a headline method advantage survives
common training and evaluation conditions.
To the best of our knowledge, this is the first study to explicitly quantify
how reported performance gaps across AR, block-parallel, and semi-AR drafters
change through both released-checkpoint remeasurement and shared-data
retraining. We find that the observed gaps narrow substantially, providing
a common reference for assessing the remaining design trade-offs.
On this basis, we distinguish acceptance improvements with little additional
inference overhead from more expensive conditioning, and examine how task
uncertainty and serving conditions determine whether those gains repay their
execution cost in throughput.

%% file: section/controlled-protocol.tex
\section{Controlled Protocol}
\label{sec:evaluation}

Our protocol distinguishes \emph{remeasurement} of released checkpoints
under common inference and evaluation conditions from \emph{retraining}
of drafter recipes on shared target-generated data under the same conditions.
Together, these stages quantify how method gaps change with alignment while
retaining each recipe's defining architectural components, objective, and
training strategy. The protocol connects proposal quality and execution
costs to throughput across tasks and serving loads. Apps.~\ref{app:published-protocols}
and~\ref{app:published-tau} document checkpoints and source-paper conditions.

\subsection{Training}
Targets are Qwen3-4B and Qwen3-8B \citep{yang2025qwen3}, with thinking
disabled. We choose this pair because five of the six evaluated methods
report results on both models (App.~\ref{app:published-tau}). Their
shared target coverage enables direct comparison with the original papers
and provides a common reference for our released-checkpoint remeasurement
and shared-data retraining.
Each target regenerates responses to the same Open-PerfectBlend prompts
\citep{labonne2024openperfectblend}, yielding one shared corpus per target.
Block-parallel and semi-AR drafters use five-layer backbones, the default
depth for most of these recipes; DFlare defaults to seven layers and is
retrained at five layers to align backbone depth.
Eagle-3 retains its single-layer AR backbone and training schedule.
Tab.~\ref{tab:drafter-capacity} reports parameter counts and their differences
across drafters.

Beyond the shared settings above, we retain each method's defining
architectural components, loss terms, and training procedures.
App.~\ref{sec:appendix-training} documents recipe-specific settings,
hyperparameters, and data details.

\subsection{Inference}
\label{subsec:inference}
All methods run in SGLang \citep{zheng2024sglang}, with common settings
within each sweep and CUDA-graph optimization of draft paths.
We compare fixed verification budgets $B\in\{8,16\}$, excluding adaptive
budget control. We label Eagle-3 configurations as (draft steps / top-$k$ /
verification budget); for example, $7/10/16$ denotes seven draft steps,
$k{=}10$, and $B{=}16$.
App.~\ref{app:serving-impl} documents budget conventions,
method-specific settings, graph coverage, and timing boundaries.

\paragraph{Lossless sampling.}
We evaluate target/draft temperatures $T_t|T_d$ of $0|0$, $1|0$, and
$1|1$ with lossless verification, excluding Eagle-3 trees from $1|1$
(App.~\ref{sec:appendix-protocol}).

\subsection{Evaluation}
\label{subsec:evaluation}
We evaluate ten benchmarks spanning mathematics, code, chat, and agentic
tasks: GSM8K \citep{cobbe2021gsm8k}, MATH-500
\citep{hendrycks2021math,lightman2024verify}, AIME25 \citep{maa2025aime},
HumanEval \citep{chen2021humaneval}, MBPP \citep{austin2021mbpp},
LiveCodeBench \citep{jain2025livecodebench}, MT-Bench \citep{zheng2023mtbench},
AlpacaEval \citep{li2023alpacaeval}, Arena-Hard \citep{li2024arenahard},
and SWE-Bench \citep{jimenez2024swebench}.

We group benchmarks by task entropy rather than domain to characterize the
difficulty of obtaining accepted drafts, using a median split
(App.~\ref{sec:appendix-entropy}).
Task entropy reflects variation in continuation predictability that broad
domain labels can obscure. For example, LiveCodeBench and SWE-Bench are
both code tasks, yet exhibit the lowest and highest measured entropy,
respectively.

To emulate serving traffic that mixes requests from different tasks, we
construct \textsc{SpecMix} by pooling $3331$ prompts from all ten benchmarks
and shuffling them into a fixed mixture. We denote the number of concurrent
requests by $C$ and sweep $C\in\{1,2,4,8,16,32,64\}$.

\paragraph{Metrics and comparison rule.}
We report acceptance length $\tau$, defined as committed tokens per
verification step including the bonus token, proposal latency $\lambda_D$,
and throughput in tokens per second (TPS). We denote speedup over
target-only decoding by $\mathcal{S}$. With a sampled target ($1|0$ and
$1|1$), TPS and $\tau$ are means over three repeats with seeds $0$, $1$,
and $42$; greedy targets use a single run. Let $\lambda_{\mathrm{step}}$ be the
complete step time, including drafting, verification, and other overhead,
and $\lambda_{\mathrm{base}}$ the target-only time per token. For the same
workload and concurrency, with compatible token/step aggregation,
\begin{equation}
\mathcal{S}=\tau\frac{\lambda_{\mathrm{base}}}{\lambda_{\mathrm{step}}}
\equiv\tau\eta.
\label{eq:eta}
\end{equation}
Here $\eta$ denotes effective efficiency. For two schemes $A$ and $B$
under these conditions,
\begin{equation}
\mathrm{TPS}_A>\mathrm{TPS}_B\ \Longleftrightarrow\
\frac{\tau_A}{\tau_B}>\frac{\lambda_{\mathrm{step},A}}{\lambda_{\mathrm{step},B}}.
\label{eq:payoff}
\end{equation}
This rule organizes the evidence: additional acceptance must cover additional
complete step cost.

%% file: section/results.tex
\section{Alignment Narrows the Observed Gaps}
\label{sec:results}

Comparing a newly trained drafter with competitors' released checkpoints
without aligning their training data is not a fair test of drafter design.
Rerunning those checkpoints under a common evaluation protocol does not
remove the training mismatch: their performance differences still combine
recipe and training conditions. We therefore
distinguish published results, released-checkpoint remeasurement, and
shared-data retraining to examine how much of the reported advantage remains.

Fig.~\ref{fig:published_vs_controlled} summarizes how the observed gaps
change through these three stages for four recipes spanning all three
families. On MT-Bench, the maximum-minus-minimum speedup gap falls from
$1.36$ in published results to $0.85$ after remeasurement and $0.23$ after
retraining, in speedup multiples. The corresponding GSM8K gaps are $4.60$,
$2.84$, and $1.53$. The measured stages share the inference protocol;
published rows retain their source-paper conditions.
The substantial narrowing shows that released-checkpoint
comparisons can overstate the differences between recipes under shared
training and evaluation conditions.

App. Tab.~\ref{tab:published-tau} expands the three-stage comparison
summarized in Fig.~\ref{fig:published_vs_controlled}, providing detailed
acceptance lengths on seven benchmarks, both targets, and the reported
temperature pairs. Across both targets, Eagle-3 gains acceptance length on
every benchmark shown after retraining, whereas DFlare's acceptance length
decreases on most benchmarks, narrowing its advantage over DFlash under
matched greedy settings.

\paragraph{From alignment to design and deployment.}
Aligning training and evaluation narrows the observed gaps, but the remaining
differences still matter for throughput.
\begin{itemize}[leftmargin=*,nosep]
    \item \emph{How do different designs obtain additional acceptance, and at
    what execution cost?} (Section~\ref{sec:design})
    \item \emph{Under which task and serving conditions does that additional
    acceptance translate into higher throughput?} (Section~\ref{sec:analysis})
\end{itemize}

%% file: section/design-costs.tex
\section{Acceptance Gains and Execution Costs}
\label{sec:design}

To answer the first question, we examine how components and training
objectives change acceptance length, proposal latency, and speedup in the
design ablation roadmap (Fig.~\ref{fig:roadmap}). We organize these
contrasts by whether acceptance gains add measured proposal latency.

\input{figure/Figure-design-roadmap}

\input{table/Table-main-results.tex}

\paragraph{Additional acceptance follows two cost routes.}
\textbf{Improving supervision and context} can raise acceptance with little
additional measured proposal latency. Starting from DFlash's $3.004\times$
speedup, D-PACE reaches $3.205\times$ at unchanged $1.79$~ms latency;
KL$+$TV reaches $3.180\times$ at $1.78$~ms. DFlare's richer target-state
context reaches $3.109\times$ at $1.79$~ms with the same five-layer depth.
\textbf{Adding computation} can also raise acceptance, but its value depends
on the cost. Increasing either parallel backbone from five to seven layers
raises acceptance yet increases proposal latency to $2.15$~ms and lowers
speedup. Replacing DSpark's Markov head with a recurrent neural network
(RNN) raises acceptance from $4.91$ to $5.05$ and latency from $2.71$ to
$2.88$~ms, leaving speedup nearly unchanged. Conversely, widening Eagle-3's
tree from $7/10/16$ to $7/10/60$ raises acceptance from $4.51$ to $5.26$
and speedup from $2.484\times$ to $2.67\times$, with only a small latency
increase ($3.13$ to $3.19$~ms). Extra computation can therefore pay when
the acceptance gain covers the added complete step cost. Execution cost
can also be improved independently: Domino's candidate pool preserves
acceptance while reducing latency from $2.64$ to $2.35$~ms and raising
speedup from $3.150\times$ to $3.241\times$.
We suggest prioritizing the low-overhead route and testing combinations:
D-PACE lifts DFlash to $3.205\times$, but on DSpark, even paired with KL
(D-PAKL), it reaches only $3.153\times$ against $3.146\times$.
App.~\ref{sec:appendix-loss-ablation} details objective interactions;
App.~\ref{sec:appendix-dcore-roadmap} documents configurations and
distinguishes exploratory runs from repeated anchors.

\paragraph{DCore improves acceptance at DFlash's low proposal latency.}
We therefore start from DFlash and retain two improvements: DFlare's
target-layer context mixture and the D-PACE objective. With the five-layer
backbone, this combination yields DCore, reaching $3.261\times$ speedup
and acceptance length $4.74$ at unchanged $1.79$~ms proposal latency in
the design search. App.~\ref{sec:dcore} gives the full recipe.
Tab.~\ref{tab:main_results_temps} evaluates this selected recipe alongside
all six shared-data retrained drafters across the full benchmark suite,
pairing throughput with acceptance length. DCore achieves the highest
mean throughput in $15$ of $24$ single-request settings, defined by
target, budget, temperature pair, and entropy group, including all
greedy-draft $1|0$ settings; DSpark leads all eight $1|1$ settings.
D-PACE changes only the training weights, so DCore's proposals remain
prefix-independent. On the larger target, $74$--$77\%$ of the $1|1$
acceptance gain at $j{\ge}1$ of D-PACE over DFlash and of DCore over DFlare
comes after argmax predecessors (previous accepted tokens that are the
target's argmax), and acceptance after non-argmax predecessors rises only
from $0.32$--$0.34$ to $0.40$--$0.41$ (App.~\ref{sec:app-sd-results}).
At $B{=}16$, $1|0$ on the larger target, DCore raises mean speedup over DFlash from
$3.011\times$ to $3.242\times$. Matched greedy profiles also show nearly
unchanged proposal latency (Tab.~\ref{tab:serving-stage-latency}).
This broader comparison supports DCore as a strong baseline obtained
through the low-overhead route identified in the roadmap.

\paragraph{Higher acceptance does not guarantee higher throughput.}
On the larger target's high-entropy half at $B{=}16$, $1|0$, Eagle-3 achieves
$\tau=3.67$ versus DFlash's $3.19$, but only $2.05\times$ speedup versus
$2.24\times$. In the larger-target, $B{=}16$ summary, greedy-draft effective
efficiency $\eta$ is approximately $0.69$ for block-parallel recipes,
$0.63$ for semi-AR recipes, and $0.55$ for Eagle-3. With sampled drafts,
both values hold: about $0.69$ for block-parallel recipes and $0.63$ for
Domino and DSpark.
We next deepen this analysis by examining how task and serving conditions
change the acceptance--cost trade-off.

%% file: figure/Figure-design-roadmap.tex
\begin{figure}[!t]
\centering

\definecolor{rmArch}{HTML}{5C6FBF}   %
\definecolor{rmLoss}{HTML}{3E7A47}   %
\definecolor{rmTau}{HTML}{E2E2E4}    %
\definecolor{rmStar}{HTML}{5A6FB5}   %
\definecolor{rmStarLbl}{HTML}{8FA0D8}%
\definecolor{rmGold}{HTML}{D9A400}   %
\definecolor{rmAxis}{HTML}{333333}   %
\definecolor{rmRowLbl}{HTML}{111111} %
\definecolor{rmValOut}{HTML}{555555} %
\definecolor{rmGridA}{HTML}{F1F1F1}  %
\definecolor{rmGridB}{HTML}{E3E3E3}  %
\definecolor{rmEdge}{HTML}{BFBFBF}   %
\definecolor{rmTree}{HTML}{9A9A9A}   %

\def\rmW{139.905}      %
\def\rmBarH{8.9}       %
\def\rmPitch{11.6}     %
\def\rmFamGap{1.8}     %
\def\rmTop{-10.05}     %
\def\rmBot{-302.85}    %
\def\rmValIn{4.2}      %
\def\rmValOutGap{3.5}  %
\def\rmLatGap{5.6}     %
\def\rmGoldGap{7.0}    %
\def\rmBaseDrop{3.05}  %
\def\rmValDrop{2.05}   %
\def\rmLatDrop{1.94}   %
\def\rmTickUp{4.85}    %
\def\rmTickDown{11.27} %
\def\rmLabA{-74.5}     %
\def\rmLabB{-61.5}     %
\def\rmLabC{-50.5}     %
\def\rmElbowVa{-68.0}  %
\def\rmElbowHa{-63.5}  %
\def\rmElbowVb{-57.0}  %
\def\rmElbowHb{-52.5}
\def\rmTreeTop{2.90}   %

\def\rmxtau#1{\rmW*((#1)-2.5)/3}
\def\rmxspd#1{\rmW*((#1)-2)/2}
\def\rmxlat#1{\rmW*(1.23735+0.55062*(#1))/3}

\def\rmFRow{\fontsize{8.5}{10}\selectfont}
\def\rmFVal{\fontsize{7.5}{8}\selectfont}
\def\rmFLat{\fontsize{7}{7.5}\selectfont}
\def\rmFAx{\fontsize{8.5}{9}\selectfont}

\newcommand{\rmtimes}{\kern-0.08em\ensuremath{\times}\kern-0.05em}

\newcommand{\rmreset}{\gdef\rmcury{0}\gdef\rmfirst{1}\gdef\rmstarco{}%
    \gdef\rmpA{0}\gdef\rmpB{0}}
\newcommand{\rmrow}[9]{%
  \ifnum\rmfirst=1
    \pgfmathsetmacro{\rmy}{\rmTop}\gdef\rmfirst{0}%
  \else
    \ifnum#1=0 \pgfmathsetmacro{\rmy}{\rmcury-\rmPitch-\rmFamGap}%
    \else      \pgfmathsetmacro{\rmy}{\rmcury-\rmPitch}\fi
  \fi
  \global\let\rmcury\rmy
  \ifnum#1=0 \global\let\rmpA\rmy \fi
  \ifnum#1=1 \global\let\rmpB\rmy \fi
  \pgfmathsetmacro{\rmgx}{min(\rmxtau{#4},\rmW)}%
  \begin{pgfonlayer}{rmbg}
    \fill[rmTau] (0,{\rmy-\rmBarH/2}) rectangle (\rmgx,{\rmy+\rmBarH/2});
  \end{pgfonlayer}
  \ifnum#6=1 \colorlet{rmBar}{rmLoss}\else\colorlet{rmBar}{rmArch}\fi
  \pgfmathsetmacro{\rmsx}{\rmxspd{#3}}%
  \fill[rmBar] (0,{\rmy-\rmBarH/2}) rectangle (\rmsx,{\rmy+\rmBarH/2});
  \ifnum#7=0
    \node[anchor=base east, text=white, font=\rmFVal, inner sep=0pt]
        at ({\rmsx-\rmValIn},{\rmy-\rmValDrop}) {#3\rmtimes};
  \else
    \node[anchor=base west, text=rmValOut, font=\rmFVal, inner sep=0pt]
        at ({\rmsx+\rmValOutGap},{\rmy-\rmValDrop}) {#3\rmtimes};
  \fi
  \pgfmathsetmacro{\rmlx}{\rmxlat{#5}}%
  \xdef\rmstarco{\rmstarco (\rmlx,\rmy)}%
  \ifnum#8=0
    \node[rmstarshape, fill=rmStar, minimum size=7.4pt] at (\rmlx,\rmy) {};
    \ifnum#9=0
      \node[anchor=base east, text=rmStarLbl, font=\rmFLat, inner sep=0pt]
          at ({\rmlx-\rmLatGap},{\rmy-\rmLatDrop}) {#5};
    \else
      \node[anchor=base west, text=rmStarLbl, font=\rmFLat, inner sep=0pt]
          at ({\rmlx+\rmLatGap},{\rmy-\rmLatDrop}) {#5};
    \fi
  \else
    \node[rmstarshape, fill=rmGold, minimum size=11.6pt] at (\rmlx,\rmy) {};
    \node[anchor=base west, text=rmGold, font=\rmFVal\bfseries, inner sep=0pt]
        at ({\rmlx+\rmGoldGap},{\rmy-\rmValDrop}) {DCore};
  \fi
  \ifnum#1>0
    \ifnum#1=1 \edef\rmvx{\rmElbowVa}\edef\rmhx{\rmElbowHa}\let\rmpy\rmpA
    \else      \edef\rmvx{\rmElbowVb}\edef\rmhx{\rmElbowHb}\let\rmpy\rmpB\fi
    \draw[rmTree, line width=0.72pt]
        (\rmvx,{\rmpy-\rmTreeTop}) -- (\rmvx,\rmy) -- (\rmhx,\rmy);
  \fi
  \ifcase#1\relax \let\rmlbx\rmLabA \def\rmbf{\bfseries}%
    \or           \let\rmlbx\rmLabB \def\rmbf{}%
    \or           \let\rmlbx\rmLabC \def\rmbf{}%
  \fi
  \node[anchor=base west, text=rmRowLbl, font=\rmFRow\rmbf, inner sep=0pt]
      at (\rmlbx,{\rmy-\rmBaseDrop}) {#2};
}

\pgfdeclarelayer{rmbg}
\pgfsetlayers{rmbg,main}
\tikzset{rmstarshape/.style={star, star points=5, star point ratio=2.1,
    draw=none, inner sep=0pt, outer sep=0pt}}

\begin{tikzpicture}[x=1pt, y=1pt, line cap=butt]
\rmreset

\rmrow{0}{Eagle-3 (7/10/16)}           {2.484}{4.512}{3.13}{0}{1}{0}{0}
\rmrow{1}{(7/10/60)}                   {2.670}{5.258}{3.19}{0}{0}{0}{0}
\rmrow{1}{(7/1/8)}                     {2.144}{3.537}{2.66}{0}{1}{0}{0}
\rmrow{0}{DFlash}                      {3.004}{4.340}{1.79}{0}{0}{0}{0}
\rmrow{1}{5L $\rightarrow$ 3L}         {2.877}{3.996}{1.42}{0}{0}{0}{0}
\rmrow{1}{5L $\rightarrow$ 7L}         {2.966}{4.466}{2.15}{0}{0}{0}{0} %
\rmrow{1}{w/ Anchor Pred.}                {3.050}{4.416}{1.79}{0}{0}{0}{0}
\rmrow{1}{CE $\rightarrow$ D-PACE}      {3.205}{4.640}{1.79}{1}{0}{0}{1}
\rmrow{1}{CE $\rightarrow$ KL+TV}                       {3.180}{4.557}{1.78}{1}{0}{0}{1}
\rmrow{1}{B=16$\rightarrow$B=8}      {2.768}{3.929}{1.77}{0}{0}{0}{1}
\rmrow{0}{DFlare}                      {3.109}{4.504}{1.79}{0}{0}{0}{0}
\rmrow{1}{5L $\rightarrow$ 7L}         {3.016}{4.555}{2.15}{0}{0}{0}{0} %
\rmrow{1}{9HS $\rightarrow$ 5HS}       {3.103}{4.485}{1.79}{0}{0}{0}{0}
\rmrow{2}{{\fontsize{7.2}{8.5}\selectfont CE $\rightarrow$ D-PACE}}                       {3.136}{4.536}{1.78}{1}{0}{0}{1}
\rmrow{1}{CE $\rightarrow$ D-PACE}                       {3.261}{4.735}{1.79}{1}{0}{1}{0}
\rmrow{2}{{\fontsize{7.2}{8.5}\selectfont w/ Anchor Pred.}}                {3.251}{4.716}{1.79}{0}{0}{0}{1}
\rmrow{0}{Domino}                      {3.150}{4.984}{2.64}{0}{0}{0}{0}
\rmrow{1}{w/ Cand. Pool}               {3.241}{4.984}{2.35}{0}{0}{0}{0}
\rmrow{1}{w/o Curr.}                   {3.116}{4.931}{2.65}{0}{0}{0}{0}
\rmrow{2}{{\fontsize{7.2}{8.5}\selectfont CE $\rightarrow$ D-PACE}}                       {3.102}{4.902}{2.64}{1}{0}{0}{0} %
\rmrow{0}{DSpark}                      {3.146}{4.911}{2.71}{0}{0}{0}{0}
\rmrow{1}{Markov $\rightarrow$ RNN}    {3.139}{5.047}{2.88}{0}{0}{0}{0}
\rmrow{1}{{\fontsize{7.2}{8.5}\selectfont CE+TV $\rightarrow$ KL+TV}}                       {3.181}{4.974}{2.71}{1}{0}{0}{0}
\rmrow{1}{{\fontsize{7.2}{8.5}\selectfont CE+TV $\rightarrow$ D-PAKL}}                       {3.153}{4.926}{2.71}{1}{0}{0}{0}
\rmrow{1}{B=16$\rightarrow$B=8}      {2.854}{4.195}{2.25}{0}{0}{0}{0}

\begin{pgfonlayer}{rmbg}
  \foreach \t in {3,4,5}
    \draw[rmGridA, line width=0.62pt] ({\rmxtau{\t}},0) -- ({\rmxtau{\t}},\rmBot);
  \foreach \s in {3,4}
    \draw[rmGridB, line width=0.62pt] ({\rmxspd{\s}},0) -- ({\rmxspd{\s}},\rmBot);
  \draw[rmEdge, line width=0.72pt] (0,0) -- (0,\rmBot);
  \draw[rmStar, line width=0.98pt, dash pattern=on 3.0pt off 2.0pt]
      plot coordinates {\rmstarco};
\end{pgfonlayer}

\draw[rmAxis, line width=0.92pt] (0,0) -- (\rmW,0);
\draw[rmAxis, line width=0.92pt] (0,\rmBot) -- (\rmW,\rmBot);
\node[anchor=base east, text=rmAxis, font=\rmFAx, inner sep=0pt]
    at (-5.7,\rmTickUp) {Speedup};
\foreach \s in {3,4}
  \node[anchor=base, text=rmAxis, font=\rmFAx, inner sep=0pt]
      at ({\rmxspd{\s}},\rmTickUp) {\s\rmtimes};
\node[anchor=base east, text=rmAxis, font=\rmFAx, inner sep=0pt]
    at (-5.4,{\rmBot-\rmTickDown}) {$\tau$};
\foreach \t in {3,4,5}
  \node[anchor=base, text=rmAxis, font=\rmFAx, inner sep=0pt]
      at ({\rmxtau{\t}},{\rmBot-\rmTickDown}) {\t};
\end{tikzpicture}

\caption{Key design ablation roadmap on Qwen3-8B ($B{=}16, T_t \vert T_d{=}1 \vert 0$, weight-averaged over ten benchmarks).
Most points are single design-search runs; branches illustrate design changes.
\textcolor{rmArch}{\ensuremath{\blacksquare}}/\textcolor{rmLoss}{\ensuremath{\blacksquare}}:
architecture/loss-ablation speedup;
\textcolor{rmTau}{\ensuremath{\blacksquare}}: acceptance length;
\textcolor{rmStar}{\ensuremath{\bigstar}}/\textcolor{rmGold}{\ensuremath{\bigstar}}:
proposal latency, gold $=$ DCore.
L: backbone layer count; HS: number of target layers supplying hidden states;
Curr.: curriculum learning; Anchor Pred.: anchor-position prediction.
Cand. Pool restricts conditioned predictions to top-$K$ marginal candidates
following \citet{oda2026treesfrommarginals}.
KL$+$TV is an LK-style objective~\citep{samarin2026lk};
D-PAKL pairs KL supervision with D-PACE position weights.}
\label{fig:roadmap}
\end{figure}
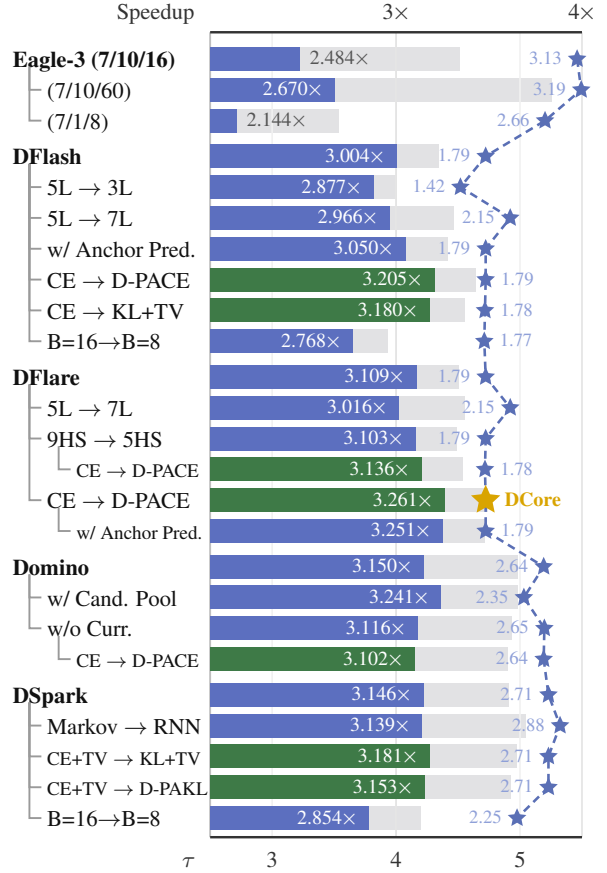

%% file: table/Table-main-results.tex
\providecommand{\ulm}[1]{\smash{\underline{#1}}}
\begin{table*}[!t]
\centering
\caption{\textbf{Main results.} Cells are $\textsc{TPS}_{(\tau)}$: tokens/s with acceptance length $\tau$ subscripted, averaged over each entropy half's five benchmarks. Entropy split, temperature pair $T_t|T_d$, budget \textsc{Bgt.}, Eagle-3 configuration triple and the missing $1|1$ trees (the dashes) all follow \S\ref{sec:evaluation}. \textbf{Bold}/\underline{underline}: best/second-best in a column of one target--\textsc{Bgt.}\ block, the only comparable unit, \textsc{TPS} and $\tau$ ranked apart. Per-benchmark values: App.~\ref{sec:appendix-per-benchmark}.}
\label{tab:main_results_temps}
\setlength{\tabcolsep}{6pt}
\scriptsize
\resizebox{\ifdim\width>\textwidth \textwidth\else \width\fi}{!}{%
\begin{tabular}{cclcccccc}
\toprule
& & & \multicolumn{2}{c}{$T_t|T_d=0|0$} & \multicolumn{2}{c}{$T_t|T_d=1|0$} & \multicolumn{2}{c}{$T_t|T_d=1|1$} \\
\cmidrule(lr){4-5} \cmidrule(lr){6-7} \cmidrule(lr){8-9}
\textsc{Tgt.} & \textsc{Bgt.} & Method & \textsc{Low-Ent.} & \textsc{High-Ent.} & \textsc{Low-Ent.} & \textsc{High-Ent.} & \textsc{Low-Ent.} & \textsc{High-Ent.} \\
\midrule
\multirow{17}{*}{\rotatebox[origin=c]{90}{Qwen3-4B}} & \multirow{8}{*}{8} & Eagle-3 (7/1/8) & $616.1_{(4.69)}$ & $435.0_{(3.20)}$ & $514.9_{(4.18)}$ & $347.9_{(2.77)}$ & $526.6_{(4.16)}$ & $358.5_{(2.76)}$ \\
& & Eagle-3 (7/4/8) & $591.9_{(4.93)}$ & $439.6_{(3.57)}$ & $509.5_{(4.48)}$ & $366.7_{(3.16)}$ & --- & --- \\
& & DFlash & $809.3_{(5.37)}$ & $595.4_{(3.86)}$ & $697.4_{(4.73)}$ & $487.7_{(3.23)}$ & $703.3_{(4.69)}$ & $487.4_{(3.19)}$ \\
& & DFlare & $815.5_{(5.43)}$ & $604.3_{(3.94)}$ & $707.4_{(4.77)}$ & $495.7_{(3.28)}$ & $712.5_{(4.75)}$ & $497.2_{(3.26)}$ \\
& & D-PACE & $\ulm{825.8}_{(5.49)}$ & $611.0_{(3.96)}$ & $\ulm{715.0}_{(4.83)}$ & $494.3_{(3.27)}$ & $729.6_{(4.88)}$ & $510.7_{(3.35)}$ \\
& & Domino & $807.4_{(\ulm{5.71})}$ & $\ulm{616.8}_{(\mathbf{4.26})}$ & $699.3_{(\ulm{5.02})}$ & $495.5_{(\mathbf{3.49})}$ & $729.0_{(\ulm{5.20})}$ & $\ulm{528.6}_{(\ulm{3.70})}$ \\
& & DSpark & $810.2_{(\mathbf{5.76})}$ & $607.4_{(\ulm{4.21})}$ & $708.5_{(\mathbf{5.04})}$ & $\ulm{496.1}_{(\ulm{3.45})}$ & $\mathbf{753.3}_{(\mathbf{5.38})}$ & $\mathbf{546.3}_{(\mathbf{3.83})}$ \\
& & DCore & $\mathbf{837.7}_{(5.57)}$ & $\mathbf{622.0}_{(4.05)}$ & $\mathbf{725.0}_{(4.91)}$ & $\mathbf{502.7}_{(3.33)}$ & $\ulm{743.3}_{(4.97)}$ & $521.8_{(3.42)}$ \\
\cmidrule(l){2-9}
& \multirow{9}{*}{16} & Eagle-3 (15/1/16) & $546.6_{(5.53)}$ & $351.3_{(3.43)}$ & $443.4_{(4.77)}$ & $274.8_{(2.88)}$ & $438.4_{(4.69)}$ & $273.9_{(2.85)}$ \\
& & Eagle-3 (7/10/16) & $661.0_{(5.71)}$ & $495.9_{(4.14)}$ & $570.8_{(5.23)}$ & $408.6_{(3.65)}$ & --- & --- \\
& & Eagle-3 (7/10/60) & $696.2_{(6.53)}$ & $543.6_{(\mathbf{4.91})}$ & $597.6_{(\ulm{6.07})}$ & $440.0_{(\mathbf{4.34})}$ & --- & --- \\
& & DFlash & $968.7_{(6.60)}$ & $624.4_{(4.14)}$ & $799.6_{(5.52)}$ & $489.3_{(3.30)}$ & $776.6_{(5.30)}$ & $482.5_{(3.22)}$ \\
& & DFlare & $980.9_{(6.69)}$ & $630.7_{(4.19)}$ & $809.4_{(5.60)}$ & $498.7_{(3.37)}$ & $793.9_{(5.42)}$ & $490.4_{(3.27)}$ \\
& & D-PACE & $\ulm{1009.0}_{(6.92)}$ & $\ulm{645.6}_{(4.28)}$ & $\ulm{831.2}_{(5.75)}$ & $\ulm{501.8}_{(3.38)}$ & $847.2_{(5.80)}$ & $510.5_{(3.40)}$ \\
& & Domino & $975.5_{(\mathbf{7.56})}$ & $639.2_{(\ulm{4.79})}$ & $794.9_{(\mathbf{6.21})}$ & $484.6_{(\ulm{3.69})}$ & $828.0_{(\ulm{6.48})}$ & $511.2_{(\ulm{3.90})}$ \\
& & DSpark & $979.4_{(\mathbf{7.56})}$ & $626.2_{(4.67)}$ & $807.0_{(\mathbf{6.21})}$ & $487.8_{(3.66)}$ & $\mathbf{870.8}_{(\mathbf{6.72})}$ & $\mathbf{538.6}_{(\mathbf{4.05})}$ \\
& & DCore & $\mathbf{1037.6}_{(\ulm{7.10})}$ & $\mathbf{654.7}_{(4.35)}$ & $\mathbf{850.7}_{(5.89)}$ & $\mathbf{511.2}_{(3.45)}$ & $\ulm{866.8}_{(5.94)}$ & $\ulm{524.7}_{(3.50)}$ \\
\midrule
\multirow{17}{*}{\rotatebox[origin=c]{90}{Qwen3-8B}} & \multirow{8}{*}{8} & Eagle-3 (7/1/8) & $465.2_{(4.94)}$ & $322.0_{(3.33)}$ & $384.6_{(4.30)}$ & $252.4_{(2.78)}$ & $398.4_{(4.37)}$ & $271.2_{(2.91)}$ \\
& & Eagle-3 (7/4/8) & $458.4_{(5.16)}$ & $336.3_{(3.69)}$ & $386.5_{(4.61)}$ & $271.1_{(3.18)}$ & --- & --- \\
& & DFlash & $574.3_{(5.48)}$ & $409.7_{(3.83)}$ & $490.0_{(4.73)}$ & $328.4_{(3.11)}$ & $486.5_{(4.66)}$ & $335.5_{(3.16)}$ \\
& & DFlare & $583.5_{(5.58)}$ & $417.9_{(3.91)}$ & $496.0_{(4.79)}$ & $333.6_{(3.16)}$ & $500.6_{(4.80)}$ & $345.0_{(3.25)}$ \\
& & D-PACE & $\ulm{599.2}_{(5.72)}$ & $426.2_{(3.99)}$ & $\ulm{505.1}_{(4.89)}$ & $338.5_{(3.21)}$ & $522.3_{(5.02)}$ & $358.6_{(3.38)}$ \\
& & Domino & $592.8_{(\ulm{5.93})}$ & $431.8_{(\ulm{4.22})}$ & $500.5_{(\ulm{5.05})}$ & $339.3_{(\mathbf{3.35})}$ & $\ulm{544.3}_{(\ulm{5.48})}$ & $\ulm{384.0}_{(\ulm{3.79})}$ \\
& & DSpark & $594.3_{(\mathbf{5.96})}$ & $\ulm{432.0}_{(\mathbf{4.23})}$ & $505.0_{(\mathbf{5.06})}$ & $\ulm{339.7}_{(\ulm{3.34})}$ & $\mathbf{550.9}_{(\mathbf{5.58})}$ & $\mathbf{391.5}_{(\mathbf{3.88})}$ \\
& & DCore & $\mathbf{601.5}_{(5.76)}$ & $\mathbf{433.2}_{(4.06)}$ & $\mathbf{510.5}_{(4.93)}$ & $\mathbf{341.1}_{(3.23)}$ & $528.5_{(5.08)}$ & $364.0_{(3.43)}$ \\
\cmidrule(l){2-9}
& \multirow{9}{*}{16} & Eagle-3 (15/1/16) & $412.9_{(5.77)}$ & $261.8_{(3.56)}$ & $329.7_{(4.82)}$ & $201.4_{(2.88)}$ & $330.2_{(4.83)}$ & $209.0_{(2.99)}$ \\
& & Eagle-3 (7/10/16) & $506.8_{(5.98)}$ & $374.0_{(4.30)}$ & $433.6_{(5.35)}$ & $304.2_{(\ulm{3.67})}$ & --- & --- \\
& & Eagle-3 (7/10/60) & $533.5_{(6.73)}$ & $407.5_{(\mathbf{5.00})}$ & $462.2_{(6.19)}$ & $331.1_{(\mathbf{4.33})}$ & --- & --- \\
& & DFlash & $693.9_{(6.78)}$ & $425.0_{(4.03)}$ & $562.3_{(5.53)}$ & $332.2_{(3.19)}$ & $518.6_{(5.07)}$ & $322.7_{(3.09)}$ \\
& & DFlare & $720.9_{(7.06)}$ & $443.7_{(4.22)}$ & $576.7_{(5.68)}$ & $342.6_{(3.29)}$ & $564.0_{(5.52)}$ & $345.8_{(3.31)}$ \\
& & D-PACE & $\ulm{758.7}_{(7.46)}$ & $445.9_{(4.30)}$ & $\ulm{602.5}_{(5.95)}$ & $\ulm{347.1}_{(3.34)}$ & $618.9_{(6.09)}$ & $364.6_{(3.49)}$ \\
& & Domino & $755.8_{(\mathbf{8.10})}$ & $\mathbf{465.7}_{(\ulm{4.83})}$ & $592.9_{(\mathbf{6.39})}$ & $339.8_{(3.56)}$ & $\ulm{651.4}_{(\ulm{7.06})}$ & $\ulm{388.7}_{(\ulm{4.09})}$ \\
& & DSpark & $752.7_{(\ulm{8.04})}$ & $\ulm{464.0}_{(4.79)}$ & $594.2_{(\ulm{6.34})}$ & $342.3_{(3.55)}$ & $\mathbf{660.9}_{(\mathbf{7.12})}$ & $\mathbf{399.0}_{(\mathbf{4.18})}$ \\
& & DCore & $\mathbf{772.6}_{(7.59)}$ & $462.4_{(4.39)}$ & $\mathbf{612.5}_{(6.05)}$ & $\mathbf{350.7}_{(3.37)}$ & $630.2_{(6.20)}$ & $370.7_{(3.55)}$ \\
\bottomrule
\end{tabular}%
}
\end{table*}

%% file: section/analysis.tex
\section{Operating Conditions and Throughput Returns}
\label{sec:analysis}

\input{figure/Figure-accepted-paths}

To answer the second question, we examine when the acceptance gains
identified by the component analysis repay their cost under different
workloads and execution environments. We first
examine uncertainty on the acceptance side, progressing from accepted paths
to task-level differences and budget returns. We then examine serving
conditions on the execution side, relating changes in load and target scale
to drafter comparisons and execution costs.

\subsection{Task uncertainty and budget returns}

\paragraph{Accepted paths motivate uncertainty-aware budgets.}
Across Eagle-3, DFlash, and DSpark, accepted prefixes tend to have
lower target entropy than the rejected position ending the path. This
contrast recurs at both budgets on both target under $1|0$
(Figs.~\ref{fig:draftability} and~\ref{fig:pq_entropy_heatmap}), suggesting uncertainty as a signal for
where further drafting may yield diminishing returns.

One deployment hypothesis is to use draft entropy as a proxy for target
uncertainty: shorten the proposal horizon and reduce the verification
budget when draft entropy is high, while allowing more speculation when
it is low. This could limit computation on drafts likely to be rejected.
The proxy and adaptive policy remain to be evaluated against end-to-end
throughput, since acceptance also depends on draft--target agreement
(App.~\ref{sec:appendix-entropy}).

\paragraph{Across tasks: uncertainty and acceptance.}
We next examine benchmark-level uncertainty as a complementary view of
proposal acceptance. Task entropy aggregates the target's raw $T{=}1$
uncertainty on separate greedy trajectories, whereas the heatmaps measure
uncertainty at positions on realized decoding paths. On the larger target
at $B{=}16$, $1|0$, $C{=}1$, mean $\tau$ falls from $5.35$ to $3.67$ for
Eagle-3, $5.72$ to $3.27$ for the block-parallel family, and $6.36$ to $3.55$
for the semi-AR family between the low- and high-entropy groups.

Across all ten benchmarks, higher task entropy is consistently associated
with shorter acceptance lengths across the evaluated drafter families
(App. Fig.~\ref{fig:entropy-continuous}). The next question is whether
it also helps identify where a larger budget improves throughput.

\paragraph{Budget returns depend on uncertainty and proposal structure.}
Block-parallel recipes best preserve effective efficiency as $B$ increases
from $8$ to $16$: efficiency falls by only $1.3$--$2.2\%$, compared with
$5.1$--$6.4\%$ for semi-AR recipes. Both families are nevertheless sensitive
to task entropy, gaining up to about $20\%$ in throughput on low-entropy
tasks but at most $2.8\%$ on high-entropy tasks. Eagle-3's tree instead
achieves similar throughput gains of about $12\%$ in both entropy groups,
suggesting that wider candidate coverage may make budget returns less
sensitive to task uncertainty. Extending its token-by-token chain from
$7$ to $15$ steps lowers throughput in both groups, as the additional
sequential drafting cost outweighs the acceptance gain
(Fig.~\ref{fig:elasticity_budget_entropy_qwen3_8b}).

\input{figure/Figure-budget-returns.tex}

\subsection{Serving conditions and execution costs}
Task uncertainty helps characterize acceptance returns. We deepen the
comparison by examining how serving load changes the preferred drafter
and budget, and how target scale changes their relative execution costs.

\input{figure/Figure-concurrency-8b-b8.tex}

\paragraph{Concurrency: the preferred budget and drafter change with load.}
At high concurrency, a smaller budget preserves more of the speculative
speedup on the measured mixed workload. Under $1|0$ on the larger target,
budget-$8$ non-AR recipes retain about $1.43$--$1.50\times$ speedup at
$C{=}64$, compared with $0.98$--$1.10\times$ for budget $16$
(Fig.~\ref{fig:serving_tps_qwen3_8b_bgt8};
App. Fig.~\ref{fig:serving_tps_concurrency}). The task-level benefit
of a larger budget therefore needs to be assessed under the intended load.

At $B{=}8$, DCore achieves the highest or tied speedup under $0|0$ and
$1|0$ at the measured concurrencies $C\le16$, reaching $3.27\times$ and
$2.67\times$, respectively, at $C{=}1$
(Fig.~\ref{fig:serving_tps_qwen3_8b_bgt8}). At $C{=}32$ and $64$, it
remains within about $1.3\%$ of the best result in these two settings;
for example, under $1|0$ at $C{=}64$, DCore reaches $1.493\times$ versus
Domino's $1.501\times$, a $0.5\%$ gap. Under $1|1$, semi-AR recipes
instead lead, with Domino and DSpark occupying the top two positions across
the measured concurrency range. For example, DSpark's advantage over DCore
grows from about $6.2\%$ at $C{=}1$ ($2.96\times$ versus $2.78\times$)
to $10.5\%$ at $C{=}64$ ($1.71\times$ versus $1.55\times$).
DSpark's edge over Domino may partly reflect its TV-related loss alongside
CE, which explicitly promotes distributional overlap.
These results highlight DCore's competitiveness under greedy drafting and the semi-AR
family's advantage under sampled drafting, with the ordering also varying
with load.
Eagle-3 is chain-decoded here.
App. Fig.~\ref{fig:serving_profile_bgt60} provides the separate chain/tree comparison.

\input{figure/Figure-sampling-gap}

\paragraph{Sampled drafts expose prefix-independent proposals.}
A block-parallel drafter proposes position $j$ from $q_j$ without seeing
the tokens drafted before it. When those tokens are sampled ($T_d{=}1$),
its rejection rate at $j$ is at least half the mean total-variation
distance between the target's next-token distributions after two
independently reached prefixes; under $T_d{=}0$ the only reached prefix is
the greedy path and the bound is zero (App.~\ref{sec:app-sd-theory}).
Fig.~\ref{fig:sampling_gap} localizes the change from $1|0$ to $1|1$ on
the larger target. At $j{=}0$, where both settings condition
on the same context, sampling raises entropy-standardized acceptance by
$0.056$--$0.071$ for every non-AR recipe. At $j{\ge}1$, the Shapley term
is $-0.06$ to $-0.41$ committed tokens for block-parallel recipes and
$+0.14$ to $+0.25$ for semi-AR recipes, accounting for $76$--$94\%$ of
each semi-AR versus block-parallel difference in $\Delta\tau$. After a
non-argmax predecessor, block-parallel acceptance falls $0.38$--$0.43$ below the
rate after an argmax predecessor, against $0.08$--$0.09$ for
semi-AR recipes, and $\mathrm{KL}(q\Vert p)$ rises by $9.0$--$10.1$ versus
$0.9$--$1.0$ nats. These measurements are consistent with the
marginalization over prior tokens noted in prior work
\citep{cheng2026dspark,karimi2026dblast,rusanovsky2026lilicorr}.

\paragraph{Target scale: higher acceptance can become more valuable.}
When drafting latency grows more slowly than target latency, the relative
drafting premium can shrink, making higher acceptance more valuable.
Matched $0|0$, $C{=}1$ profiles illustrate this pattern between the two
evaluated target sizes. At $B{=}16$, DCore's target-forward time
rises from $4.63$ to $6.97$~ms, while its proposal latency rises from $1.33$
to $1.78$~ms (Tab.~\ref{tab:serving-stage-latency}). DSpark's recorded
device-step premium over DCore narrows from $13.3\%$ to $9.5\%$; at $B{=}8$,
it narrows from $7.2\%$ to $5.0\%$.
These measurements motivate a conditional interpretation: a more expressive
drafter becomes more competitive if it retains its acceptance advantage
while its relative complete step cost falls (Eq.~\eqref{eq:payoff}).

%% file: figure/Figure-accepted-paths.tex
\begin{figure*}[t]
    \centering
    \def\hcell{7.0pt}
    \begin{tikzpicture}[x=\hcell, y=\hcell,
        frontier/.style={draw=black, line width=0.45pt},
        insetframe/.style={draw=black!55, line width=0.4pt},
        panelframe/.style={draw=black!35, line width=0.4pt},
        sharebar/.style={fill=black!45},
        sharegrid/.style={draw=black!20, line width=0.3pt},
        ticklab/.style={font=\tiny, inner sep=1pt},
        axlab/.style={font=\scriptsize, inner sep=1pt},
    ]
    \begin{scope}[shift={(0,0)}]
    \fill[vir76] (0,15) rectangle ++(1,1);
    \fill[vir29] (0,14) rectangle ++(1,1);
    \fill[vir72] (1,14) rectangle ++(1,1);
    \fill[vir23] (0,13) rectangle ++(1,1);
    \fill[vir27] (1,13) rectangle ++(1,1);
    \fill[vir68] (2,13) rectangle ++(1,1);
    \fill[vir20] (0,12) rectangle ++(1,1);
    \fill[vir20] (1,12) rectangle ++(1,1);
    \fill[vir24] (2,12) rectangle ++(1,1);
    \fill[vir65] (3,12) rectangle ++(1,1);
    \fill[vir18] (0,11) rectangle ++(1,1);
    \fill[vir18] (1,11) rectangle ++(1,1);
    \fill[vir18] (2,11) rectangle ++(1,1);
    \fill[vir21] (3,11) rectangle ++(1,1);
    \fill[vir62] (4,11) rectangle ++(1,1);
    \fill[vir16] (0,10) rectangle ++(1,1);
    \fill[vir15] (1,10) rectangle ++(1,1);
    \fill[vir15] (2,10) rectangle ++(1,1);
    \fill[vir16] (3,10) rectangle ++(1,1);
    \fill[vir19] (4,10) rectangle ++(1,1);
    \fill[vir60] (5,10) rectangle ++(1,1);
    \fill[vir14] (0,9) rectangle ++(1,1);
    \fill[vir13] (1,9) rectangle ++(1,1);
    \fill[vir12] (2,9) rectangle ++(1,1);
    \fill[vir12] (3,9) rectangle ++(1,1);
    \fill[vir15] (4,9) rectangle ++(1,1);
    \fill[vir18] (5,9) rectangle ++(1,1);
    \fill[vir55] (6,9) rectangle ++(1,1);
    \fill[vir13] (0,8) rectangle ++(1,1);
    \fill[vir12] (1,8) rectangle ++(1,1);
    \fill[vir11] (2,8) rectangle ++(1,1);
    \fill[vir10] (3,8) rectangle ++(1,1);
    \fill[vir11] (4,8) rectangle ++(1,1);
    \fill[vir13] (5,8) rectangle ++(1,1);
    \fill[vir17] (6,8) rectangle ++(1,1);
    \fill[vir51] (7,8) rectangle ++(1,1);
    \fill[vir12] (0,7) rectangle ++(1,1);
    \fill[vir11] (1,7) rectangle ++(1,1);
    \fill[vir10] (2,7) rectangle ++(1,1);
    \fill[vir8] (3,7) rectangle ++(1,1);
    \fill[vir9] (4,7) rectangle ++(1,1);
    \fill[vir10] (5,7) rectangle ++(1,1);
    \fill[vir12] (6,7) rectangle ++(1,1);
    \fill[vir14] (7,7) rectangle ++(1,1);
    \fill[vir45] (8,7) rectangle ++(1,1);
    \fill[vir11] (0,6) rectangle ++(1,1);
    \fill[vir10] (1,6) rectangle ++(1,1);
    \fill[vir9] (2,6) rectangle ++(1,1);
    \fill[vir8] (3,6) rectangle ++(1,1);
    \fill[vir8] (4,6) rectangle ++(1,1);
    \fill[vir8] (5,6) rectangle ++(1,1);
    \fill[vir8] (6,6) rectangle ++(1,1);
    \fill[vir10] (7,6) rectangle ++(1,1);
    \fill[vir13] (8,6) rectangle ++(1,1);
    \fill[vir40] (9,6) rectangle ++(1,1);
    \fill[vir11] (0,5) rectangle ++(1,1);
    \fill[vir9] (1,5) rectangle ++(1,1);
    \fill[vir8] (2,5) rectangle ++(1,1);
    \fill[vir7] (3,5) rectangle ++(1,1);
    \fill[vir6] (4,5) rectangle ++(1,1);
    \fill[vir6] (5,5) rectangle ++(1,1);
    \fill[vir6] (6,5) rectangle ++(1,1);
    \fill[vir7] (7,5) rectangle ++(1,1);
    \fill[vir8] (8,5) rectangle ++(1,1);
    \fill[vir11] (9,5) rectangle ++(1,1);
    \fill[vir35] (10,5) rectangle ++(1,1);
    \fill[vir10] (0,4) rectangle ++(1,1);
    \fill[vir8] (1,4) rectangle ++(1,1);
    \fill[vir7] (2,4) rectangle ++(1,1);
    \fill[vir6] (3,4) rectangle ++(1,1);
    \fill[vir6] (4,4) rectangle ++(1,1);
    \fill[vir6] (5,4) rectangle ++(1,1);
    \fill[vir5] (6,4) rectangle ++(1,1);
    \fill[vir5] (7,4) rectangle ++(1,1);
    \fill[vir6] (8,4) rectangle ++(1,1);
    \fill[vir7] (9,4) rectangle ++(1,1);
    \fill[vir10] (10,4) rectangle ++(1,1);
    \fill[vir31] (11,4) rectangle ++(1,1);
    \fill[vir9] (0,3) rectangle ++(1,1);
    \fill[vir8] (1,3) rectangle ++(1,1);
    \fill[vir7] (2,3) rectangle ++(1,1);
    \fill[vir6] (3,3) rectangle ++(1,1);
    \fill[vir5] (4,3) rectangle ++(1,1);
    \fill[vir5] (5,3) rectangle ++(1,1);
    \fill[vir5] (6,3) rectangle ++(1,1);
    \fill[vir5] (7,3) rectangle ++(1,1);
    \fill[vir5] (8,3) rectangle ++(1,1);
    \fill[vir5] (9,3) rectangle ++(1,1);
    \fill[vir6] (10,3) rectangle ++(1,1);
    \fill[vir9] (11,3) rectangle ++(1,1);
    \fill[vir27] (12,3) rectangle ++(1,1);
    \fill[vir9] (0,2) rectangle ++(1,1);
    \fill[vir8] (1,2) rectangle ++(1,1);
    \fill[vir6] (2,2) rectangle ++(1,1);
    \fill[vir5] (3,2) rectangle ++(1,1);
    \fill[vir5] (4,2) rectangle ++(1,1);
    \fill[vir5] (5,2) rectangle ++(1,1);
    \fill[vir5] (6,2) rectangle ++(1,1);
    \fill[vir4] (7,2) rectangle ++(1,1);
    \fill[vir4] (8,2) rectangle ++(1,1);
    \fill[vir4] (9,2) rectangle ++(1,1);
    \fill[vir5] (10,2) rectangle ++(1,1);
    \fill[vir5] (11,2) rectangle ++(1,1);
    \fill[vir8] (12,2) rectangle ++(1,1);
    \fill[vir23] (13,2) rectangle ++(1,1);
    \fill[vir9] (0,1) rectangle ++(1,1);
    \fill[vir7] (1,1) rectangle ++(1,1);
    \fill[vir6] (2,1) rectangle ++(1,1);
    \fill[vir5] (3,1) rectangle ++(1,1);
    \fill[vir4] (4,1) rectangle ++(1,1);
    \fill[vir5] (5,1) rectangle ++(1,1);
    \fill[vir4] (6,1) rectangle ++(1,1);
    \fill[vir4] (7,1) rectangle ++(1,1);
    \fill[vir4] (8,1) rectangle ++(1,1);
    \fill[vir4] (9,1) rectangle ++(1,1);
    \fill[vir4] (10,1) rectangle ++(1,1);
    \fill[vir4] (11,1) rectangle ++(1,1);
    \fill[vir4] (12,1) rectangle ++(1,1);
    \fill[vir6] (13,1) rectangle ++(1,1);
    \fill[vir20] (14,1) rectangle ++(1,1);
    \fill[vir9] (0,0) rectangle ++(1,1);
    \fill[vir8] (1,0) rectangle ++(1,1);
    \fill[vir6] (2,0) rectangle ++(1,1);
    \fill[vir5] (3,0) rectangle ++(1,1);
    \fill[vir5] (4,0) rectangle ++(1,1);
    \fill[vir4] (5,0) rectangle ++(1,1);
    \fill[vir4] (6,0) rectangle ++(1,1);
    \fill[vir4] (7,0) rectangle ++(1,1);
    \fill[vir4] (8,0) rectangle ++(1,1);
    \fill[vir3] (9,0) rectangle ++(1,1);
    \fill[vir4] (10,0) rectangle ++(1,1);
    \fill[vir3] (11,0) rectangle ++(1,1);
    \fill[vir3] (12,0) rectangle ++(1,1);
    \fill[vir4] (13,0) rectangle ++(1,1);
    \fill[vir4] (14,0) rectangle ++(1,1);
    \draw[frontier] (0,15) rectangle ++(1,1);
    \draw[frontier] (1,14) rectangle ++(1,1);
    \draw[frontier] (2,13) rectangle ++(1,1);
    \draw[frontier] (3,12) rectangle ++(1,1);
    \draw[frontier] (4,11) rectangle ++(1,1);
    \draw[frontier] (5,10) rectangle ++(1,1);
    \draw[frontier] (6,9) rectangle ++(1,1);
    \draw[frontier] (7,8) rectangle ++(1,1);
    \draw[frontier] (8,7) rectangle ++(1,1);
    \draw[frontier] (9,6) rectangle ++(1,1);
    \draw[frontier] (10,5) rectangle ++(1,1);
    \draw[frontier] (11,4) rectangle ++(1,1);
    \draw[frontier] (12,3) rectangle ++(1,1);
    \draw[frontier] (13,2) rectangle ++(1,1);
    \draw[frontier] (14,1) rectangle ++(1,1);
    \draw[sharegrid] (16.0111,0) -- (16.0111,16);
    \draw[sharegrid] (16.6222,0) -- (16.6222,16);
    \draw[sharegrid] (17.2333,0) -- (17.2333,16);
    \fill[sharebar] (15.4000,15.1400) rectangle ++(2.1160,0.72);
    \fill[sharebar] (15.4000,14.1400) rectangle ++(1.2761,0.72);
    \fill[sharebar] (15.4000,13.1400) rectangle ++(0.7929,0.72);
    \fill[sharebar] (15.4000,12.1400) rectangle ++(0.5078,0.72);
    \fill[sharebar] (15.4000,11.1400) rectangle ++(0.3427,0.72);
    \fill[sharebar] (15.4000,10.1400) rectangle ++(0.2447,0.72);
    \fill[sharebar] (15.4000,9.1400) rectangle ++(0.1854,0.72);
    \fill[sharebar] (15.4000,8.1400) rectangle ++(0.1446,0.72);
    \fill[sharebar] (15.4000,7.1400) rectangle ++(0.1166,0.72);
    \fill[sharebar] (15.4000,6.1400) rectangle ++(0.0933,0.72);
    \fill[sharebar] (15.4000,5.1400) rectangle ++(0.0766,0.72);
    \fill[sharebar] (15.4000,4.1400) rectangle ++(0.0607,0.72);
    \fill[sharebar] (15.4000,3.1400) rectangle ++(0.0472,0.72);
    \fill[sharebar] (15.4000,2.1400) rectangle ++(0.0360,0.72);
    \fill[sharebar] (15.4000,1.1400) rectangle ++(0.0269,0.72);
    \fill[sharebar] (15.4000,0.1400) rectangle ++(0.0436,0.72);
    \draw[panelframe] (15.4000,0) -- (15.4000,16);
    \begin{scope}[shift={(8.35,8.5)}, xscale=0.68, yscale=0.68]
    \fill[white] (-0.2,-0.2) rectangle (9.1,8.2);
    \fill[vir75] (0,7) rectangle ++(1,1);
    \fill[vir29] (0,6) rectangle ++(1,1);
    \fill[vir71] (1,6) rectangle ++(1,1);
    \fill[vir22] (0,5) rectangle ++(1,1);
    \fill[vir26] (1,5) rectangle ++(1,1);
    \fill[vir68] (2,5) rectangle ++(1,1);
    \fill[vir19] (0,4) rectangle ++(1,1);
    \fill[vir19] (1,4) rectangle ++(1,1);
    \fill[vir23] (2,4) rectangle ++(1,1);
    \fill[vir65] (3,4) rectangle ++(1,1);
    \fill[vir17] (0,3) rectangle ++(1,1);
    \fill[vir17] (1,3) rectangle ++(1,1);
    \fill[vir17] (2,3) rectangle ++(1,1);
    \fill[vir20] (3,3) rectangle ++(1,1);
    \fill[vir62] (4,3) rectangle ++(1,1);
    \fill[vir15] (0,2) rectangle ++(1,1);
    \fill[vir14] (1,2) rectangle ++(1,1);
    \fill[vir14] (2,2) rectangle ++(1,1);
    \fill[vir15] (3,2) rectangle ++(1,1);
    \fill[vir19] (4,2) rectangle ++(1,1);
    \fill[vir59] (5,2) rectangle ++(1,1);
    \fill[vir14] (0,1) rectangle ++(1,1);
    \fill[vir12] (1,1) rectangle ++(1,1);
    \fill[vir12] (2,1) rectangle ++(1,1);
    \fill[vir12] (3,1) rectangle ++(1,1);
    \fill[vir14] (4,1) rectangle ++(1,1);
    \fill[vir17] (5,1) rectangle ++(1,1);
    \fill[vir55] (6,1) rectangle ++(1,1);
    \fill[vir11] (0,0) rectangle ++(1,1);
    \fill[vir10] (1,0) rectangle ++(1,1);
    \fill[vir9] (2,0) rectangle ++(1,1);
    \fill[vir8] (3,0) rectangle ++(1,1);
    \fill[vir8] (4,0) rectangle ++(1,1);
    \fill[vir8] (5,0) rectangle ++(1,1);
    \fill[vir10] (6,0) rectangle ++(1,1);
    \draw[frontier] (0,7) rectangle ++(1,1);
    \draw[frontier] (1,6) rectangle ++(1,1);
    \draw[frontier] (2,5) rectangle ++(1,1);
    \draw[frontier] (3,4) rectangle ++(1,1);
    \draw[frontier] (4,3) rectangle ++(1,1);
    \draw[frontier] (5,2) rectangle ++(1,1);
    \draw[frontier] (6,1) rectangle ++(1,1);
    \draw[sharegrid] (7.7444,0) -- (7.7444,8);
    \draw[sharegrid] (8.1889,0) -- (8.1889,8);
    \draw[sharegrid] (8.6333,0) -- (8.6333,8);
    \fill[sharebar] (7.3000,7.1400) rectangle ++(1.4856,0.72);
    \fill[sharebar] (7.3000,6.1400) rectangle ++(0.9038,0.72);
    \fill[sharebar] (7.3000,5.1400) rectangle ++(0.5694,0.72);
    \fill[sharebar] (7.3000,4.1400) rectangle ++(0.3697,0.72);
    \fill[sharebar] (7.3000,3.1400) rectangle ++(0.2547,0.72);
    \fill[sharebar] (7.3000,2.1400) rectangle ++(0.1851,0.72);
    \fill[sharebar] (7.3000,1.1400) rectangle ++(0.1418,0.72);
    \fill[sharebar] (7.3000,0.1400) rectangle ++(0.5341,0.72);
    \draw[insetframe] (7.3000,0) -- (7.3000,8);
    \draw[insetframe] (-0.2,-0.2) rectangle (9.1,8.2);
    \end{scope}
    \node[ticklab, anchor=south west] at (8.214,14.196) {$B{=}8$};
    \node[ticklab, anchor=west] at (1.25,15.5) {$B{=}16$};
    \draw[panelframe] (0,0) rectangle (15,16);
    \node[ticklab, anchor=north] at (0.5,-0.1) {0};
    \node[ticklab, anchor=north] at (2.5,-0.1) {2};
    \node[ticklab, anchor=north] at (4.5,-0.1) {4};
    \node[ticklab, anchor=north] at (6.5,-0.1) {6};
    \node[ticklab, anchor=north] at (8.5,-0.1) {8};
    \node[ticklab, anchor=north] at (10.5,-0.1) {10};
    \node[ticklab, anchor=north] at (12.5,-0.1) {12};
    \node[ticklab, anchor=north] at (14.5,-0.1) {14};
    \node[axlab, anchor=north] at (7.5,-1.05) {Draft position $j$};
    \node[ticklab, anchor=north] at (15.4,-0.1) {0};
    \node[ticklab, anchor=north] at (17.2333,-0.1) {30\%};
    \node[axlab, anchor=north] at (16.5000,-1.05) {Share};
    \node[ticklab, anchor=east] at (-0.15,15.5) {1};
    \node[ticklab, anchor=east] at (-0.15,14.5) {2};
    \node[ticklab, anchor=east] at (-0.15,12.5) {4};
    \node[ticklab, anchor=east] at (-0.15,10.5) {6};
    \node[ticklab, anchor=east] at (-0.15,8.5) {8};
    \node[ticklab, anchor=east] at (-0.15,6.5) {10};
    \node[ticklab, anchor=east] at (-0.15,4.5) {12};
    \node[ticklab, anchor=east] at (-0.15,2.5) {14};
    \node[ticklab, anchor=east] at (-0.15,0.5) {16};
    \node[axlab, anchor=south, rotate=90] at (-1.35,8) {Committed tokens $n_c{+}1$};
    \node[font=\small\bfseries, anchor=base] at (7.5,16.55) {Eagle-3};
    \end{scope}

    \begin{scope}[shift={(19.1,0)}]
    \fill[vir89] (0,15) rectangle ++(1,1);
    \fill[vir39] (0,14) rectangle ++(1,1);
    \fill[vir77] (1,14) rectangle ++(1,1);
    \fill[vir27] (0,13) rectangle ++(1,1);
    \fill[vir33] (1,13) rectangle ++(1,1);
    \fill[vir71] (2,13) rectangle ++(1,1);
    \fill[vir22] (0,12) rectangle ++(1,1);
    \fill[vir23] (1,12) rectangle ++(1,1);
    \fill[vir30] (2,12) rectangle ++(1,1);
    \fill[vir67] (3,12) rectangle ++(1,1);
    \fill[vir19] (0,11) rectangle ++(1,1);
    \fill[vir18] (1,11) rectangle ++(1,1);
    \fill[vir20] (2,11) rectangle ++(1,1);
    \fill[vir26] (3,11) rectangle ++(1,1);
    \fill[vir64] (4,11) rectangle ++(1,1);
    \fill[vir16] (0,10) rectangle ++(1,1);
    \fill[vir15] (1,10) rectangle ++(1,1);
    \fill[vir15] (2,10) rectangle ++(1,1);
    \fill[vir17] (3,10) rectangle ++(1,1);
    \fill[vir24] (4,10) rectangle ++(1,1);
    \fill[vir62] (5,10) rectangle ++(1,1);
    \fill[vir14] (0,9) rectangle ++(1,1);
    \fill[vir12] (1,9) rectangle ++(1,1);
    \fill[vir12] (2,9) rectangle ++(1,1);
    \fill[vir12] (3,9) rectangle ++(1,1);
    \fill[vir16] (4,9) rectangle ++(1,1);
    \fill[vir22] (5,9) rectangle ++(1,1);
    \fill[vir59] (6,9) rectangle ++(1,1);
    \fill[vir13] (0,8) rectangle ++(1,1);
    \fill[vir11] (1,8) rectangle ++(1,1);
    \fill[vir10] (2,8) rectangle ++(1,1);
    \fill[vir10] (3,8) rectangle ++(1,1);
    \fill[vir12] (4,8) rectangle ++(1,1);
    \fill[vir15] (5,8) rectangle ++(1,1);
    \fill[vir21] (6,8) rectangle ++(1,1);
    \fill[vir57] (7,8) rectangle ++(1,1);
    \fill[vir12] (0,7) rectangle ++(1,1);
    \fill[vir10] (1,7) rectangle ++(1,1);
    \fill[vir9] (2,7) rectangle ++(1,1);
    \fill[vir8] (3,7) rectangle ++(1,1);
    \fill[vir9] (4,7) rectangle ++(1,1);
    \fill[vir10] (5,7) rectangle ++(1,1);
    \fill[vir13] (6,7) rectangle ++(1,1);
    \fill[vir20] (7,7) rectangle ++(1,1);
    \fill[vir54] (8,7) rectangle ++(1,1);
    \fill[vir10] (0,6) rectangle ++(1,1);
    \fill[vir9] (1,6) rectangle ++(1,1);
    \fill[vir8] (2,6) rectangle ++(1,1);
    \fill[vir7] (3,6) rectangle ++(1,1);
    \fill[vir7] (4,6) rectangle ++(1,1);
    \fill[vir8] (5,6) rectangle ++(1,1);
    \fill[vir9] (6,6) rectangle ++(1,1);
    \fill[vir12] (7,6) rectangle ++(1,1);
    \fill[vir18] (8,6) rectangle ++(1,1);
    \fill[vir50] (9,6) rectangle ++(1,1);
    \fill[vir9] (0,5) rectangle ++(1,1);
    \fill[vir7] (1,5) rectangle ++(1,1);
    \fill[vir7] (2,5) rectangle ++(1,1);
    \fill[vir6] (3,5) rectangle ++(1,1);
    \fill[vir6] (4,5) rectangle ++(1,1);
    \fill[vir6] (5,5) rectangle ++(1,1);
    \fill[vir7] (6,5) rectangle ++(1,1);
    \fill[vir8] (7,5) rectangle ++(1,1);
    \fill[vir11] (8,5) rectangle ++(1,1);
    \fill[vir16] (9,5) rectangle ++(1,1);
    \fill[vir46] (10,5) rectangle ++(1,1);
    \fill[vir8] (0,4) rectangle ++(1,1);
    \fill[vir7] (1,4) rectangle ++(1,1);
    \fill[vir6] (2,4) rectangle ++(1,1);
    \fill[vir5] (3,4) rectangle ++(1,1);
    \fill[vir5] (4,4) rectangle ++(1,1);
    \fill[vir5] (5,4) rectangle ++(1,1);
    \fill[vir5] (6,4) rectangle ++(1,1);
    \fill[vir6] (7,4) rectangle ++(1,1);
    \fill[vir7] (8,4) rectangle ++(1,1);
    \fill[vir10] (9,4) rectangle ++(1,1);
    \fill[vir15] (10,4) rectangle ++(1,1);
    \fill[vir42] (11,4) rectangle ++(1,1);
    \fill[vir7] (0,3) rectangle ++(1,1);
    \fill[vir6] (1,3) rectangle ++(1,1);
    \fill[vir5] (2,3) rectangle ++(1,1);
    \fill[vir4] (3,3) rectangle ++(1,1);
    \fill[vir4] (4,3) rectangle ++(1,1);
    \fill[vir5] (5,3) rectangle ++(1,1);
    \fill[vir5] (6,3) rectangle ++(1,1);
    \fill[vir5] (7,3) rectangle ++(1,1);
    \fill[vir5] (8,3) rectangle ++(1,1);
    \fill[vir7] (9,3) rectangle ++(1,1);
    \fill[vir9] (10,3) rectangle ++(1,1);
    \fill[vir13] (11,3) rectangle ++(1,1);
    \fill[vir38] (12,3) rectangle ++(1,1);
    \fill[vir6] (0,2) rectangle ++(1,1);
    \fill[vir5] (1,2) rectangle ++(1,1);
    \fill[vir5] (2,2) rectangle ++(1,1);
    \fill[vir4] (3,2) rectangle ++(1,1);
    \fill[vir4] (4,2) rectangle ++(1,1);
    \fill[vir4] (5,2) rectangle ++(1,1);
    \fill[vir4] (6,2) rectangle ++(1,1);
    \fill[vir4] (7,2) rectangle ++(1,1);
    \fill[vir4] (8,2) rectangle ++(1,1);
    \fill[vir5] (9,2) rectangle ++(1,1);
    \fill[vir6] (10,2) rectangle ++(1,1);
    \fill[vir8] (11,2) rectangle ++(1,1);
    \fill[vir12] (12,2) rectangle ++(1,1);
    \fill[vir34] (13,2) rectangle ++(1,1);
    \fill[vir6] (0,1) rectangle ++(1,1);
    \fill[vir4] (1,1) rectangle ++(1,1);
    \fill[vir4] (2,1) rectangle ++(1,1);
    \fill[vir3] (3,1) rectangle ++(1,1);
    \fill[vir3] (4,1) rectangle ++(1,1);
    \fill[vir3] (5,1) rectangle ++(1,1);
    \fill[vir3] (6,1) rectangle ++(1,1);
    \fill[vir4] (7,1) rectangle ++(1,1);
    \fill[vir4] (8,1) rectangle ++(1,1);
    \fill[vir4] (9,1) rectangle ++(1,1);
    \fill[vir4] (10,1) rectangle ++(1,1);
    \fill[vir6] (11,1) rectangle ++(1,1);
    \fill[vir7] (12,1) rectangle ++(1,1);
    \fill[vir11] (13,1) rectangle ++(1,1);
    \fill[vir31] (14,1) rectangle ++(1,1);
    \fill[vir4] (0,0) rectangle ++(1,1);
    \fill[vir3] (1,0) rectangle ++(1,1);
    \fill[vir3] (2,0) rectangle ++(1,1);
    \fill[vir2] (3,0) rectangle ++(1,1);
    \fill[vir2] (4,0) rectangle ++(1,1);
    \fill[vir2] (5,0) rectangle ++(1,1);
    \fill[vir2] (6,0) rectangle ++(1,1);
    \fill[vir2] (7,0) rectangle ++(1,1);
    \fill[vir2] (8,0) rectangle ++(1,1);
    \fill[vir3] (9,0) rectangle ++(1,1);
    \fill[vir3] (10,0) rectangle ++(1,1);
    \fill[vir3] (11,0) rectangle ++(1,1);
    \fill[vir3] (12,0) rectangle ++(1,1);
    \fill[vir4] (13,0) rectangle ++(1,1);
    \fill[vir5] (14,0) rectangle ++(1,1);
    \draw[frontier] (0,15) rectangle ++(1,1);
    \draw[frontier] (1,14) rectangle ++(1,1);
    \draw[frontier] (2,13) rectangle ++(1,1);
    \draw[frontier] (3,12) rectangle ++(1,1);
    \draw[frontier] (4,11) rectangle ++(1,1);
    \draw[frontier] (5,10) rectangle ++(1,1);
    \draw[frontier] (6,9) rectangle ++(1,1);
    \draw[frontier] (7,8) rectangle ++(1,1);
    \draw[frontier] (8,7) rectangle ++(1,1);
    \draw[frontier] (9,6) rectangle ++(1,1);
    \draw[frontier] (10,5) rectangle ++(1,1);
    \draw[frontier] (11,4) rectangle ++(1,1);
    \draw[frontier] (12,3) rectangle ++(1,1);
    \draw[frontier] (13,2) rectangle ++(1,1);
    \draw[frontier] (14,1) rectangle ++(1,1);
    \draw[sharegrid] (16.0111,0) -- (16.0111,16);
    \draw[sharegrid] (16.6222,0) -- (16.6222,16);
    \draw[sharegrid] (17.2333,0) -- (17.2333,16);
    \fill[sharebar] (15.4000,15.1400) rectangle ++(1.8166,0.72);
    \fill[sharebar] (15.4000,14.1400) rectangle ++(1.3372,0.72);
    \fill[sharebar] (15.4000,13.1400) rectangle ++(0.8516,0.72);
    \fill[sharebar] (15.4000,12.1400) rectangle ++(0.5377,0.72);
    \fill[sharebar] (15.4000,11.1400) rectangle ++(0.3563,0.72);
    \fill[sharebar] (15.4000,10.1400) rectangle ++(0.2494,0.72);
    \fill[sharebar] (15.4000,9.1400) rectangle ++(0.1814,0.72);
    \fill[sharebar] (15.4000,8.1400) rectangle ++(0.1383,0.72);
    \fill[sharebar] (15.4000,7.1400) rectangle ++(0.1071,0.72);
    \fill[sharebar] (15.4000,6.1400) rectangle ++(0.0864,0.72);
    \fill[sharebar] (15.4000,5.1400) rectangle ++(0.0708,0.72);
    \fill[sharebar] (15.4000,4.1400) rectangle ++(0.0591,0.72);
    \fill[sharebar] (15.4000,3.1400) rectangle ++(0.0506,0.72);
    \fill[sharebar] (15.4000,2.1400) rectangle ++(0.0439,0.72);
    \fill[sharebar] (15.4000,1.1400) rectangle ++(0.0394,0.72);
    \fill[sharebar] (15.4000,0.1400) rectangle ++(0.1855,0.72);
    \draw[panelframe] (15.4000,0) -- (15.4000,16);
    \begin{scope}[shift={(8.35,8.5)}, xscale=0.68, yscale=0.68]
    \fill[white] (-0.2,-0.2) rectangle (9.1,8.2);
    \fill[vir91] (0,7) rectangle ++(1,1);
    \fill[vir39] (0,6) rectangle ++(1,1);
    \fill[vir79] (1,6) rectangle ++(1,1);
    \fill[vir27] (0,5) rectangle ++(1,1);
    \fill[vir33] (1,5) rectangle ++(1,1);
    \fill[vir72] (2,5) rectangle ++(1,1);
    \fill[vir22] (0,4) rectangle ++(1,1);
    \fill[vir23] (1,4) rectangle ++(1,1);
    \fill[vir29] (2,4) rectangle ++(1,1);
    \fill[vir67] (3,4) rectangle ++(1,1);
    \fill[vir18] (0,3) rectangle ++(1,1);
    \fill[vir18] (1,3) rectangle ++(1,1);
    \fill[vir20] (2,3) rectangle ++(1,1);
    \fill[vir25] (3,3) rectangle ++(1,1);
    \fill[vir63] (4,3) rectangle ++(1,1);
    \fill[vir15] (0,2) rectangle ++(1,1);
    \fill[vir14] (1,2) rectangle ++(1,1);
    \fill[vir15] (2,2) rectangle ++(1,1);
    \fill[vir17] (3,2) rectangle ++(1,1);
    \fill[vir23] (4,2) rectangle ++(1,1);
    \fill[vir59] (5,2) rectangle ++(1,1);
    \fill[vir13] (0,1) rectangle ++(1,1);
    \fill[vir12] (1,1) rectangle ++(1,1);
    \fill[vir11] (2,1) rectangle ++(1,1);
    \fill[vir12] (3,1) rectangle ++(1,1);
    \fill[vir15] (4,1) rectangle ++(1,1);
    \fill[vir20] (5,1) rectangle ++(1,1);
    \fill[vir53] (6,1) rectangle ++(1,1);
    \fill[vir8] (0,0) rectangle ++(1,1);
    \fill[vir7] (1,0) rectangle ++(1,1);
    \fill[vir6] (2,0) rectangle ++(1,1);
    \fill[vir6] (3,0) rectangle ++(1,1);
    \fill[vir6] (4,0) rectangle ++(1,1);
    \fill[vir7] (5,0) rectangle ++(1,1);
    \fill[vir9] (6,0) rectangle ++(1,1);
    \draw[frontier] (0,7) rectangle ++(1,1);
    \draw[frontier] (1,6) rectangle ++(1,1);
    \draw[frontier] (2,5) rectangle ++(1,1);
    \draw[frontier] (3,4) rectangle ++(1,1);
    \draw[frontier] (4,3) rectangle ++(1,1);
    \draw[frontier] (5,2) rectangle ++(1,1);
    \draw[frontier] (6,1) rectangle ++(1,1);
    \draw[sharegrid] (7.7444,0) -- (7.7444,8);
    \draw[sharegrid] (8.1889,0) -- (8.1889,8);
    \draw[sharegrid] (8.6333,0) -- (8.6333,8);
    \fill[sharebar] (7.3000,7.1400) rectangle ++(1.1838,0.72);
    \fill[sharebar] (7.3000,6.1400) rectangle ++(0.8985,0.72);
    \fill[sharebar] (7.3000,5.1400) rectangle ++(0.5999,0.72);
    \fill[sharebar] (7.3000,4.1400) rectangle ++(0.4021,0.72);
    \fill[sharebar] (7.3000,3.1400) rectangle ++(0.2801,0.72);
    \fill[sharebar] (7.3000,2.1400) rectangle ++(0.2096,0.72);
    \fill[sharebar] (7.3000,1.1400) rectangle ++(0.1683,0.72);
    \fill[sharebar] (7.3000,0.1400) rectangle ++(0.7021,0.72);
    \draw[insetframe] (7.3000,0) -- (7.3000,8);
    \draw[insetframe] (-0.2,-0.2) rectangle (9.1,8.2);
    \end{scope}
    \node[ticklab, anchor=south west] at (8.214,14.196) {$B{=}8$};
    \node[ticklab, anchor=west] at (1.25,15.5) {$B{=}16$};
    \draw[panelframe] (0,0) rectangle (15,16);
    \node[ticklab, anchor=north] at (0.5,-0.1) {0};
    \node[ticklab, anchor=north] at (2.5,-0.1) {2};
    \node[ticklab, anchor=north] at (4.5,-0.1) {4};
    \node[ticklab, anchor=north] at (6.5,-0.1) {6};
    \node[ticklab, anchor=north] at (8.5,-0.1) {8};
    \node[ticklab, anchor=north] at (10.5,-0.1) {10};
    \node[ticklab, anchor=north] at (12.5,-0.1) {12};
    \node[ticklab, anchor=north] at (14.5,-0.1) {14};
    \node[axlab, anchor=north] at (7.5,-1.05) {Draft position $j$};
    \node[ticklab, anchor=north] at (15.4,-0.1) {0};
    \node[ticklab, anchor=north] at (17.2333,-0.1) {30\%};
    \node[axlab, anchor=north] at (16.5000,-1.05) {Share};
    \node[font=\small\bfseries, anchor=base] at (7.5,16.55) {DFlash};
    \end{scope}

    \begin{scope}[shift={(38.2,0)}]
    \fill[vir95] (0,15) rectangle ++(1,1);
    \fill[vir36] (0,14) rectangle ++(1,1);
    \fill[vir89] (1,14) rectangle ++(1,1);
    \fill[vir30] (0,13) rectangle ++(1,1);
    \fill[vir33] (1,13) rectangle ++(1,1);
    \fill[vir85] (2,13) rectangle ++(1,1);
    \fill[vir27] (0,12) rectangle ++(1,1);
    \fill[vir26] (1,12) rectangle ++(1,1);
    \fill[vir30] (2,12) rectangle ++(1,1);
    \fill[vir81] (3,12) rectangle ++(1,1);
    \fill[vir24] (0,11) rectangle ++(1,1);
    \fill[vir23] (1,11) rectangle ++(1,1);
    \fill[vir23] (2,11) rectangle ++(1,1);
    \fill[vir26] (3,11) rectangle ++(1,1);
    \fill[vir77] (4,11) rectangle ++(1,1);
    \fill[vir21] (0,10) rectangle ++(1,1);
    \fill[vir20] (1,10) rectangle ++(1,1);
    \fill[vir20] (2,10) rectangle ++(1,1);
    \fill[vir21] (3,10) rectangle ++(1,1);
    \fill[vir25] (4,10) rectangle ++(1,1);
    \fill[vir75] (5,10) rectangle ++(1,1);
    \fill[vir20] (0,9) rectangle ++(1,1);
    \fill[vir18] (1,9) rectangle ++(1,1);
    \fill[vir17] (2,9) rectangle ++(1,1);
    \fill[vir17] (3,9) rectangle ++(1,1);
    \fill[vir19] (4,9) rectangle ++(1,1);
    \fill[vir23] (5,9) rectangle ++(1,1);
    \fill[vir72] (6,9) rectangle ++(1,1);
    \fill[vir19] (0,8) rectangle ++(1,1);
    \fill[vir16] (1,8) rectangle ++(1,1);
    \fill[vir15] (2,8) rectangle ++(1,1);
    \fill[vir14] (3,8) rectangle ++(1,1);
    \fill[vir16] (4,8) rectangle ++(1,1);
    \fill[vir18] (5,8) rectangle ++(1,1);
    \fill[vir22] (6,8) rectangle ++(1,1);
    \fill[vir69] (7,8) rectangle ++(1,1);
    \fill[vir17] (0,7) rectangle ++(1,1);
    \fill[vir15] (1,7) rectangle ++(1,1);
    \fill[vir14] (2,7) rectangle ++(1,1);
    \fill[vir12] (3,7) rectangle ++(1,1);
    \fill[vir13] (4,7) rectangle ++(1,1);
    \fill[vir14] (5,7) rectangle ++(1,1);
    \fill[vir17] (6,7) rectangle ++(1,1);
    \fill[vir20] (7,7) rectangle ++(1,1);
    \fill[vir66] (8,7) rectangle ++(1,1);
    \fill[vir16] (0,6) rectangle ++(1,1);
    \fill[vir14] (1,6) rectangle ++(1,1);
    \fill[vir12] (2,6) rectangle ++(1,1);
    \fill[vir11] (3,6) rectangle ++(1,1);
    \fill[vir11] (4,6) rectangle ++(1,1);
    \fill[vir11] (5,6) rectangle ++(1,1);
    \fill[vir13] (6,6) rectangle ++(1,1);
    \fill[vir16] (7,6) rectangle ++(1,1);
    \fill[vir19] (8,6) rectangle ++(1,1);
    \fill[vir62] (9,6) rectangle ++(1,1);
    \fill[vir15] (0,5) rectangle ++(1,1);
    \fill[vir12] (1,5) rectangle ++(1,1);
    \fill[vir11] (2,5) rectangle ++(1,1);
    \fill[vir10] (3,5) rectangle ++(1,1);
    \fill[vir10] (4,5) rectangle ++(1,1);
    \fill[vir10] (5,5) rectangle ++(1,1);
    \fill[vir10] (6,5) rectangle ++(1,1);
    \fill[vir12] (7,5) rectangle ++(1,1);
    \fill[vir14] (8,5) rectangle ++(1,1);
    \fill[vir18] (9,5) rectangle ++(1,1);
    \fill[vir59] (10,5) rectangle ++(1,1);
    \fill[vir13] (0,4) rectangle ++(1,1);
    \fill[vir11] (1,4) rectangle ++(1,1);
    \fill[vir10] (2,4) rectangle ++(1,1);
    \fill[vir9] (3,4) rectangle ++(1,1);
    \fill[vir9] (4,4) rectangle ++(1,1);
    \fill[vir9] (5,4) rectangle ++(1,1);
    \fill[vir9] (6,4) rectangle ++(1,1);
    \fill[vir9] (7,4) rectangle ++(1,1);
    \fill[vir10] (8,4) rectangle ++(1,1);
    \fill[vir13] (9,4) rectangle ++(1,1);
    \fill[vir16] (10,4) rectangle ++(1,1);
    \fill[vir55] (11,4) rectangle ++(1,1);
    \fill[vir12] (0,3) rectangle ++(1,1);
    \fill[vir10] (1,3) rectangle ++(1,1);
    \fill[vir9] (2,3) rectangle ++(1,1);
    \fill[vir8] (3,3) rectangle ++(1,1);
    \fill[vir8] (4,3) rectangle ++(1,1);
    \fill[vir8] (5,3) rectangle ++(1,1);
    \fill[vir8] (6,3) rectangle ++(1,1);
    \fill[vir7] (7,3) rectangle ++(1,1);
    \fill[vir8] (8,3) rectangle ++(1,1);
    \fill[vir9] (9,3) rectangle ++(1,1);
    \fill[vir12] (10,3) rectangle ++(1,1);
    \fill[vir15] (11,3) rectangle ++(1,1);
    \fill[vir51] (12,3) rectangle ++(1,1);
    \fill[vir11] (0,2) rectangle ++(1,1);
    \fill[vir9] (1,2) rectangle ++(1,1);
    \fill[vir8] (2,2) rectangle ++(1,1);
    \fill[vir7] (3,2) rectangle ++(1,1);
    \fill[vir7] (4,2) rectangle ++(1,1);
    \fill[vir7] (5,2) rectangle ++(1,1);
    \fill[vir7] (6,2) rectangle ++(1,1);
    \fill[vir7] (7,2) rectangle ++(1,1);
    \fill[vir7] (8,2) rectangle ++(1,1);
    \fill[vir7] (9,2) rectangle ++(1,1);
    \fill[vir9] (10,2) rectangle ++(1,1);
    \fill[vir11] (11,2) rectangle ++(1,1);
    \fill[vir14] (12,2) rectangle ++(1,1);
    \fill[vir47] (13,2) rectangle ++(1,1);
    \fill[vir10] (0,1) rectangle ++(1,1);
    \fill[vir8] (1,1) rectangle ++(1,1);
    \fill[vir7] (2,1) rectangle ++(1,1);
    \fill[vir6] (3,1) rectangle ++(1,1);
    \fill[vir6] (4,1) rectangle ++(1,1);
    \fill[vir6] (5,1) rectangle ++(1,1);
    \fill[vir6] (6,1) rectangle ++(1,1);
    \fill[vir6] (7,1) rectangle ++(1,1);
    \fill[vir6] (8,1) rectangle ++(1,1);
    \fill[vir6] (9,1) rectangle ++(1,1);
    \fill[vir7] (10,1) rectangle ++(1,1);
    \fill[vir8] (11,1) rectangle ++(1,1);
    \fill[vir10] (12,1) rectangle ++(1,1);
    \fill[vir12] (13,1) rectangle ++(1,1);
    \fill[vir43] (14,1) rectangle ++(1,1);
    \fill[vir7] (0,0) rectangle ++(1,1);
    \fill[vir6] (1,0) rectangle ++(1,1);
    \fill[vir5] (2,0) rectangle ++(1,1);
    \fill[vir4] (3,0) rectangle ++(1,1);
    \fill[vir4] (4,0) rectangle ++(1,1);
    \fill[vir4] (5,0) rectangle ++(1,1);
    \fill[vir4] (6,0) rectangle ++(1,1);
    \fill[vir4] (7,0) rectangle ++(1,1);
    \fill[vir4] (8,0) rectangle ++(1,1);
    \fill[vir4] (9,0) rectangle ++(1,1);
    \fill[vir4] (10,0) rectangle ++(1,1);
    \fill[vir5] (11,0) rectangle ++(1,1);
    \fill[vir5] (12,0) rectangle ++(1,1);
    \fill[vir6] (13,0) rectangle ++(1,1);
    \fill[vir7] (14,0) rectangle ++(1,1);
    \draw[frontier] (0,15) rectangle ++(1,1);
    \draw[frontier] (1,14) rectangle ++(1,1);
    \draw[frontier] (2,13) rectangle ++(1,1);
    \draw[frontier] (3,12) rectangle ++(1,1);
    \draw[frontier] (4,11) rectangle ++(1,1);
    \draw[frontier] (5,10) rectangle ++(1,1);
    \draw[frontier] (6,9) rectangle ++(1,1);
    \draw[frontier] (7,8) rectangle ++(1,1);
    \draw[frontier] (8,7) rectangle ++(1,1);
    \draw[frontier] (9,6) rectangle ++(1,1);
    \draw[frontier] (10,5) rectangle ++(1,1);
    \draw[frontier] (11,4) rectangle ++(1,1);
    \draw[frontier] (12,3) rectangle ++(1,1);
    \draw[frontier] (13,2) rectangle ++(1,1);
    \draw[frontier] (14,1) rectangle ++(1,1);
    \draw[sharegrid] (16.0111,0) -- (16.0111,16);
    \draw[sharegrid] (16.6222,0) -- (16.6222,16);
    \draw[sharegrid] (17.2333,0) -- (17.2333,16);
    \fill[sharebar] (15.4000,15.1400) rectangle ++(1.7479,0.72);
    \fill[sharebar] (15.4000,14.1400) rectangle ++(1.1502,0.72);
    \fill[sharebar] (15.4000,13.1400) rectangle ++(0.7689,0.72);
    \fill[sharebar] (15.4000,12.1400) rectangle ++(0.5299,0.72);
    \fill[sharebar] (15.4000,11.1400) rectangle ++(0.3811,0.72);
    \fill[sharebar] (15.4000,10.1400) rectangle ++(0.2812,0.72);
    \fill[sharebar] (15.4000,9.1400) rectangle ++(0.2158,0.72);
    \fill[sharebar] (15.4000,8.1400) rectangle ++(0.1679,0.72);
    \fill[sharebar] (15.4000,7.1400) rectangle ++(0.1329,0.72);
    \fill[sharebar] (15.4000,6.1400) rectangle ++(0.1099,0.72);
    \fill[sharebar] (15.4000,5.1400) rectangle ++(0.0900,0.72);
    \fill[sharebar] (15.4000,4.1400) rectangle ++(0.0763,0.72);
    \fill[sharebar] (15.4000,3.1400) rectangle ++(0.0636,0.72);
    \fill[sharebar] (15.4000,2.1400) rectangle ++(0.0552,0.72);
    \fill[sharebar] (15.4000,1.1400) rectangle ++(0.0503,0.72);
    \fill[sharebar] (15.4000,0.1400) rectangle ++(0.2902,0.72);
    \draw[panelframe] (15.4000,0) -- (15.4000,16);
    \begin{scope}[shift={(8.35,8.5)}, xscale=0.68, yscale=0.68]
    \fill[white] (-0.2,-0.2) rectangle (9.1,8.2);
    \fill[vir95] (0,7) rectangle ++(1,1);
    \fill[vir35] (0,6) rectangle ++(1,1);
    \fill[vir89] (1,6) rectangle ++(1,1);
    \fill[vir29] (0,5) rectangle ++(1,1);
    \fill[vir33] (1,5) rectangle ++(1,1);
    \fill[vir85] (2,5) rectangle ++(1,1);
    \fill[vir26] (0,4) rectangle ++(1,1);
    \fill[vir25] (1,4) rectangle ++(1,1);
    \fill[vir29] (2,4) rectangle ++(1,1);
    \fill[vir80] (3,4) rectangle ++(1,1);
    \fill[vir22] (0,3) rectangle ++(1,1);
    \fill[vir22] (1,3) rectangle ++(1,1);
    \fill[vir22] (2,3) rectangle ++(1,1);
    \fill[vir25] (3,3) rectangle ++(1,1);
    \fill[vir77] (4,3) rectangle ++(1,1);
    \fill[vir20] (0,2) rectangle ++(1,1);
    \fill[vir18] (1,2) rectangle ++(1,1);
    \fill[vir19] (2,2) rectangle ++(1,1);
    \fill[vir20] (3,2) rectangle ++(1,1);
    \fill[vir23] (4,2) rectangle ++(1,1);
    \fill[vir73] (5,2) rectangle ++(1,1);
    \fill[vir18] (0,1) rectangle ++(1,1);
    \fill[vir16] (1,1) rectangle ++(1,1);
    \fill[vir15] (2,1) rectangle ++(1,1);
    \fill[vir15] (3,1) rectangle ++(1,1);
    \fill[vir18] (4,1) rectangle ++(1,1);
    \fill[vir21] (5,1) rectangle ++(1,1);
    \fill[vir66] (6,1) rectangle ++(1,1);
    \fill[vir12] (0,0) rectangle ++(1,1);
    \fill[vir10] (1,0) rectangle ++(1,1);
    \fill[vir9] (2,0) rectangle ++(1,1);
    \fill[vir9] (3,0) rectangle ++(1,1);
    \fill[vir9] (4,0) rectangle ++(1,1);
    \fill[vir10] (5,0) rectangle ++(1,1);
    \fill[vir11] (6,0) rectangle ++(1,1);
    \draw[frontier] (0,7) rectangle ++(1,1);
    \draw[frontier] (1,6) rectangle ++(1,1);
    \draw[frontier] (2,5) rectangle ++(1,1);
    \draw[frontier] (3,4) rectangle ++(1,1);
    \draw[frontier] (4,3) rectangle ++(1,1);
    \draw[frontier] (5,2) rectangle ++(1,1);
    \draw[frontier] (6,1) rectangle ++(1,1);
    \draw[sharegrid] (7.7444,0) -- (7.7444,8);
    \draw[sharegrid] (8.1889,0) -- (8.1889,8);
    \draw[sharegrid] (8.6333,0) -- (8.6333,8);
    \fill[sharebar] (7.3000,7.1400) rectangle ++(1.1591,0.72);
    \fill[sharebar] (7.3000,6.1400) rectangle ++(0.7797,0.72);
    \fill[sharebar] (7.3000,5.1400) rectangle ++(0.5356,0.72);
    \fill[sharebar] (7.3000,4.1400) rectangle ++(0.3810,0.72);
    \fill[sharebar] (7.3000,3.1400) rectangle ++(0.2800,0.72);
    \fill[sharebar] (7.3000,2.1400) rectangle ++(0.2169,0.72);
    \fill[sharebar] (7.3000,1.1400) rectangle ++(0.1759,0.72);
    \fill[sharebar] (7.3000,0.1400) rectangle ++(0.9163,0.72);
    \draw[insetframe] (7.3000,0) -- (7.3000,8);
    \draw[insetframe] (-0.2,-0.2) rectangle (9.1,8.2);
    \end{scope}
    \node[ticklab, anchor=south west] at (8.214,14.196) {$B{=}8$};
    \node[ticklab, anchor=west] at (1.25,15.5) {$B{=}16$};
    \draw[panelframe] (0,0) rectangle (15,16);
    \node[ticklab, anchor=north] at (0.5,-0.1) {0};
    \node[ticklab, anchor=north] at (2.5,-0.1) {2};
    \node[ticklab, anchor=north] at (4.5,-0.1) {4};
    \node[ticklab, anchor=north] at (6.5,-0.1) {6};
    \node[ticklab, anchor=north] at (8.5,-0.1) {8};
    \node[ticklab, anchor=north] at (10.5,-0.1) {10};
    \node[ticklab, anchor=north] at (12.5,-0.1) {12};
    \node[ticklab, anchor=north] at (14.5,-0.1) {14};
    \node[axlab, anchor=north] at (7.5,-1.05) {Draft position $j$};
    \node[ticklab, anchor=north] at (15.4,-0.1) {0};
    \node[ticklab, anchor=north] at (17.2333,-0.1) {30\%};
    \node[axlab, anchor=north] at (16.5000,-1.05) {Share};
    \node[font=\small\bfseries, anchor=base] at (7.5,16.55) {DSpark};
    \end{scope}

    \begin{scope}[shift={(19.9,-4.1)}]
    \fill[vir0] (0.0000,0) rectangle ++(0.1616,0.55);
    \fill[vir1] (0.1584,0) rectangle ++(0.1616,0.55);
    \fill[vir2] (0.3168,0) rectangle ++(0.1616,0.55);
    \fill[vir3] (0.4752,0) rectangle ++(0.1616,0.55);
    \fill[vir4] (0.6337,0) rectangle ++(0.1616,0.55);
    \fill[vir5] (0.7921,0) rectangle ++(0.1616,0.55);
    \fill[vir6] (0.9505,0) rectangle ++(0.1616,0.55);
    \fill[vir7] (1.1089,0) rectangle ++(0.1616,0.55);
    \fill[vir8] (1.2673,0) rectangle ++(0.1616,0.55);
    \fill[vir9] (1.4257,0) rectangle ++(0.1616,0.55);
    \fill[vir10] (1.5842,0) rectangle ++(0.1616,0.55);
    \fill[vir11] (1.7426,0) rectangle ++(0.1616,0.55);
    \fill[vir12] (1.9010,0) rectangle ++(0.1616,0.55);
    \fill[vir13] (2.0594,0) rectangle ++(0.1616,0.55);
    \fill[vir14] (2.2178,0) rectangle ++(0.1616,0.55);
    \fill[vir15] (2.3762,0) rectangle ++(0.1616,0.55);
    \fill[vir16] (2.5347,0) rectangle ++(0.1616,0.55);
    \fill[vir17] (2.6931,0) rectangle ++(0.1616,0.55);
    \fill[vir18] (2.8515,0) rectangle ++(0.1616,0.55);
    \fill[vir19] (3.0099,0) rectangle ++(0.1616,0.55);
    \fill[vir20] (3.1683,0) rectangle ++(0.1616,0.55);
    \fill[vir21] (3.3267,0) rectangle ++(0.1616,0.55);
    \fill[vir22] (3.4851,0) rectangle ++(0.1616,0.55);
    \fill[vir23] (3.6436,0) rectangle ++(0.1616,0.55);
    \fill[vir24] (3.8020,0) rectangle ++(0.1616,0.55);
    \fill[vir25] (3.9604,0) rectangle ++(0.1616,0.55);
    \fill[vir26] (4.1188,0) rectangle ++(0.1616,0.55);
    \fill[vir27] (4.2772,0) rectangle ++(0.1616,0.55);
    \fill[vir28] (4.4356,0) rectangle ++(0.1616,0.55);
    \fill[vir29] (4.5941,0) rectangle ++(0.1616,0.55);
    \fill[vir30] (4.7525,0) rectangle ++(0.1616,0.55);
    \fill[vir31] (4.9109,0) rectangle ++(0.1616,0.55);
    \fill[vir32] (5.0693,0) rectangle ++(0.1616,0.55);
    \fill[vir33] (5.2277,0) rectangle ++(0.1616,0.55);
    \fill[vir34] (5.3861,0) rectangle ++(0.1616,0.55);
    \fill[vir35] (5.5446,0) rectangle ++(0.1616,0.55);
    \fill[vir36] (5.7030,0) rectangle ++(0.1616,0.55);
    \fill[vir37] (5.8614,0) rectangle ++(0.1616,0.55);
    \fill[vir38] (6.0198,0) rectangle ++(0.1616,0.55);
    \fill[vir39] (6.1782,0) rectangle ++(0.1616,0.55);
    \fill[vir40] (6.3366,0) rectangle ++(0.1616,0.55);
    \fill[vir41] (6.4950,0) rectangle ++(0.1616,0.55);
    \fill[vir42] (6.6535,0) rectangle ++(0.1616,0.55);
    \fill[vir43] (6.8119,0) rectangle ++(0.1616,0.55);
    \fill[vir44] (6.9703,0) rectangle ++(0.1616,0.55);
    \fill[vir45] (7.1287,0) rectangle ++(0.1616,0.55);
    \fill[vir46] (7.2871,0) rectangle ++(0.1616,0.55);
    \fill[vir47] (7.4455,0) rectangle ++(0.1616,0.55);
    \fill[vir48] (7.6040,0) rectangle ++(0.1616,0.55);
    \fill[vir49] (7.7624,0) rectangle ++(0.1616,0.55);
    \fill[vir50] (7.9208,0) rectangle ++(0.1616,0.55);
    \fill[vir51] (8.0792,0) rectangle ++(0.1616,0.55);
    \fill[vir52] (8.2376,0) rectangle ++(0.1616,0.55);
    \fill[vir53] (8.3960,0) rectangle ++(0.1616,0.55);
    \fill[vir54] (8.5545,0) rectangle ++(0.1616,0.55);
    \fill[vir55] (8.7129,0) rectangle ++(0.1616,0.55);
    \fill[vir56] (8.8713,0) rectangle ++(0.1616,0.55);
    \fill[vir57] (9.0297,0) rectangle ++(0.1616,0.55);
    \fill[vir58] (9.1881,0) rectangle ++(0.1616,0.55);
    \fill[vir59] (9.3465,0) rectangle ++(0.1616,0.55);
    \fill[vir60] (9.5050,0) rectangle ++(0.1616,0.55);
    \fill[vir61] (9.6634,0) rectangle ++(0.1616,0.55);
    \fill[vir62] (9.8218,0) rectangle ++(0.1616,0.55);
    \fill[vir63] (9.9802,0) rectangle ++(0.1616,0.55);
    \fill[vir64] (10.1386,0) rectangle ++(0.1616,0.55);
    \fill[vir65] (10.2970,0) rectangle ++(0.1616,0.55);
    \fill[vir66] (10.4554,0) rectangle ++(0.1616,0.55);
    \fill[vir67] (10.6139,0) rectangle ++(0.1616,0.55);
    \fill[vir68] (10.7723,0) rectangle ++(0.1616,0.55);
    \fill[vir69] (10.9307,0) rectangle ++(0.1616,0.55);
    \fill[vir70] (11.0891,0) rectangle ++(0.1616,0.55);
    \fill[vir71] (11.2475,0) rectangle ++(0.1616,0.55);
    \fill[vir72] (11.4059,0) rectangle ++(0.1616,0.55);
    \fill[vir73] (11.5644,0) rectangle ++(0.1616,0.55);
    \fill[vir74] (11.7228,0) rectangle ++(0.1616,0.55);
    \fill[vir75] (11.8812,0) rectangle ++(0.1616,0.55);
    \fill[vir76] (12.0396,0) rectangle ++(0.1616,0.55);
    \fill[vir77] (12.1980,0) rectangle ++(0.1616,0.55);
    \fill[vir78] (12.3564,0) rectangle ++(0.1616,0.55);
    \fill[vir79] (12.5149,0) rectangle ++(0.1616,0.55);
    \fill[vir80] (12.6733,0) rectangle ++(0.1616,0.55);
    \fill[vir81] (12.8317,0) rectangle ++(0.1616,0.55);
    \fill[vir82] (12.9901,0) rectangle ++(0.1616,0.55);
    \fill[vir83] (13.1485,0) rectangle ++(0.1616,0.55);
    \fill[vir84] (13.3069,0) rectangle ++(0.1616,0.55);
    \fill[vir85] (13.4653,0) rectangle ++(0.1616,0.55);
    \fill[vir86] (13.6238,0) rectangle ++(0.1616,0.55);
    \fill[vir87] (13.7822,0) rectangle ++(0.1616,0.55);
    \fill[vir88] (13.9406,0) rectangle ++(0.1616,0.55);
    \fill[vir89] (14.0990,0) rectangle ++(0.1616,0.55);
    \fill[vir90] (14.2574,0) rectangle ++(0.1616,0.55);
    \fill[vir91] (14.4158,0) rectangle ++(0.1616,0.55);
    \fill[vir92] (14.5743,0) rectangle ++(0.1616,0.55);
    \fill[vir93] (14.7327,0) rectangle ++(0.1616,0.55);
    \fill[vir94] (14.8911,0) rectangle ++(0.1616,0.55);
    \fill[vir95] (15.0495,0) rectangle ++(0.1616,0.55);
    \fill[vir96] (15.2079,0) rectangle ++(0.1616,0.55);
    \fill[vir97] (15.3663,0) rectangle ++(0.1616,0.55);
    \fill[vir98] (15.5248,0) rectangle ++(0.1616,0.55);
    \fill[vir99] (15.6832,0) rectangle ++(0.1616,0.55);
    \fill[vir100] (15.8416,0) rectangle ++(0.1616,0.55);
    \draw[panelframe] (0,0) rectangle (16,0.55);
    \draw[panelframe] (0,0) -- (0,-0.2);
    \node[ticklab, anchor=north] at (0,-0.22) {0.00};
    \draw[panelframe] (8,0) -- (8,-0.2);
    \node[ticklab, anchor=north] at (8,-0.22) {0.47};
    \draw[panelframe] (16,0) -- (16,-0.2);
    \node[ticklab, anchor=north] at (16,-0.22) {0.95};
    \node[axlab, anchor=east] at (-0.45,0.275) {Target entropy $H$ (nats)};
    \end{scope}
    \end{tikzpicture}
    \caption{\textbf{Accepted paths tend to have lower target entropy than their
    rejected endpoints across three drafter families.}
    Qwen3-8B at $1|0$: Eagle-3 (chain), DFlash, and DSpark.
    Cells show mean raw-$T{=}1$ target entropy at draft position $j$;
    rows show committed tokens $n_c+1$, where $n_c$ is the number of accepted drafts.
    Black borders mark rejected positions; grey bars show the share of steps
    at each length (full strip $36\%$). $B{=}16$ is the main grid and $B{=}8$
    the inset. Other targets and temperatures are in Fig.~\ref{fig:pq_entropy_heatmap}.}
    \label{fig:draftability}
\end{figure*}
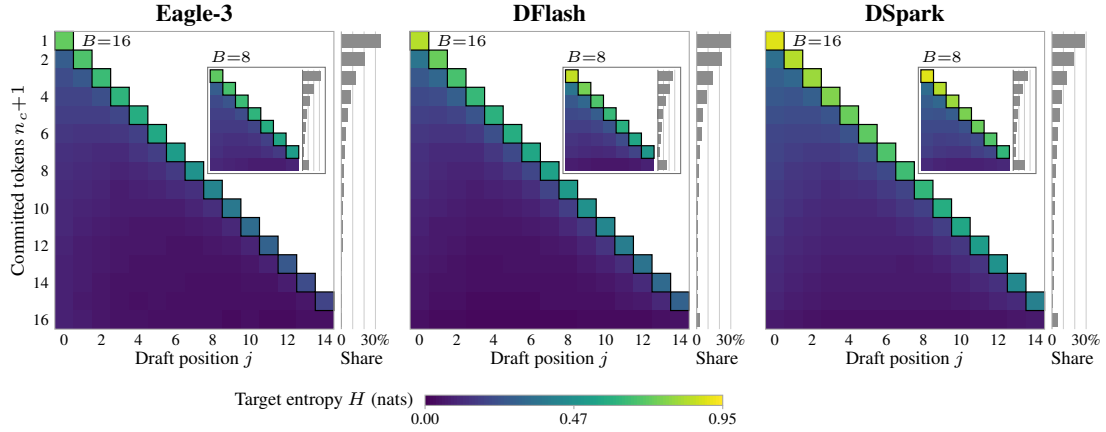

%% file: figure/Figure-budget-returns.tex
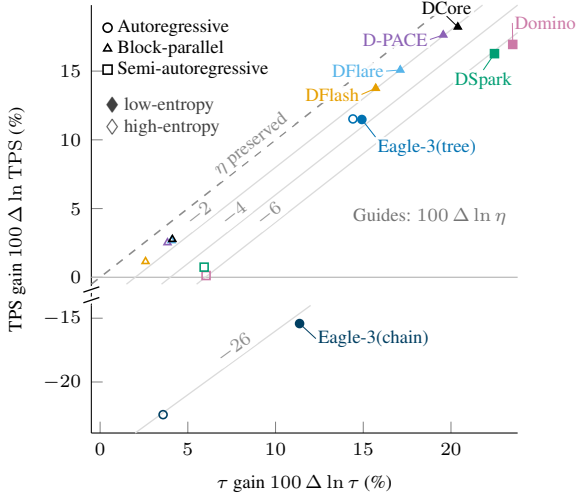
\begin{figure}[!t]
    \centering
    \colorlet{cEagleChain}{cEagle!55!black}
    \tikzset{
        guidelab/.style={font=\scriptsize, color=black!45, inner sep=1pt,
            fill=white, rotate=38.1, anchor=south},
        methodlab/.style={font=\scriptsize, inner sep=0.7pt,
            fill=white, fill opacity=0.9, text opacity=1},
    }
    \pgfplotsset{
        dot/.style={only marks, mark=*, mark size=1.5pt,
            mark options={fill=#1, draw=#1, line width=0.6pt}},
        ring/.style={only marks, mark=o, mark size=1.5pt,
            mark options={draw=#1, line width=0.6pt}},
        dotT/.style={only marks, mark=triangle*, mark size=1.5pt,
            mark options={fill=#1, draw=#1, line width=0.6pt}},
        ringT/.style={only marks, mark=triangle, mark size=1.5pt,
            mark options={draw=#1, line width=0.6pt}},
        dotS/.style={only marks, mark=square*, mark size=1.5pt,
            mark options={fill=#1, draw=#1, line width=0.6pt}},
        ringS/.style={only marks, mark=square, mark size=1.5pt,
            mark options={draw=#1, line width=0.6pt}},
        ray/.style={color=#1, line width=0.6pt, opacity=0.5},
    }
    \begin{tikzpicture}
    \begin{groupplot}[
        group style={group size=1 by 2, vertical sep=8pt},
        width=160.4pt, scale only axis,
        xmin=-0.5, xmax=23.8,
        xticklabel style={font=\scriptsize}, yticklabel style={font=\scriptsize},
        xlabel style={font=\scriptsize},
        axis y line*=left,
        clip=false,
    ]
    \nextgroupplot[
        height=105pt, ymin=-0.5, ymax=19.8,
        axis x line=none, xtick=\empty, ytick={0,5,10,15},
        legend style={at={(0.02,0.98)}, anchor=north west,
            draw=none, fill=none, font=\scriptsize,
            inner sep=1pt, row sep=-2.5pt},
        legend cell align=left,
    ]
    \addlegendimage{only marks, mark=o, mark size=1.7pt,
        mark options={draw=black, line width=0.6pt}}
    \addlegendentry{Autoregressive}
    \addlegendimage{only marks, mark=triangle, mark size=1.7pt,
        mark options={draw=black, line width=0.6pt}}
    \addlegendentry{Block-parallel}
    \addlegendimage{only marks, mark=square, mark size=1.7pt,
        mark options={draw=black, line width=0.6pt}}
    \addlegendentry{Semi-autoregressive}
    \node[font=\scriptsize, anchor=north west, inner sep=1pt, align=left,
        text=black!75] at (axis cs: 0.1, 13.3)
        {$\blacklozenge$~low-entropy\\$\lozenge$~high-entropy};
    \addplot[color=black!25, line width=0.5pt] coordinates {(-0.5,0) (23.8,0)};
    \addplot[color=black!14, line width=0.5pt] coordinates {(1.5,-0.5) (21.8,19.8)};
    \addplot[color=black!14, line width=0.5pt] coordinates {(3.5,-0.5) (23.8,19.8)};
    \addplot[color=black!14, line width=0.5pt] coordinates {(5.5,-0.5) (23.8,17.8)};
    \addplot[color=black!45, line width=0.6pt, dashed] coordinates {(-0.5,-0.5) (19.8,19.8)};
    \node[guidelab] at (axis cs: 9, 9) {$\eta$ preserved};
    \node[guidelab] at (axis cs: 6, 4) {$-2$};
    \node[guidelab] at (axis cs: 8, 4) {$-4$};
    \node[guidelab] at (axis cs: 10, 4) {$-6$};
    \node[font=\scriptsize, text=black!55, anchor=east, inner sep=0.7pt]
        at (axis cs: 23.3,4.4) {Guides: $100\,\Delta\ln\eta$};
    \addplot[ring=cEagle  ] coordinates {(14.41,11.52)};
    \addplot[ringT=cDFlash] coordinates {( 2.59, 1.15)};
    \addplot[ringT=cDFlare] coordinates {( 4.07, 2.67)};
    \addplot[ringT=cDPace ] coordinates {( 3.84, 2.52)};
    \addplot[ringS=cDomino] coordinates {( 6.04, 0.13)};
    \addplot[ringS=cDSpark] coordinates {( 5.93, 0.74)};
    \addplot[ringT=cDCore ] coordinates {( 4.12, 2.76)};
    \addplot[dot=cEagle  ] coordinates {(14.92,11.48)};
    \addplot[dotT=cDFlash] coordinates {(15.71,13.75)};
    \addplot[dotT=cDFlare] coordinates {(17.11,15.08)};
    \addplot[dotT=cDPace ] coordinates {(19.56,17.63)};
    \addplot[dotS=cDomino] coordinates {(23.52,16.94)};
    \addplot[dotS=cDSpark] coordinates {(22.48,16.27)};
    \addplot[dotT=cDCore ] coordinates {(20.39,18.21)};
    \node[methodlab, text=cEagle, anchor=north west] (labelEagleTree)
        at (axis cs: 15.7,10.2) {Eagle-3(tree)};
    \node[methodlab, text=cDFlash, anchor=east] (labelDFlash)
        at (axis cs: 14.9,13.3) {DFlash};
    \node[methodlab, text=cDFlare, anchor=east] (labelDFlare)
        at (axis cs: 16.3,15.0) {DFlare};
    \node[methodlab, text=cDPace, anchor=east] (labelDPace)
        at (axis cs: 18.8,17.3) {D-PACE};
    \node[methodlab, text=cDCore, anchor=south east] (labelDCore)
        at (axis cs: 21.3,19.0) {DCore};
    \node[methodlab, text=cDSpark, anchor=east] (labelDSpark)
        at (axis cs: 23.6,14.3) {DSpark};
    \node[methodlab, text=cDomino, anchor=west] (labelDomino)
        at (axis cs: 23.5,18.5) {Domino};
    \draw[cEagle, line width=0.35pt] (axis cs: 14.92,11.48) -- (labelEagleTree.north west);
    \draw[cDFlash, line width=0.35pt] (axis cs: 15.71,13.75) -- (labelDFlash.east);
    \draw[cDFlare, line width=0.35pt] (axis cs: 17.11,15.08) -- (labelDFlare.east);
    \draw[cDPace, line width=0.35pt] (axis cs: 19.56,17.63) -- (labelDPace.east);
    \draw[cDCore, line width=0.35pt] (axis cs: 20.39,18.21) -- (labelDCore.south);
    \draw[cDSpark, line width=0.35pt] (axis cs: 22.48,16.27) -- (labelDSpark.north);
    \draw[cDomino, line width=0.35pt] (axis cs: 23.52,16.94) -- (labelDomino.west);
    \nextgroupplot[
        height=48pt, ymin=-24, ymax=-14,
        xtick={0,5,10,15,20}, ytick={-20,-15},
        xlabel={$\tau$ gain $100\,\Delta\ln\tau$ (\%)},
        axis x line*=bottom,
    ]
    \addplot[color=black!14, line width=0.5pt] coordinates {(2,-24) (12,-14)};
    \node[font=\scriptsize, color=black!45, inner sep=1pt, fill=white, rotate=36, anchor=south] at (axis cs: 8, -18) {$-26$};
    \addplot[ring=cEagleChain] coordinates {(3.59,-22.56)};
    \addplot[dot=cEagleChain] coordinates {(11.37,-15.42)};
    \node[methodlab, text=cEagleChain, anchor=west] (labelEagleChain)
        at (axis cs: 12.3,-16.6) {Eagle-3(chain)};
    \draw[cEagleChain, line width=0.35pt]
        (axis cs: 11.37,-15.42) -- (labelEagleChain.west);
    \end{groupplot}
    \draw[line width=0.4pt] (group c1r1.south west) -- (group c1r2.north west);
    \draw[white, line width=2.5pt]
        ([xshift=-3pt,yshift=-3pt]group c1r1.south west) --
        ([xshift=3pt,yshift=-1pt]group c1r1.south west);
    \draw[line width=0.5pt]
        ([xshift=-3pt,yshift=-3pt]group c1r1.south west) --
        ([xshift=3pt,yshift=-1pt]group c1r1.south west);
    \draw[white, line width=2.5pt]
        ([xshift=-3pt,yshift=-7pt]group c1r1.south west) --
        ([xshift=3pt,yshift=-5pt]group c1r1.south west);
    \draw[line width=0.5pt]
        ([xshift=-3pt,yshift=-7pt]group c1r1.south west) --
        ([xshift=3pt,yshift=-5pt]group c1r1.south west);
    \path (group c1r1.north west) -- (group c1r2.south west)
        node[midway, xshift=-28pt, rotate=90, font=\scriptsize]
        {TPS gain $100\,\Delta\ln\mathrm{TPS}$ (\%)};
    \end{tikzpicture}
    \caption{\textbf{Budget elasticity: what a larger budget buys, and what it
    costs} (Qwen3-8B, $1|0$). We use log changes to express throughput gains
    as the sum of acceptance gains and efficiency changes, revealing whether
    additional acceptance repays its execution cost.
    Axes: log change in acceptance length
    $\tau$ and throughput from $B{=}8$ to $16$. Eagle-3(tree) compares
    $7/4/8\rightarrow7/10/16$; Eagle-3(chain) compares $7/1/8\rightarrow15/1/16$.
    Parallel guide labels give $100\,\Delta\ln\eta$ via Eq.~\eqref{eq:eta},
    not ordinary percentage changes. Negative TPS changes indicate throughput loss.}
    \label{fig:elasticity_budget_entropy_qwen3_8b}
\end{figure}

%% file: figure/Figure-concurrency-8b-b8.tex
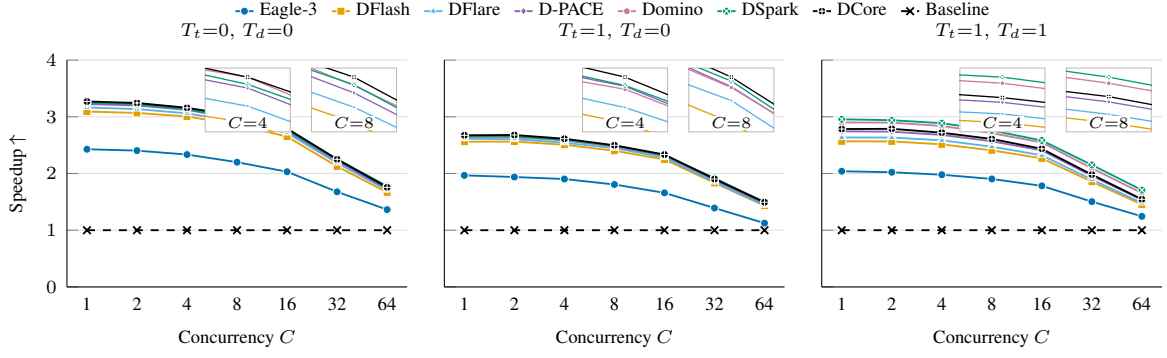
\begin{figure*}[t]
    \centering
    \def\panelw{128pt}
    \pgfplotsset{
        spline/.style={
            color=#1, line width=0.7pt, mark size=1.5pt,
            mark options={fill=#1, draw=white, line width=0.3pt},
        },
        tpspanel/.style={
            width=\panelw, height=3.0cm, scale only axis,
            xmin=0.6, xmax=7.4,
            xtick={1,2,3,4,5,6,7},
            xticklabels={1,2,4,8,16,32,64},
            ymin=0, ymax=4,
            ytick={0,1,2,3,4},
            yticklabels={$0$,$1$,$2$,$3$,$4$},
            scaled y ticks=false,
            ymajorgrids=true, grid style={gray!25},
            xtick style={draw=none},
            axis x line*=bottom, axis y line*=left,
            xlabel={Concurrency $C$},
            xticklabel style={font=\scriptsize},
            yticklabel style={font=\scriptsize},
            xlabel style={font=\scriptsize},
            ylabel style={font=\scriptsize},
            title style={font=\scriptsize, yshift=-2pt},
        },
        zoombox/.style={
            width=32pt, height=24pt, scale only axis,
            enlargelimits=false, clip=true,
            xtick=\empty, ytick=\empty,
            axis line style={gray!55, line width=0.3pt},
            axis background/.style={fill=white},
            xlabel style={font=\tiny, at={(axis description cs:0.5,0.025)},
                          anchor=south, fill=white, inner sep=0.5pt},
            spline/.style={
                color=##1, line width=0.5pt, mark size=1.0pt,
                mark options={fill=##1, draw=white, line width=0.2pt},
            },
        },
    }
    \def\sixa{%
        \addplot[spline=cDFlash, mark=square*   ] coordinates {(1,3.092105) (2,3.067797) (3,3.005217) (4,2.859982) (5,2.650387) (6,2.125445) (7,1.667499)};
        \addplot[spline=cDFlare, mark=triangle* ] coordinates {(1,3.164474) (2,3.135593) (3,3.060870) (4,2.913279) (5,2.715184) (6,2.185155) (7,1.707959)};
        \addplot[spline=cDPace,  mark=diamond*  ] coordinates {(1,3.223684) (2,3.200000) (3,3.121739) (4,2.962963) (5,2.749033) (6,2.222679) (7,1.732268)};
        \addplot[spline=cDomino, mark=pentagon* ] coordinates {(1,3.256579) (2,3.227119) (3,3.158261) (4,2.988257) (5,2.788685) (6,2.253355) (7,1.768731)};
        \addplot[spline=cDSpark, mark=otimes*   ] coordinates {(1,3.250000) (2,3.223729) (3,3.133913) (4,2.990967) (5,2.765957) (6,2.250616) (7,1.778555)};
        \addplot[spline=cDCore,  mark=oplus*    ] coordinates {(1,3.269737) (2,3.244068) (3,3.158261) (4,3.017164) (5,2.792070) (6,2.250616) (7,1.755744)};
    }
    \def\sixb{%
        \addplot[spline=cDFlash, mark=square*   ] coordinates {(1,2.560000) (2,2.562069) (3,2.507067) (4,2.400735) (5,2.247291) (6,1.833937) (7,1.432682)};
        \addplot[spline=cDFlare, mark=triangle* ] coordinates {(1,2.613333) (2,2.603448) (3,2.542403) (4,2.447610) (5,2.280788) (6,1.850098) (7,1.441761)};
        \addplot[spline=cDPace,  mark=diamond*  ] coordinates {(1,2.646667) (2,2.648276) (3,2.591873) (4,2.477022) (5,2.314778) (6,1.884648) (7,1.471223)};
        \addplot[spline=cDomino, mark=pentagon* ] coordinates {(1,2.646667) (2,2.634483) (3,2.584806) (4,2.475184) (5,2.318719) (6,1.904987) (7,1.500685)};
        \addplot[spline=cDSpark, mark=otimes*   ] coordinates {(1,2.660000) (2,2.655172) (3,2.593640) (4,2.488971) (5,2.318227) (6,1.919755) (7,1.487324)};
        \addplot[spline=cDCore,  mark=oplus*    ] coordinates {(1,2.673333) (2,2.679310) (3,2.613074) (4,2.497243) (5,2.333990) (6,1.903594) (7,1.493320)};
    }
    \def\sixc{%
        \addplot[spline=cDFlash, mark=square*   ] coordinates {(1,2.566606) (2,2.565911) (3,2.513740) (4,2.407918) (5,2.267362) (6,1.854632) (7,1.452865)};
        \addplot[spline=cDFlare, mark=triangle* ] coordinates {(1,2.635834) (2,2.633570) (3,2.585501) (4,2.472132) (5,2.321274) (6,1.891454) (7,1.480854)};
        \addplot[spline=cDPace,  mark=diamond*  ] coordinates {(1,2.743425) (2,2.737427) (3,2.680773) (4,2.567938) (5,2.407649) (6,1.958402) (7,1.535534)};
        \addplot[spline=cDomino, mark=pentagon* ] coordinates {(1,2.900670) (2,2.896664) (3,2.837875) (4,2.711322) (5,2.546688) (6,2.092169) (7,1.653725)};
        \addplot[spline=cDSpark, mark=otimes*   ] coordinates {(1,2.955873) (2,2.939388) (3,2.887431) (4,2.758074) (5,2.582901) (6,2.150884) (7,1.708007)};
        \addplot[spline=cDCore,  mark=oplus*    ] coordinates {(1,2.783793) (2,2.788019) (3,2.719414) (4,2.608385) (5,2.435138) (6,1.980793) (7,1.545672)};
    }
    \def\tobase#1{%
        \addplot[black, dashed, line width=0.7pt, mark=x, mark size=2.0pt,
                 mark options={solid}] coordinates {#1};
    }
    \def\bZ{(1,1.000000) (2,1.000000) (3,1.000000) (4,1.000000) (5,1.000000) (6,1.000000) (7,1.000000)}
    \def\bS{(1,1.000000) (2,1.000000) (3,1.000000) (4,1.000000) (5,1.000000) (6,1.000000) (7,1.000000)}
    \begin{tikzpicture}
    \begin{groupplot}[
        tpspanel,
        group style={
            group size=3 by 1,
            group name=tq,
            horizontal sep=14pt,
            ylabels at=edge left, yticklabels at=edge left,
        },
    ]
    \nextgroupplot[
        title={$T_t{=}0,\,T_d{=}0$},
        ylabel={Speedup\,$\uparrow$},
        legend entries={Eagle-3, DFlash, DFlare, D-PACE, Domino, DSpark,
                        DCore, Baseline},
        legend style={
            at={(1.61,1)}, anchor=south, yshift=11pt,
            draw=none, fill=none, legend columns=8, font=\scriptsize,
            /tikz/every even column/.append style={column sep=3pt},
        },
        legend image code/.code={
            \draw[mark repeat=2, mark phase=2, #1]
                plot coordinates {(0cm,0cm) (0.12cm,0cm) (0.24cm,0cm)};
        },
    ]
    \addplot[spline=cEagle, mark=*] coordinates {(1,2.427632) (2,2.403390) (3,2.333913) (4,2.199639) (5,2.030948) (6,1.677349) (7,1.362304)};
    \sixa
    \tobase{\bZ}
    \nextgroupplot[title={$T_t{=}1,\,T_d{=}0$}]
    \addplot[spline=cEagle, mark=*] coordinates {(1,1.966667) (2,1.937931) (3,1.902827) (4,1.806066) (5,1.658621) (6,1.391195) (7,1.124700)};
    \sixb
    \tobase{\bS}
    \nextgroupplot[title={$T_t{=}1,\,T_d{=}1$}]
    \addplot[spline=cEagle, mark=*] coordinates {(1,2.039928) (2,2.022204) (3,1.977984) (4,1.903790) (5,1.781782) (6,1.503337) (7,1.244524)};
    \sixc
    \tobase{\bS}
    \end{groupplot}
    \def\zm#1#2#3#4#5#6#7#8#9{%
        \begin{axis}[zoombox, at={(#1.north east)}, anchor=north east,
                     xshift=#2, yshift=-3pt,
                     xmin=#3, xmax=#4, ymin=#5, ymax=#6,
                     xlabel={$C{=}#7$}]
        #9
        \end{axis}%
    }
    \zm{tq c1r1}{-44pt}{2.650}{3.350}{2.974609}{3.188870}{4}{0.21}{\sixa}
    \zm{tq c1r1}{-4pt}{3.650}{4.350}{2.828546}{3.048600}{8}{0.22}{\sixa}
    \zm{tq c2r1}{-44pt}{2.650}{3.350}{2.485866}{2.634276}{4}{0.15}{\sixb}
    \zm{tq c2r1}{-4pt}{3.650}{4.350}{2.381434}{2.516544}{8}{0.14}{\sixb}
    \zm{tq c3r1}{-44pt}{2.650}{3.350}{2.439002}{2.962169}{4}{0.52}{\sixc}
    \zm{tq c3r1}{-4pt}{3.650}{4.350}{2.337886}{2.828105}{8}{0.49}{\sixc}
    \end{tikzpicture}
    \caption{\textbf{Speedup under concurrency (Qwen3-8B, budget $8$).}
    Request output throughput relative to the target-only baseline at the
    same temperature and concurrency; dashed is $1\times$. $C$ is per replica
    and doubles at each tick. Eagle-3 is chain-decoded.
    Insets use local $y$ ranges to resolve method ordering at $C=4$ and $8$. Fig.~\ref{fig:serving_tps_concurrency} adds
    the other target and budget; App.~\ref{sec:appendix-concurrency}
    defines the metric and documents repeats.}
    \label{fig:serving_tps_qwen3_8b_bgt8}
\end{figure*}

%% file: figure/Figure-sampling-gap.tex
\begin{figure*}[t]
    \centering
    \pgfplotsset{
        sgpanel/.style={
            width=176pt, height=2.9cm, scale only axis,
            xmin=0.3, xmax=13.7,
            xtick={1,2,3,4,5,6,8,9,10,11,12,13},
            xticklabels={DFlash,DFlare,D-PACE,DCore,Domino,DSpark,
                         DFlash,DFlare,D-PACE,DCore,Domino,DSpark},
            xticklabel style={font=\tiny, rotate=55, anchor=north east,
                              inner sep=1pt},
            xtick style={draw=none},
            ymajorgrids=true, grid style={gray!25},
            axis x line*=bottom, axis y line*=left,
            yticklabel style={font=\scriptsize},
            ylabel style={font=\scriptsize},
            title style={font=\scriptsize, yshift=-2pt},
            legend style={draw=none, fill=none, font=\scriptsize,
                          inner sep=1pt},
            legend cell align=left,
        },
    }
    \tikzset{dtaumark/.style={black, line width=0.9pt}}
    \def\dumbbell#1#2#3#4{%
        \draw[#2, line width=0.8pt] (#1,#4) -- (#1,#3);
        \node[circle, fill=#2, draw=white, line width=0.3pt, inner sep=0pt,
              minimum size=4.4pt] at (#1,#3) {};
        \node[circle, fill=white, draw=#2, line width=0.8pt, inner sep=0pt,
              minimum size=3.8pt] at (#1,#4) {};
    }
    \def\budgetlabels#1#2{%
        \draw[gray!60, dashed, line width=0.4pt] (7,#1) -- (7,#2);
        \node[font=\scriptsize, anchor=north] at (3.5,#2) {$B{=}8$};
        \node[font=\scriptsize, anchor=north] at (10.5,#2) {$B{=}16$};
    }
    \begin{tikzpicture}
    \begin{groupplot}[
        sgpanel,
        group style={group size=2 by 1, group name=sg,
                     horizontal sep=40pt},
    ]
    \nextgroupplot[
        title={(a) $\Delta\tau$ from $1|0$ to $1|1$, split by position},
        ylabel={Committed tokens},
        ymin=-0.45, ymax=0.75,
        ytick={-0.4,-0.2,0,0.2,0.4,0.6},
        yticklabels={$-0.4$,$-0.2$,$0$,$0.2$,$0.4$,$0.6$},
        legend style={at={(0.5,-0.36)}, anchor=north, legend columns=3,
            /tikz/every even column/.append style={column sep=6pt}},
    ]
    \addlegendimage{area legend, fill=black!30, draw=none}
    \addlegendentry{$j{=}0$ term}
    \addlegendimage{area legend, fill=black!85, draw=none}
    \addlegendentry{$j{\ge}1$ term}
    \addlegendimage{dtaumark}
    \addlegendentry{$\Delta\tau$}
    \draw[black!60, line width=0.4pt] (0.3,0) -- (13.7,0);
    \fill[cDFlash!35] (0.64,0) rectangle (0.98,0.2455); \fill[cDFlash] (1.02,0) rectangle (1.36,-0.2195); \draw[dtaumark] (0.56,0.0260) -- (1.44,0.0260); %
    \fill[cDFlare!35] (1.64,0) rectangle (1.98,0.2714); \fill[cDFlare] (2.02,0) rectangle (2.36,-0.1954); \draw[dtaumark] (1.56,0.0760) -- (2.44,0.0760); %
    \fill[cDPace!35] (2.64,0) rectangle (2.98,0.2481); \fill[cDPace] (3.02,0) rectangle (3.36,-0.0781); \draw[dtaumark] (2.56,0.1700) -- (3.44,0.1700); %
    \fill[cDCore!35] (3.64,0) rectangle (3.98,0.2566); \fill[cDCore] (4.02,0) rectangle (4.36,-0.0578); \draw[dtaumark] (3.56,0.1988) -- (4.44,0.1988); %
    \fill[cDomino!35] (4.64,0) rectangle (4.98,0.2946); \fill[cDomino] (5.02,0) rectangle (5.36,0.1409); \draw[dtaumark] (4.56,0.4355) -- (5.44,0.4355); %
    \fill[cDSpark!35] (5.64,0) rectangle (5.98,0.3318); \fill[cDSpark] (6.02,0) rectangle (6.36,0.2023); \draw[dtaumark] (5.56,0.5341) -- (6.44,0.5341); %
    \fill[cDFlash!35] (7.64,0) rectangle (7.98,0.2439); \fill[cDFlash] (8.02,0) rectangle (8.36,-0.4146); \draw[dtaumark] (7.56,-0.1706) -- (8.44,-0.1706); %
    \fill[cDFlare!35] (8.64,0) rectangle (8.98,0.3123); \fill[cDFlare] (9.02,0) rectangle (9.36,-0.3265); \draw[dtaumark] (8.56,-0.0141) -- (9.44,-0.0141); %
    \fill[cDPace!35] (9.64,0) rectangle (9.98,0.2868); \fill[cDPace] (10.02,0) rectangle (10.36,-0.1244); \draw[dtaumark] (9.56,0.1624) -- (10.44,0.1624); %
    \fill[cDCore!35] (10.64,0) rectangle (10.98,0.2992); \fill[cDCore] (11.02,0) rectangle (11.36,-0.1387); \draw[dtaumark] (10.56,0.1605) -- (11.44,0.1605); %
    \fill[cDomino!35] (11.64,0) rectangle (11.98,0.3559); \fill[cDomino] (12.02,0) rectangle (12.36,0.1703); \draw[dtaumark] (11.56,0.5262) -- (12.44,0.5262); %
    \fill[cDSpark!35] (12.64,0) rectangle (12.98,0.4070); \fill[cDSpark] (13.02,0) rectangle (13.36,0.2499); \draw[dtaumark] (12.56,0.6569) -- (13.44,0.6569); %
    \budgetlabels{-0.45}{0.75}
    \nextgroupplot[
        title={(b) $1|1$ acceptance at $j{\ge}1$ by previous token},
        ylabel={Acceptance},
        ymin=0.25, ymax=0.95,
        ytick={0.3,0.4,0.5,0.6,0.7,0.8,0.9},
        yticklabels={$0.3$,$0.4$,$0.5$,$0.6$,$0.7$,$0.8$,$0.9$},
        legend style={at={(0.5,-0.36)}, anchor=north, legend columns=2,
            /tikz/every even column/.append style={column sep=6pt}},
    ]
    \addlegendimage{only marks, mark=*, mark size=2.0pt,
                    mark options={fill=black!70, draw=black!70}}
    \addlegendentry{argmax predecessor}
    \addlegendimage{only marks, mark=o, mark size=1.7pt,
                    mark options={draw=black!70, line width=0.8pt}}
    \addlegendentry{non-argmax predecessor}
    \dumbbell{1}{cDFlash}{0.7578}{0.3332} %
    \dumbbell{2}{cDFlare}{0.7676}{0.3373} %
    \dumbbell{3}{cDPace}{0.7957}{0.3959} %
    \dumbbell{4}{cDCore}{0.8010}{0.3964} %
    \dumbbell{5}{cDomino}{0.8266}{0.7393} %
    \dumbbell{6}{cDSpark}{0.8341}{0.7550} %
    \dumbbell{8}{cDFlash}{0.7423}{0.3240} %
    \dumbbell{9}{cDFlare}{0.7617}{0.3434} %
    \dumbbell{10}{cDPace}{0.7922}{0.4112} %
    \dumbbell{11}{cDCore}{0.7954}{0.4131} %
    \dumbbell{12}{cDomino}{0.8215}{0.7400} %
    \dumbbell{13}{cDSpark}{0.8241}{0.7480} %
    \budgetlabels{0.25}{0.95}
    \end{groupplot}
    \end{tikzpicture}
    \caption{\textbf{Where sampled drafting gains or loses acceptance
    (Qwen3-8B).}
    (a)~Two-factor Shapley split of the change in mean committed tokens
    $\Delta\tau$ from $1|0$ to $1|1$ into the first draft position ($j{=}0$,
    light) and all later positions ($j{\ge}1$, solid); the bar pair sums to
    $\Delta\tau$ (black line).
    (b)~Acceptance at $j{\ge}1$ under $1|1$ after an argmax predecessor
    (filled: the previous accepted token is the target's argmax) or a
    non-argmax predecessor (open: it is not), standardized to a common
    target-entropy mix.
    Block-parallel recipes (first four of each group) lose at $j{\ge}1$ and
    lose far more after non-argmax predecessors than semi-AR recipes.
    App.~\ref{sec:appendix-sampled-drafts} gives the estimators, intervals,
    and per-position curves.}
    \label{fig:sampling_gap}
\end{figure*}
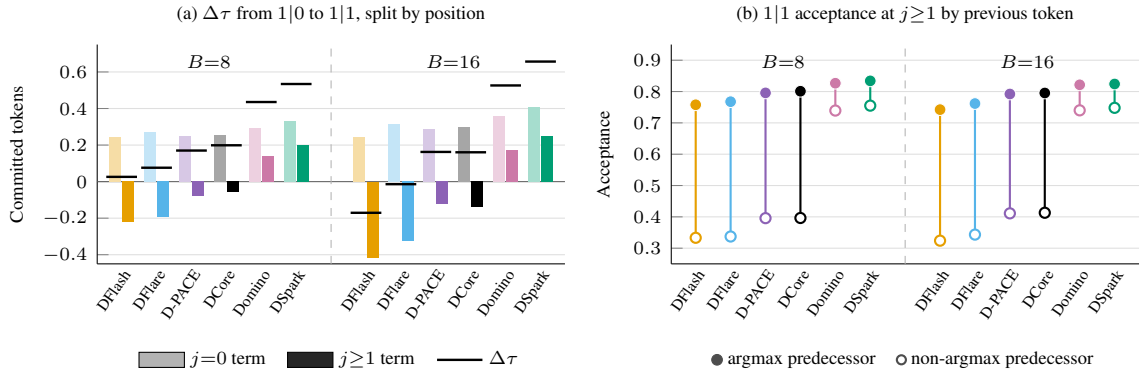

%% file: section/conclusion.tex
\section{Conclusion}
\label{sec:conclusion}

Published drafter comparisons often combine differences in training
conditions, proposal design, and inference protocols, making reported
rankings difficult to interpret as evidence of design advantages. We
establish a common reference across autoregressive, block-parallel, and
semi-autoregressive drafting through a protocol spanning training,
inference, and evaluation. Remeasuring available checkpoints and retraining
six recipes on shared target-generated data substantially narrows the
observed gaps. The remaining differences can then be assessed through
their acceptance gains and execution costs.

Against this reference, component ablations identify two routes to better
acceptance: improvements to supervision and context with little additional
measured drafting overhead, exemplified by DCore, and more expensive
conditioning whose benefits must repay its execution cost, as
prefix-conditioned proposals do under sampled drafting. Task uncertainty
and proposal structure shape acceptance and budget returns, while serving
load and target scale change relative costs and preferred configurations.
Together, these findings connect controlled design comparisons to practical
deployment decisions: prioritize low-overhead improvements, and select the
drafter and budget jointly by end-to-end throughput under the intended
workload and serving conditions. The resulting contribution is an empirical
basis for judging both which advantages remain under common conditions and
when those advantages become useful throughput gains.

%% file: section/limitations.tex
\section*{Limitations}

\paragraph{Coverage and recipe-level comparison.}
The study covers two target sizes within one model family, chosen because
most evaluated methods report results on both, enabling comparison with
the original papers. To strengthen the comparison within this scope, we
evaluate six drafter recipes across
three families and ten benchmarks, remeasure released checkpoints, and
retrain on shared target-generated data. We preserve defining recipe
differences, so the findings concern complete recipes rather than isolated
architectures. Matched profiles of both target sizes
(Tab.~\ref{tab:serving-stage-latency}) suggest that costlier drafters may
become more competitive as their latency premium shrinks relative to target
execution, provided their acceptance advantage persists. Experiments on
larger targets and other model families remain necessary to assess this
transfer and are left to future work.

\paragraph{Serving portability.}
Throughput depends on hardware, kernels, and workload. We use a common
engine, optimize draft paths, and compare speedups against matched
target-only baselines across budgets and concurrency. End-to-end throughput
establishes performance differences; separate stage profiles diagnose costs
without substituting for complete step measurements. Implementation
boundaries, reference baselines, and evaluation repeats are
documented in Apps.~\ref{app:serving-impl},
\ref{sec:appendix-per-benchmark}, and~\ref{sec:appendix-concurrency}.

%% file: section/acknowledgments.tex
\section*{Acknowledgments}

Our implementation is built on
\href{https://github.com/sgl-project/SpecForge}{SpecForge} and
\href{https://github.com/sgl-project/sglang}{SGLang}. We thank their authors
and maintainers for the training and serving foundations of this work.

We also thank the authors and maintainers of
\href{https://github.com/SafeAILab/EAGLE}{EAGLE} (EAGLE-3),
\href{https://github.com/z-lab/dflash}{DFlash},
\href{https://github.com/Lucas-TY/D-PACE}{D-PACE},
\href{https://github.com/Tencent/AngelSlim}{AngelSlim} (DFlare),
\href{https://github.com/jianuo-huang/Domino}{Domino}, and
\href{https://github.com/deepseek-ai/DeepSpec}{DeepSpec} (DSpark), whose
open-source implementations informed our training and evaluation code.

%% file: section/Appendix-training.tex
\section{Training Details}
\label{sec:appendix-training}

\paragraph{Optimization.}
Tab.~\ref{tab:training_hparams} lists the shared optimization settings
and Eagle-3's family-specific schedule.

\begin{table}[H]
\centering
\footnotesize
\setlength{\tabcolsep}{4pt}
\begin{tabular}{@{}lcc@{}}
\toprule
 & BP / Semi-AR & Eagle-3 \\
\midrule
Epochs                & $3$    & $3$ \\
Learning rate         & $6\mathrm{e}{-4}$ & $1\mathrm{e}{-4}$ \\
Warmup ratio          & $0.04$ & $0.015$ \\
Gradient clip         & $1.0$  & $0.5$ \\
Optimiser             & \multicolumn{2}{c}{AdamW $(0.9,\,0.999)$} \\
Weight decay          & \multicolumn{2}{c}{$0$} \\
LR schedule           & \multicolumn{2}{c}{cosine $\to 0$} \\
Precision             & \multicolumn{2}{c}{bf16, FSDP} \\
\bottomrule
\end{tabular}
\caption{Training settings. Common settings such as epochs, optimizer, and
precision are unified across the two columns. The left column
is shared by DFlash, DFlare, D-PACE, Domino, DSpark, and DCore; Eagle-3 keeps
its family-specific defaults.}
\label{tab:training_hparams}
\end{table}

\paragraph{Evaluated recipes.}
Tab.~\ref{tab:drafter-summary} summarizes the recipes used in our comparison;
the original papers provide their full architectural and objective definitions
\citep{NEURIPS2025_c7b5a35e,chen2026dflash,zhang2026dflare,
wu2026dpacedynamicpositionawarecrossentropy,
huang2026dominodecouplingcausalmodeling,cheng2026dspark}.
DFlash, D-PACE, Domino, and DSpark share a projection over five target layers;
DFlare uses draft-layer-specific projections over nine. Eagle-3 fuses three
target layers and uses a reduced $32$k draft vocabulary.
App.~\ref{sec:dcore} specifies DCore's combined recipe.
Parallel and semi-AR checkpoints are trained separately for each budget.

\input{table/Table-drafter-recipes.tex}

\paragraph{Objective settings.}
Eagle-3 uses target-distribution KL without feature regression.
DFlash and the evaluated DFlare use fixed exponential position decay with
$\gamma=7$ for a block of $16$; D-PACE retains its published position weights.
Domino retains its base/final cross-entropy curriculum, with the base-loss
coefficient decreasing linearly from one to zero. DSpark keeps decayed
cross-entropy, L1 distribution matching, and confidence BCE with coefficients
$(0.1,0.9,1.0)$ and a detached soft confidence target, following
\citet{cheng2026dspark}. Its confidence head remains part of training, while
its serving policy is excluded as described in App.~\ref{app:serving-impl}.
The separate loss-ablation sweep is specified in
App.~\ref{sec:appendix-loss-ablation}.

\input{table/Table-drafter-parameters.tex}

\paragraph{Response regeneration and prompts.}
Each target regenerates responses to the Open-PerfectBlend prompts with
thinking disabled, temperature $0$, a $4096$-token generation cap, and seed
$42$, yielding one shared training corpus per target.

\paragraph{Training-seed robustness.}
We retrain DCore on Qwen3-4B at budget $8$ with three seeds, each controlling
both data shuffling and initialization. Across seeds, entropy-half mean $\tau$
varies by at most $0.03$ under every temperature pair
(Tab.~\ref{tab:seed_consistency}), so the recipe comparison is not driven by
a favorable seed. We deliberately reduce these runs to $256$ block anchors per
sequence, half of the $512$ used for the Tab.~\ref{tab:main_results_temps}
checkpoint; they remain within $0.1$ of it, so the anchor count also has a
limited effect.

\input{table/Table-seed-consistency.tex}

%% file: table/Table-drafter-recipes.tex
\begin{table*}[t]
\centering
\small
\setlength{\tabcolsep}{5pt}
\caption{The evaluated drafter recipes viewed through the common framework. We use AR for autoregressive, Block Parallel for block-parallel, and Semi-AR for semi-autoregressive.}
\label{tab:drafter-summary}
\begin{tabular}{@{}p{0.14\textwidth}p{0.16\textwidth}p{0.34\textwidth}p{0.25\textwidth}@{}}
\toprule
\textbf{Drafter} & \textbf{Family} & \textbf{Architectural distinction} & \textbf{Training target} \\
\midrule
Eagle-3 & AR & Token-conditioned decoder with target-layer fusion and a reduced draft vocabulary & KL to target next-token distributions \\
DFlash & Block Parallel & Shared projection over selected target layers; masked block input & Decayed cross-entropy \\
DFlare & Block Parallel & Per-draft-layer target-state mixing & Decayed cross-entropy \\
D-PACE & Block Parallel & Parallel backbone with acceptance-oriented position weighting & D-PACE weighted cross-entropy \\
Domino & Semi-AR & Parallel backbone plus a correction head and curriculum & Base/final cross-entropy curriculum \\
DSpark & Semi-AR & Parallel backbone with a token-dependent correction path & Decayed CE, distribution matching, and confidence BCE \\
DCore & Block Parallel & Five-layer DFlash backbone with DFlare target-layer mixing & D-PACE weighted cross-entropy \\
\bottomrule
\end{tabular}
\end{table*}

%% file: table/Table-drafter-parameters.tex
\begin{table*}[t]
\centering
\footnotesize
\setlength{\tabcolsep}{4pt}
\caption{Drafter parameters (millions) under both targets.
\textsc{Layers} is the number of drafter decoder layers.
\textsc{Fusion} is the projection from the selected target layers into the
drafter width: a shared projection over five layers for DFlash, D-PACE,
Domino, and DSpark; per-draft-layer projections over nine layers for DFlare
and DCore; a three-layer fusion FC for Eagle-3.
\textsc{Backbone} is the transformer decoder stack.
\textsc{Corr.} is the within-block correction module (Domino's correction
head, DSpark's token-dependent correction path).
\textsc{Head} is the output unembedding; entries marked with $^\dagger$ are
the frozen LM head shared with the target model---not trained, but executed
in every draft forward pass and therefore included in \textsc{Total}.
All other columns are trainable. The target embedding table reused by all
drafters is excluded. The counts are independent of the draft block size,
which only sets the number of mask placeholders per drafted block.}
\label{tab:drafter-capacity}
\begin{tabular}{@{}lc ccccc ccccc@{}}
\toprule
& & \multicolumn{5}{c}{Qwen3-4B target ($d{=}2560$)}
  & \multicolumn{5}{c}{Qwen3-8B target ($d{=}4096$)} \\
\cmidrule(lr){3-7}\cmidrule(l){8-12}
\textbf{Drafter} & \textsc{Layers}
& \textsc{Fusion} & \textsc{Backbone} & \textsc{Corr.} & \textsc{Head} & \textsc{Total}
& \textsc{Fusion} & \textsc{Backbone} & \textsc{Corr.} & \textsc{Head} & \textsc{Total} \\
\midrule
Eagle-3 & 1 & 19.7 & 116.7 & ---   & 81.9              & 218.2  & 50.3 & 218.1 & ---   & 131.1             & 399.5 \\
DFlash  & 5 & 32.8 & 504.7 & ---   & 389.0$^\dagger$   & 926.4  & 83.9 & 964.7 & ---   & 622.3$^\dagger$   & 1671.0 \\
DFlare  & 5 & 26.2 & 504.7 & ---   & 389.0$^\dagger$   & 919.8  & 41.9 & 964.7 & ---   & 622.3$^\dagger$   & 1629.0 \\
D-PACE  & 5 & 32.8 & 504.7 & ---   & 389.0$^\dagger$   & 926.4  & 83.9 & 964.7 & ---   & 622.3$^\dagger$   & 1671.0 \\
Domino  & 5 & 32.8 & 504.7 & 50.8  & 389.0$^\dagger$   & 977.2  & 83.9 & 964.7 & 55.9  & 622.3$^\dagger$   & 1726.9 \\
DSpark  & 5 & 32.8 & 504.7 & 77.8  & 389.0$^\dagger$   & 1004.2 & 83.9 & 964.7 & 77.8  & 622.3$^\dagger$   & 1748.8 \\
\midrule
DCore   & 5 & 26.2 & 504.7 & ---   & 389.0$^\dagger$   & 919.8  & 41.9 & 964.7 & ---   & 622.3$^\dagger$   & 1629.0 \\
\midrule
Target  & \multicolumn{1}{c}{---}
& \multicolumn{4}{r}{total parameters:} & 4022.5
& \multicolumn{4}{r}{total parameters:} & 8190.7 \\
\bottomrule
\end{tabular}
\end{table*}

%% file: table/Table-seed-consistency.tex
\begin{table}[H]
\centering
\footnotesize
\setlength{\tabcolsep}{3.2pt}
\begin{tabular}{@{}lc cc cc cc@{}}
\toprule
 & & \multicolumn{2}{c}{$0|0$} & \multicolumn{2}{c}{$1|0$} & \multicolumn{2}{c}{$1|1$} \\
\cmidrule(lr){3-4}\cmidrule(lr){5-6}\cmidrule(lr){7-8}
Seed & Anchors & Low & High & Low & High & Low & High \\
\midrule
$0$         & $256$ & 5.53 & 3.96 & 4.84 & 3.27 & 4.88 & 3.34 \\
$1$         & $256$ & 5.53 & 3.96 & 4.83 & 3.27 & 4.87 & 3.33 \\
$42$        & $256$ & 5.53 & 3.96 & 4.82 & 3.29 & 4.89 & 3.34 \\
\midrule
Tab.~\ref{tab:main_results_temps} & $512$ & 5.57 & 4.05 & 4.91 & 3.33 & 4.97 & 3.42 \\
\bottomrule
\end{tabular}
\caption{\textbf{Training-seed robustness (DCore, Qwen3-4B, budget $8$).}
Mean $\tau$ over each entropy half. Each seed sets both data shuffling and
initialization. The last row is the Tab.~\ref{tab:main_results_temps}
checkpoint.}
\label{tab:seed_consistency}
\end{table}

%% file: section/Appendix-measurement.tex
\section{Serving and Measurement Details}
\label{sec:appendix-protocol}

\paragraph{Lossless verification.}
For target distribution $p$ and a sampled proposal $\hat x\sim q$, rejection
sampling accepts with probability $\min\{1,p(\hat x)/q(\hat x)\}$ and,
on rejection, draws from $p'(v)\propto[p(v)-q(v)]_+$, preserving the target
distribution \citep{leviathan2023fast,chen2023accelerating}.
Expected token acceptance is
\begin{equation}
\alpha(p,q)=\sum_v\min\{p(v),q(v)\}=1-\operatorname{TV}(p,q).
\label{eq:appendix-acceptance-tv}
\end{equation}
For a greedy draft, $q_j$ is a point mass at $\hat x_j$, giving
\begin{align}
\Pr[\text{accept }\hat x_j]&=p_j(\hat x_j),
\label{eq:accept}\\
p'_j(v)&=\frac{p_j(v)\mathbf{1}[v\ne\hat x_j]}{1-p_j(\hat x_j)}.
\label{eq:residual}
\end{align}
The residual is used only after rejection, when its denominator is nonzero.
At $T_t=0$, acceptance reduces to greedy-token agreement. For sampled
proposals it reaches one at $q=p$, without generally ordering sampled and
greedy drafts. For comparison, independent sampling followed by token-equality
verification yields
\begin{equation}
\mathbb E[\alpha_{\mathrm{indep}}]=\sum_v p(v)q(v),
\label{eq:app-couple-indep}
\end{equation}
which is no greater than maximal-coupling acceptance for fixed $p,q$.
Our $1|1$ comparison excludes Eagle-3 trees, whose deterministic candidate
selection requires separate tree verification.

\paragraph{Serving configuration.}
Measurements use eight single-GPU SGLang replicas, with concurrency $C$
per replica and $8C$ requests in flight. All methods and the target-only
baseline share bf16 precision, FlashInfer attention and
CUDA graphs within each sweep.
App.~\ref{app:serving-impl} specifies execution and timing boundaries;
App.~\ref{sec:appendix-concurrency} defines the request-throughput metric.

\paragraph{Benchmark configuration and SpecMix.}
Benchmarks run zero-shot with a $2048$-token generation cap.
\textsc{SpecMix} pools GSM8K ($500$), MATH-500 ($500$),
AIME25 ($30$), HumanEval ($164$), MBPP ($257$), LiveCodeBench ($500$),
MT-Bench ($80$), AlpacaEval ($500$), Arena-Hard ($500$), and SWE-Bench
($300$). Its $3331$ prompts are shuffled once with a fixed seed and shared
across methods and concurrencies.

%% file: section/Appendix-serving.tex
\section{Serving Implementation and Stage Costs}
\label{app:serving-impl}

All methods use a common SGLang fork. Tab.~\ref{tab:serving-graph-coverage}
documents step execution; Tab.~\ref{tab:serving-stage-latency} reports
matched device-stage medians at $0|0$, $C{=}1$ on both targets and budgets.

\input{table/Table-execution-stages.tex}
\input{table/Table-stage-latency.tex}

\paragraph{Engine and configuration.}
We use SGLang v0.5.14, FlashInfer 0.6.12, PyTorch 2.9.1, and CUDA 12.8 on
H800 GPUs. Both greedy and sampled draft paths use CUDA-graph replay across
the measured batch sizes, with device-resident acceptance handling.
DSpark's confidence head and adaptive-length policy are disabled at inference,
excluding general serving budget control from the fixed-budget comparison.

\paragraph{Budget convention.}
The verification budget $B$ counts the window including its committed/bonus
token convention. For context-only anchors, block size is $B$; the anchor
supplies context and is not a draft position. DSpark's predictive anchor uses
block size $B-1$, aligning verification windows across serving paths.
Eagle-3's default $B{=}60$ is reported separately.

\paragraph{Timing and profiling scope.}
CUDA-event timers measure proposal generation, verification, acceptance,
and the complete device step; Tab.~\ref{tab:serving-stage-latency} reports
stage medians. Throughput is measured separately without stage timers.
Timers follow worker boundaries: Eagle-3's verification region includes
auxiliary mask and bonus handling. \emph{Resid.} covers device work outside
named stages, included in the complete device step.

\paragraph{Matched cost observations.}
Across the $32$ greedy $C{=}1$ configurations, the host/device step-time gap
$\Delta$ is at most $0.10$~ms ($1.6\%$ of the recorded step). Target-forward
medians among the six parallel and semi-AR recipes differ by at most
$0.01$~ms within a configuration. These diagnostics assess launch overhead
and verification costs under matched configurations.

%% file: table/Table-execution-stages.tex
\begin{table*}[t]
\centering
\footnotesize
\setlength{\tabcolsep}{4pt}
\caption{Execution stages of one speculative decode step, using the labels
in Tab.~\ref{tab:serving-stage-latency}. AR and parallel/semi-AR drafters
differ in proposal generation and cache resynchronization; a dash denotes
an absent stage. Model forwards use CUDA-graph replay and accepted-length
bookkeeping stays on device. The table describes greedy drafts; sampled
paths also use graph replay, with distribution-dependent sampling tails
outside the rollout graph.}
\label{tab:serving-graph-coverage}
\begin{tabular}{@{}l p{0.275\textwidth} p{0.275\textwidth} p{0.23\textwidth}@{}}
\toprule
\textbf{Stage} & \textsc{Autoregressive drafter} & \textsc{Block-parallel and semi-AR drafters}
& \textsc{Execution} \\
\midrule
Setup
& Allocates the drafter's key--value slots for the whole rollout and builds its
positions and forward metadata in one pass, sized from an upper bound so that no
per-step count is read back.
& Writes the draft block --- the previous step's bonus token, then mask tokens
--- with its positions and cache slots, and gathers its input embeddings from
the target's embedding table.
& Metadata kernels ahead of the replay. Timed for the autoregressive drafter,
part of \textsc{Resid.}\ for the others. \\
\addlinespace[2pt]
Rollout
& Every drafter forward and, between consecutive forwards, the top-$k$ expansion
that turns one step's logits into the next step's input, with the running score
and parent bookkeeping.
& One forward over the whole block, with the output head and argmax that turn its
hidden states into the proposal; Domino's and DSpark's correction paths run here
as fused kernels.
& One replayed graph, holding all of the work in this row. \\
\addlinespace[2pt]
Proposal assembly
& One kernel writes the verification mask in the attention backend's bit-packed
layout, the proposal's positions, and the three retrieval arrays acceptance
indexes with.
& --- The block is contiguous and causal, so verification needs no custom mask
and no retrieval arrays.
& One kernel launch, writing straight into the verification graph's buffers. \\
\midrule
Target forward
& \multicolumn{2}{p{0.56\textwidth}}{The target verifies the entire proposal in
one forward using the shared target model and serving framework.}
& CUDA-graph replay for the target forward. \\
\addlinespace[2pt]
Acceptance
& Constructs the target's distribution where the protocol needs one, runs the
tree or chain acceptance kernel, then extracts the bonus token; above top-$k$
one the accepted path is compacted to the front of the cache.
& Compares the block against the target's argmax, or against its distribution at
a sampled target; one kernel writes the accepted length, the commit length, the
bonus token, the output tokens and the new sequence lengths.
& Device-resident kernels writing preallocated buffers. The host's view of the
accepted length is deliberately lagging. \\
\addlinespace[2pt]
Resync
& A drafter forward over the tokens verification has just committed, which both
refreshes its key--value cache and emits the first candidate of the next
rollout.
& The target's hidden states for the committed prefix, projected to the drafter's
width and written into the drafter's key--value cache.
& A third replayed graph for the autoregressive drafter, one fused kernel for the
others. \\
\bottomrule
\end{tabular}
\end{table*}

%% file: table/Table-stage-latency.tex
\begin{table*}[!t]
\centering
\footnotesize
\setlength{\tabcolsep}{4pt}
\caption{Device-stage latency (ms), greedy decoding ($0|0$), one request
per replica. Entries are medians over timed steps with CUDA graphs enabled.
\emph{Rollout} measures proposal compute; \emph{Total} includes its bracketing
stages. \emph{Drafter resync} updates the cache through KV writes or an AR forward.
A dash denotes an absent stage. Eagle-3's \emph{Target fwd.} includes bonus
handling and tree cache compaction, giving an upper bound on forward time.
\emph{Resid.} captures device work outside named timers; \emph{Device} records
the complete device step. $\Delta$ is host minus device step time.
These profiles diagnose stage costs; Fig.~\ref{fig:serving_tps_qwen3_8b_bgt8}
reports end-to-end speedup from unprofiled runs. App.~\ref{app:serving-impl}
defines the timing boundaries.}
\label{tab:serving-stage-latency}
\begin{tabular}{@{}l cc cc cc ccc@{}}
\toprule
& \multicolumn{2}{c}{\textsc{Proposal}} & \multicolumn{2}{c}{\textsc{Verification}}
& \multicolumn{2}{c}{\textsc{Drafter resync}} & \multicolumn{3}{c}{\textsc{Step}} \\
\cmidrule(lr){2-3}\cmidrule(lr){4-5}\cmidrule(lr){6-7}\cmidrule(l){8-10}
\textbf{Drafter} & Rollout & Total & Target fwd. & Acceptance
& KV write & Resync fwd. & Resid. & Device & $\Delta$ \\
\midrule
\multicolumn{10}{@{}l}{\textsc{Qwen3-4B}, budget $8$} \\
Eagle-3 (7/1/8)   & $2.02$ & $2.14$ & $4.57$ & $0.025$ & ---     & $0.320$ & $0.010$ & $7.07$ & $0.020$ \\
Eagle-3 (7/4/8)   & $2.44$ & $2.47$ & $4.82$ & $0.026$ & ---     & $0.430$ & $0.000$ & $7.75$ & $0.020$ \\
DFlash            & $1.28$ & $1.31$ & $4.54$ & $0.018$ & $0.076$ & ---     & $0.266$ & $6.21$ & $0.090$ \\
DFlare            & $1.28$ & $1.31$ & $4.55$ & $0.018$ & $0.070$ & ---     & $0.262$ & $6.21$ & $0.090$ \\
D-PACE            & $1.28$ & $1.31$ & $4.54$ & $0.018$ & $0.077$ & ---     & $0.255$ & $6.20$ & $0.100$ \\
Domino            & $1.66$ & $1.69$ & $4.54$ & $0.018$ & $0.077$ & ---     & $0.265$ & $6.59$ & $0.090$ \\
DSpark            & $1.74$ & $1.77$ & $4.54$ & $0.046$ & $0.140$ & ---     & $0.164$ & $6.66$ & $0.030$ \\
\midrule
DCore             & $1.28$ & $1.31$ & $4.54$ & $0.018$ & $0.070$ & ---     & $0.272$ & $6.21$ & $0.100$ \\
\midrule
\multicolumn{10}{@{}l}{\textsc{Qwen3-4B}, budget $16$} \\
Eagle-3 (15/1/16) & $4.26$ & $4.39$ & $4.66$ & $0.028$ & ---     & $0.330$ & $0.010$ & $9.42$ & $0.020$ \\
Eagle-3 (7/10/16) & $2.48$ & $2.52$ & $4.91$ & $0.029$ & ---     & $0.440$ & $0.010$ & $7.91$ & $0.020$ \\
DFlash            & $1.29$ & $1.32$ & $4.62$ & $0.021$ & $0.079$ & ---     & $0.270$ & $6.31$ & $0.090$ \\
DFlare            & $1.29$ & $1.32$ & $4.63$ & $0.021$ & $0.084$ & ---     & $0.265$ & $6.32$ & $0.080$ \\
D-PACE            & $1.29$ & $1.32$ & $4.62$ & $0.021$ & $0.079$ & ---     & $0.270$ & $6.31$ & $0.090$ \\
Domino            & $2.12$ & $2.15$ & $4.62$ & $0.021$ & $0.080$ & ---     & $0.259$ & $7.13$ & $0.100$ \\
DSpark            & $2.15$ & $2.18$ & $4.62$ & $0.047$ & $0.140$ & ---     & $0.183$ & $7.17$ & $0.030$ \\
\midrule
DCore             & $1.29$ & $1.33$ & $4.63$ & $0.021$ & $0.085$ & ---     & $0.264$ & $6.33$ & $0.080$ \\
\midrule
\multicolumn{10}{@{}l}{\textsc{Qwen3-8B}, budget $8$} \\
Eagle-3 (7/1/8)   & $2.58$ & $2.61$ & $6.89$ & $0.025$ & ---     & $0.440$ & $0.010$ & $9.98$ & $0.020$ \\
Eagle-3 (7/4/8)   & $2.91$ & $2.94$ & $7.18$ & $0.026$ & ---     & $0.550$ & $0.000$ & $10.7$ & $0.000$ \\
DFlash            & $1.74$ & $1.77$ & $6.84$ & $0.018$ & $0.120$ & ---     & $0.262$ & $9.01$ & $0.060$ \\
DFlare            & $1.74$ & $1.77$ & $6.85$ & $0.018$ & $0.088$ & ---     & $0.264$ & $8.99$ & $0.080$ \\
D-PACE            & $1.74$ & $1.77$ & $6.84$ & $0.018$ & $0.120$ & ---     & $0.262$ & $9.01$ & $0.060$ \\
Domino            & $2.12$ & $2.15$ & $6.84$ & $0.018$ & $0.120$ & ---     & $0.272$ & $9.40$ & $0.050$ \\
DSpark            & $2.21$ & $2.24$ & $6.84$ & $0.046$ & $0.200$ & ---     & $0.114$ & $9.44$ & $0.030$ \\
\midrule
DCore             & $1.74$ & $1.77$ & $6.84$ & $0.018$ & $0.089$ & ---     & $0.273$ & $8.99$ & $0.080$ \\
\midrule
\multicolumn{10}{@{}l}{\textsc{Qwen3-8B}, budget $16$} \\
Eagle-3 (15/1/16) & $5.63$ & $5.67$ & $6.99$ & $0.027$ & ---     & $0.450$ & $0.060$ & $13.2$ & $0.000$ \\
Eagle-3 (7/10/16) & $3.01$ & $3.05$ & $7.30$ & $0.029$ & ---     & $0.570$ & $0.050$ & $11.0$ & $0.000$ \\
DFlash            & $1.75$ & $1.79$ & $6.96$ & $0.021$ & $0.120$ & ---     & $0.259$ & $9.15$ & $0.060$ \\
DFlare            & $1.75$ & $1.78$ & $6.97$ & $0.021$ & $0.081$ & ---     & $0.268$ & $9.12$ & $0.090$ \\
D-PACE            & $1.75$ & $1.79$ & $6.96$ & $0.021$ & $0.120$ & ---     & $0.259$ & $9.15$ & $0.060$ \\
Domino            & $2.59$ & $2.62$ & $6.96$ & $0.020$ & $0.120$ & ---     & $0.260$ & $9.98$ & $0.020$ \\
DSpark            & $2.62$ & $2.66$ & $6.96$ & $0.047$ & $0.200$ & ---     & $0.123$ & $9.99$ & $0.010$ \\
\midrule
DCore             & $1.75$ & $1.78$ & $6.97$ & $0.021$ & $0.081$ & ---     & $0.268$ & $9.12$ & $0.090$ \\
\bottomrule
\end{tabular}
\end{table*}

%% file: section/Appendix-source-protocols.tex
\section{Verification Protocols of the Published Baselines}
\label{app:published-protocols}

Tab.~\ref{tab:published-protocols} records source-method verification rules,
training data, and released checkpoints at the listed repository revisions.

\input{table/Table-source-protocols.tex}

\paragraph{Source-specific verification.}
DFlare uses deterministic proposals with a sampled target
(Eq.~\eqref{eq:accept}). Eagle-3 constructs deterministic top-$k$ trees,
whose acceptance also depends on candidate traversal. Domino's Transformers
path uses independent sampling and token equality
(Eq.~\eqref{eq:app-couple-indep}); DFlash and DSpark use maximal coupling
(Eq.~\eqref{eq:appendix-acceptance-tv}), and D-PACE inherits DFlash's verifier.
App.~\ref{sec:appendix-protocol} gives our common verification rules.

\paragraph{Checkpoint provenance.}
Released-checkpoint measurements use the linked weights, including AngelSlim's
Qwen3 Eagle-3 checkpoints. D-PACE has no released weights and enters only
the retrained comparison. Its reported training corpus reuses DFlash data
for Qwen3-4B and ShareGPT for Qwen3-8B. Dataset sizes are transcribed from
the source reports.

%% file: table/Table-source-protocols.tex
\begin{table*}[!t]
\centering
\footnotesize
\setlength{\tabcolsep}{2pt}
\newcolumntype{L}[1]{>{\raggedright\arraybackslash}p{#1}}
\providecommand{\hf}[2]{\href{https://huggingface.co/#1}{#2}}
\caption{\textbf{Verification protocols, training sets and released weights of the
published baselines.} $T_t$ and $T_d$ are the target and draft temperatures the
pinned commit's code actually reaches, not the flag value; one row per pair.
$\hat x$ is the drafted token, $p,q$ the target and drafter distributions,
$R$ the rejected set. NPT2/CA/S3.5/OPB $=$ Nemotron-Post-Training-Dataset-v2 /
CodeAlpaca / Step-3.5-Flash-SFT / Open-PerfectBlend; \mbox{n/r} $=$ unreported. $^*$D-PACE's 4B setting; its 8B corpus is ShareGPT.}
\label{tab:published-protocols}
\begin{tabular}{@{}ccL{0.275\textwidth}L{0.20\textwidth}ccL{0.10\textwidth}r@{}}
\toprule
$T_t$ & $T_d$ & \textsc{Acceptance rule} & \textsc{Resampling rule} & \textsc{Lossless} & \textsc{Coupling} & \textsc{Dataset} & \textsc{Samples} \\
\midrule
\multicolumn{8}{@{}l}{\textbf{Eagle-3} --- \href{https://github.com/SafeAILab/EAGLE}{SafeAILab/EAGLE} @ \texttt{cb7e0841}; third-party weights \hf{AngelSlim/Qwen3-4B_eagle3}{Qwen3-4B}, \hf{AngelSlim/Qwen3-8B_eagle3}{Qwen3-8B}} \\
$0$ & $0$ &
  $\mathbf{1}[\hat x = \arg\max p]$ &
  --- (bonus $=\arg\max p$) &
  \checkmark & $\delta$ &
  \multirow{2}{=}{n/r} & \multirow{2}{*}{n/r} \\
$T$ & $0$ &
  $\min\{1,\,p(\hat x)/1\} = p(\hat x)$; $q(\hat x)$ is the constant $1$ &
  $\dfrac{p(v)\,\mathbf{1}[v\notin R]}{1-\sum_{u\in R} p(u)}$ &
  \checkmark & $\delta$ & & \\
\addlinespace
\multicolumn{8}{@{}l}{\textbf{DFlash}, authors' repository --- \href{https://github.com/z-lab/dflash}{z-lab/dflash} @ \texttt{07ebd93d}; weights \hf{z-lab/Qwen3-4B-DFlash-b16}{Qwen3-4B}, \hf{z-lab/Qwen3-8B-DFlash-b16}{Qwen3-8B}} \\
$0$ & $0$ &
  $\mathbf{1}[\hat x = \arg\max p]$ &
  --- (bonus $=\arg\max p$) &
  \checkmark & $\delta$ &
  \multirow{2}{=}{NPT2$+$CA} & \multirow{2}{*}{$800$K} \\
$1$ & $1$ &
  $\min\{1,\,p(\hat x)/q(\hat x)\}$ &
  $\dfrac{(p-q)_+}{\sum_v (p-q)_+}$ &
  \checkmark & maximal & & \\
\addlinespace
\multicolumn{8}{@{}l}{\textbf{DFlare} --- \href{https://github.com/Tencent/AngelSlim}{Tencent/AngelSlim} @ \texttt{9366604d}; weights \hf{AngelSlim/Qwen3-4b-dflare}{Qwen3-4B}, \hf{AngelSlim/Qwen3-8b-dflare}{Qwen3-8B}} \\
$0$ & $0$ &
  $\mathbf{1}[\hat x = \arg\max p]$ &
  --- (bonus $=\arg\max p$) &
  \checkmark & $\delta$ &
  \multirow{2}{=}{NPT2$+$CA $+$S3.5} & \multirow{2}{*}{$2.4$M} \\
$1$ & $0$ &
  $p(\hat x)$, i.e.\ $\mathbf{1}[\hat x = y]$ with $y\sim p$ and $q=\delta_{\hat x}$ &
  --- (bonus $=y\sim p$) &
  \checkmark & $\delta$ & & \\
\addlinespace
\multicolumn{8}{@{}l}{\textbf{D-PACE}, training code only --- \href{https://github.com/Lucas-TY/D-PACE}{Lucas-TY/D-PACE} @ \texttt{f36bad6e}; \emph{no public weights}} \\
$0$ & $0$ &
  $\mathbf{1}[\hat x = \arg\max p]$ &
  --- (bonus $=\arg\max p$) &
  \checkmark & $\delta$ &
  \multirow{2}{=}{NPT2$+$CA$^*$} & \multirow{2}{*}{$100$K$^*$} \\
$1$ & $1$ &
  $\min\{1,\,p(\hat x)/q(\hat x)\}$ &
  $\dfrac{(p-q)_+}{\sum_v (p-q)_+}$ &
  \checkmark & maximal & & \\
\addlinespace
\multicolumn{8}{@{}l}{\textbf{Domino}, Transformers path --- \href{https://github.com/jianuo-huang/Domino}{jianuo-huang/Domino} @ \texttt{44e1e691}; weights \hf{Huang2020/Qwen3-4B-Domino-b16}{Qwen3-4B}, \hf{Huang2020/Qwen3-8B-Domino-b16}{Qwen3-8B}} \\
$0$ & $0$ &
  $\mathbf{1}[\hat x = \arg\max p]$ &
  --- (bonus $=\arg\max p$) &
  \checkmark & $\delta$ &
  \multirow{2}{=}{OPB} & \multirow{2}{*}{$1.42$M} \\
$1$ & $1$ &
  $\mathbf{1}[\hat x = y]$ with $y\sim p$, so $\mathbb{E}[\alpha]=\sum_v p(v)q(v)$ &
  --- (bonus $=y\sim p$) &
  \checkmark & independent & & \\
\addlinespace
\multicolumn{8}{@{}l}{\textbf{DSpark} --- \href{https://github.com/deepseek-ai/DeepSpec}{deepseek-ai/DeepSpec} @ \texttt{005e03b8}; weights \hf{deepseek-ai/dspark_qwen3_4b_block7}{Qwen3-4B}, \hf{deepseek-ai/dspark_qwen3_8b_block7}{Qwen3-8B}} \\
$1$ & $1$ &
  $\min\{1,\,p(\hat x)/q(\hat x)\}$ &
  $\dfrac{(p-q)_+}{\sum_v (p-q)_+}$ &
  \checkmark & maximal &
  OPB & $1.3$M \\
\bottomrule
\end{tabular}
\end{table*}

%% file: section/Appendix-alignment.tex
\section{Published and Aligned Acceptance Lengths}
\label{app:published-tau}

Tab.~\ref{tab:published-tau} distinguishes published acceptance lengths
(\mRep), released-checkpoint remeasurement (\mMeas), and shared-data recipe
retraining (\mRetr). Published values are transcribed from the source
main tables, with Eagle-3's Qwen3 results taken from DFlash.
\footnote{Eagle-3(16), Eagle-3(60), and DFlash: \citet{chen2026dflash},
Tab.~1; DFlare: \citet{zhang2026dflare}, Tab.~1; D-PACE:
\citet{wu2026dpacedynamicpositionawarecrossentropy}, Tab.~1, depth $5$L;
Domino: \citet{huang2026dominodecouplingcausalmodeling}, Tab.~1;
DSpark: \citet{cheng2026dspark}, Tab.~1.}
Blue and orange shading indicate acceptance increases and decreases relative
to the preceding column.

\input{table/Table-acceptance-comparison.tex}

Fig.~\ref{fig:published_vs_controlled} summarizes four budget-$16$ recipes
at target temperature $1$. The measured stages use $1|0$, three-repeat
means, and a shared $149.9$~TPS target-only reference; published bars retain
their source protocols and references.

%% file: table/Table-acceptance-comparison.tex
\newcommand{\tupA}[1]{\cellcolor{taupos!13}$#1$}
\newcommand{\tupB}[1]{\cellcolor{taupos!24}$#1$}
\newcommand{\tupC}[1]{\cellcolor{taupos!39}$#1$}
\newcommand{\tupD}[1]{\cellcolor{taupos!55}$#1$}
\newcommand{\tupE}[1]{\cellcolor{taupos!74}$#1$}
\newcommand{\tdnA}[1]{\cellcolor{tauneg!13}$#1$}
\newcommand{\tdnB}[1]{\cellcolor{tauneg!24}$#1$}
\newcommand{\tdnC}[1]{\cellcolor{tauneg!39}$#1$}
\newcommand{\tdnD}[1]{\cellcolor{tauneg!55}$#1$}
\newcommand{\tdnE}[1]{\cellcolor{tauneg!74}$#1$}
\newcommand{\tfl}[1]{$#1$}
\newcommand{\tnr}[1]{\cellcolor{black!12}$#1$}
\newcommand{\tsw}[1]{\raisebox{-0.2ex}{\setlength{\fboxsep}{0pt}%
  \fcolorbox{black!30}{#1}{\rule{9pt}{0pt}\rule{0pt}{4.4pt}}}}
\begin{table*}[!t]
\centering
\scriptsize
\setlength{\tabcolsep}{1.8pt}
\caption{\textbf{Acceptance length, as published, as re-measured and as
re-trained.} Three columns per benchmark: \mRep\ is $\tau$ exactly as reported by
the source paper, the only column taken from the literature; \mMeas\ re-runs that
method's released checkpoint, and \mRetr\ a drafter we retrain on one common
training set, both under the single protocol of
App.~\ref{sec:appendix-protocol}. $T_t|T_d$ is the target and draft
temperature pair set by each method's own evaluation code
(Tab.~\ref{tab:published-protocols}); parenthesised
numbers are verification budgets. `---' marks a configuration the paper does not
report (\mRep) or a drafter with no released weights (\mMeas).
\textbf{Shading shows within-row changes}: each measured cell is
filled by its percentage change in $\tau$ against the cell to its left (legend
below), grey where that cell is `---'. Cross-method comparisons use
matched budgets and temperature pairs.}
\label{tab:published-tau}
\begin{tabular}{@{}cl*{6}{ccc@{\hspace{4pt}}}ccc@{}}
\toprule
& & \multicolumn{15}{c}{\textsc{Low-Entropy}} & \multicolumn{6}{c}{\textsc{High-Entropy}} \\
\cmidrule(lr){3-17} \cmidrule(lr){18-23}
& & \multicolumn{3}{c}{GSM8K} & \multicolumn{3}{c}{MATH} & \multicolumn{3}{c}{HE} & \multicolumn{3}{c}{MBPP} & \multicolumn{3}{c}{LCB} & \multicolumn{3}{c}{AIME25} & \multicolumn{3}{c}{MT-B} \\
\cmidrule(lr){3-5} \cmidrule(lr){6-8} \cmidrule(lr){9-11} \cmidrule(lr){12-14} \cmidrule(lr){15-17} \cmidrule(lr){18-20} \cmidrule(lr){21-23}
$T_t|T_d$ & Method
 & \mRep & \mMeas & \mRetr & \mRep & \mMeas & \mRetr & \mRep & \mMeas & \mRetr
 & \mRep & \mMeas & \mRetr & \mRep & \mMeas & \mRetr & \mRep & \mMeas & \mRetr
 & \mRep & \mMeas & \mRetr \\
\midrule
\multicolumn{23}{@{}l}{\textbf{Qwen3-4B}} \\
\multirow{6}{*}{$0|0$} & Eagle-3(16) & $3.30$ & \tfl{3.33} & \tupE{6.33} & $3.08$ & \tfl{3.14} & \tupE{6.09} & $3.05$ & \tfl{3.11} & \tupE{5.51} & $2.95$ & \tfl{2.98} & \tupE{5.34} & $2.91$ & \tupA{3.01} & \tupE{5.28} & $3.05$ & \tfl{3.10} & \tupE{5.52} & $3.02$ & \tdnA{2.93} & \tupD{3.75} \\
 & Eagle-3(60) & $3.77$ & \tupA{3.85} & \tupE{7.05} & $3.52$ & \tupA{3.63} & \tupE{6.91} & $3.47$ & \tfl{3.53} & \tupE{6.36} & $3.38$ & \tfl{3.43} & \tupE{6.21} & $3.22$ & \tupB{3.39} & \tupE{6.11} & $3.51$ & \tupA{3.59} & \tupE{6.49} & $3.49$ & \tdnA{3.33} & \tupD{4.51} \\
 & DFlash(16) & $6.53$ & \tdnA{6.23} & \tupD{8.05} & $7.84$ & \tfl{7.87} & \tdnA{7.56} & $6.64$ & \tdnA{6.49} & \tdnB{6.00} & $6.09$ & \tdnB{5.70} & \tdnA{5.50} & $7.09$ & \tdnB{6.70} & \tdnC{5.88} & $7.27$ & \tdnA{7.08} & \tdnB{6.41} & $4.35$ & \tdnC{3.27} & \tupB{3.50} \\
 & DFlare(16) & $7.93$ & \tdnA{7.66} & \tupB{8.17} & $8.97$ & \tfl{8.96} & \tdnC{7.68} & $7.17$ & \tdnA{6.97} & \tdnC{6.07} & $7.08$ & \tdnB{6.55} & \tdnC{5.58} & --- & \tnr{7.64} & \tdnC{5.96} & $8.35$ & \tdnA{8.07} & \tdnC{6.37} & $5.30$ & \tdnC{4.04} & \tdnC{3.59} \\
 & D-PACE(16) & $5.54$ & --- & \tnr{8.48} & $6.61$ & --- & \tnr{7.97} & $5.54$ & --- & \tnr{6.16} & $4.99$ & --- & \tnr{5.71} & --- & --- & \tnr{6.28} & --- & --- & \tnr{6.87} & $3.79$ & --- & \tnr{3.51} \\
 & Domino(16) & $10.12$ & \tdnC{8.96} & \tupA{9.14} & $9.01$ & \tdnB{8.36} & \tupA{8.59} & $7.02$ & \tdnA{6.80} & \tupA{6.95} & $7.11$ & \tdnC{6.34} & \tupA{6.48} & $6.86$ & \tdnA{6.53} & \tfl{6.65} & $7.49$ & \tdnA{7.18} & \tupA{7.47} & $5.12$ & \tdnD{3.84} & \tupA{4.03} \\
\addlinespace
\multirow{3}{*}{$1|0$} & Eagle-3(16) & $3.22$ & \tfl{3.24} & \tupE{5.91} & $2.99$ & \tfl{2.97} & \tupE{5.29} & $3.01$ & \tfl{3.03} & \tupE{5.21} & $2.89$ & \tfl{2.92} & \tupE{4.98} & $2.77$ & \tfl{2.76} & \tupE{4.74} & $2.79$ & \tfl{2.76} & \tupD{4.41} & $2.95$ & \tdnB{2.77} & \tupD{3.51} \\
 & Eagle-3(60) & $3.68$ & \tfl{3.75} & \tupE{6.74} & $3.44$ & \tfl{3.43} & \tupE{6.13} & $3.39$ & \tfl{3.45} & \tupE{6.08} & $3.33$ & \tfl{3.36} & \tupE{5.91} & $3.14$ & \tfl{3.13} & \tupE{5.51} & $3.20$ & \tfl{3.23} & \tupE{5.24} & $3.36$ & \tdnA{3.20} & \tupD{4.19} \\
 & DFlare(16) & $7.14$ & \tdnB{6.69} & \tupA{6.91} & $7.34$ & \tdnC{6.40} & \tdnC{5.69} & $6.43$ & \tdnA{6.12} & \tdnC{5.38} & $6.39$ & \tdnC{5.75} & \tdnC{5.07} & --- & \tnr{6.27} & \tdnC{4.97} & $5.20$ & \tdnB{4.86} & \tdnC{4.25} & $4.77$ & \tdnD{3.55} & \tdnC{3.19} \\
\addlinespace
\multirow{4}{*}{$1|1$} & DFlash(16) & $6.00$ & \tdnC{5.34} & \tupD{6.71} & $6.67$ & \tdnC{5.85} & \tdnA{5.58} & $6.04$ & \tdnB{5.46} & \tdnB{5.13} & $5.66$ & \tdnC{4.80} & \tdnB{4.49} & $6.50$ & \tdnC{5.16} & \tdnC{4.59} & $4.92$ & \tdnB{4.65} & \tdnB{4.28} & $4.07$ & \tdnD{2.80} & \tupB{2.96} \\
 & D-PACE(16) & $5.31$ & --- & \tnr{7.27} & $5.73$ & --- & \tnr{6.05} & $5.19$ & --- & \tnr{5.49} & $4.73$ & --- & \tnr{5.02} & --- & --- & \tnr{5.17} & --- & --- & \tnr{4.60} & $3.49$ & --- & \tnr{3.04} \\
 & Domino(16) & $8.61$ & \tdnB{7.97} & \tupA{8.14} & $7.35$ & \tdnC{6.59} & \tupA{6.75} & $6.30$ & \tdnA{6.09} & \tupA{6.34} & $6.36$ & \tdnC{5.50} & \tupA{5.75} & $6.45$ & \tdnC{5.25} & \tupA{5.39} & $4.83$ & \tupB{5.15} & \tfl{5.22} & $4.56$ & \tdnD{3.39} & \tupA{3.54} \\
 & DSpark(7) & $6.11$ & \tfl{6.13} & \tfl{6.09} & $5.70$ & \tfl{5.74} & \tdnA{5.54} & $5.38$ & \tupA{5.49} & \tdnA{5.32} & $5.13$ & \tfl{5.15} & \tdnA{5.01} & $4.86$ & \tupA{5.04} & \tfl{4.94} & $4.89$ & \tfl{4.87} & \tfl{4.79} & $3.64$ & \tfl{3.64} & \tfl{3.57} \\
\midrule
\multicolumn{23}{@{}l}{\textbf{Qwen3-8B}} \\
\multirow{5}{*}{$0|0$} & Eagle-3(16) & $3.23$ & \tfl{3.26} & \tupE{6.42} & $3.02$ & \tupA{3.14} & \tupE{6.30} & $3.17$ & \tfl{3.23} & \tupE{5.86} & $2.82$ & \tupA{2.91} & \tupE{5.55} & $2.65$ & \tupC{2.95} & \tupE{5.75} & $3.00$ & \tupA{3.10} & \tupE{5.80} & $2.83$ & \tdnA{2.73} & \tupD{3.97} \\
 & Eagle-3(60) & $3.71$ & \tfl{3.78} & \tupE{7.12} & $3.49$ & \tupA{3.63} & \tupE{7.05} & $3.65$ & \tupA{3.76} & \tupE{6.61} & $3.25$ & \tupA{3.41} & \tupE{6.46} & $3.03$ & \tupC{3.34} & \tupE{6.42} & $3.44$ & \tupA{3.53} & \tupE{6.61} & $3.26$ & \tdnA{3.15} & \tupD{4.67} \\
 & DFlash(16) & $6.54$ & \tdnA{6.24} & \tupD{8.04} & $7.87$ & \tfl{7.77} & \tdnA{7.60} & $6.50$ & \tfl{6.47} & \tdnA{6.21} & $5.95$ & \tdnB{5.65} & \tdnA{5.49} & $7.27$ & \tdnA{7.08} & \tdnB{6.54} & $7.08$ & \tdnA{6.73} & \tdnB{6.20} & $4.24$ & \tdnC{3.22} & \tupB{3.40} \\
 & DFlare(16) & $7.88$ & \tdnA{7.59} & \tupC{8.36} & $8.95$ & \tfl{8.85} & \tdnC{7.88} & $7.08$ & \tfl{7.04} & \tdnB{6.54} & $6.86$ & \tdnB{6.41} & \tdnC{5.75} & --- & \tnr{7.97} & \tdnC{6.80} & $8.12$ & \tdnA{7.87} & \tdnC{6.49} & $5.09$ & \tdnC{3.95} & \tdnB{3.57} \\
 & Domino(16) & $10.03$ & \tdnB{9.19} & \tupA{9.57} & $9.43$ & \tdnB{8.52} & \tupB{9.04} & $7.39$ & \tdnA{7.11} & \tupB{7.60} & $7.04$ & \tdnB{6.43} & \tupA{6.71} & $7.04$ & \tfl{7.06} & \tupB{7.60} & $7.41$ & \tdnA{7.10} & \tupB{7.49} & $5.18$ & \tdnC{3.91} & \tupA{4.06} \\
\addlinespace
\multirow{3}{*}{$1|0$} & Eagle-3(16) & $3.12$ & \tfl{3.14} & \tupE{5.94} & $2.91$ & \tfl{2.90} & \tupE{5.39} & $3.05$ & \tfl{3.05} & \tupE{5.21} & $2.74$ & \tfl{2.79} & \tupE{5.06} & $2.57$ & \tfl{2.59} & \tupE{5.18} & $2.74$ & \tdnA{2.67} & \tupE{4.64} & $2.70$ & \tdnB{2.53} & \tupD{3.59} \\
 & Eagle-3(60) & $3.59$ & \tfl{3.66} & \tupE{6.73} & $3.38$ & \tfl{3.37} & \tupE{6.27} & $3.54$ & \tfl{3.55} & \tupE{6.06} & $3.16$ & \tupA{3.25} & \tupE{5.95} & $2.92$ & \tfl{2.95} & \tupE{5.94} & $3.18$ & \tdnA{3.11} & \tupE{5.37} & $3.05$ & \tdnA{2.91} & \tupD{4.28} \\
 & DFlare(16) & $7.07$ & \tdnB{6.49} & \tupB{6.88} & $7.18$ & \tdnC{6.20} & \tdnB{5.78} & $5.76$ & \tdnA{5.60} & \tdnB{5.31} & $5.77$ & \tdnB{5.28} & \tdnB{4.87} & --- & \tnr{6.49} & \tdnC{5.58} & $5.24$ & \tdnC{4.69} & \tdnB{4.34} & $4.37$ & \tdnC{3.37} & \tdnB{3.13} \\
\addlinespace
\multirow{3}{*}{$1|1$} & DFlash(16) & $5.98$ & \tdnC{5.18} & \tupC{6.16} & $6.40$ & \tdnC{5.71} & \tdnB{5.35} & $5.52$ & \tdnB{5.18} & \tdnB{4.87} & $5.21$ & \tdnC{4.55} & \tdnB{4.27} & $6.69$ & \tdnC{5.29} & \tdnC{4.68} & $4.73$ & \tfl{4.65} & \tdnC{4.14} & $3.80$ & \tdnD{2.79} & \tupA{2.85} \\
 & Domino(16) & $8.34$ & \tdnA{7.97} & \tupC{8.82} & $7.20$ & \tdnB{6.59} & \tupC{7.46} & $5.98$ & \tupA{6.10} & \tupB{6.70} & $6.02$ & \tdnB{5.64} & \tupB{6.07} & $6.72$ & \tdnC{5.72} & \tupB{6.23} & $4.92$ & \tupB{5.23} & \tupC{5.85} & $4.61$ & \tdnC{3.49} & \tupB{3.67} \\
 & DSpark(7) & $6.17$ & \tfl{6.22} & \tfl{6.21} & $5.78$ & \tfl{5.80} & \tfl{5.78} & $5.52$ & \tfl{5.50} & \tfl{5.43} & $5.16$ & \tfl{5.15} & \tfl{5.16} & $5.17$ & \tupA{5.28} & \tfl{5.30} & $5.01$ & \tfl{4.98} & \tfl{4.98} & $3.72$ & \tfl{3.68} & \tdnA{3.60} \\
\bottomrule
\end{tabular}

\vspace{2pt}
{\footnotesize
Fill: change in $\tau$ against the cell to its left. Rose
\tsw{taupos!13}\tsw{taupos!24}\tsw{taupos!39}\tsw{taupos!55}\tsw{taupos!74}
/ fell
\tsw{tauneg!13}\tsw{tauneg!24}\tsw{tauneg!39}\tsw{tauneg!55}\tsw{tauneg!74},
from $2/5/10/25/60\%$; ${<}2\%$ unshaded.}
\end{table*}

%% file: section/Appendix-task-entropy.tex
\section{Task Entropy and Acceptance}
\label{sec:appendix-entropy}

\paragraph{Entropy measurement and grouping.}
The target generates greedily, and we compute entropy $H$ from its raw
$T{=}1$ next-token probabilities at each generated position. We average
within each response, then across at most $500$ prompts per benchmark,
giving responses equal weight. The available top-$64$ log-probabilities
yield a lower bound on full-vocabulary entropy. Tab.~\ref{tab:model_entropy}
reports both targets; the five lowest-entropy benchmarks form
\textsc{Low-Entropy}, and the remaining five form \textsc{High-Entropy}.

\input{table/Table-task-entropy.tex}

\paragraph{Acceptance under the three temperature settings.}
The rules in App.~\ref{sec:appendix-protocol} show that $0|0$ tests
greedy-token agreement, $1|0$ depends on target probability at the proposed
token, and $1|1$ depends on distributional overlap
(Eq.~\eqref{eq:appendix-acceptance-tv}). Task entropy describes continuation
uncertainty; together with draft--target agreement, it helps interpret
the workload patterns in Section~\ref{sec:analysis}.

\input{figure/Figure-entropy-returns}

\paragraph{Continuous analysis.}
Fig.~\ref{fig:entropy-continuous} complements the grouping with
benchmark-level Pearson correlations. Entropy--acceptance correlations
are $-0.86$, $-0.84$, and $-0.86$ for Eagle-3, DFlash, and DSpark,
respectively. The accompanying script records the benchmark means and
correlations; the caption specifies the measured configurations and budget
return calculation.

%% file: table/Table-task-entropy.tex
\begin{table}[t]
\centering
\small
\setlength{\tabcolsep}{5pt}
\caption{Target-model token-distribution entropy $H$ per benchmark, in nats ($\pm$ std.\ across prompts), sorted by the Qwen3-8B column. The rule separates the five \textsc{Low-Entropy} from the five \textsc{High-Entropy} benchmarks of Tab.~\ref{tab:main_results_temps}: that grouping is exactly the top and bottom half of this ordering, under either target.}
\label{tab:model_entropy}
\begin{tabular}{lcc}
\toprule
\textsc{Dataset} & \textsc{Qwen3-4B} & \textsc{Qwen3-8B} \\
\midrule
LCB         & $0.064 \pm 0.057$ & $0.069 \pm 0.065$ \\
GSM8K       & $0.124 \pm 0.054$ & $0.142 \pm 0.059$ \\
MATH-500    & $0.171 \pm 0.088$ & $0.192 \pm 0.094$ \\
MBPP        & $0.128 \pm 0.066$ & $0.195 \pm 0.076$ \\
HumanEval   & $0.133 \pm 0.050$ & $0.209 \pm 0.071$ \\
\midrule
Alpaca      & $0.261 \pm 0.149$ & $0.313 \pm 0.172$ \\
MT-Bench    & $0.270 \pm 0.174$ & $0.331 \pm 0.191$ \\
AIME25      & $0.326 \pm 0.143$ & $0.341 \pm 0.128$ \\
Arena-Hard  & $0.308 \pm 0.117$ & $0.398 \pm 0.147$ \\
SWE-Bench   & $0.407 \pm 0.079$ & $0.525 \pm 0.102$ \\
\bottomrule
\end{tabular}
\end{table}

%% file: figure/Figure-entropy-returns.tex
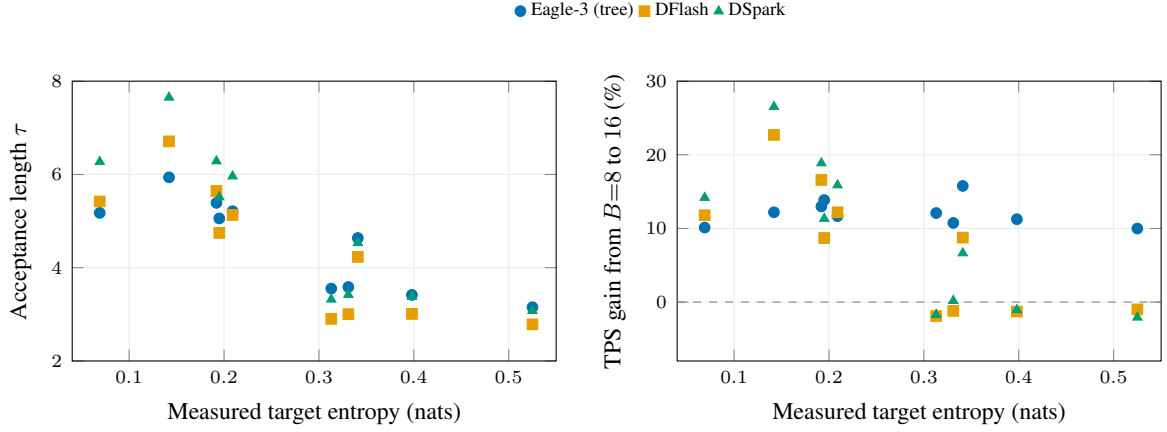
\begin{figure*}[t]
\centering
\begin{tikzpicture}
\begin{groupplot}[group style={group size=2 by 1,horizontal sep=1.6cm},
width=0.40\textwidth,height=3.7cm,scale only axis,
xmin=0.04,xmax=0.55,xtick={0.1,0.2,0.3,0.4,0.5},
xlabel={Measured target entropy (nats)},
label style={font=\small},tick label style={font=\scriptsize},
legend style={font=\scriptsize,draw=none},legend columns=3,
grid=major,grid style={gray!15}]
\nextgroupplot[ylabel={Acceptance length $\tau$},ymin=2,ymax=8,
legend style={at={(1.2,1.18)},anchor=south}]
\addplot[only marks,color=cEagle,mark=*,mark size=2pt] coordinates {(0.142,5.938) (0.192,5.391) (0.209,5.212) (0.195,5.057) (0.069,5.176) (0.341,4.637) (0.525,3.153) (0.331,3.586) (0.313,3.553) (0.398,3.417)};
\addlegendentry{Eagle-3 (tree)}
\addplot[only marks,color=cDFlash,mark=square*,mark size=2pt] coordinates {(0.142,6.709) (0.192,5.646) (0.209,5.131) (0.195,4.746) (0.069,5.421) (0.341,4.232) (0.525,2.786) (0.331,3.003) (0.313,2.904) (0.398,3.010)};
\addlegendentry{DFlash}
\addplot[only marks,color=cDSpark,mark=triangle*,mark size=2pt] coordinates {(0.142,7.652) (0.192,6.285) (0.209,5.961) (0.195,5.513) (0.069,6.272) (0.341,4.531) (0.525,3.077) (0.331,3.422) (0.313,3.320) (0.398,3.380)};
\addlegendentry{DSpark}
\nextgroupplot[ylabel={TPS gain from $B{=}8$ to $16$ (\%)},ymin=-8,ymax=30]
\addplot[gray,dashed,no marks,forget plot] coordinates {(0.04,0) (0.55,0)};
\addplot[only marks,color=cEagle,mark=*,mark size=2pt] coordinates {(0.142,12.1937) (0.192,12.9886) (0.209,11.6692) (0.195,13.8573) (0.069,10.1197) (0.341,15.7727) (0.525,9.9842) (0.331,10.7484) (0.313,12.0916) (0.398,11.2522)};
\addplot[only marks,color=cDFlash,mark=square*,mark size=2pt] coordinates {(0.142,22.7078) (0.192,16.5872) (0.209,12.1909) (0.195,8.7030) (0.069,11.7934) (0.341,8.7498) (0.525,-0.9932) (0.331,-1.2038) (0.313,-1.8975) (0.398,-1.2948)};
\addplot[only marks,color=cDSpark,mark=triangle*,mark size=2pt] coordinates {(0.142,26.5227) (0.192,18.8719) (0.209,15.8796) (0.195,11.3331) (0.069,14.1809) (0.341,6.6487) (0.525,-2.0808) (0.331,0.1793) (0.313,-1.7135) (0.398,-1.0665)};
\end{groupplot}
\end{tikzpicture}
\caption{\textbf{Continuous task variation, without an entropy split.}
Each method has one point per benchmark (Qwen3-8B, $1|0$).
Left: $B{=}16$ acceptance length. Right: throughput change
$100(\mathrm{TPS}_{16}/\mathrm{TPS}_{8}-1)$, using three-repeat means from
the evaluation logs. Entropy follows Tab.~\ref{tab:model_entropy}.
Eagle-3 widens a tree using one checkpoint; DFlash and DSpark use
budget-specific checkpoints. Points summarize benchmark-level associations.}
\label{fig:entropy-continuous}
\end{figure*}

%% file: section/Appendix-dcore.tex
\section{DCore Recipe}
\label{sec:dcore}

\input{figure/Figure-dcore}

DCore combines DFlash's five-layer block-parallel backbone,
DFlare's draft-layer-specific projections over nine target layers,
and D-PACE's position-weighted cross-entropy objective
(Fig.~\ref{fig:dcore}). It retains the context-only anchor and frozen target
LM head, with a separate checkpoint per verification budget. The objective
adds no inference operations.

%% file: figure/Figure-dcore.tex
\begin{figure}[!t]
\centering

\definecolor{dcInk}{HTML}{111111}
\definecolor{dcGrey}{HTML}{666666}

\def\dcFT{\fontsize{8}{9.4}\selectfont}    %
\def\dcFL{\fontsize{7}{8.2}\selectfont}    %
\def\dcFS{\fontsize{6.4}{7.4}\selectfont}  %

\def\dcCW{7.2}     %
\def\dcX{106}      %
\def\dcY{184}      %
\def\dcS{2.0714}   %
\def\dcFw{4.6}     %
\def\dcFh{6.6}     %
\def\dcFx{32}      %
\def\dcFy{76}      %
\def\dcIx{20}      %
\def\dcIy{188}     %
\def\dcIw{4.4}     %
\def\dcIh{28}      %

\newcommand{\dcfcell}[3]{%
  \pgfmathsetmacro{\dcp}{6+84*(#3)}%
  \fill[cDFlare!\dcp!white, draw=white, line width=0.35pt]
    ({\dcFx+(#2-1)*\dcFw},{\dcFy+(#1-1)*\dcFh})
    rectangle ({\dcFx+#2*\dcFw},{\dcFy+#1*\dcFh});
}

\begin{tikzpicture}[x=1pt, y=1pt, >=latex, line join=round]
\path (0,13) rectangle (216,258);

\node[anchor=base west, text=dcInk, font=\dcFT, inner sep=0pt] at (0,16)
  {\textbf{(a)} one target-layer mixture \emph{per draft layer}};
\node[anchor=base west, text=dcInk, font=\dcFT, inner sep=0pt] at (0,250)
  {\textbf{(b)} loss weight per block position};

\draw[rounded corners=2pt, fill=black!3, draw=black!30, line width=0.5pt]
  (2,50) rectangle (18,135);
\foreach \dck/\dctap in {1/0,2/1,3/1,4/1,5/1,6/1,7/1,8/1,9/1,10/1,11/0,12/0}{%
  \pgfmathsetmacro{\dcy}{52+(\dck-1)*7}%
  \ifnum\dctap=1
    \fill[black!13, draw=black!32, line width=0.35pt]
      (5,\dcy) rectangle (15,{\dcy+5});
    \draw[cDFlare!80!black, line width=0.45pt] (15,{\dcy+2.5}) -- (24,{\dcy+2.5});
    \fill[cDFlare!80!black] (15,{\dcy+2.5}) circle (0.9pt);
  \else
    \fill[white, draw=black!22, line width=0.35pt]
      (5,\dcy) rectangle (15,{\dcy+5});
  \fi
}
\draw[cDFlare!80!black, line width=1pt] (24,61.5) -- (24,117.5);
\draw[->, cDFlare!80!black, line width=1pt] (24,92.5) -- (30.5,92.5);
\node[anchor=north, align=center, text=dcGrey, font=\dcFS, inner sep=1pt]
  at (10,48) {target\\(frozen)};

\foreach \dcv [count=\dcc] in {0.86,1.00,0.86,0.54,0.25,0.09,0.02,0.00,0.00}
  {\dcfcell{1}{\dcc}{\dcv}}
\foreach \dcv [count=\dcc] in {0.38,0.71,0.96,0.96,0.71,0.38,0.15,0.04,0.01}
  {\dcfcell{2}{\dcc}{\dcv}}
\foreach \dcv [count=\dcc] in {0.09,0.25,0.54,0.86,1.00,0.86,0.54,0.25,0.09}
  {\dcfcell{3}{\dcc}{\dcv}}
\foreach \dcv [count=\dcc] in {0.01,0.04,0.15,0.38,0.71,0.96,0.96,0.71,0.38}
  {\dcfcell{4}{\dcc}{\dcv}}
\foreach \dcv [count=\dcc] in {0.00,0.00,0.02,0.09,0.25,0.54,0.86,1.00,0.86}
  {\dcfcell{5}{\dcc}{\dcv}}
\draw[cDFlare!70!black, line width=0.5pt] (32,76) rectangle (73.4,109);
\node[anchor=south, text=cDFlare!55!black, font=\dcFS, inner sep=1pt]
  at (52.7,110) {$W_{\mathrm{fuse}}$};
\foreach \dcc in {1,...,5}{%
  \fill[black!13, draw=white, line width=0.35pt]
    ({32+(\dcc-1)*\dcFw},44) rectangle ({32+\dcc*\dcFw},50);
}
\draw[black!35, line width=0.45pt] (32,44) rectangle (55,50);
\node[anchor=base west, text=dcGrey, font=\dcFS, inner sep=1pt]
  at (57.5,45) {DFlash: $c_t$};

\foreach \dci in {1,...,5}{%
  \pgfmathsetmacro{\dcy}{50+(\dci-1)*18}%
  \pgfmathsetmacro{\dcyc}{\dcy+6.5}%
  \draw[->, cDFlare!80!black, line width=0.6pt]
    (74.5,{\dcFy+(\dci-1)*\dcFh+3.3}) -- (94.8,\dcyc);
  \fill[cDFlare!35, draw=cDFlare!80!black, line width=0.45pt]
    (96,\dcy) rectangle (104,{\dcy+13});
  \fill[cDFlash!22, draw=cDFlash!80!black, line width=0.45pt]
    (104,\dcy) rectangle (214,{\dcy+13});
  \node[text=dcInk, font=\dcFL, inner sep=0pt] at (159,\dcyc)
    {draft layer \dci};
}
\fill[cDFlash!35, draw=cDFlash!80!black, line width=0.45pt]
  (96,34) rectangle (106,45);
\node[text=dcInk, font=\dcFL, inner sep=0pt] at (101,39.5) {$x_t$};
\foreach \dcc in {1,...,15}{%
  \fill[black!5, draw=black!30, line width=0.3pt]
    ({\dcX+(\dcc-1)*\dcCW},34) rectangle ({\dcX+\dcc*\dcCW},45);
}
\node[anchor=north, text=dcGrey, font=\dcFS, inner sep=1pt] at (160,32)
  {$1$ anchor $+$ $15$ masked positions, one pass};
\draw[->, dcInk!55, line width=0.6pt] (160,45) -- (160,49.5);
\draw[->, dcInk!55, line width=0.6pt] (160,135) -- (160,139.5);
\fill[black!8, draw=black!35, line width=0.45pt] (104,140) rectangle (214,152);
\node[text=dcGrey, font=\dcFL, inner sep=0pt] at (159,146)
  {target LM head (frozen)};
\draw[->, dcInk!55, line width=0.6pt] (160,152) -- (160,156.5);
\foreach \dcc in {1,...,15}{%
  \fill[cDPace!14, draw=cDPace!55, line width=0.3pt]
    ({\dcX+(\dcc-1)*\dcCW},157) rectangle ({\dcX+\dcc*\dcCW},168);
}
\node[anchor=east, text=cDPace!70!black, font=\dcFS, inner sep=2pt]
  at (105,162.5) {$q_j$};

\foreach \dcc in {1,...,15}{%
  \draw[black!22, dotted, line width=0.35pt]
    ({\dcX+(\dcc-0.5)*\dcCW},169) -- ({\dcX+(\dcc-0.5)*\dcCW},183);
}
\draw[->, cDPace!60, line width=0.7pt] (100,170) -- (100,182);

\foreach \dcg in {10,20}{%
  \draw[black!10, line width=0.35pt]
    (104,{\dcY+\dcg*\dcS}) -- (214,{\dcY+\dcg*\dcS});
}
\draw[black!45, line width=0.5pt] (104,\dcY) -- (104,244);
\draw[black!45, line width=0.5pt] (104,\dcY) -- (214,\dcY);
\node[anchor=east, text=dcGrey, font=\dcFS, inner sep=1.5pt] at (103,\dcY) {0};
\node[anchor=east, text=dcGrey, font=\dcFS, inner sep=1.5pt]
  at (103,{\dcY+10*\dcS}) {10};
\node[anchor=east, text=dcGrey, font=\dcFS, inner sep=1.5pt]
  at (103,{\dcY+20*\dcS}) {20\%};
\foreach \dcw [count=\dcc] in
  {26.34,19.66,14.66,10.91,8.09,5.98,4.39,3.21,2.32,1.65,1.15,0.77,0.49,0.28,0.12}{%
  \fill[cDPace!22, draw=cDPace!50, line width=0.25pt]
    ({\dcX+(\dcc-1)*\dcCW+0.9},\dcY)
    rectangle ({\dcX+(\dcc-1)*\dcCW+3.5},{\dcY+\dcw*\dcS});
}
\foreach \dcw [count=\dcc] in
  {13.63,12.48,11.36,10.29,9.26,8.26,7.29,6.36,5.47,4.60,3.76,2.96,2.18,1.43,0.70}{%
  \fill[cDPace!85, draw=cDPace!95!black, line width=0.25pt]
    ({\dcX+(\dcc-1)*\dcCW+3.7},\dcY)
    rectangle ({\dcX+(\dcc-1)*\dcCW+6.3},{\dcY+\dcw*\dcS});
}
\draw[dcGrey!85, dash pattern=on 1.6pt off 1.2pt, line width=0.6pt]
  (109.6,215.2) -- (116.8,211.1) -- (124.0,207.5) -- (131.2,204.3) --
  (138.4,201.7) -- (145.6,199.3) -- (152.8,197.3) -- (160.0,195.5) --
  (167.2,194.0) -- (174.4,192.6) -- (181.6,191.5) -- (188.8,190.5) --
  (196.0,189.6) -- (203.2,188.9) -- (210.4,188.2);

\draw[dcGrey!85, dash pattern=on 1.6pt off 1.2pt, line width=0.6pt]
  (4,244.5) -- (14,244.5);
\node[anchor=west, text=dcGrey, font=\dcFS, inner sep=1.2pt]
  at (16,244.5) {DFlash: fixed decay};
\fill[cDPace!22, draw=cDPace!50, line width=0.25pt]
  (4,234.8) rectangle (14,238.2);
\node[anchor=west, text=cDPace!70!black, font=\dcFS, inner sep=1.2pt]
  at (16,236.5) {D-PACE, early training};
\fill[cDPace!85, draw=cDPace!95!black, line width=0.25pt]
  (4,226.8) rectangle (14,230.2);
\node[anchor=west, text=cDPace!70!black, font=\dcFS, inner sep=1.2pt]
  at (16,228.5) {D-PACE, late training};

\fill[black!2] (\dcIx,\dcIy) rectangle ({\dcIx+15*\dcIw},{\dcIy+\dcIh});
\draw[black!20, dotted, line width=0.35pt]
  (\dcIx,{\dcIy+\dcIh}) -- ({\dcIx+15*\dcIw},{\dcIy+\dcIh});
\draw[black!35, line width=0.4pt]
  (\dcIx,{\dcIy+\dcIh}) -- (\dcIx,\dcIy) -- ({\dcIx+15*\dcIw},\dcIy);
\node[anchor=east, text=dcGrey, font=\dcFS, inner sep=1.5pt]
  at ({\dcIx-1},{\dcIy+\dcIh}) {1};
\node[anchor=east, text=dcGrey, font=\dcFS, inner sep=1.5pt]
  at ({\dcIx-1},\dcIy) {0};
\node[anchor=base west, text=dcGrey, font=\dcFS, inner sep=0pt]
  at (\dcIx,{\dcIy+\dcIh+3}) {reach probability};
\draw[cDPace!55, line width=0.7pt]
  (20,216.0) -- (24.4,209.0) -- (28.8,203.8) -- (33.2,199.8) -- (37.6,196.9) --
  (42,194.6) -- (46.4,193.0) -- (50.8,191.7) -- (55.2,190.8) -- (59.6,190.1) --
  (64,189.6) -- (68.4,189.2) -- (72.8,188.9) -- (77.2,188.7) -- (81.6,188.5) --
  (86,188.4);
\draw[cDPace!95!black, line width=0.7pt]
  (20,216.0) -- (24.4,215.0) -- (28.8,214.1) -- (33.2,213.2) -- (37.6,212.3) --
  (42,211.4) -- (46.4,210.6) -- (50.8,209.8) -- (55.2,209.1) -- (59.6,208.3) --
  (64,207.6) -- (68.4,206.9) -- (72.8,206.3) -- (77.2,205.6) -- (81.6,205.0) --
  (86,204.4);

\end{tikzpicture}

\caption{\textbf{DCore recipe.} Orange: DFlash backbone; blue: DFlare context mixture;
purple: D-PACE objective; grey: frozen target and replaced defaults.
\textbf{(a)} Draft layer $i$ mixes nine target layers using its own
$\alpha^{(i)}=\mathrm{softmax}(W_{\mathrm{fuse}}[i,:])$.
\textbf{(b)} Position weights
$\bar w_j=\sum_{m\ge j}\prod_{i\le m}\tilde q_i$ use smoothed draft confidences
$\tilde q_i$. Bars illustrate the normalized weights for uniform early/late
confidences; the inset shows the corresponding prefix products.
Fusion shading and confidence curves are schematic, not measured.
App.~\ref{sec:dcore} specifies the recipe.}
\label{fig:dcore}
\end{figure}
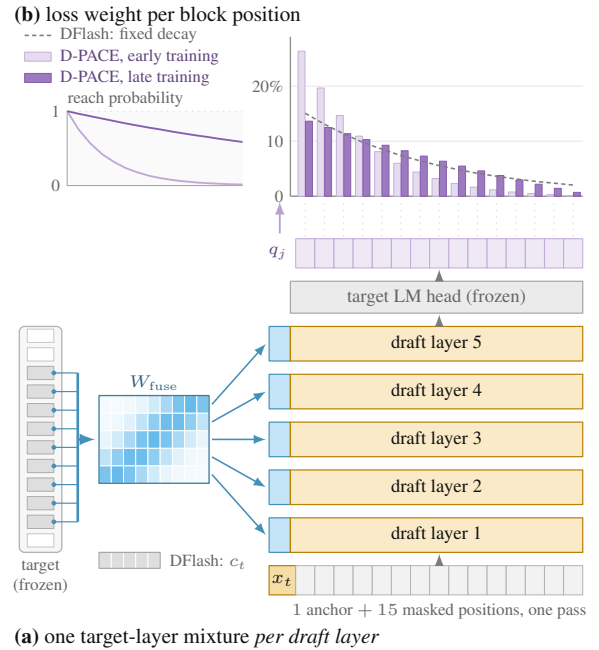

%% file: section/Appendix-design-notes.tex
\section{Design Ablation Notes}
\label{sec:appendix-dcore-roadmap}

\paragraph{Search and evaluation results.}
The roadmap uses single-run design-search measurements, with three-repeat
means for Eagle-3 anchors. Tab.~\ref{tab:main_results_temps} evaluates
the selected DCore recipe using the full benchmark protocol. The objective sweep is in
App.~\ref{sec:appendix-loss-ablation}.

\paragraph{Configuration details.}
Most roadmap edges change one component. Eagle-3's $7/10/16$ tree to
$7/1/8$ chain changes both budget and decoding mode; the intermediate
$7/4/8$ tree is reported in Tab.~\ref{tab:per_benchmark_b8}.
Domino's candidate-pool variant restricts conditioned predictions to
the top-$2048$ marginal candidates, following
\citet{oda2026treesfrommarginals}.

%% file: section/Appendix-eagle-concurrency.tex
\section{Eagle-3 Chains and Trees Under Concurrency}
\label{sec:appendix-bgt60}

Fig.~\ref{fig:serving_profile_bgt60} compares five decoding configurations
using one Eagle-3 checkpoint.

\paragraph{Load changes the preferred configuration.}
The budget-$60$ tree has the highest speedup at $C=1$ ($2.49\times$),
the budget-$16$ tree at $C=4,8$, and the budget-$8$ chain from $C=16$
onward. At budget $8$, the tree is about $5\%$ faster at $C=1$ but the
chain is $33\%$ faster at $C=64$. At budget $16$, the tree leads by $45\%$
at $C=1$, while the chain leads by $9\%$ at $C=64$.
The widest tree falls below $1\times$ between $C=8$ and $16$ and reaches
$0.34\times$ at $C=64$; only the budget-$8$ chain stays above the baseline
throughout.

\paragraph{Execution costs.}
The budget-$8$ pair shares draft depth, while the budget-$16$ pair compares
$15$-step chains with $7$-step trees. A similar load-dependent reversal is
reported by \citet{liu2026illusion}.

\input{figure/Figure-eagle-concurrency.tex}

%% file: figure/Figure-eagle-concurrency.tex
\begin{figure*}[t]
    \centering
    \colorlet{cEa}{cEagle!60!white}
    \colorlet{cEb}{cEagle!85!white}
    \colorlet{cEc}{cEagle}
    \colorlet{cEd}{cEagle!70!black}
    \colorlet{cEe}{cEagle!45!black}
    \pgfplotsset{
        totbar/.style={ybar, bar width=6pt, fill=#1!75, draw=#1!95},
        tvbar/.style={ybar, bar width=6pt, fill=#1!30, draw=#1!95},
        splineF/.style={
            color=#1, line width=0.7pt, mark size=1.6pt,
            mark options={solid, fill=#1, draw=white, line width=0.3pt},
        },
        splineO/.style={
            color=#1, line width=0.7pt, mark size=1.9pt,
            mark options={solid, fill=white, draw=#1, line width=0.6pt},
        },
        panel/.style={
            width=360pt, height=3.8cm, scale only axis,
            xmin=0.5, xmax=7.5,
            xtick={1,2,3,4,5,6,7},
            xticklabels={1,2,4,8,16,32,64},
            xticklabel style={font=\scriptsize},
            yticklabel style={font=\scriptsize},
            ylabel style={font=\scriptsize},
            xlabel style={font=\scriptsize},
        },
    }
    \begin{tikzpicture}
    \begin{axis}[
        panel,
        ymin=0, ymax=90, ytick={0,30,60,90},
        ylabel={$\lambda_V{+}\lambda_D$ (ms)\,$\downarrow$},
        ymajorgrids=true, grid style={gray!25},
        xtick style={draw=none},
        axis x line*=bottom,
        axis y line*=left,
        xlabel={Concurrency $C$},
    ]
    \addplot[totbar=cEa, bar shift=-14pt] coordinates {(1,10.11) (2,10.34) (3,10.65) (4,11.14) (5,11.98) (6,14.16) (7,16.70)};
    \addplot[tvbar =cEa, bar shift=-14pt] coordinates {(1, 7.46) (2, 7.63) (3, 7.91) (4, 8.37) (5, 9.13) (6,11.30) (7,13.70)};
    \addplot[totbar=cEb, bar shift= -7pt] coordinates {(1,10.72) (2,10.96) (3,11.53) (4,12.08) (5,13.28) (6,18.38) (7,24.19)};
    \addplot[tvbar =cEb, bar shift= -7pt] coordinates {(1, 7.76) (2, 7.93) (3, 8.39) (4, 8.88) (5, 9.84) (6,14.53) (7,19.80)};
    \addplot[totbar=cEc, bar shift=  0pt] coordinates {(1,13.35) (2,13.81) (3,14.31) (4,15.04) (5,17.24) (6,22.65) (7,23.89)};
    \addplot[tvbar =cEc, bar shift=  0pt] coordinates {(1, 7.60) (2, 7.90) (3, 8.35) (4, 9.01) (5,11.06) (6,16.40) (7,17.20)};
    \addplot[totbar=cEd, bar shift=  7pt] coordinates {(1,10.98) (2,11.49) (3,12.12) (4,13.12) (5,15.96) (6,25.60) (7,36.09)};
    \addplot[tvbar =cEd, bar shift=  7pt] coordinates {(1, 7.93) (2, 8.38) (3, 8.83) (4, 9.55) (5,11.81) (6,19.85) (7,28.60)};
    \addplot[totbar=cEe, bar shift= 14pt] coordinates {(1,11.90) (2,12.83) (3,14.93) (4,20.88) (5,36.03) (6,66.10) (7,88.52)};
    \addplot[tvbar =cEe, bar shift= 14pt] coordinates {(1, 8.79) (2, 9.64) (3,11.57) (4,17.19) (5,31.70) (6,60.03) (7,80.80)};
    \end{axis}
    \begin{axis}[
        panel,
        ymin=0, ymax=3, ytick={0,1,2,3},
        yticklabels={$0\times$,$1\times$,$2\times$,$3\times$},
        ylabel={Speedup over baseline\,$\uparrow$},
        axis x line=none,
        axis y line*=right,
        legend style={
            at={(0.5,1.03)}, anchor=south,
            draw=none, fill=none, legend columns=6, font=\scriptsize,
            /tikz/every even column/.append style={column sep=6pt},
        },
        legend image code/.code={
            \draw[mark repeat=2, mark phase=2, #1]
                plot coordinates {(0cm,0cm) (0.15cm,0cm) (0.3cm,0cm)};
        },
    ]
    \addplot[splineF=cEa, mark=*        ] coordinates {(1,1.97) (2,1.94) (3,1.90) (4,1.81) (5,1.66) (6,1.39) (7,1.12)};
    \addlegendentry{$7/1/8$}
    \addplot[splineO=cEb, mark=o        ] coordinates {(1,2.06) (2,2.05) (3,1.94) (4,1.80) (5,1.58) (6,1.11) (7,0.85)};
    \addlegendentry{$7/4/8$}
    \addplot[splineF=cEc, mark=square*  ] coordinates {(1,1.59) (2,1.57) (3,1.51) (4,1.44) (5,1.27) (6,0.99) (7,0.75)};
    \addlegendentry{$15/1/16$}
    \addplot[splineO=cEd, mark=square   ] coordinates {(1,2.31) (2,2.23) (3,2.10) (4,1.90) (5,1.53) (6,0.98) (7,0.68)};
    \addlegendentry{$7/10/16$}
    \addplot[splineO=cEe, mark=triangle, mark size=2.2pt] coordinates {(1,2.49) (2,2.32) (3,1.95) (4,1.38) (5,0.80) (6,0.50) (7,0.34)};
    \addlegendentry{$7/10/60$}
    \addplot[black, dashed, line width=0.7pt, domain=0.5:7.5, samples=2] {1};
    \addlegendentry{Baseline}
    \end{axis}
    \end{tikzpicture}
    \caption{\textbf{Eagle-3 configurations under concurrency
    (Qwen3-8B, $1|0$).} Settings are depth/top-$k$/budget; filled markers
    denote chains and open markers trees. Lines show unprofiled speedup
    relative to the target-only baseline at the same concurrency. Bars
    show drafting and verification GPU-stage medians ($\lambda_D+\lambda_V$)
    through $C=64$, measured separately from the throughput runs;
    they provide supplementary stage-cost diagnostics. Acceptance is
    approximately stable across load within each setting ($3.17$, $3.56$,
    $3.34$, $4.09$, $4.80$ in legend order).}
    \label{fig:serving_profile_bgt60}
\end{figure*}
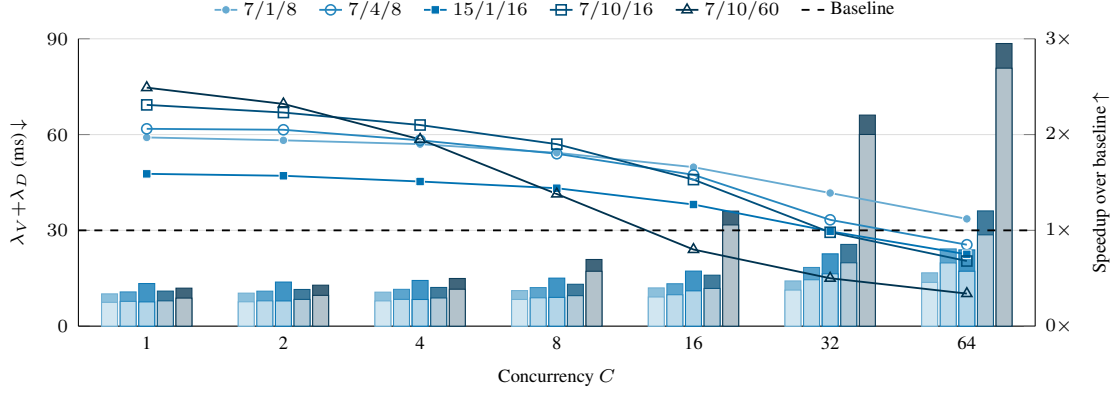

%% file: section/Appendix-benchmark-results.tex
\section{Per-Benchmark Results on Both Targets}
\label{sec:appendix-per-benchmark}

\input{table/Table-benchmark-b8.tex}
\input{table/Table-benchmark-b16.tex}

Tabs.~\ref{tab:per_benchmark_b8} and~\ref{tab:per_benchmark} expand
the main summaries into ten benchmarks at $B=8$ and $16$, respectively,
for both targets at $C=1$. The larger-budget table also includes the
budget-$60$ Eagle-3 tree; budget-$16$ chain results appear in
Tab.~\ref{tab:main_results_temps}.
Sampled-target cells are three-repeat means; greedy cells are single runs.
Tree Eagle-3 is excluded from $1|1$ because its deterministic top-$k$
candidate selection is not a sampled proposal distribution.

\paragraph{Speedup normalization.}
Speedups normalize measured throughput by the target-only reference for
each target--temperature block. Within each benchmark column, all methods
share the same divisor, so normalization preserves their throughput ranking.

\paragraph{Task variation within entropy groups.}
AIME25's unusually high greedy speedup changes when the target samples,
illustrating variation beyond the entropy-group means.

\paragraph{Acceptance and throughput remain distinct.}
On 4B at $B=16$, $1|1$, Domino has longer acceptance length than DCore on all
ten benchmarks, yet has lower speedup on nine.
The budget-$8$ table also compares Eagle-3's depth-matched tree and chain.
Across greedy-draft blocks, the tree improves speedup on chat and agentic
benchmarks by $1.9$--$9.7\%$, while low-entropy tree/chain ratios range
from about $0.95$ to $1.02$.

%% file: table/Table-benchmark-b8.tex
\providecommand{\ulm}[1]{\smash{\underline{#1}}}
\begin{table*}[!tp]
\centering
\scriptsize
\setlength{\tabcolsep}{2.0pt}
\caption{\textbf{Per-benchmark speedup and acceptance length on both targets}
($B{=}8$, $C{=}1$). Cells and ranking marks follow
Tab.~\ref{tab:per_benchmark}, using the same target-only references.
Eagle-3 includes both the $7/1/8$ chain and the $7/4/8$ tree; sampled
$1|1$ results use the chain.}
\label{tab:per_benchmark_b8}
\resizebox{\ifdim\width>\textwidth \textwidth\else \width\fi}{!}{%
\begin{tabular}{c c l cccccccccc}
\toprule
& & & \multicolumn{5}{c}{\textsc{Low-Entropy}} & \multicolumn{5}{c}{\textsc{High-Entropy}} \\
\cmidrule(lr){4-8} \cmidrule(lr){9-13}
\textsc{Tgt.} & $T_t|T_d$ & Method & GSM8K & MATH & HE & MBPP & LCB & AIME25 & SWE-B & MT-B & Alpaca & Arena-H \\
\midrule
\multirow{26}{*}{\rotatebox[origin=c]{90}{Qwen3-4B}}
& \multirow{9}{*}{$0|0$} & \textit{Target only} & \multicolumn{10}{c}{$229.6$} \\
& & Eagle-3\,(7/1/8)$^{\textsc{a}}$ & $2.97_{(5.29)}$ & $2.94_{(5.00)}$ & $2.62_{(4.53)}$ & $2.42_{(4.22)}$ & $2.46_{(4.41)}$ & $2.60_{(4.37)}$ & $1.72_{(2.91)}$ & $1.68_{(2.84)}$ & $1.66_{(2.85)}$ & $1.81_{(3.04)}$ \\
& & Eagle-3\,(7/4/8)$^{\textsc{a}}$ & $2.82_{(5.47)}$ & $2.82_{(5.26)}$ & $2.49_{(4.72)}$ & $2.37_{(4.51)}$ & $2.39_{(4.70)}$ & $2.47_{(4.59)}$ & $1.78_{(3.31)}$ & $1.76_{(3.28)}$ & $1.73_{(3.25)}$ & $1.84_{(3.40)}$ \\
& & DFlash$^{\textsc{b}}$        & $3.91_{(5.99)}$ & $3.90_{(5.83)}$ & $3.38_{(5.09)}$ & $3.21_{(4.84)}$ & $3.23_{(5.08)}$ & $3.57_{(5.29)}$ & $2.43_{(3.63)}$ & $2.29_{(3.41)}$ & $2.19_{(3.28)}$ & $2.49_{(3.69)}$ \\
& & DFlare$^{\textsc{b}}$        & $3.94_{(6.08)}$ & $3.94_{(5.93)}$ & $3.40_{(5.14)}$ & $3.23_{(4.90)}$ & $3.24_{(5.10)}$ & $3.61_{(5.37)}$ & $2.45_{(3.68)}$ & $2.33_{(3.50)}$ & $2.23_{(3.34)}$ & $2.53_{(3.77)}$ \\
& & D-PACE$^{\textsc{b}}$         & $\ulm{3.97}_{(6.10)}$ & $\ulm{3.99}_{(5.97)}$ & $\ulm{3.45}_{(5.21)}$ & $\ulm{3.28}_{(4.96)}$ & $\ulm{3.29}_{(5.19)}$ & $\ulm{3.66}_{(5.45)}$ & $\ulm{2.51}_{(3.74)}$ & $2.34_{(3.49)}$ & $2.23_{(3.33)}$ & $2.57_{(3.80)}$ \\
& & Domino$^{\textsc{s}}$        & $3.85_{(\ulm{6.30})}$ & $3.87_{(\ulm{6.17})}$ & $3.41_{(\ulm{5.50})}$ & $3.27_{(\mathbf{5.26})}$ & $3.18_{(\ulm{5.34})}$ & $3.63_{(\mathbf{5.73})}$ & $2.49_{(\mathbf{3.96})}$ & $\mathbf{2.42}_{(\mathbf{3.84})}$ & $\mathbf{2.31}_{(\mathbf{3.69})}$ & $\ulm{2.59}_{(\mathbf{4.08})}$ \\
& & DSpark$^{\textsc{s}}$        & $3.86_{(\mathbf{6.34})}$ & $3.87_{(\mathbf{6.18})}$ & $3.42_{(\mathbf{5.53})}$ & $3.25_{(\mathbf{5.26})}$ & $3.25_{(\mathbf{5.47})}$ & $3.59_{(\ulm{5.71})}$ & $2.47_{(\ulm{3.94})}$ & $\ulm{2.37}_{(\ulm{3.75})}$ & $2.25_{(\ulm{3.60})}$ & $2.55_{(\ulm{4.04})}$ \\
& & DCore$^{\textsc{b}}$         & $\mathbf{4.02}_{(6.17)}$ & $\mathbf{4.04}_{(6.06)}$ & $\mathbf{3.51}_{(5.31)}$ & $\mathbf{3.32}_{(\ulm{5.04})}$ & $\mathbf{3.35}_{(5.29)}$ & $\mathbf{3.75}_{(5.59)}$ & $\mathbf{2.54}_{(3.80)}$ & $\ulm{2.37}_{(3.56)}$ & $\ulm{2.27}_{(3.40)}$ & $\mathbf{2.62}_{(3.90)}$ \\
\cmidrule(l){2-13}
& \multirow{9}{*}{$1|0$} & \textit{Target only} & \multicolumn{10}{c}{$224.2$} \\
& & Eagle-3\,(7/1/8)$^{\textsc{a}}$ & $2.63_{(4.85)}$ & $2.36_{(4.22)}$ & $2.30_{(4.15)}$ & $2.17_{(3.93)}$ & $2.03_{(3.77)}$ & $1.89_{(3.37)}$ & $1.40_{(2.51)}$ & $1.49_{(2.67)}$ & $1.46_{(2.64)}$ & $1.51_{(2.67)}$ \\
& & Eagle-3\,(7/4/8)$^{\textsc{a}}$ & $2.55_{(5.08)}$ & $2.35_{(4.54)}$ & $2.29_{(4.45)}$ & $2.16_{(4.25)}$ & $2.03_{(4.07)}$ & $1.96_{(3.77)}$ & $1.49_{(2.89)}$ & $1.59_{(3.08)}$ & $1.55_{(3.03)}$ & $1.58_{(3.06)}$ \\
& & DFlash$^{\textsc{b}}$        & $3.52_{(5.36)}$ & $3.17_{(4.72)}$ & $3.04_{(4.65)}$ & $2.96_{(4.45)}$ & $2.87_{(4.47)}$ & $2.59_{(3.84)}$ & $2.01_{(3.00)}$ & $2.09_{(3.11)}$ & $2.02_{(3.03)}$ & $2.16_{(3.19)}$ \\
& & DFlare$^{\textsc{b}}$        & $\ulm{3.59}_{(5.46)}$ & $3.20_{(4.77)}$ & $3.13_{(4.69)}$ & $2.98_{(4.47)}$ & $2.87_{(4.47)}$ & $\mathbf{2.66}_{(3.93)}$ & $2.03_{(3.03)}$ & $2.11_{(3.14)}$ & $2.07_{(3.08)}$ & $2.18_{(3.23)}$ \\
& & D-PACE$^{\textsc{b}}$         & $3.58_{(5.45)}$ & $\ulm{3.22}_{(4.79)}$ & $\ulm{3.17}_{(4.76)}$ & $\ulm{3.04}_{(4.57)}$ & $\ulm{2.93}_{(4.57)}$ & $2.60_{(3.85)}$ & $\ulm{2.06}_{(3.06)}$ & $2.11_{(3.13)}$ & $2.07_{(3.08)}$ & $2.19_{(3.24)}$ \\
& & Domino$^{\textsc{s}}$        & $3.48_{(\ulm{5.62})}$ & $3.14_{(\ulm{4.96})}$ & $3.12_{(\ulm{4.98})}$ & $2.99_{(\mathbf{4.78})}$ & $2.86_{(\ulm{4.74})}$ & $2.56_{(\mathbf{4.03})}$ & $2.03_{(\mathbf{3.20})}$ & $\mathbf{2.16}_{(\mathbf{3.41})}$ & $\mathbf{2.11}_{(\mathbf{3.35})}$ & $2.19_{(\mathbf{3.45})}$ \\
& & DSpark$^{\textsc{s}}$        & $3.52_{(\mathbf{5.64})}$ & $3.18_{(\mathbf{4.98})}$ & $3.16_{(\mathbf{5.00})}$ & $3.01_{(\mathbf{4.78})}$ & $\ulm{2.93}_{(\mathbf{4.81})}$ & $2.59_{(\ulm{4.01})}$ & $2.04_{(\mathbf{3.20})}$ & $\ulm{2.14}_{(\ulm{3.34})}$ & $2.08_{(\ulm{3.27})}$ & $\ulm{2.21}_{(\ulm{3.44})}$ \\
& & DCore$^{\textsc{b}}$         & $\mathbf{3.63}_{(5.52)}$ & $\mathbf{3.26}_{(4.86)}$ & $\mathbf{3.20}_{(4.82)}$ & $\mathbf{3.09}_{(\ulm{4.66})}$ & $\mathbf{2.99}_{(4.67)}$ & $\ulm{2.63}_{(3.90)}$ & $\mathbf{2.09}_{(\ulm{3.11})}$ & $\mathbf{2.16}_{(3.21)}$ & $\ulm{2.10}_{(3.14)}$ & $\mathbf{2.24}_{(3.32)}$ \\
\cmidrule(l){2-13}
& \multirow{8}{*}{$1|1$} & \textit{Target only} & \multicolumn{10}{c}{$224.2$} \\
& & Eagle-3\,(7/1/8)$^{\textsc{a}}$ & $2.74_{(4.95)}$ & $2.55_{(4.42)}$ & $2.34_{(4.11)}$ & $2.13_{(3.76)}$ & $1.99_{(3.58)}$ & $2.10_{(3.61)}$ & $1.46_{(2.53)}$ & $1.47_{(2.55)}$ & $1.44_{(2.50)}$ & $1.52_{(2.61)}$ \\
& & DFlash$^{\textsc{b}}$        & $3.61_{(5.43)}$ & $3.27_{(4.82)}$ & $3.13_{(4.65)}$ & $2.85_{(4.23)}$ & $2.82_{(4.34)}$ & $2.70_{(3.96)}$ & $2.05_{(3.02)}$ & $2.03_{(2.99)}$ & $1.96_{(2.89)}$ & $2.13_{(3.12)}$ \\
& & DFlare$^{\textsc{b}}$        & $3.65_{(5.49)}$ & $3.31_{(4.87)}$ & $3.15_{(4.69)}$ & $2.93_{(4.35)}$ & $2.85_{(4.38)}$ & $2.77_{(4.06)}$ & $2.08_{(3.07)}$ & $2.07_{(3.05)}$ & $2.00_{(2.96)}$ & $2.16_{(3.17)}$ \\
& & D-PACE$^{\textsc{b}}$         & $3.68_{(5.53)}$ & $3.36_{(4.95)}$ & $3.25_{(4.83)}$ & $3.02_{(4.49)}$ & $2.96_{(4.58)}$ & $2.80_{(4.10)}$ & $2.15_{(3.18)}$ & $2.13_{(3.13)}$ & $2.06_{(3.03)}$ & $2.25_{(3.30)}$ \\
& & Domino$^{\textsc{s}}$        & $3.67_{(\ulm{5.91})}$ & $3.40_{(\ulm{5.34})}$ & $3.23_{(\ulm{5.13})}$ & $3.06_{(\ulm{4.87})}$ & $2.89_{(\ulm{4.75})}$ & $\ulm{2.92}_{(\ulm{4.56})}$ & $\ulm{2.22}_{(\ulm{3.50})}$ & $2.18_{(\ulm{3.43})}$ & $\ulm{2.16}_{(\ulm{3.40})}$ & $\ulm{2.31}_{(\ulm{3.61})}$ \\
& & DSpark$^{\textsc{s}}$        & $\mathbf{3.79}_{(\mathbf{6.09})}$ & $\mathbf{3.52}_{(\mathbf{5.54})}$ & $\mathbf{3.35}_{(\mathbf{5.32})}$ & $\mathbf{3.15}_{(\mathbf{5.01})}$ & $\ulm{2.99}_{(\mathbf{4.94})}$ & $\mathbf{3.06}_{(\mathbf{4.79})}$ & $\mathbf{2.28}_{(\mathbf{3.60})}$ & $\mathbf{2.26}_{(\mathbf{3.57})}$ & $\mathbf{2.20}_{(\mathbf{3.47})}$ & $\mathbf{2.39}_{(\mathbf{3.73})}$ \\
& & DCore$^{\textsc{b}}$         & $\ulm{3.73}_{(5.61)}$ & $\ulm{3.42}_{(5.04)}$ & $\ulm{3.31}_{(4.92)}$ & $\ulm{3.10}_{(4.61)}$ & $\mathbf{3.02}_{(4.65)}$ & $2.83_{(4.15)}$ & $2.20_{(3.25)}$ & $\ulm{2.19}_{(3.22)}$ & $2.10_{(3.11)}$ & $\ulm{2.31}_{(3.37)}$ \\
\midrule
\multirow{26}{*}{\rotatebox[origin=c]{90}{Qwen3-8B}}
& \multirow{9}{*}{$0|0$} & \textit{Target only} & \multicolumn{10}{c}{$151.0$} \\
& & Eagle-3\,(7/1/8)$^{\textsc{a}}$ & $3.33_{(5.41)}$ & $3.32_{(5.22)}$ & $3.03_{(4.82)}$ & $2.80_{(4.48)}$ & $2.92_{(4.78)}$ & $2.97_{(4.62)}$ & $1.86_{(2.90)}$ & $1.93_{(3.03)}$ & $1.88_{(2.97)}$ & $2.02_{(3.14)}$ \\
& & Eagle-3\,(7/4/8)$^{\textsc{a}}$ & $3.25_{(5.59)}$ & $3.27_{(5.45)}$ & $2.99_{(5.05)}$ & $2.79_{(4.72)}$ & $2.88_{(5.00)}$ & $2.98_{(4.91)}$ & $1.98_{(3.28)}$ & $2.06_{(3.43)}$ & $2.01_{(3.37)}$ & $2.11_{(3.48)}$ \\
& & DFlash$^{\textsc{b}}$        & $4.16_{(6.05)}$ & $4.14_{(5.87)}$ & $3.68_{(5.25)}$ & $3.40_{(4.86)}$ & $3.63_{(5.39)}$ & $3.75_{(5.28)}$ & $2.50_{(3.54)}$ & $2.39_{(3.38)}$ & $2.31_{(3.28)}$ & $2.61_{(3.67)}$ \\
& & DFlare$^{\textsc{b}}$        & $4.22_{(6.13)}$ & $4.19_{(5.95)}$ & $3.76_{(5.37)}$ & $3.47_{(4.97)}$ & $3.68_{(5.48)}$ & $3.79_{(5.32)}$ & $2.56_{(3.62)}$ & $2.44_{(3.46)}$ & $2.37_{(3.37)}$ & $2.68_{(3.77)}$ \\
& & D-PACE$^{\textsc{b}}$         & $\ulm{4.32}_{(6.27)}$ & $\ulm{4.30}_{(6.09)}$ & $\ulm{3.85}_{(5.50)}$ & $\ulm{3.57}_{(5.11)}$ & $\ulm{3.79}_{(5.63)}$ & $\ulm{3.90}_{(5.47)}$ & $2.61_{(3.71)}$ & $2.49_{(3.51)}$ & $2.39_{(3.39)}$ & $2.73_{(3.85)}$ \\
& & Domino$^{\textsc{s}}$        & $4.24_{(\ulm{6.46})}$ & $4.25_{(\ulm{6.30})}$ & $\ulm{3.85}_{(\ulm{5.76})}$ & $\mathbf{3.59}_{(\ulm{5.38})}$ & $3.69_{(\ulm{5.76})}$ & $3.83_{(\ulm{5.64})}$ & $2.64_{(\ulm{3.91})}$ & $\mathbf{2.56}_{(\mathbf{3.78})}$ & $\mathbf{2.49}_{(\mathbf{3.70})}$ & $\mathbf{2.78}_{(\mathbf{4.08})}$ \\
& & DSpark$^{\textsc{s}}$        & $4.24_{(\mathbf{6.47})}$ & $4.25_{(\mathbf{6.31})}$ & $\mathbf{3.87}_{(\mathbf{5.79})}$ & $\mathbf{3.59}_{(\mathbf{5.39})}$ & $3.74_{(\mathbf{5.83})}$ & $3.88_{(\mathbf{5.71})}$ & $\ulm{2.65}_{(\mathbf{3.93})}$ & $\ulm{2.53}_{(\ulm{3.75})}$ & $\ulm{2.47}_{(\ulm{3.68})}$ & $\ulm{2.77}_{(\ulm{4.07})}$ \\
& & DCore$^{\textsc{b}}$         & $\mathbf{4.33}_{(6.30)}$ & $\mathbf{4.33}_{(6.15)}$ & $\mathbf{3.87}_{(5.53)}$ & $\mathbf{3.59}_{(5.16)}$ & $\mathbf{3.81}_{(5.67)}$ & $\mathbf{3.98}_{(5.60)}$ & $\mathbf{2.66}_{(3.78)}$ & $2.49_{(3.55)}$ & $2.42_{(3.45)}$ & $\mathbf{2.78}_{(3.92)}$ \\
\cmidrule(l){2-13}
& \multirow{9}{*}{$1|0$} & \textit{Target only} & \multicolumn{10}{c}{$148.5$} \\
& & Eagle-3\,(7/1/8)$^{\textsc{a}}$ & $2.90_{(4.85)}$ & $2.64_{(4.31)}$ & $2.52_{(4.15)}$ & $2.39_{(3.96)}$ & $2.49_{(4.21)}$ & $2.11_{(3.45)}$ & $1.46_{(2.38)}$ & $1.67_{(2.73)}$ & $1.64_{(2.69)}$ & $1.62_{(2.62)}$ \\
& & Eagle-3\,(7/4/8)$^{\textsc{a}}$ & $2.86_{(5.12)}$ & $2.65_{(4.63)}$ & $2.57_{(4.51)}$ & $2.43_{(4.30)}$ & $2.50_{(4.51)}$ & $2.25_{(3.89)}$ & $1.60_{(2.77)}$ & $1.80_{(3.15)}$ & $1.75_{(3.08)}$ & $1.73_{(2.99)}$ \\
& & DFlash$^{\textsc{b}}$        & $3.71_{(5.34)}$ & $3.38_{(4.76)}$ & $3.17_{(4.51)}$ & $3.02_{(4.29)}$ & $3.21_{(4.74)}$ & $2.74_{(3.84)}$ & $1.97_{(2.77)}$ & $2.13_{(3.00)}$ & $2.07_{(2.91)}$ & $2.15_{(3.00)}$ \\
& & DFlare$^{\textsc{b}}$        & $3.74_{(5.38)}$ & $3.41_{(4.81)}$ & $3.23_{(4.59)}$ & $3.06_{(4.35)}$ & $3.26_{(4.83)}$ & $2.79_{(3.91)}$ & $2.00_{(2.82)}$ & $2.15_{(3.02)}$ & $2.12_{(3.00)}$ & $2.17_{(3.05)}$ \\
& & D-PACE$^{\textsc{b}}$         & $\ulm{3.82}_{(5.50)}$ & $\ulm{3.44}_{(4.87)}$ & $3.26_{(4.65)}$ & $3.14_{(4.47)}$ & $3.35_{(4.98)}$ & $\mathbf{2.83}_{(3.96)}$ & $\ulm{2.06}_{(2.90)}$ & $2.18_{(3.08)}$ & $2.13_{(3.01)}$ & $2.20_{(3.09)}$ \\
& & Domino$^{\textsc{s}}$        & $3.75_{(\ulm{5.63})}$ & $3.39_{(\mathbf{4.99})}$ & $3.28_{(\mathbf{4.87})}$ & $3.11_{(\ulm{4.62})}$ & $3.32_{(\ulm{5.12})}$ & $2.77_{(\ulm{4.04})}$ & $2.03_{(\mathbf{2.98})}$ & $\mathbf{2.23}_{(\mathbf{3.28})}$ & $\mathbf{2.20}_{(\mathbf{3.24})}$ & $2.21_{(\ulm{3.23})}$ \\
& & DSpark$^{\textsc{s}}$        & $3.78_{(\mathbf{5.64})}$ & $3.42_{(\mathbf{4.99})}$ & $\ulm{3.30}_{(\mathbf{4.87})}$ & $\ulm{3.15}_{(\mathbf{4.65})}$ & $\ulm{3.36}_{(\mathbf{5.15})}$ & $2.77_{(\mathbf{4.05})}$ & $2.04_{(\mathbf{2.98})}$ & $\ulm{2.22}_{(\ulm{3.24})}$ & $\ulm{2.18}_{(\ulm{3.20})}$ & $\ulm{2.23}_{(\mathbf{3.24})}$ \\
& & DCore$^{\textsc{b}}$         & $\mathbf{3.84}_{(5.52)}$ & $\mathbf{3.47}_{(\ulm{4.89})}$ & $\mathbf{3.31}_{(\ulm{4.70})}$ & $\mathbf{3.18}_{(4.53)}$ & $\mathbf{3.39}_{(5.03)}$ & $\ulm{2.81}_{(3.94)}$ & $\mathbf{2.07}_{(\ulm{2.91})}$ & $2.21_{(3.11)}$ & $2.16_{(3.05)}$ & $\mathbf{2.24}_{(3.13)}$ \\
\cmidrule(l){2-13}
& \multirow{8}{*}{$1|1$} & \textit{Target only} & \multicolumn{10}{c}{$148.5$} \\
& & Eagle-3\,(7/1/8)$^{\textsc{a}}$ & $3.05_{(5.01)}$ & $2.84_{(4.53)}$ & $2.67_{(4.31)}$ & $2.47_{(4.01)}$ & $2.39_{(3.97)}$ & $2.43_{(3.85)}$ & $1.59_{(2.53)}$ & $1.73_{(2.77)}$ & $1.67_{(2.69)}$ & $1.71_{(2.71)}$ \\
& & DFlash$^{\textsc{b}}$        & $3.71_{(5.30)}$ & $3.45_{(4.83)}$ & $3.18_{(4.49)}$ & $2.96_{(4.18)}$ & $3.07_{(4.50)}$ & $2.85_{(3.97)}$ & $2.10_{(2.94)}$ & $2.12_{(2.97)}$ & $2.05_{(2.87)}$ & $2.18_{(3.03)}$ \\
& & DFlare$^{\textsc{b}}$        & $3.80_{(5.43)}$ & $3.52_{(4.93)}$ & $3.30_{(4.65)}$ & $3.08_{(4.36)}$ & $3.16_{(4.63)}$ & $2.91_{(4.06)}$ & $2.15_{(3.02)}$ & $2.19_{(3.06)}$ & $2.12_{(2.98)}$ & $2.25_{(3.13)}$ \\
& & D-PACE$^{\textsc{b}}$         & $3.96_{(5.67)}$ & $3.65_{(5.11)}$ & $3.42_{(4.83)}$ & $3.22_{(4.56)}$ & $\ulm{3.34}_{(4.92)}$ & $3.05_{(4.25)}$ & $2.24_{(3.14)}$ & $2.26_{(3.17)}$ & $2.17_{(3.05)}$ & $2.35_{(3.27)}$ \\
& & Domino$^{\textsc{s}}$        & $\ulm{4.11}_{(\ulm{6.18})}$ & $\ulm{3.87}_{(\ulm{5.69})}$ & $\ulm{3.62}_{(\ulm{5.37})}$ & $\ulm{3.40}_{(\ulm{5.05})}$ & $3.32_{(\ulm{5.11})}$ & $\ulm{3.33}_{(\ulm{4.86})}$ & $\ulm{2.38}_{(\ulm{3.50})}$ & $\ulm{2.40}_{(\ulm{3.53})}$ & $\ulm{2.34}_{(\ulm{3.45})}$ & $\ulm{2.47}_{(\ulm{3.61})}$ \\
& & DSpark$^{\textsc{s}}$        & $\mathbf{4.12}_{(\mathbf{6.21})}$ & $\mathbf{3.92}_{(\mathbf{5.78})}$ & $\mathbf{3.63}_{(\mathbf{5.43})}$ & $\mathbf{3.45}_{(\mathbf{5.16})}$ & $\mathbf{3.42}_{(\mathbf{5.30})}$ & $\mathbf{3.40}_{(\mathbf{4.98})}$ & $\mathbf{2.43}_{(\mathbf{3.58})}$ & $\mathbf{2.44}_{(\mathbf{3.60})}$ & $\mathbf{2.38}_{(\mathbf{3.53})}$ & $\mathbf{2.52}_{(\mathbf{3.70})}$ \\
& & DCore$^{\textsc{b}}$         & $3.97_{(5.69)}$ & $3.68_{(5.16)}$ & $3.47_{(4.90)}$ & $3.26_{(4.62)}$ & $\mathbf{3.42}_{(5.04)}$ & $3.09_{(4.31)}$ & $2.26_{(3.17)}$ & $2.30_{(3.22)}$ & $2.23_{(3.14)}$ & $2.38_{(3.32)}$ \\
\bottomrule
\end{tabular}%
}
\end{table*}

%% file: table/Table-benchmark-b16.tex
\providecommand{\ulm}[1]{\smash{\underline{#1}}}
\begin{table*}[!tp]
\centering
\scriptsize
\setlength{\tabcolsep}{2.0pt}
\caption{\textbf{Per-benchmark speedup and acceptance length on both targets}
($B{=}16$, $C{=}1$), with an additional budget-$60$ Eagle-3 tree.
Cells are $\mathcal{S}_{(\tau)}$, normalized to the \emph{Target only}
reference in the same target--temperature block. \textbf{Bold}/\underline{underline}
mark best/second-best values within a benchmark column; speedup and acceptance
are ranked separately. Superscripts denote AR ($\textsc{a}$), block-parallel
($\textsc{b}$), and semi-AR ($\textsc{s}$). All recipes use shared-data
retraining. The Eagle-3 rows use tree decoding; its budget-$16$ chain
is summarized in Tab.~\ref{tab:main_results_temps}.}
\label{tab:per_benchmark}
\resizebox{\ifdim\width>\textwidth \textwidth\else \width\fi}{!}{%
\begin{tabular}{c c l cccccccccc}
\toprule
& & & \multicolumn{5}{c}{\textsc{Low-Entropy}} & \multicolumn{5}{c}{\textsc{High-Entropy}} \\
\cmidrule(lr){4-8} \cmidrule(lr){9-13}
\textsc{Tgt.} & $T_t|T_d$ & Method & GSM8K & MATH & HE & MBPP & LCB & AIME25 & SWE-B & MT-B & Alpaca & Arena-H \\
\midrule
\multirow{25}{*}{\rotatebox[origin=c]{90}{Qwen3-4B}}
& \multirow{9}{*}{$0|0$} & \textit{Target only} & \multicolumn{10}{c}{$229.6$} \\
& & Eagle-3$^{\textsc{a}}$   & $3.14_{(6.33)}$ & $3.15_{(6.09)}$ & $2.81_{(5.51)}$ & $2.70_{(5.34)}$ & $2.60_{(5.28)}$ & $2.90_{(5.52)}$ & $1.97_{(3.77)}$ & $1.94_{(3.75)}$ & $1.93_{(3.74)}$ & $2.06_{(3.90)}$ \\
& & Eagle-3(60)$^{\textsc{a}}$ & $3.21_{(7.05)}$ & $3.31_{(6.91)}$ & $2.99_{(\ulm{6.36})}$ & $2.89_{(6.21)}$ & $2.75_{(6.11)}$ & $3.15_{(6.49)}$ & $2.14_{(\mathbf{4.45})}$ & $2.17_{(\mathbf{4.51})}$ & $2.13_{(\mathbf{4.48})}$ & $2.24_{(\mathbf{4.60})}$ \\
& & DFlash$^{\textsc{b}}$    & $5.07_{(8.05)}$ & $4.94_{(7.56)}$ & $3.89_{(6.00)}$ & $3.56_{(5.50)}$ & $3.63_{(5.88)}$ & $4.21_{(6.41)}$ & $2.43_{(3.71)}$ & $2.30_{(3.50)}$ & $2.14_{(3.27)}$ & $2.52_{(3.81)}$ \\
& & DFlare$^{\textsc{b}}$    & $5.15_{(8.17)}$ & $5.00_{(7.68)}$ & $3.92_{(6.07)}$ & $3.59_{(5.58)}$ & $3.70_{(5.96)}$ & $4.18_{(6.37)}$ & $2.45_{(3.75)}$ & $\ulm{2.36}_{(3.59)}$ & $\ulm{2.18}_{(3.33)}$ & $2.57_{(3.89)}$ \\
& & D-PACE$^{\textsc{b}}$     & $\ulm{5.30}_{(8.48)}$ & $\ulm{5.18}_{(7.97)}$ & $3.96_{(6.16)}$ & $3.67_{(5.71)}$ & $\ulm{3.86}_{(6.28)}$ & $\mathbf{4.53}_{(6.87)}$ & $\ulm{2.52}_{(3.85)}$ & $2.31_{(3.51)}$ & $2.12_{(3.25)}$ & $\ulm{2.58}_{(3.90)}$ \\
& & Domino$^{\textsc{s}}$    & $5.06_{(\mathbf{9.14})}$ & $4.95_{(\mathbf{8.59})}$ & $3.95_{(\mathbf{6.95})}$ & $3.66_{(\mathbf{6.48})}$ & $3.63_{(\ulm{6.65})}$ & $4.36_{(\mathbf{7.47})}$ & $2.43_{(\ulm{4.19})}$ & $2.34_{(\ulm{4.03})}$ & $\mathbf{2.23}_{(\ulm{3.85})}$ & $2.56_{(\ulm{4.39})}$ \\
& & DSpark$^{\textsc{s}}$    & $5.05_{(\ulm{9.07})}$ & $4.91_{(\ulm{8.51})}$ & $\ulm{3.97}_{(\mathbf{6.95})}$ & $\ulm{3.68}_{(\ulm{6.45})}$ & $3.73_{(\mathbf{6.81})}$ & $4.16_{(\ulm{7.11})}$ & $2.44_{(4.18)}$ & $2.32_{(3.98)}$ & $2.17_{(3.74)}$ & $2.55_{(4.34)}$ \\
& & DCore$^{\textsc{b}}$     & $\mathbf{5.44}_{(8.67)}$ & $\mathbf{5.32}_{(8.20)}$ & $\mathbf{4.08}_{(6.32)}$ & $\mathbf{3.80}_{(5.90)}$ & $\mathbf{3.95}_{(6.41)}$ & $\ulm{4.48}_{(6.83)}$ & $\mathbf{2.57}_{(3.93)}$ & $\mathbf{2.37}_{(3.60)}$ & $\ulm{2.18}_{(3.34)}$ & $\mathbf{2.66}_{(4.03)}$ \\
\cmidrule(l){2-13}
& \multirow{9}{*}{$1|0$} & \textit{Target only} & \multicolumn{10}{c}{$224.2$} \\
& & Eagle-3$^{\textsc{a}}$   & $2.84_{(5.91)}$ & $2.63_{(5.29)}$ & $2.56_{(5.21)}$ & $2.44_{(4.98)}$ & $2.26_{(4.74)}$ & $2.22_{(4.41)}$ & $1.65_{(\ulm{3.31})}$ & $1.75_{(3.51)}$ & $1.73_{(\ulm{3.51})}$ & $1.77_{(3.52)}$ \\
& & Eagle-3(60)$^{\textsc{a}}$ & $2.91_{(6.74)}$ & $2.76_{(\mathbf{6.13})}$ & $2.70_{(6.08)}$ & $2.59_{(\mathbf{5.91})}$ & $2.37_{(\ulm{5.51})}$ & $2.39_{(\mathbf{5.24})}$ & $1.78_{(\mathbf{3.93})}$ & $1.88_{(\mathbf{4.19})}$ & $1.87_{(\mathbf{4.20})}$ & $1.90_{(\mathbf{4.16})}$ \\
& & DFlash$^{\textsc{b}}$    & $4.38_{(6.83)}$ & $3.66_{(5.55)}$ & $3.48_{(5.33)}$ & $3.24_{(4.98)}$ & $3.08_{(4.91)}$ & $2.74_{(4.14)}$ & $1.99_{(3.02)}$ & $2.05_{(3.10)}$ & $1.99_{(3.01)}$ & $2.14_{(3.22)}$ \\
& & DFlare$^{\textsc{b}}$    & $4.41_{(6.91)}$ & $3.74_{(5.69)}$ & $3.51_{(5.38)}$ & $3.29_{(5.07)}$ & $3.12_{(4.97)}$ & $\ulm{2.82}_{(4.25)}$ & $2.02_{(3.06)}$ & $\ulm{2.11}_{(3.19)}$ & $2.01_{(3.06)}$ & $2.17_{(3.27)}$ \\
& & D-PACE$^{\textsc{b}}$     & $\ulm{4.52}_{(7.09)}$ & $\ulm{3.84}_{(5.83)}$ & $\ulm{3.61}_{(5.53)}$ & $\ulm{3.32}_{(5.11)}$ & $\ulm{3.26}_{(5.21)}$ & $\mathbf{2.91}_{(4.39)}$ & $\ulm{2.04}_{(3.10)}$ & $2.08_{(3.15)}$ & $1.97_{(2.99)}$ & $\ulm{2.18}_{(3.29)}$ \\
& & Domino$^{\textsc{s}}$    & $4.33_{(\mathbf{7.66})}$ & $3.54_{(6.08)}$ & $3.51_{(\ulm{6.10})}$ & $3.31_{(\ulm{5.77})}$ & $3.03_{(5.45)}$ & $2.67_{(\ulm{4.55})}$ & $1.92_{(3.28)}$ & $2.08_{(\ulm{3.56})}$ & $\ulm{2.02}_{(3.47)}$ & $2.12_{(\ulm{3.61})}$ \\
& & DSpark$^{\textsc{s}}$    & $4.36_{(\ulm{7.60})}$ & $3.63_{(\ulm{6.12})}$ & $3.58_{(\mathbf{6.11})}$ & $3.29_{(5.63)}$ & $3.14_{(\mathbf{5.58})}$ & $2.71_{(4.54)}$ & $1.96_{(3.30)}$ & $2.07_{(3.48)}$ & $2.00_{(3.38)}$ & $2.14_{(3.58)}$ \\
& & DCore$^{\textsc{b}}$     & $\mathbf{4.65}_{(7.28)}$ & $\mathbf{3.91}_{(5.94)}$ & $\mathbf{3.70}_{(5.69)}$ & $\mathbf{3.41}_{(5.25)}$ & $\mathbf{3.30}_{(5.28)}$ & $\mathbf{2.91}_{(4.38)}$ & $\mathbf{2.09}_{(3.16)}$ & $\mathbf{2.12}_{(3.22)}$ & $\mathbf{2.03}_{(3.08)}$ & $\mathbf{2.25}_{(3.39)}$ \\
\cmidrule(l){2-13}
& \multirow{7}{*}{$1|1$} & \textit{Target only} & \multicolumn{10}{c}{$224.2$} \\
& & DFlash$^{\textsc{b}}$    & $4.34_{(6.71)}$ & $3.71_{(5.58)}$ & $3.39_{(5.13)}$ & $2.97_{(4.49)}$ & $2.92_{(4.59)}$ & $2.87_{(4.28)}$ & $1.99_{(2.98)}$ & $1.97_{(2.96)}$ & $1.87_{(2.80)}$ & $2.06_{(3.07)}$ \\
& & DFlare$^{\textsc{b}}$    & $4.40_{(6.79)}$ & $3.76_{(5.65)}$ & $3.51_{(5.32)}$ & $3.06_{(4.64)}$ & $2.98_{(4.67)}$ & $2.86_{(4.26)}$ & $2.03_{(3.05)}$ & $2.02_{(3.02)}$ & $1.91_{(2.87)}$ & $2.12_{(3.16)}$ \\
& & D-PACE$^{\textsc{b}}$     & $4.69_{(7.27)}$ & $4.02_{(6.05)}$ & $3.62_{(5.49)}$ & $3.30_{(5.02)}$ & $\ulm{3.27}_{(5.17)}$ & $3.08_{(4.60)}$ & $2.12_{(3.19)}$ & $2.03_{(3.04)}$ & $1.94_{(2.92)}$ & $2.20_{(3.28)}$ \\
& & Domino$^{\textsc{s}}$    & $4.59_{(\ulm{8.14})}$ & $3.93_{(\ulm{6.75})}$ & $3.65_{(\ulm{6.34})}$ & $3.30_{(\ulm{5.75})}$ & $3.00_{(\ulm{5.39})}$ & $3.06_{(\ulm{5.22})}$ & $2.09_{(\ulm{3.58})}$ & $2.06_{(\ulm{3.54})}$ & $\ulm{2.00}_{(\ulm{3.44})}$ & $2.19_{(\ulm{3.73})}$ \\
& & DSpark$^{\textsc{s}}$    & $\ulm{4.75}_{(\mathbf{8.32})}$ & $\mathbf{4.17}_{(\mathbf{7.07})}$ & $\mathbf{3.79}_{(\mathbf{6.49})}$ & $\mathbf{3.45}_{(\mathbf{5.93})}$ & $3.26_{(\mathbf{5.81})}$ & $\mathbf{3.23}_{(\mathbf{5.43})}$ & $\mathbf{2.19}_{(\mathbf{3.71})}$ & $\mathbf{2.17}_{(\mathbf{3.66})}$ & $\mathbf{2.09}_{(\mathbf{3.54})}$ & $\mathbf{2.32}_{(\mathbf{3.90})}$ \\
& & DCore$^{\textsc{b}}$     & $\mathbf{4.78}_{(7.41)}$ & $\ulm{4.07}_{(6.12)}$ & $\ulm{3.75}_{(5.71)}$ & $\ulm{3.40}_{(5.17)}$ & $\mathbf{3.33}_{(5.26)}$ & $\ulm{3.16}_{(4.72)}$ & $\ulm{2.17}_{(3.25)}$ & $\ulm{2.11}_{(3.16)}$ & $1.99_{(2.99)}$ & $\ulm{2.27}_{(3.39)}$ \\
\midrule
\multirow{25}{*}{\rotatebox[origin=c]{90}{Qwen3-8B}}
& \multirow{9}{*}{$0|0$} & \textit{Target only} & \multicolumn{10}{c}{$151.0$} \\
& & Eagle-3$^{\textsc{a}}$   & $3.56_{(6.42)}$ & $3.62_{(6.30)}$ & $3.32_{(5.86)}$ & $3.13_{(5.55)}$ & $3.16_{(5.75)}$ & $3.36_{(5.80)}$ & $2.17_{(3.77)}$ & $2.28_{(3.97)}$ & $2.23_{(3.91)}$ & $2.34_{(4.03)}$ \\
& & Eagle-3(60)$^{\textsc{a}}$ & $3.69_{(7.12)}$ & $3.79_{(7.05)}$ & $3.51_{(6.61)}$ & $3.40_{(6.46)}$ & $3.28_{(6.42)}$ & $3.58_{(6.61)}$ & $2.37_{(\mathbf{4.39})}$ & $2.51_{(\mathbf{4.67})}$ & $\ulm{2.48}_{(\mathbf{4.66})}$ & $2.55_{(\mathbf{4.68})}$ \\
& & DFlash$^{\textsc{b}}$    & $5.39_{(8.04)}$ & $5.26_{(7.60)}$ & $4.27_{(6.21)}$ & $3.76_{(5.49)}$ & $4.29_{(6.54)}$ & $4.35_{(6.20)}$ & $2.48_{(3.57)}$ & $2.37_{(3.40)}$ & $2.24_{(3.23)}$ & $2.62_{(3.74)}$ \\
& & DFlare$^{\textsc{b}}$    & $5.58_{(8.36)}$ & $5.43_{(7.88)}$ & $4.49_{(6.54)}$ & $3.92_{(5.75)}$ & $4.45_{(6.80)}$ & $4.53_{(6.49)}$ & $2.60_{(3.73)}$ & $2.48_{(3.57)}$ & $2.34_{(3.39)}$ & $2.74_{(3.92)}$ \\
& & D-PACE$^{\textsc{b}}$     & $\ulm{5.94}_{(8.94)}$ & $\ulm{5.76}_{(8.35)}$ & $4.64_{(6.78)}$ & $4.07_{(5.95)}$ & $\ulm{4.72}_{(7.28)}$ & $4.56_{(6.79)}$ & $2.67_{(3.85)}$ & $2.45_{(3.52)}$ & $2.29_{(3.34)}$ & $2.79_{(4.01)}$ \\
& & Domino$^{\textsc{s}}$    & $5.82_{(\mathbf{9.57})}$ & $5.71_{(\mathbf{9.04})}$ & $\mathbf{4.77}_{(\mathbf{7.60})}$ & $\mathbf{4.18}_{(\mathbf{6.71})}$ & $4.54_{(\ulm{7.60})}$ & $\ulm{4.80}_{(\mathbf{7.49})}$ & $2.69_{(\ulm{4.21})}$ & $\mathbf{2.59}_{(\ulm{4.06})}$ & $\mathbf{2.49}_{(\ulm{3.92})}$ & $\ulm{2.86}_{(\ulm{4.46})}$ \\
& & DSpark$^{\textsc{s}}$    & $5.80_{(\ulm{9.49})}$ & $5.65_{(\ulm{8.92})}$ & $4.72_{(\ulm{7.50})}$ & $\mathbf{4.18}_{(\ulm{6.68})}$ & $4.56_{(\mathbf{7.61})}$ & $4.79_{(\ulm{7.44})}$ & $\ulm{2.70}_{(\ulm{4.21})}$ & $\ulm{2.56}_{(4.00)}$ & $2.46_{(3.87)}$ & $2.85_{(4.43)}$ \\
& & DCore$^{\textsc{b}}$     & $\mathbf{6.02}_{(9.05)}$ & $\mathbf{5.86}_{(8.50)}$ & $\ulm{4.75}_{(6.94)}$ & $\ulm{4.17}_{(6.11)}$ & $\mathbf{4.78}_{(7.36)}$ & $\mathbf{4.83}_{(6.89)}$ & $\mathbf{2.73}_{(3.92)}$ & $2.52_{(3.62)}$ & $2.37_{(3.43)}$ & $\mathbf{2.87}_{(4.10)}$ \\
\cmidrule(l){2-13}
& \multirow{9}{*}{$1|0$} & \textit{Target only} & \multicolumn{10}{c}{$148.5$} \\
& & Eagle-3$^{\textsc{a}}$   & $3.21_{(5.94)}$ & $3.00_{(5.39)}$ & $2.86_{(5.21)}$ & $2.77_{(5.06)}$ & $2.76_{(5.18)}$ & $2.60_{(\ulm{4.64})}$ & $1.76_{(\ulm{3.15})}$ & $1.99_{(\ulm{3.59})}$ & $1.96_{(\ulm{3.55})}$ & $1.92_{(\ulm{3.42})}$ \\
& & Eagle-3(60)$^{\textsc{a}}$ & $3.36_{(6.73)}$ & $3.22_{(6.27)}$ & $3.08_{(\mathbf{6.06})}$ & $3.00_{(\mathbf{5.95})}$ & $2.90_{(5.94)}$ & $2.78_{(\mathbf{5.37})}$ & $1.91_{(\mathbf{3.70})}$ & $\ulm{2.20}_{(\mathbf{4.28})}$ & $\mathbf{2.17}_{(\mathbf{4.25})}$ & $2.09_{(\mathbf{4.03})}$ \\
& & DFlash$^{\textsc{b}}$    & $4.56_{(6.71)}$ & $3.94_{(5.65)}$ & $3.56_{(5.13)}$ & $3.28_{(4.75)}$ & $3.59_{(5.42)}$ & $2.98_{(4.23)}$ & $1.95_{(2.79)}$ & $2.11_{(3.00)}$ & $2.03_{(2.90)}$ & $2.12_{(3.01)}$ \\
& & DFlare$^{\textsc{b}}$    & $4.65_{(6.88)}$ & $4.03_{(5.78)}$ & $3.68_{(5.31)}$ & $3.36_{(4.87)}$ & $3.69_{(5.58)}$ & $3.05_{(4.34)}$ & $2.01_{(2.87)}$ & $2.19_{(3.13)}$ & $2.10_{(3.01)}$ & $2.18_{(3.10)}$ \\
& & D-PACE$^{\textsc{b}}$     & $\ulm{4.92}_{(7.28)}$ & $\ulm{4.22}_{(6.06)}$ & $3.81_{(5.51)}$ & $\ulm{3.46}_{(5.02)}$ & $\ulm{3.87}_{(5.88)}$ & $\mathbf{3.19}_{(4.54)}$ & $\ulm{2.04}_{(2.92)}$ & $2.17_{(3.10)}$ & $2.06_{(2.96)}$ & $\ulm{2.22}_{(3.16)}$ \\
& & Domino$^{\textsc{s}}$    & $4.75_{(\mathbf{7.67})}$ & $4.06_{(\mathbf{6.34})}$ & $3.79_{(\ulm{5.99})}$ & $\mathbf{3.51}_{(\ulm{5.55})}$ & $3.85_{(\mathbf{6.37})}$ & $2.93_{(4.54)}$ & $1.96_{(3.07)}$ & $\ulm{2.20}_{(3.43)}$ & $\ulm{2.15}_{(3.37)}$ & $2.19_{(3.40)}$ \\
& & DSpark$^{\textsc{s}}$    & $4.78_{(\ulm{7.65})}$ & $4.06_{(\ulm{6.28})}$ & $\ulm{3.82}_{(5.96)}$ & $\mathbf{3.51}_{(5.51)}$ & $3.83_{(\ulm{6.27})}$ & $2.96_{(4.53)}$ & $2.00_{(3.08)}$ & $\mathbf{2.22}_{(3.42)}$ & $2.14_{(3.32)}$ & $2.20_{(3.38)}$ \\
& & DCore$^{\textsc{b}}$     & $\mathbf{4.98}_{(7.36)}$ & $\mathbf{4.24}_{(6.08)}$ & $\mathbf{3.85}_{(5.57)}$ & $\mathbf{3.51}_{(5.09)}$ & $\mathbf{4.04}_{(6.15)}$ & $\ulm{3.14}_{(4.46)}$ & $\mathbf{2.07}_{(2.96)}$ & $\mathbf{2.22}_{(3.17)}$ & $2.11_{(3.03)}$ & $\mathbf{2.25}_{(3.20)}$ \\
\cmidrule(l){2-13}
& \multirow{7}{*}{$1|1$} & \textit{Target only} & \multicolumn{10}{c}{$148.5$} \\
& & DFlash$^{\textsc{b}}$    & $4.21_{(6.16)}$ & $3.75_{(5.35)}$ & $3.39_{(4.87)}$ & $2.97_{(4.27)}$ & $3.14_{(4.68)}$ & $2.92_{(4.14)}$ & $1.98_{(2.81)}$ & $2.00_{(2.85)}$ & $1.91_{(2.72)}$ & $2.06_{(2.91)}$ \\
& & DFlare$^{\textsc{b}}$    & $4.56_{(6.69)}$ & $4.02_{(5.74)}$ & $3.66_{(5.27)}$ & $3.28_{(4.73)}$ & $3.47_{(5.20)}$ & $3.16_{(4.47)}$ & $2.11_{(3.00)}$ & $2.15_{(3.06)}$ & $2.03_{(2.90)}$ & $2.20_{(3.12)}$ \\
& & D-PACE$^{\textsc{b}}$     & $5.06_{(7.48)}$ & $4.46_{(6.37)}$ & $3.96_{(5.71)}$ & $3.54_{(5.11)}$ & $\ulm{3.82}_{(5.77)}$ & $3.44_{(4.88)}$ & $2.22_{(3.16)}$ & $2.21_{(3.15)}$ & $2.08_{(2.98)}$ & $2.32_{(3.29)}$ \\
& & Domino$^{\textsc{s}}$    & $\mathbf{5.41}_{(\mathbf{8.82})}$ & $\ulm{4.74}_{(\mathbf{7.46})}$ & $\ulm{4.22}_{(\ulm{6.70})}$ & $\ulm{3.81}_{(\ulm{6.07})}$ & $3.76_{(\ulm{6.23})}$ & $\mathbf{3.75}_{(\mathbf{5.85})}$ & $\ulm{2.31}_{(\ulm{3.63})}$ & $\ulm{2.34}_{(\ulm{3.67})}$ & $\ulm{2.25}_{(\ulm{3.55})}$ & $\ulm{2.42}_{(\ulm{3.78})}$ \\
& & DSpark$^{\textsc{s}}$    & $\ulm{5.40}_{(\ulm{8.73})}$ & $\mathbf{4.76}_{(\ulm{7.44})}$ & $\mathbf{4.27}_{(\mathbf{6.74})}$ & $\mathbf{3.91}_{(\mathbf{6.19})}$ & $\mathbf{3.91}_{(\mathbf{6.50})}$ & $\mathbf{3.75}_{(\ulm{5.81})}$ & $\mathbf{2.39}_{(\mathbf{3.73})}$ & $\mathbf{2.42}_{(\mathbf{3.79})}$ & $\mathbf{2.33}_{(\mathbf{3.64})}$ & $\mathbf{2.54}_{(\mathbf{3.94})}$ \\
& & DCore$^{\textsc{b}}$     & $5.16_{(7.63)}$ & $4.50_{(6.43)}$ & $4.02_{(5.80)}$ & $3.63_{(5.25)}$ & $\mathbf{3.91}_{(5.91)}$ & $\ulm{3.47}_{(4.92)}$ & $2.24_{(3.19)}$ & $2.26_{(3.22)}$ & $2.14_{(3.07)}$ & $2.37_{(3.35)}$ \\
\bottomrule
\end{tabular}%
}
\end{table*}

%% file: section/Appendix-concurrency.tex
\section{Speedup Under Concurrency}
\label{sec:appendix-concurrency}

\input{figure/Figure-concurrency.tex}

Fig.~\ref{fig:serving_tps_concurrency} compares seven drafters on
\textsc{SpecMix}, using both targets, budgets $8$ and $16$, and three
temperature pairs. Its third row also appears in
Fig.~\ref{fig:serving_tps_qwen3_8b_bgt8}.
Eagle-3 is chain-decoded throughout ($7/1/8$ or $15/1/16$).

\paragraph{Speedup definition and repeats.}
Each point divides mean request output throughput by the target-only mean
at the same target, temperature, and concurrency. Within a run, this metric
is $\sum_i n_i/\sum_i\ell_i$, using output tokens $n_i$ and request latencies
$\ell_i$. Curves use unprofiled evaluations: one run for greedy targets,
three for sampled settings.

\paragraph{The preferred budget depends on load.}
At $B{=}16$, greedy speedup for the six parallel and semi-AR recipes falls
from $3.08$--$3.27\times$ at $C{=}1$ to $1.30$--$1.43\times$ at $C{=}64$
on 4B, and from $3.31$--$3.63\times$ to $1.19$--$1.36\times$ on 8B.
At $B{=}8$, the corresponding high-load ranges are
$1.65$--$1.74\times$ and $1.67$--$1.78\times$.
Budget $16$ benefits most parallel recipes at $C{=}1$ under greedy decoding,
whereas budget $8$ is faster for every method at the sampled loads
$C\ge16$ in all twelve panels. Eagle-3's chain prefers budget $8$
throughout.

\paragraph{Sampling changes the acceptance--cost balance.}
On 4B at $B{=}16$, $C{=}64$, DSpark exceeds DFlash by
$9.8\%$ under $0|0$ and $9.2\%$ under $1|0$.
Under $1|1$, high-load speedups at $B{=}16$ span $1.09$--$1.34\times$ for
the six parallel and semi-AR recipes on 4B and $0.94$--$1.25\times$ on 8B.
DSpark has the highest speedup at every concurrency in all four $1|1$
panels.

App.~\ref{sec:appendix-bgt60} reports the separate chain/tree comparison.

%% file: figure/Figure-concurrency.tex
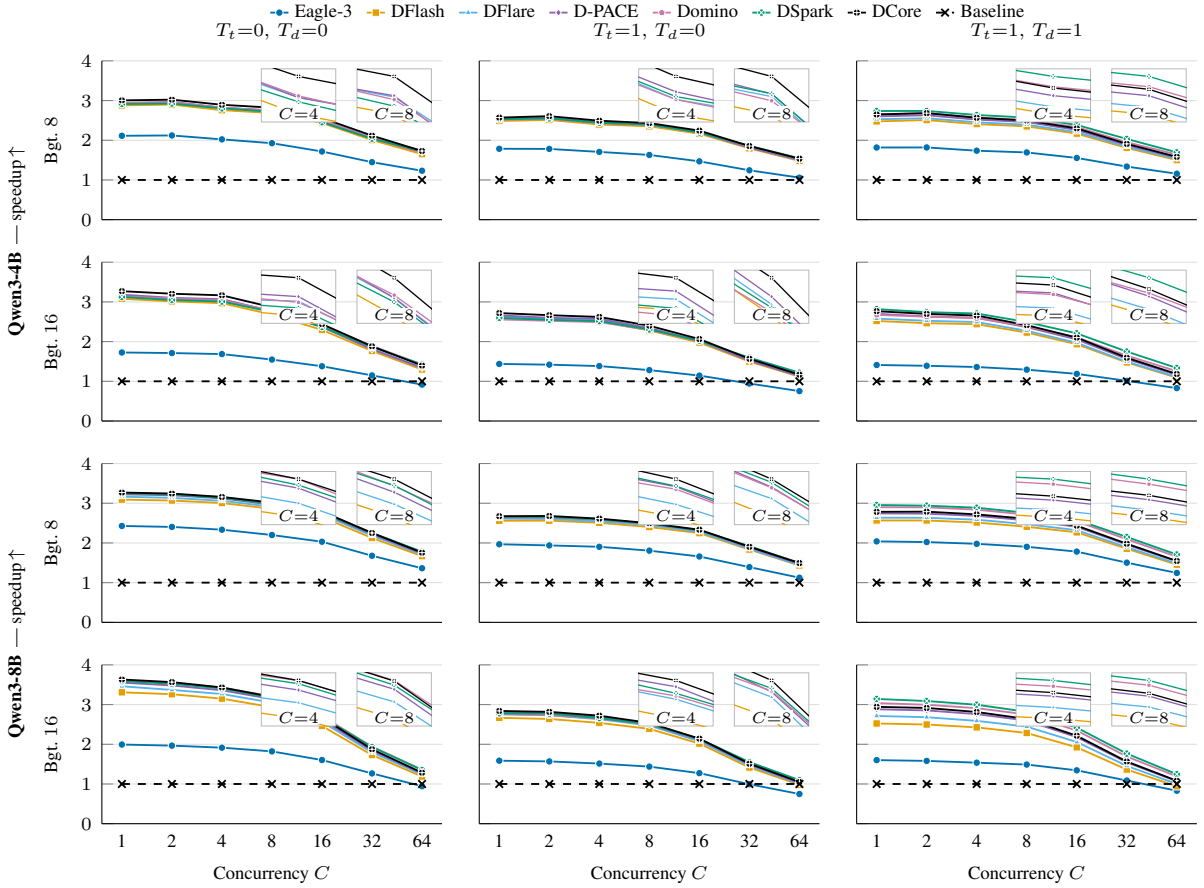
\begin{figure*}[!t]
    \centering
    \def\panelw{128pt}
    \pgfplotsset{
        spline/.style={
            color=#1, line width=0.7pt, mark size=1.4pt,
            mark options={fill=#1, draw=white, line width=0.3pt},
        },
        tpspanel/.style={
            width=\panelw, height=2.1cm, scale only axis,
            xmin=0.6, xmax=7.4,
            xtick={1,2,3,4,5,6,7},
            xticklabels={1,2,4,8,16,32,64},
            ymin=0, ymax=4,
            ytick={0,1,2,3,4},
            yticklabels={$0$,$1$,$2$,$3$,$4$},
            scaled y ticks=false,
            ymajorgrids=true, grid style={gray!25},
            xtick style={draw=none},
            axis x line*=bottom, axis y line*=left,
            xlabel={Concurrency $C$},
            xticklabel style={font=\scriptsize},
            yticklabel style={font=\scriptsize},
            xlabel style={font=\scriptsize},
            ylabel style={font=\scriptsize},
            title style={font=\scriptsize, yshift=-2pt},
        },
        zoombox/.style={
            width=28pt, height=20pt, scale only axis,
            enlargelimits=false, clip=true,
            xtick=\empty, ytick=\empty,
            axis line style={gray!55, line width=0.3pt},
            axis background/.style={fill=white},
            xlabel style={font=\tiny, at={(axis description cs:0.5,0.025)},
                          anchor=south, fill=white, inner sep=0.5pt},
            spline/.style={
                color=##1, line width=0.5pt, mark size=1.0pt,
                mark options={fill=##1, draw=white, line width=0.2pt},
            },
        },
    }
    \def\sixIa{%
        \addplot[spline=cDFlash, mark=square*   ] coordinates {(1,2.878261) (2,2.899314) (3,2.757683) (4,2.686177) (5,2.436860) (6,2.005299) (7,1.650238)};
        \addplot[spline=cDFlare, mark=triangle* ] coordinates {(1,2.913043) (2,2.951945) (3,2.817967) (4,2.747821) (5,2.504778) (6,2.065009) (7,1.719856)};
        \addplot[spline=cDPace,  mark=diamond*  ] coordinates {(1,2.934783) (2,2.965675) (3,2.817967) (4,2.752179) (5,2.465529) (6,2.044426) (7,1.698108)};
        \addplot[spline=cDomino, mark=pentagon* ] coordinates {(1,2.939130) (2,2.961098) (3,2.823877) (4,2.738481) (5,2.502730) (6,2.073976) (7,1.736713)};
        \addplot[spline=cDSpark, mark=otimes*   ] coordinates {(1,2.913043) (2,2.924485) (3,2.802600) (4,2.715442) (5,2.455973) (6,2.040758) (7,1.729250)};
        \addplot[spline=cDCore,  mark=oplus*    ] coordinates {(1,3.004348) (2,3.022883) (3,2.893617) (4,2.820672) (5,2.560068) (6,2.117180) (7,1.729378)};
    }
    \def\sixIb{%
        \addplot[spline=cDFlash, mark=square*   ] coordinates {(1,2.486607) (2,2.510539) (3,2.397094) (4,2.351213) (5,2.164905) (6,1.795863) (7,1.480128)};
        \addplot[spline=cDFlare, mark=triangle* ] coordinates {(1,2.526786) (2,2.548009) (3,2.434625) (4,2.389527) (5,2.193094) (6,1.815921) (7,1.497466)};
        \addplot[spline=cDPace,  mark=diamond*  ] coordinates {(1,2.540179) (2,2.559719) (3,2.453995) (4,2.395275) (5,2.200141) (6,1.814250) (7,1.508536)};
        \addplot[spline=cDomino, mark=pentagon* ] coordinates {(1,2.517857) (2,2.538642) (3,2.434625) (4,2.380587) (5,2.206836) (6,1.841621) (7,1.539611)};
        \addplot[spline=cDSpark, mark=otimes*   ] coordinates {(1,2.531250) (2,2.550351) (3,2.441889) (4,2.395913) (5,2.196265) (6,1.860217) (7,1.546679)};
        \addplot[spline=cDCore,  mark=oplus*    ] coordinates {(1,2.571429) (2,2.606557) (3,2.492736) (4,2.432950) (5,2.243129) (6,1.854576) (7,1.536410)};
    }
    \def\sixIc{%
        \addplot[spline=cDFlash, mark=square*   ] coordinates {(1,2.475003) (2,2.508536) (3,2.407349) (4,2.350038) (5,2.167364) (6,1.808048) (7,1.500414)};
        \addplot[spline=cDFlare, mark=triangle* ] coordinates {(1,2.528430) (2,2.554849) (3,2.452315) (4,2.391781) (5,2.205629) (6,1.839019) (7,1.528182)};
        \addplot[spline=cDPace,  mark=diamond*  ] coordinates {(1,2.601826) (2,2.629999) (3,2.520448) (4,2.459122) (5,2.260305) (6,1.880653) (7,1.561986)};
        \addplot[spline=cDomino, mark=pentagon* ] coordinates {(1,2.654948) (2,2.685185) (3,2.574053) (4,2.511659) (5,2.330818) (6,1.950072) (7,1.640545)};
        \addplot[spline=cDSpark, mark=otimes*   ] coordinates {(1,2.737210) (2,2.739101) (3,2.636784) (4,2.571989) (5,2.386249) (6,2.033089) (7,1.695385)};
        \addplot[spline=cDCore,  mark=oplus*    ] coordinates {(1,2.649118) (2,2.682799) (3,2.566740) (4,2.496495) (5,2.298489) (6,1.909148) (7,1.582873)};
    }
    \def\sixIIa{%
        \addplot[spline=cDFlash, mark=square*   ] coordinates {(1,3.078261) (2,3.013730) (3,2.968085) (4,2.682441) (5,2.293515) (6,1.764011) (7,1.297516)};
        \addplot[spline=cDFlare, mark=triangle* ] coordinates {(1,3.147826) (2,3.059497) (3,3.043735) (4,2.768991) (5,2.376451) (6,1.813328) (7,1.335221)};
        \addplot[spline=cDPace,  mark=diamond*  ] coordinates {(1,3.182609) (2,3.102975) (3,3.067376) (4,2.767123) (5,2.366553) (6,1.808845) (7,1.342298)};
        \addplot[spline=cDomino, mark=pentagon* ] coordinates {(1,3.134783) (2,3.089245) (3,3.037825) (4,2.781445) (5,2.413652) (6,1.892399) (7,1.421181)};
        \addplot[spline=cDSpark, mark=otimes*   ] coordinates {(1,3.121739) (2,3.048055) (3,3.008274) (4,2.748443) (5,2.368601) (6,1.864887) (7,1.425042)};
        \addplot[spline=cDCore,  mark=oplus*    ] coordinates {(1,3.269565) (2,3.205950) (3,3.165485) (4,2.864882) (5,2.441297) (6,1.880986) (7,1.396474)};
    }
    \def\sixIIb{%
        \addplot[spline=cDFlash, mark=square*   ] coordinates {(1,2.593750) (2,2.536300) (3,2.509685) (4,2.281609) (5,1.976039) (6,1.496657) (7,1.115231)};
        \addplot[spline=cDFlare, mark=triangle* ] coordinates {(1,2.633929) (2,2.569087) (3,2.547215) (4,2.316731) (5,2.005285) (6,1.526536) (7,1.129368)};
        \addplot[spline=cDPace,  mark=diamond*  ] coordinates {(1,2.660714) (2,2.599532) (3,2.575061) (4,2.340996) (5,2.020437) (6,1.518596) (7,1.138704)};
        \addplot[spline=cDomino, mark=pentagon* ] coordinates {(1,2.558036) (2,2.524590) (3,2.490315) (4,2.293103) (5,2.013742) (6,1.576055) (7,1.192585)};
        \addplot[spline=cDSpark, mark=otimes*   ] coordinates {(1,2.598214) (2,2.548009) (3,2.515738) (4,2.306513) (5,2.022199) (6,1.594651) (7,1.217391)};
        \addplot[spline=cDCore,  mark=oplus*    ] coordinates {(1,2.718750) (2,2.665105) (3,2.621065) (4,2.399745) (5,2.064130) (6,1.567280) (7,1.167378)};
    }
    \def\sixIIc{%
        \addplot[spline=cDFlash, mark=square*   ] coordinates {(1,2.518885) (2,2.463606) (3,2.438598) (4,2.228240) (5,1.933959) (6,1.475408) (7,1.092462)};
        \addplot[spline=cDFlare, mark=triangle* ] coordinates {(1,2.578756) (2,2.523832) (3,2.492560) (4,2.272299) (5,1.969668) (6,1.494180) (7,1.109985)};
        \addplot[spline=cDPace,  mark=diamond*  ] coordinates {(1,2.699786) (2,2.634212) (3,2.600866) (4,2.364211) (5,2.054700) (6,1.562084) (7,1.160214)};
        \addplot[spline=cDomino, mark=pentagon* ] coordinates {(1,2.673623) (2,2.628005) (3,2.589549) (4,2.385690) (5,2.108377) (6,1.644930) (7,1.255256)};
        \addplot[spline=cDSpark, mark=otimes*   ] coordinates {(1,2.819587) (2,2.742195) (3,2.708375) (4,2.492468) (5,2.207920) (6,1.748994) (7,1.336076)};
        \addplot[spline=cDCore,  mark=oplus*    ] coordinates {(1,2.765919) (2,2.697964) (3,2.656471) (4,2.413940) (5,2.100102) (6,1.593090) (7,1.184412)};
    }
    \def\sixIIIa{%
        \addplot[spline=cDFlash, mark=square*   ] coordinates {(1,3.092105) (2,3.067797) (3,3.005217) (4,2.859982) (5,2.650387) (6,2.125445) (7,1.667499)};
        \addplot[spline=cDFlare, mark=triangle* ] coordinates {(1,3.164474) (2,3.135593) (3,3.060870) (4,2.913279) (5,2.715184) (6,2.185155) (7,1.707959)};
        \addplot[spline=cDPace,  mark=diamond*  ] coordinates {(1,3.223684) (2,3.200000) (3,3.121739) (4,2.962963) (5,2.749033) (6,2.222679) (7,1.732268)};
        \addplot[spline=cDomino, mark=pentagon* ] coordinates {(1,3.256579) (2,3.227119) (3,3.158261) (4,2.988257) (5,2.788685) (6,2.253355) (7,1.768731)};
        \addplot[spline=cDSpark, mark=otimes*   ] coordinates {(1,3.250000) (2,3.223729) (3,3.133913) (4,2.990967) (5,2.765957) (6,2.250616) (7,1.778555)};
        \addplot[spline=cDCore,  mark=oplus*    ] coordinates {(1,3.269737) (2,3.244068) (3,3.158261) (4,3.017164) (5,2.792070) (6,2.250616) (7,1.755744)};
    }
    \def\sixIIIb{%
        \addplot[spline=cDFlash, mark=square*   ] coordinates {(1,2.560000) (2,2.562069) (3,2.507067) (4,2.400735) (5,2.247291) (6,1.833937) (7,1.432682)};
        \addplot[spline=cDFlare, mark=triangle* ] coordinates {(1,2.613333) (2,2.603448) (3,2.542403) (4,2.447610) (5,2.280788) (6,1.850098) (7,1.441761)};
        \addplot[spline=cDPace,  mark=diamond*  ] coordinates {(1,2.646667) (2,2.648276) (3,2.591873) (4,2.477022) (5,2.314778) (6,1.884648) (7,1.471223)};
        \addplot[spline=cDomino, mark=pentagon* ] coordinates {(1,2.646667) (2,2.634483) (3,2.584806) (4,2.475184) (5,2.318719) (6,1.904987) (7,1.500685)};
        \addplot[spline=cDSpark, mark=otimes*   ] coordinates {(1,2.660000) (2,2.655172) (3,2.593640) (4,2.488971) (5,2.318227) (6,1.919755) (7,1.487324)};
        \addplot[spline=cDCore,  mark=oplus*    ] coordinates {(1,2.673333) (2,2.679310) (3,2.613074) (4,2.497243) (5,2.333990) (6,1.903594) (7,1.493320)};
    }
    \def\sixIIIc{%
        \addplot[spline=cDFlash, mark=square*   ] coordinates {(1,2.566606) (2,2.565911) (3,2.513740) (4,2.407918) (5,2.267362) (6,1.854632) (7,1.452865)};
        \addplot[spline=cDFlare, mark=triangle* ] coordinates {(1,2.635834) (2,2.633570) (3,2.585501) (4,2.472132) (5,2.321274) (6,1.891454) (7,1.480854)};
        \addplot[spline=cDPace,  mark=diamond*  ] coordinates {(1,2.743425) (2,2.737427) (3,2.680773) (4,2.567938) (5,2.407649) (6,1.958402) (7,1.535534)};
        \addplot[spline=cDomino, mark=pentagon* ] coordinates {(1,2.900670) (2,2.896664) (3,2.837875) (4,2.711322) (5,2.546688) (6,2.092169) (7,1.653725)};
        \addplot[spline=cDSpark, mark=otimes*   ] coordinates {(1,2.955873) (2,2.939388) (3,2.887431) (4,2.758074) (5,2.582901) (6,2.150884) (7,1.708007)};
        \addplot[spline=cDCore,  mark=oplus*    ] coordinates {(1,2.783793) (2,2.788019) (3,2.719414) (4,2.608385) (5,2.435138) (6,1.980793) (7,1.545672)};
    }
    \def\sixIVa{%
        \addplot[spline=cDFlash, mark=square*   ] coordinates {(1,3.309211) (2,3.261017) (3,3.149565) (4,2.934056) (5,2.461315) (6,1.729389) (7,1.187646)};
        \addplot[spline=cDFlare, mark=triangle* ] coordinates {(1,3.460526) (2,3.369492) (3,3.266087) (4,3.049684) (5,2.564313) (6,1.796494) (7,1.237096)};
        \addplot[spline=cDPace,  mark=diamond*  ] coordinates {(1,3.552632) (2,3.481356) (3,3.360000) (4,3.138211) (5,2.618956) (6,1.844426) (7,1.266400)};
        \addplot[spline=cDomino, mark=pentagon* ] coordinates {(1,3.605263) (2,3.542373) (3,3.429565) (4,3.199639) (5,2.709381) (6,1.941386) (7,1.359973)};
        \addplot[spline=cDSpark, mark=otimes*   ] coordinates {(1,3.592105) (2,3.525424) (3,3.406957) (4,3.167118) (5,2.680368) (6,1.932347) (7,1.351482)};
        \addplot[spline=cDCore,  mark=oplus*    ] coordinates {(1,3.631579) (2,3.566102) (3,3.431304) (4,3.195122) (5,2.672147) (6,1.874281) (7,1.285215)};
    }
    \def\sixIVb{%
        \addplot[spline=cDFlash, mark=square*   ] coordinates {(1,2.666667) (2,2.637931) (3,2.538869) (4,2.388787) (5,2.017734) (6,1.414043) (7,0.980987)};
        \addplot[spline=cDFlare, mark=triangle* ] coordinates {(1,2.753333) (2,2.731034) (3,2.630742) (4,2.472426) (5,2.073892) (6,1.461410) (7,1.006680)};
        \addplot[spline=cDPace,  mark=diamond*  ] coordinates {(1,2.806667) (2,2.779310) (3,2.690813) (4,2.492647) (5,2.116256) (6,1.492338) (7,1.029633)};
        \addplot[spline=cDomino, mark=pentagon* ] coordinates {(1,2.766667) (2,2.731034) (3,2.646643) (4,2.496324) (5,2.134483) (6,1.528838) (7,1.076225)};
        \addplot[spline=cDSpark, mark=otimes*   ] coordinates {(1,2.780000) (2,2.772414) (3,2.659011) (4,2.506434) (5,2.137931) (6,1.551964) (7,1.097294)};
        \addplot[spline=cDCore,  mark=oplus*    ] coordinates {(1,2.840000) (2,2.817241) (3,2.720848) (4,2.537684) (5,2.141872) (6,1.510727) (7,1.048476)};
    }
    \def\sixIVc{%
        \addplot[spline=cDFlash, mark=square*   ] coordinates {(1,2.524282) (2,2.497586) (3,2.426318) (4,2.284234) (5,1.923867) (6,1.363650) (7,0.942967)};
        \addplot[spline=cDFlare, mark=triangle* ] coordinates {(1,2.713992) (2,2.681766) (3,2.589807) (4,2.444084) (5,2.057561) (6,1.455614) (7,0.999638)};
        \addplot[spline=cDPace,  mark=diamond*  ] coordinates {(1,2.882654) (2,2.852774) (3,2.760126) (4,2.596525) (5,2.179976) (6,1.544001) (7,1.067911)};
        \addplot[spline=cDomino, mark=pentagon* ] coordinates {(1,3.037873) (2,2.995762) (3,2.909330) (4,2.751140) (5,2.346090) (6,1.691246) (7,1.191120)};
        \addplot[spline=cDSpark, mark=otimes*   ] coordinates {(1,3.142308) (2,3.089750) (3,2.996145) (4,2.817625) (5,2.419597) (6,1.764886) (7,1.246226)};
        \addplot[spline=cDCore,  mark=oplus*    ] coordinates {(1,2.942946) (2,2.915091) (3,2.813221) (4,2.636761) (5,2.215144) (6,1.566569) (7,1.082251)};
    }
    \def\tobase#1{%
        \addplot[black, dashed, line width=0.7pt, mark=x, mark size=2.0pt,
                 mark options={solid}] coordinates {#1};
    }
    \def\bfourZ{(1,1.000000) (2,1.000000) (3,1.000000) (4,1.000000) (5,1.000000) (6,1.000000) (7,1.000000)}
    \def\bfourS{(1,1.000000) (2,1.000000) (3,1.000000) (4,1.000000) (5,1.000000) (6,1.000000) (7,1.000000)}
    \def\beightZ{(1,1.000000) (2,1.000000) (3,1.000000) (4,1.000000) (5,1.000000) (6,1.000000) (7,1.000000)}
    \def\beightS{(1,1.000000) (2,1.000000) (3,1.000000) (4,1.000000) (5,1.000000) (6,1.000000) (7,1.000000)}
    \begin{tikzpicture}
    \begin{groupplot}[
        tpspanel,
        group style={
            group size=3 by 4,
            group name=tps,
            horizontal sep=14pt, vertical sep=16pt,
            xlabels at=edge bottom, xticklabels at=edge bottom,
            ylabels at=edge left,  yticklabels at=edge left,
        },
    ]
    \nextgroupplot[
        title={$T_t{=}0,\,T_d{=}0$},
        ylabel={Bgt.\ $8$},
        legend entries={Eagle-3, DFlash, DFlare, D-PACE, Domino, DSpark,
                        DCore, Baseline},
        legend style={
            at={(1.61,1.18)}, anchor=south,
            draw=none, fill=none, legend columns=8, font=\scriptsize,
            /tikz/every even column/.append style={column sep=3pt},
        },
        legend image code/.code={
            \draw[mark repeat=2, mark phase=2, #1]
                plot coordinates {(0cm,0cm) (0.12cm,0cm) (0.24cm,0cm)};
        },
    ]
    \addplot[spline=cEagle,  mark=*] coordinates {(1,2.113043) (2,2.121281) (3,2.022459) (4,1.925903) (5,1.717406) (6,1.448135) (7,1.232531)};
    \sixIa
    \tobase{\bfourZ}
    \nextgroupplot[title={$T_t{=}1,\,T_d{=}0$}]
    \addplot[spline=cEagle,  mark=*] coordinates {(1,1.785714) (2,1.782201) (3,1.707022) (4,1.630268) (5,1.469345) (6,1.243627) (7,1.058549)};
    \sixIb
    \tobase{\bfourS}
    \nextgroupplot[title={$T_t{=}1,\,T_d{=}1$}]
    \addplot[spline=cEagle,  mark=*] coordinates {(1,1.817341) (2,1.819237) (3,1.737016) (4,1.693238) (5,1.554861) (6,1.337409) (7,1.157170)};
    \sixIc
    \tobase{\bfourS}
    \nextgroupplot[ylabel={Bgt.\ $16$}]
    \addplot[spline=cEagle,  mark=*] coordinates {(1,1.726087) (2,1.711670) (3,1.685579) (4,1.547323) (5,1.379522) (6,1.148563) (7,0.916742)};
    \sixIIa
    \tobase{\bfourZ}
    \nextgroupplot
    \addplot[spline=cEagle,  mark=*] coordinates {(1,1.437500) (2,1.419204) (3,1.383777) (4,1.282248) (5,1.144116) (6,0.944212) (7,0.753134)};
    \sixIIb
    \tobase{\bfourS}
    \nextgroupplot
    \addplot[spline=cEagle,  mark=*] coordinates {(1,1.410815) (2,1.391100) (3,1.359758) (4,1.293296) (5,1.187389) (6,1.011494) (7,0.828739)};
    \sixIIc
    \tobase{\bfourS}
    \nextgroupplot[ylabel={Bgt.\ $8$}]
    \addplot[spline=cEagle,  mark=*] coordinates {(1,2.427632) (2,2.403390) (3,2.333913) (4,2.199639) (5,2.030948) (6,1.677349) (7,1.362304)};
    \sixIIIa
    \tobase{\beightZ}
    \nextgroupplot
    \addplot[spline=cEagle,  mark=*] coordinates {(1,1.966667) (2,1.937931) (3,1.902827) (4,1.806066) (5,1.658621) (6,1.391195) (7,1.124700)};
    \sixIIIb
    \tobase{\beightS}
    \nextgroupplot
    \addplot[spline=cEagle,  mark=*] coordinates {(1,2.039928) (2,2.022204) (3,1.977984) (4,1.903790) (5,1.781782) (6,1.503337) (7,1.244524)};
    \sixIIIc
    \tobase{\beightS}
    \nextgroupplot[ylabel={Bgt.\ $16$}]
    \addplot[spline=cEagle,  mark=*] coordinates {(1,1.993421) (2,1.966102) (3,1.914783) (4,1.821138) (5,1.603482) (6,1.266228) (7,0.943556)};
    \sixIVa
    \tobase{\beightZ}
    \nextgroupplot
    \addplot[spline=cEagle,  mark=*] coordinates {(1,1.586667) (2,1.568966) (3,1.514134) (4,1.437500) (5,1.272906) (6,0.994149) (7,0.748030)};
    \sixIVb
    \tobase{\beightS}
    \nextgroupplot
    \addplot[spline=cEagle,  mark=*]
        coordinates {(1,1.601275) (2,1.581393) (3,1.536831) (4,1.489066) (5,1.344129) (6,1.087003) (7,0.830567)};
    \sixIVc
    \tobase{\beightS}
    \end{groupplot}
    \node[rotate=90, anchor=center, font=\scriptsize]
        at ([xshift=-32pt, yshift=-8pt] tps c1r1.south west)
        {\textbf{Qwen3-4B}\, --- speedup\,$\uparrow$};
    \node[rotate=90, anchor=center, font=\scriptsize]
        at ([xshift=-32pt, yshift=-8pt] tps c1r3.south west)
        {\textbf{Qwen3-8B}\, --- speedup\,$\uparrow$};
    \def\zm#1#2#3#4#5#6#7#8#9{%
        \begin{axis}[zoombox, at={(#1.north east)}, anchor=north east,
                     xshift=#2, yshift=-3pt,
                     xmin=#3, xmax=#4, ymin=#5, ymax=#6,
                     xlabel={$C{=}#7$}]
        #9
        \end{axis}%
    }
    \zm{tps c1r1}{-40pt}{2.650}{3.350}{2.730496}{2.920804}{4}{0.19}{\sixIa}
    \zm{tps c1r1}{-4pt}{3.650}{4.350}{2.659278}{2.847572}{8}{0.19}{\sixIa}
    \zm{tps c2r1}{-40pt}{2.650}{3.350}{2.377966}{2.511864}{4}{0.13}{\sixIb}
    \zm{tps c2r1}{-4pt}{3.650}{4.350}{2.334866}{2.449298}{8}{0.11}{\sixIb}
    \zm{tps c3r1}{-40pt}{2.650}{3.350}{2.361462}{2.682671}{4}{0.32}{\sixIc}
    \zm{tps c3r1}{-4pt}{3.650}{4.350}{2.305648}{2.616379}{8}{0.31}{\sixIc}
    \zm{tps c1r2}{-40pt}{2.650}{3.350}{2.928605}{3.204965}{4}{0.28}{\sixIIa}
    \zm{tps c1r2}{-4pt}{3.650}{4.350}{2.645953}{2.901370}{8}{0.26}{\sixIIa}
    \zm{tps c2r2}{-40pt}{2.650}{3.350}{2.464165}{2.647215}{4}{0.18}{\sixIIb}
    \zm{tps c2r2}{-4pt}{3.650}{4.350}{2.257982}{2.423372}{8}{0.17}{\sixIIb}
    \zm{tps c3r2}{-40pt}{2.650}{3.350}{2.384642}{2.762331}{4}{0.38}{\sixIIc}
    \zm{tps c3r2}{-4pt}{3.650}{4.350}{2.175394}{2.545313}{8}{0.37}{\sixIIc}
    \zm{tps c1r3}{-40pt}{2.650}{3.350}{2.974609}{3.188870}{4}{0.21}{\sixIIIa}
    \zm{tps c1r3}{-4pt}{3.650}{4.350}{2.828546}{3.048600}{8}{0.22}{\sixIIIa}
    \zm{tps c2r3}{-40pt}{2.650}{3.350}{2.485866}{2.634276}{4}{0.15}{\sixIIIb}
    \zm{tps c2r3}{-4pt}{3.650}{4.350}{2.381434}{2.516544}{8}{0.14}{\sixIIIb}
    \zm{tps c3r3}{-40pt}{2.650}{3.350}{2.439002}{2.962169}{4}{0.52}{\sixIIIc}
    \zm{tps c3r3}{-4pt}{3.650}{4.350}{2.337886}{2.828105}{8}{0.49}{\sixIIIc}
    \zm{tps c1r4}{-40pt}{2.650}{3.350}{3.093217}{3.487652}{4}{0.39}{\sixIVa}
    \zm{tps c1r4}{-4pt}{3.650}{4.350}{2.880939}{3.252755}{8}{0.37}{\sixIVa}
    \zm{tps c2r4}{-40pt}{2.650}{3.350}{2.502473}{2.757244}{4}{0.25}{\sixIVb}
    \zm{tps c2r4}{-4pt}{3.650}{4.350}{2.359007}{2.567463}{8}{0.21}{\sixIVb}
    \zm{tps c3r4}{-40pt}{2.650}{3.350}{2.312353}{3.110110}{4}{0.80}{\sixIVc}
    \zm{tps c3r4}{-4pt}{3.650}{4.350}{2.177555}{2.924303}{8}{0.75}{\sixIVc}
    \end{tikzpicture}
    \caption{\textbf{Speedup under concurrency.} Request output throughput
    normalized to the target-only baseline at the same target, temperature,
    and concurrency; dashed is $1\times$. Rows show Qwen3-4B at budgets
    $8$, $16$, then Qwen3-8B at $8$, $16$; columns show temperature pairs.
    $C$ is per-replica concurrency and doubles at each tick. All panels share
    a speedup axis. Insets use local $y$ ranges to resolve method ordering at $C=4$ and $8$. Metric and
    repeat provenance are in App.~\ref{sec:appendix-concurrency}.}
    \label{fig:serving_tps_concurrency}
\end{figure*}

%% file: section/Appendix-sampled-drafts.tex
\section{Sampled Drafts and Prefix-Independent Proposals}
\label{sec:appendix-sampled-drafts}

\renewcommand{\dbltopfraction}{0.9}
\input{table/Table-sampling-gap}
\input{figure/Figure-sampling-position}

\subsection{A prefix-mismatch bound}
\label{sec:app-sd-theory}

Fix a context and draft positions $j=0,\dots,K-1$, $K{=}B{-}1$. Let
$P_j(\cdot\mid y)$ be the target's $T{=}1$ next-token distribution after
the drafted prefix $y=y_{<j}$, and $\nu_j$ the distribution of $y_{<j}$
given that $j$ is reached. A block-parallel drafter draws $y_j\sim q_j$
independently of $y_{<j}$; since reaching $j$ depends only on $y_{<j}$ and
its verification draws, Eq.~\eqref{eq:appendix-acceptance-tv} gives its
$T_d{=}1$ acceptance after $y$ as
$\alpha_j(y)=1-\operatorname{TV}(P_j(\cdot\mid y),q_j)$. For independent
$Y,Y'\sim\nu_j$, taking expectations in
$\operatorname{TV}(P_j(\cdot\mid Y),P_j(\cdot\mid Y'))\le
\operatorname{TV}(P_j(\cdot\mid Y),q_j)+\operatorname{TV}(q_j,P_j(\cdot\mid Y'))$
yields
\begin{equation}
\mathbb E_{Y\sim\nu_j}\bigl[1-\alpha_j(Y)\bigr]\ge\tfrac12 D_j,
\label{eq:app-sd-bound}
\end{equation}
where
$D_j=\mathbb E_{Y,Y'}\operatorname{TV}(P_j(\cdot\mid Y),P_j(\cdot\mid Y'))$
measures how much the target varies across prefixes that reach $j$. It
binds every prefix-independent $q_j$, including loss-reweighted ones such as
D-PACE \citep{wu2026dpacedynamicpositionawarecrossentropy}, but not semi-AR
or AR drafters, whose $q_j$ can follow $P_j(\cdot\mid y)$. Under
$T_d{=}0$, $\nu_j$ is a point mass on the greedy path and $D_j{=}0$.

At $j{=}0$ both settings share the prefix, and with $\hat x=\arg\max q$,
$\alpha^{1|1}-\alpha^{1|0}=\sum_{x\ne\hat x}\min\{p(x),q(x)\}
-(p(\hat x)-q(\hat x))_+$, which can take either sign. If the target picks
$y_0\in\{a,b\}$ w.p.\ $(.6,.4)$ and then repeats it, the
prefix-independent optimum $q_j=(.6,.4)$ gives $\tau^{1|0}=1+.6K$ but
$\tau^{1|1}<3.17$ for every $K$, whereas a semi-AR drafter attains $1+K$.

\subsection{Measurement}
\label{sec:app-sd-measure}

We log per-token draft and target log-probabilities on Qwen3-8B at
$T_t{=}1$ (top-$p$ $1$, no top-$k$) under $1|0$ and $1|1$, both budgets,
three repeats: $27$ runs and $3.89\times10^7$ verification chains after
dropping warmup and temperature-mismatched chains ($5.5\%$). Each draft
token is sampled from the $q$ passed to verification, unlike upstream SGLang
DFlash's argmax proposal \citep{wu2026dpal}. At each reached position, acceptance is the Rao--Blackwellized
$\min\{1,p(\hat x)/q(\hat x)\}$ under $1|1$ and $p(\hat x)$ under $1|0$;
per-run means match realized acceptance within $8.9\times10^{-4}$.
$\mathrm{KL}(q\Vert p)$ is the mean of $\log q(\hat x)-\log p(\hat x)$ over
drafted tokens. Deeper positions are reached mostly on easy tokens, so rates
over a position set $\mathcal J$ are entropy-standardized,
\begin{equation}
\bar\alpha^{\mathrm{std}}=\frac{\sum_{j\in\mathcal J}\sum_{b}w_{jb}\,\hat\alpha_{jb}}
{\sum_{j\in\mathcal J}\sum_{b}w_{jb}},
\label{eq:app-sd-stdz}
\end{equation}
where $\hat\alpha_{jb}$ is the mean acceptance at $j$ in target-entropy bin
$b$ (edges $0.05,0.2,0.5,1,1.5,2,3$ nats) and $w_{jb}$ counts reached
tokens in that cell, pooled over drafters under $1|0$. At $j\ge1$ the
predecessor (previous accepted token) is argmax (A) if it is the target's
argmax and non-argmax (N) otherwise; $f$ is the raw N share. With
$S_k=\Pr[\text{at least }k\text{ drafts accepted}]$, so that
$\tau=1+\sum_{k=1}^{K}S_k$, and realized $a_j=S_{j+1}/S_j$,
\begin{equation}
\tau=1+a_0G,\quad
G=1+\sum_{k=2}^{K}\prod_{j=1}^{k-1}a_j,
\label{eq:app-sd-tau}
\end{equation}
so $\Delta\tau$ splits exactly into $\Delta a_0\,\bar G$ ($j{=}0$) and
$\bar a_0\,\Delta G$ ($j\ge1$), bars averaging the two settings. Likewise
the $j\ge1$ mean $\bar\alpha=(1-f)\alpha_{\mathrm{A}}+f\alpha_{\mathrm{N}}$
splits as $(1-\bar f)\Delta\alpha_{\mathrm{A}}+\bar f\Delta\alpha_{\mathrm{N}}
+\Delta f\,\bar g$ with $g=\alpha_{\mathrm{N}}-\alpha_{\mathrm{A}}$: the
argmax, non-argmax, and share terms. $95\%$ intervals use $1000$ bootstrap
replicates over $256$ prompt clusters, shared across drafters and settings
so contrasts are paired; half-widths are $\le0.024$
($\tau$ terms), $0.006$ ($\Delta\alpha_j$), $0.005$ (predecessor terms), and
$0.10$ nats (KL).

\subsection{Per-method results}
\label{sec:app-sd-results}

\paragraph{Position and prefix state.}
Non-AR recipes gain $0.056$--$0.071$ at $j{=}0$; block-parallel
$\Delta\alpha_j$ turns negative from $j{=}1$ or $2$
(Tab.~\ref{tab:sampling_gap}, Fig.~\ref{fig:sampling_position}), and the
$j\ge1$ term accounts for $76$--$94\%$ of each semi-AR versus
block-parallel difference in $\Delta\tau$. Block-parallel versus semi-AR, $g$ is
$-0.38$ to $-0.43$ versus $-0.08$ to $-0.09$ under $1|1$ and $-0.14$ to
$-0.20$ versus $-0.04$ to $-0.05$ under $1|0$, where $f\approx2.5\%$
rather than $7$--$9\%$. Non-argmax predecessors raise
$\mathrm{KL}(q\Vert p)$ by $9.0$--$10.1$ vs.\ $0.9$--$1.0$ nats
(cf.\ Eq.~\eqref{eq:app-sd-bound}).

\paragraph{Position-weighted training.}
D-PACE's $1|1$ gain at $j\ge1$ over DFlash is $77\%$ argmax term (DCore over
DFlare: $74$--$76\%$), while N acceptance reaches only $0.40$--$0.41$:
$q$ sharpens on the greedy path, but the mismatch remains (Domino over
DFlash: non-argmax term $32$--$36\%$). Eagle-3 is semi-AR-like in $g$ and $\Delta$KL,
but its $j{=}0$ KL is $2.53$ vs.\ $0.77$--$0.95$ nats and
$\Delta\alpha_0$ only $0.035$: prefix conditioning is necessary, not
sufficient.

\renewcommand{\dbltopfraction}{0.7}

%% file: table/Table-sampling-gap.tex
\begin{table*}[t]
\centering
\footnotesize
\setlength{\tabcolsep}{4.2pt}
\caption{\textbf{Sampled-draft decomposition on Qwen3-8B} ($T_t{=}1$). $\tau$ columns and the Shapley terms (Eq.~\eqref{eq:app-sd-tau})
use realized chains; $\Delta\alpha_0$ and the prefix-state columns are
entropy-standardized (Eq.~\eqref{eq:app-sd-stdz}) and pool $j\ge1$.
A/N: $1|1$ acceptance after an argmax or non-argmax predecessor; gap is
N minus A. $f$: raw N share at $j\ge1$ (\%). $\Delta$KL: N minus A
$\mathrm{KL}(q\Vert p)$ under $1|1$, in nats.
Eagle-3 has no $B{=}16$, $1|1$ dump.}
\label{tab:sampling_gap}
\begin{tabular}{@{}l ccc cc c ccc c cc c@{}}
\toprule
& \multicolumn{3}{c}{Committed tokens}
& \multicolumn{2}{c}{Shapley terms}
& & \multicolumn{3}{c}{$1|1$, $j\ge1$}
& $1|0$ & \multicolumn{2}{c}{$f$ (\%)} & \\
\cmidrule(lr){2-4}\cmidrule(lr){5-6}\cmidrule(lr){8-10}\cmidrule(lr){11-11}\cmidrule(lr){12-13}
\textbf{Drafter} & $\tau^{1|0}$ & $\tau^{1|1}$ & $\Delta\tau$
& $j{=}0$ & $j{\ge}1$ & $\Delta\alpha_0$
& A & N & gap & gap & $1|0$ & $1|1$ & $\Delta$KL \\
\midrule
\multicolumn{14}{@{}l}{\textit{Budget $B{=}8$}} \\
Eagle-3 & $3.18$ & $3.27$ & $0.095$ & $0.138$ & ${-}0.043$ & $0.035$ & $0.738$ & $0.643$ & ${-}0.095$ & ${-}0.043$ & $2.9$ & $9.6$ & $1.16$ \\
DFlash & $3.57$ & $3.60$ & $0.026$ & $0.246$ & ${-}0.220$ & $0.064$ & $0.758$ & $0.333$ & ${-}0.425$ & ${-}0.196$ & $2.5$ & $8.7$ & $9.94$ \\
DFlare & $3.63$ & $3.71$ & $0.076$ & $0.271$ & ${-}0.195$ & $0.070$ & $0.768$ & $0.337$ & ${-}0.430$ & ${-}0.198$ & $2.5$ & $8.5$ & $10.08$ \\
D-PACE & $3.69$ & $3.86$ & $0.170$ & $0.248$ & ${-}0.078$ & $0.058$ & $0.796$ & $0.396$ & ${-}0.400$ & ${-}0.159$ & $2.5$ & $7.3$ & $9.38$ \\
DCore & $3.72$ & $3.92$ & $0.199$ & $0.257$ & ${-}0.058$ & $0.060$ & $0.801$ & $0.396$ & ${-}0.405$ & ${-}0.166$ & $2.5$ & $7.2$ & $9.48$ \\
Domino & $3.84$ & $4.27$ & $0.436$ & $0.295$ & $0.141$ & $0.061$ & $0.827$ & $0.739$ & ${-}0.087$ & ${-}0.039$ & $2.6$ & $8.5$ & $0.99$ \\
DSpark & $3.84$ & $4.38$ & $0.534$ & $0.332$ & $0.202$ & $0.070$ & $0.834$ & $0.755$ & ${-}0.079$ & ${-}0.047$ & $2.6$ & $8.1$ & $0.96$ \\
\midrule
\multicolumn{14}{@{}l}{\textit{Budget $B{=}16$}} \\
DFlash & $3.76$ & $3.59$ & ${-}0.171$ & $0.244$ & ${-}0.415$ & $0.056$ & $0.742$ & $0.324$ & ${-}0.418$ & ${-}0.196$ & $2.4$ & $8.6$ & $9.98$ \\
DFlare & $3.89$ & $3.88$ & ${-}0.014$ & $0.312$ & ${-}0.326$ & $0.070$ & $0.762$ & $0.343$ & ${-}0.418$ & ${-}0.192$ & $2.4$ & $8.2$ & $9.99$ \\
D-PACE & $3.97$ & $4.13$ & $0.162$ & $0.287$ & ${-}0.124$ & $0.057$ & $0.792$ & $0.411$ & ${-}0.381$ & ${-}0.144$ & $2.5$ & $6.9$ & $9.01$ \\
DCore & $4.03$ & $4.19$ & $0.161$ & $0.299$ & ${-}0.139$ & $0.058$ & $0.795$ & $0.413$ & ${-}0.382$ & ${-}0.143$ & $2.4$ & $6.9$ & $9.00$ \\
Domino & $4.26$ & $4.78$ & $0.526$ & $0.356$ & $0.170$ & $0.058$ & $0.822$ & $0.740$ & ${-}0.082$ & ${-}0.038$ & $2.6$ & $8.3$ & $0.90$ \\
DSpark & $4.26$ & $4.92$ & $0.657$ & $0.407$ & $0.250$ & $0.071$ & $0.824$ & $0.748$ & ${-}0.076$ & ${-}0.051$ & $2.5$ & $7.7$ & $0.91$ \\
\bottomrule
\end{tabular}
\end{table*}

%% file: figure/Figure-sampling-position.tex
\begin{figure*}[t]
    \centering
    \pgfplotsset{
        spline/.style={
            color=#1, line width=0.7pt, mark size=1.4pt,
            mark options={fill=#1, draw=white, line width=0.3pt},
        },
        sppanel/.style={
            height=2.2cm, scale only axis,
            ymajorgrids=true, grid style={gray!25},
            axis x line*=bottom, axis y line*=left,
            xticklabel style={font=\scriptsize},
            yticklabel style={font=\scriptsize},
            xlabel style={font=\scriptsize},
            ylabel style={font=\scriptsize},
            title style={font=\scriptsize, yshift=-2pt},
            scaled y ticks=false,
            yticklabel style={font=\scriptsize,
                /pgf/number format/fixed, /pgf/number format/precision=2},
            extra y ticks={0}, extra y tick style={grid=major,
                grid style={black!60, line width=0.4pt}, tick label style={opacity=0}},
        },
        posrow/.style={ymin=-0.06, ymax=0.08, ytick={-0.04,0,0.04,0.08}},
        survrow/.style={ymin=-0.04, ymax=0.12, ytick={-0.04,0,0.04,0.08,0.12}},
    }
    \begin{tikzpicture}
    \begin{groupplot}[
        sppanel,
        group style={group size=2 by 2, group name=sp,
                     horizontal sep=30pt, vertical sep=30pt},
    ]
    \nextgroupplot[posrow, width=120pt, xmin=-0.4, xmax=6.4,
        xtick={0,...,6}, xlabel={Draft position $j$},
        ylabel={$\Delta\alpha_j$}, title={$B{=}8$},
        legend entries={Eagle-3, DFlash, DFlare, D-PACE, DCore, Domino, DSpark},
        legend style={at={(1.27,1)}, anchor=south, yshift=10pt,
            draw=none, fill=none, legend columns=7, font=\scriptsize,
            /tikz/every even column/.append style={column sep=3pt}},
        legend image code/.code={
            \draw[mark repeat=2, mark phase=2, #1]
                plot coordinates {(0cm,0cm) (0.12cm,0cm) (0.24cm,0cm)};
        },
    ]
    \addplot[spline=cEagle, mark=*] coordinates {(0,0.0352) (1,0.0107) (2,0.0001) (3,-0.0072) (4,-0.0116) (5,-0.0165) (6,-0.0175)};
    \addplot[spline=cDFlash, mark=square*] coordinates {(0,0.0643) (1,-0.0123) (2,-0.0387) (3,-0.0437) (4,-0.0382) (5,-0.0331) (6,-0.0289)};
    \addplot[spline=cDFlare, mark=triangle*] coordinates {(0,0.0695) (1,-0.0061) (2,-0.0345) (3,-0.0407) (4,-0.0352) (5,-0.0325) (6,-0.0303)};
    \addplot[spline=cDPace, mark=diamond*] coordinates {(0,0.0583) (1,0.0059) (2,-0.0109) (3,-0.0149) (4,-0.0132) (5,-0.0115) (6,-0.0114)};
    \addplot[spline=cDCore, mark=oplus*] coordinates {(0,0.0598) (1,0.0089) (2,-0.0091) (3,-0.0119) (4,-0.0116) (5,-0.0110) (6,-0.0100)};
    \addplot[spline=cDomino, mark=pentagon*] coordinates {(0,0.0609) (1,0.0388) (2,0.0246) (3,0.0141) (4,0.0075) (5,0.0013) (6,-0.0006)};
    \addplot[spline=cDSpark, mark=otimes*] coordinates {(0,0.0700) (1,0.0478) (2,0.0310) (3,0.0207) (4,0.0137) (5,0.0081) (6,0.0052)};
    \nextgroupplot[posrow, width=240pt, xmin=-0.4, xmax=14.4,
        xtick={0,2,...,14}, xlabel={Draft position $j$}, title={$B{=}16$}]
    \addplot[spline=cDFlash, mark=square*] coordinates {(0,0.0560) (1,-0.0295) (2,-0.0529) (3,-0.0537) (4,-0.0444) (5,-0.0380) (6,-0.0323) (7,-0.0285) (8,-0.0229) (9,-0.0215) (10,-0.0196) (11,-0.0141) (12,-0.0178) (13,-0.0197) (14,-0.0146)};
    \addplot[spline=cDFlare, mark=triangle*] coordinates {(0,0.0702) (1,-0.0068) (2,-0.0369) (3,-0.0432) (4,-0.0373) (5,-0.0329) (6,-0.0275) (7,-0.0247) (8,-0.0226) (9,-0.0184) (10,-0.0164) (11,-0.0186) (12,-0.0163) (13,-0.0182) (14,-0.0192)};
    \addplot[spline=cDPace, mark=diamond*] coordinates {(0,0.0566) (1,0.0037) (2,-0.0123) (3,-0.0140) (4,-0.0104) (5,-0.0077) (6,-0.0041) (7,-0.0035) (8,-0.0022) (9,-0.0028) (10,-0.0032) (11,-0.0034) (12,-0.0036) (13,-0.0076) (14,-0.0089)};
    \addplot[spline=cDCore, mark=oplus*] coordinates {(0,0.0580) (1,0.0045) (2,-0.0125) (3,-0.0151) (4,-0.0110) (5,-0.0078) (6,-0.0042) (7,-0.0026) (8,-0.0035) (9,-0.0032) (10,-0.0034) (11,-0.0042) (12,-0.0036) (13,-0.0059) (14,-0.0068)};
    \addplot[spline=cDomino, mark=pentagon*] coordinates {(0,0.0582) (1,0.0385) (2,0.0235) (3,0.0152) (4,0.0093) (5,0.0053) (6,0.0041) (7,0.0016) (8,-0.0004) (9,-0.0030) (10,-0.0018) (11,-0.0037) (12,-0.0040) (13,-0.0058) (14,-0.0035)};
    \addplot[spline=cDSpark, mark=otimes*] coordinates {(0,0.0706) (1,0.0481) (2,0.0317) (3,0.0209) (4,0.0142) (5,0.0087) (6,0.0054) (7,0.0039) (8,0.0022) (9,0.0037) (10,0.0013) (11,-0.0025) (12,-0.0015) (13,-0.0018) (14,-0.0003)};
    \nextgroupplot[survrow, width=120pt, xmin=0.6, xmax=7.4,
        xtick={1,...,7}, xlabel={Accepted drafts $k$},
        ylabel={$\Delta S_k$}]
    \addplot[spline=cEagle, mark=*] coordinates {(1,0.0425) (2,0.0354) (3,0.0214) (4,0.0089) (5,0.0005) (6,-0.0052) (7,-0.0081)};
    \addplot[spline=cDFlash, mark=square*] coordinates {(1,0.0732) (2,0.0412) (3,0.0049) (4,-0.0155) (5,-0.0244) (6,-0.0270) (7,-0.0264)};
    \addplot[spline=cDFlare, mark=triangle*] coordinates {(1,0.0793) (2,0.0517) (3,0.0151) (4,-0.0075) (5,-0.0178) (6,-0.0219) (7,-0.0228)};
    \addplot[spline=cDPace, mark=diamond*] coordinates {(1,0.0686) (2,0.0532) (3,0.0303) (4,0.0144) (5,0.0054) (6,0.0006) (7,-0.0025)};
    \addplot[spline=cDCore, mark=oplus*] coordinates {(1,0.0702) (2,0.0575) (3,0.0352) (4,0.0198) (5,0.0103) (6,0.0046) (7,0.0011)};
    \addplot[spline=cDomino, mark=pentagon*] coordinates {(1,0.0747) (2,0.0894) (3,0.0815) (4,0.0671) (5,0.0526) (6,0.0401) (7,0.0301)};
    \addplot[spline=cDSpark, mark=otimes*] coordinates {(1,0.0834) (2,0.1032) (3,0.0971) (4,0.0832) (5,0.0683) (6,0.0551) (7,0.0439)};
    \nextgroupplot[survrow, width=240pt, xmin=0.6, xmax=15.4,
        xtick={1,3,...,15}, xlabel={Accepted drafts $k$}]
    \addplot[spline=cDFlash, mark=square*] coordinates {(1,0.0669) (2,0.0235) (3,-0.0112) (4,-0.0263) (5,-0.0300) (6,-0.0298) (7,-0.0277) (8,-0.0251) (9,-0.0224) (10,-0.0199) (11,-0.0176) (12,-0.0154) (13,-0.0135) (14,-0.0119) (15,-0.0102)};
    \addplot[spline=cDFlare, mark=triangle*] coordinates {(1,0.0814) (2,0.0519) (3,0.0147) (4,-0.0061) (5,-0.0146) (6,-0.0179) (7,-0.0181) (8,-0.0175) (9,-0.0166) (10,-0.0152) (11,-0.0138) (12,-0.0125) (13,-0.0111) (14,-0.0100) (15,-0.0089)};
    \addplot[spline=cDPace, mark=diamond*] coordinates {(1,0.0682) (2,0.0495) (3,0.0264) (4,0.0127) (5,0.0060) (6,0.0028) (7,0.0014) (8,0.0006) (9,0.0001) (10,-0.0001) (11,-0.0004) (12,-0.0008) (13,-0.0010) (14,-0.0014) (15,-0.0017)};
    \addplot[spline=cDCore, mark=oplus*] coordinates {(1,0.0705) (2,0.0514) (3,0.0272) (4,0.0123) (5,0.0054) (6,0.0022) (7,0.0009) (8,0.0002) (9,-0.0003) (10,-0.0008) (11,-0.0011) (12,-0.0015) (13,-0.0018) (14,-0.0019) (15,-0.0022)};
    \addplot[spline=cDomino, mark=pentagon*] coordinates {(1,0.0753) (2,0.0881) (3,0.0786) (4,0.0645) (5,0.0511) (6,0.0402) (7,0.0317) (8,0.0251) (9,0.0196) (10,0.0151) (11,0.0118) (12,0.0091) (13,0.0070) (14,0.0052) (15,0.0038)};
    \addplot[spline=cDSpark, mark=otimes*] coordinates {(1,0.0859) (2,0.1037) (3,0.0953) (4,0.0794) (5,0.0641) (6,0.0512) (7,0.0408) (8,0.0325) (9,0.0258) (10,0.0211) (11,0.0170) (12,0.0136) (13,0.0109) (14,0.0087) (15,0.0070)};
    \end{groupplot}
    \end{tikzpicture}
    \caption{\textbf{Per-position change from $1|0$ to $1|1$ (Qwen3-8B).}
    Top: acceptance change $\Delta\alpha_j$ at each reached draft position,
    standardized to a common target-entropy mix
    (Eq.~\eqref{eq:app-sd-stdz}). Bottom: change in the survival probability
    $\Delta S_k=\Delta\Pr[\text{at least }k\text{ drafts accepted}]$; each
    curve sums over $k$ to that method's $\Delta\tau$ (Eq.~\eqref{eq:app-sd-tau}).
    Every recipe gains at $j{=}0$; block-parallel curves turn negative from
    $j{=}1$ or $2$, semi-AR curves stay positive through $j{=}5$ ($B{=}8$)
    and $j{=}7$ ($B{=}16$).
    Eagle-3 has no $B{=}16$ sampled-draft dump.}
    \label{fig:sampling_position}
\end{figure*}
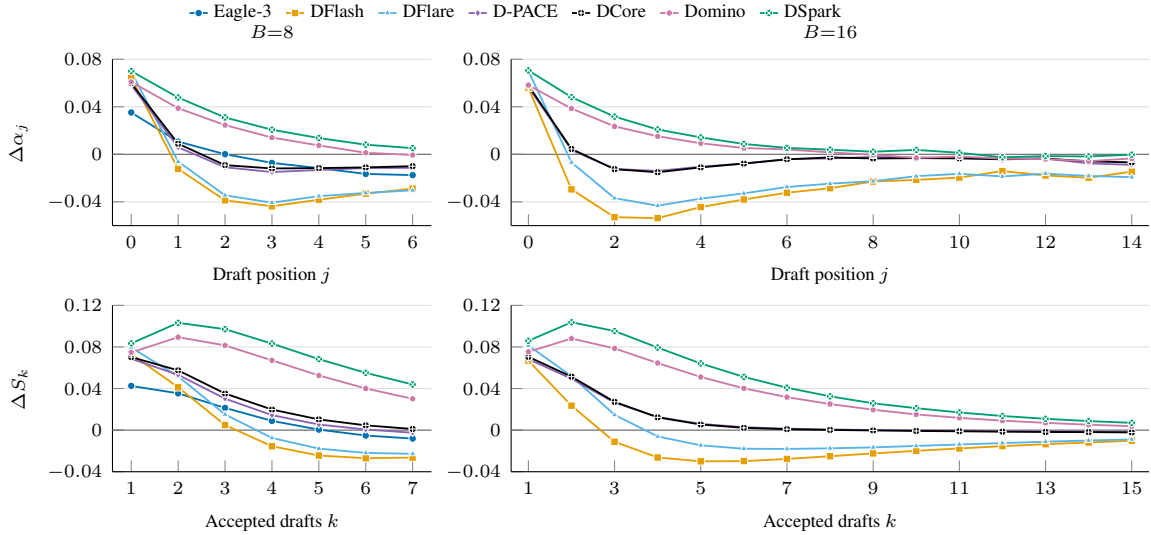

%% file: section/Appendix-loss-ablation.tex
\section{Loss Ablation Details}
\label{sec:appendix-loss-ablation}

Fig.~\ref{fig:loss-interaction} sweeps two orthogonal axes of the drafter
objective: the token-level term (CE,
KL, and TV) and the per-position weight $w_k$ ($0.8^{k}$, $e^{-k/\gamma}$, or
D-PACE). All runs use Qwen3-8B at $\textsc{Bgt.}{=}16$, $T{=}1$ with rejection
sampling, draft temperature $0$. The paired comparisons below hold one axis fixed to
isolate the other, and quote mean acceptance length $\tau$ averaged over all
ten benchmarks; the figure resolves the same configurations onto the two
entropy halves.

\paragraph{Ablated terms.}
For target distribution $p_k$, draft distribution $q_k$, and regenerated
token $x_k^\star$ at position $k$, the sweep uses
\begin{equation}
\begin{aligned}
\ell_{\mathrm{CE},k} &= -\log q_k(x_k^\star),\\
\ell_{\mathrm{KL},k} &= \mathrm{KL}(p_k\|q_k),\\
\ell_{\mathrm{TV},k} &= \lVert p_k-q_k\rVert_1.
\end{aligned}
\label{eq:ablation-terms}
\end{equation}
The TV term follows DSpark's L1 convention (twice total variation).
Position weights multiply the selected token-level terms before summing
over offsets; D-PACE weights follow \citet{wu2026dpacedynamicpositionawarecrossentropy}.

\input{figure/Figure-loss-accuracy.tex}
\paragraph{Token-level objectives.}
In paired comparisons with $e^{-k/\gamma}$ weights, replacing CE by KL adds
$0.135$ on DFlash but only $0.024$ on DSpark, so KL offers no clear advantage
over CE as target supervision. Adding TV on top of KL adds a further $0.094$
and $0.119$.

\paragraph{Position weights expose the interaction.}
With CE fixed, D-PACE adds $0.267$ of average $\tau$ on DFlash but only $0.046$
on DSpark. Its effect depends on the token-level objective: adding D-PACE
to KL$+$TV changes average $\tau$ by only $+0.013$ and $-0.146$.
The D-PACE gain over fixed-decay CE is concentrated in the low-entropy half:
$+7.9\%$ versus $+3.3\%$ on DFlash, and $+1.7\%$ versus $-0.3\%$ on DSpark.

%% file: figure/Figure-loss-accuracy.tex
\begin{figure}[t]
\input{figure/Figure-loss-ablation.tex}
\par\vspace{10pt}
\input{table/Table-accuracy.tex}
\end{figure}

%% file: figure/Figure-loss-ablation.tex
\colorlet{cLIEagle}{cEagle}
\colorlet{cLIDFlash}{cDFlash}
\colorlet{cLIDSpark}{cDSpark}

\newcommand{\lossIline}[1]{\tikz[baseline=-0.6ex]{\draw[#1, line width=1.1pt] (0,0) -- (3.0mm,0);}}
\newcommand{\lossIsched}[3]{\tikz[baseline=-0.6ex]{%
    \draw[black!60, #1, line width=0.8pt] (0,0) -- (4.0mm,0);
    \draw[black!60] plot[mark=#2, mark size=#3, mark options={fill=black!60, solid}] coordinates {(2.0mm,0)};}}

\captionsetup{hypcap=false}
\centering
\mbox{\tiny
 \lossIline{cLIEagle}\,Eagle-3\hspace{5pt}
 \lossIline{cLIDFlash}\,DFlash\hspace{5pt}
 \lossIline{cLIDSpark}\,DSpark\hspace{8pt}
 \lossIsched{densely dotted}{triangle*}{1.7pt}\,$0.8^{k}$\hspace{5pt}
 \lossIsched{solid}{diamond*}{1.7pt}\,$e^{-k/\gamma}$\hspace{5pt}
 \lossIsched{dashed}{*}{1.4pt}\,D-PACE}\\[5pt]
\begin{tikzpicture}
\pgfplotsset{
    lossmerged/.style={
        scale only axis=true,
        width=2.9cm, height=100pt,
        xmin=0.45, xmax=4.55,
        xtick={1,2,3,4},
        xticklabels={CE, KL, CE+TV, KL+TV},
        axis x line*=bottom,
        axis line style={black!55, thin},
        tick align=outside,
        ticklabel style={font=\tiny, color=black!75},
        ymajorgrids=true, grid style={black!11, thin},
        yticklabel style={/pgf/number format/fixed,
            /pgf/number format/precision=1, /pgf/number format/zerofill},
        every axis plot/.append style={line width=0.8pt, mark size=1.5pt},
        clip=false,
    },
}

\begin{axis}[lossmerged,
    name=lossL,
    axis y line*=left,
    ymin=5.1, ymax=6.5,
    ytick={5.2,5.4,5.6,5.8,6.0,6.2,6.4}]

\addplot[color=cLIEagle, densely dotted, mark=triangle*, mark size=1.9pt,
    mark options={solid}] coordinates {(0.86,5.319) (1.86,5.376) (2.86,5.296) (3.86,5.355)};
\addplot[color=cLIEagle, mark=*, mark size=1.6pt, only marks]
    coordinates {(1.14,5.218)};
\addplot[color=cLIDFlash, mark=diamond*, mark size=1.9pt]
    coordinates {(1,5.487) (2,5.618) (3,5.802) (4,5.815)};
\addplot[color=cLIDFlash, dashed, mark=*, mark size=1.6pt,
    mark options={solid}] coordinates {(1.14,5.919) (2.14,5.886) (3.14,5.795) (4.14,5.875)};
\addplot[color=cLIDSpark, mark=diamond*, mark size=1.9pt]
    coordinates {(1,6.160) (2,6.180) (3,6.279) (4,6.394)};
\addplot[color=cLIDSpark, dashed, mark=*, mark size=1.6pt,
    mark options={solid}] coordinates {(1.14,6.263) (2.14,6.331) (3.14,6.234) (4.14,6.263)};

\node[draw=black!65, thin, circle, inner sep=2.0pt] at (axis cs:1.86,5.376) {};
\node[draw=black!65, thin, circle, inner sep=2.0pt] at (axis cs:1,5.487) {};
\node[draw=black!65, thin, circle, inner sep=2.0pt] at (axis cs:3,6.279) {};
\end{axis}

\begin{axis}[lossmerged,
    name=lossH,
    at={(lossL.north east)}, anchor=north west, xshift=0.38cm,
    axis y line*=right,
    ytick pos=right,
    ymin=3.05, ymax=3.75,
    ytick={3.1,3.2,3.3,3.4,3.5,3.6,3.7}]

\addplot[color=cLIEagle, densely dotted, mark=triangle*, mark size=1.9pt,
    mark options={solid}] coordinates {(0.86,3.668) (1.86,3.646) (2.86,3.608) (3.86,3.617)};
\addplot[color=cLIEagle, mark=*, mark size=1.6pt, only marks]
    coordinates {(1.14,3.568)};
\addplot[color=cLIDFlash, mark=diamond*, mark size=1.9pt]
    coordinates {(1,3.126) (2,3.267) (3,3.230) (4,3.258)};
\addplot[color=cLIDFlash, dashed, mark=*, mark size=1.6pt,
    mark options={solid}] coordinates {(1.14,3.230) (2.14,3.327) (3.14,3.212) (4.14,3.222)};
\addplot[color=cLIDSpark, mark=diamond*, mark size=1.9pt]
    coordinates {(1,3.502) (2,3.531) (3,3.543) (4,3.554)};
\addplot[color=cLIDSpark, dashed, mark=*, mark size=1.6pt,
    mark options={solid}] coordinates {(1.14,3.491) (2.14,3.520) (3.14,3.415) (4.14,3.394)};

\node[draw=black!65, thin, circle, inner sep=2.0pt] at (axis cs:1.86,3.646) {};
\node[draw=black!65, thin, circle, inner sep=2.0pt] at (axis cs:1,3.126) {};
\node[draw=black!65, thin, circle, inner sep=2.0pt] at (axis cs:3,3.543) {};
\end{axis}

\node[font=\scriptsize, color=black!75, anchor=south, inner sep=1pt]
    at ([yshift=2pt]lossL.north west) {$\tau$};
\node[font=\scriptsize, color=black!75, anchor=south, inner sep=1pt]
    at ([yshift=2pt]lossH.north east) {$\tau$};
\node[font=\scriptsize\scshape, anchor=south, inner sep=1pt]
    at ([yshift=2pt]lossL.north) {Low-Entropy};
\node[font=\scriptsize\scshape, anchor=south, inner sep=1pt]
    at ([yshift=2pt]lossH.north) {High-Entropy};

\end{tikzpicture}
\captionof{figure}{\textbf{Training objective ablation.} Mean $\tau$ over each entropy half for every loss
configuration, on one drafter from each family; rings mark each drafter's
shipped configuration.}
\label{fig:loss-interaction}

%% file: table/Table-accuracy.tex
\captionsetup{hypcap=false}
\centering
\small
\setlength{\tabcolsep}{2.2pt}
\captionof{table}{\textbf{Answer accuracy (Qwen3-8B).} Exact-match accuracy (\%) on GSM8K (GSM; 1,319 examples) and MATH-500 (MATH; 500) at each temperature pair $T_t|T_d$ and verification budget.}
\label{tab:accuracy_qwen3_8b}
{\footnotesize
\begin{tabular}{c l cc cc cc}
\toprule
 & & \multicolumn{2}{c}{$0|0$} & \multicolumn{2}{c}{$1|0$} & \multicolumn{2}{c}{$1|1$} \\
\cmidrule(lr){3-4}\cmidrule(lr){5-6}\cmidrule(lr){7-8}
Bgt. & Method & GSM & MATH & GSM & MATH & GSM & MATH \\
\midrule
-- & Baseline & 84.3 & 59.4 & 85.4 & 58.2 & 85.4 & 58.2 \\
\midrule
\multirow{7}{*}{8}
 & Eagle-3 & 84.8 & 59.8 & 85.4 & 59.0 & 84.7 & 57.2 \\
 & DFlash  & 84.4 & 59.4 & 84.4 & 56.2 & 85.5 & 57.0 \\
 & DFlare  & 84.3 & 59.2 & 85.4 & 56.6 & 85.9 & 57.6 \\
 & D-PACE  & 84.4 & 59.2 & 84.8 & 57.4 & 83.9 & 57.8 \\
 & Domino  & 84.3 & 59.4 & 85.2 & 57.2 & 84.4 & 57.2 \\
 & DSpark  & 84.3 & 59.4 & 85.4 & 58.8 & 84.5 & 58.8 \\
 & DCore   & 84.3 & 59.4 & 85.4 & 58.2 & 84.3 & 57.8 \\
\midrule
\multirow{7}{*}{16}
 & Eagle-3 & 84.8 & 59.6 & 85.3 & 55.0 & 84.5 & 58.0 \\
 & DFlash  & 84.4 & 59.2 & 85.4 & 56.6 & 85.1 & 58.2 \\
 & DFlare  & 84.3 & 59.4 & 84.7 & 57.4 & 85.9 & 58.0 \\
 & D-PACE  & 84.3 & 59.4 & 85.3 & 57.6 & 85.6 & 59.4 \\
 & Domino  & 84.4 & 59.2 & 85.8 & 57.6 & 85.1 & 58.8 \\
 & DSpark  & 84.4 & 59.4 & 85.1 & 58.6 & 85.7 & 59.4 \\
 & DCore   & 84.4 & 59.4 & 84.5 & 58.6 & 85.0 & 58.0 \\
\bottomrule
\end{tabular}}

%% file: section/Appendix-training-curves.tex
\section{Acceptance During Training}
\label{sec:training-scaling}

\input{figure/Figure-training-curves.tex}

Fig.~\ref{fig:training-curves-subset} evaluates seven checkpoints per
recipe on six representative benchmarks.

%% file: figure/Figure-training-curves.tex
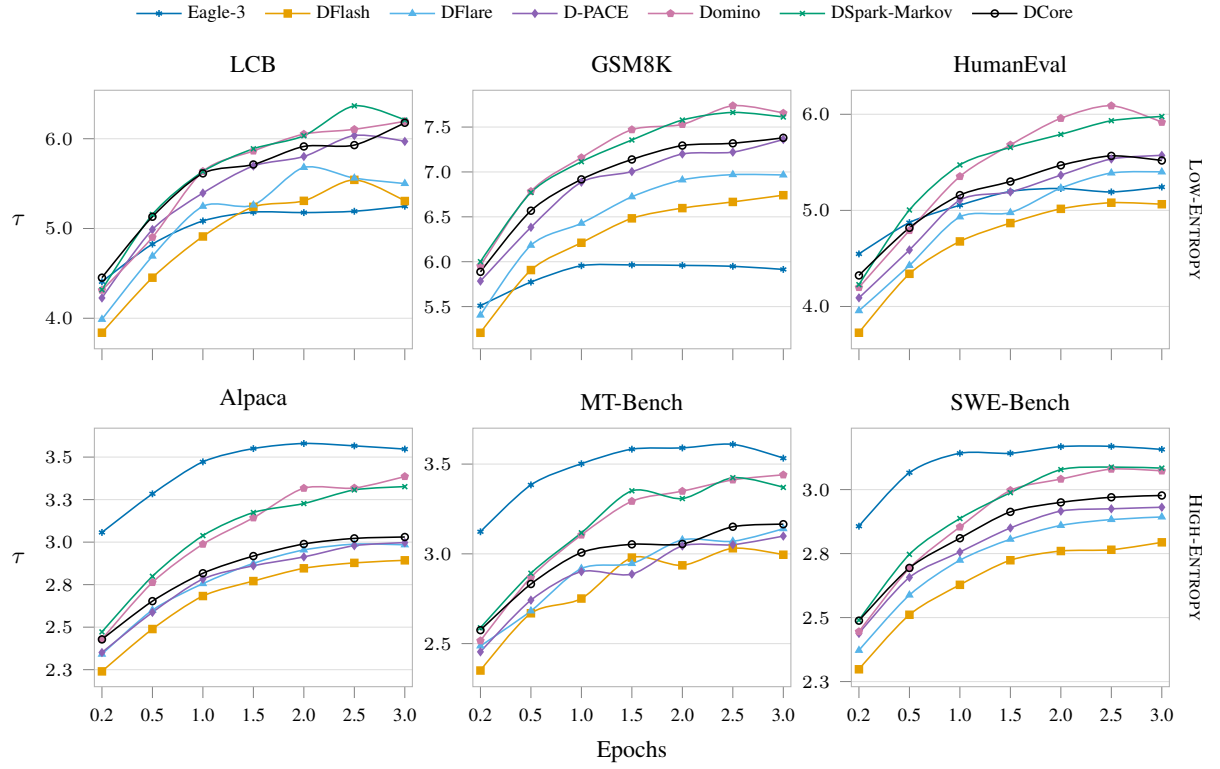
\begin{figure*}[p]
\centering
\ref*{leg:trainingcurves}\\[4pt]
\begin{tikzpicture}
\begin{groupplot}[
    group style={
        group name=tc,
        group size=3 by 2,
        horizontal sep=0.8cm,
        vertical sep=1.05cm,
        ylabels at=edge left,
        xticklabels at=edge bottom,
    },
    width=0.3617\textwidth, height=5cm,
    xmin=0.85, xmax=7.15,
    xtick={1,2,3,4,5,6,7},
    xticklabels={0.2,0.5,1.0,1.5,2.0,2.5,3.0},
    ymajorgrids=true, grid style={gray!25},
    axis line style={gray!60},
    tick pos=left,
    tick align=outside,
    ylabel={$\tau$},
    ylabel style={font=\small, rotate=-90},
    yticklabel style={/pgf/number format/fixed,
        /pgf/number format/precision=1, /pgf/number format/zerofill},
    ticklabel style={font=\scriptsize},
    title style={font=\small, yshift=-3pt},
    every axis plot/.append style={semithick, smooth, mark size=1.3pt},
]

\nextgroupplot[title={LCB},
    ymin=3.66,
    ymax=6.54,
    ytick distance=1,
    legend to name=leg:trainingcurves,
    legend columns=7,
    legend style={draw=none, fill=none, font=\scriptsize, text=black,
        /tikz/every even column/.append style={column sep=5pt}},
    legend image post style={mark size=1.3pt}]
\addplot[color=cEagle, mark=asterisk] coordinates {(1,4.406) (2,4.826) (3,5.084) (4,5.182) (5,5.177) (6,5.192) (7,5.248)};
\addlegendentry{Eagle-3}
\addplot[color=cDFlash, mark=square*] coordinates {(1,3.839) (2,4.452) (3,4.911) (4,5.241) (5,5.308) (6,5.542) (7,5.306)};
\addlegendentry{DFlash}
\addplot[color=cDFlare, mark=triangle*] coordinates {(1,3.988) (2,4.692) (3,5.247) (4,5.261) (5,5.680) (6,5.561) (7,5.501)};
\addlegendentry{DFlare}
\addplot[color=cDPace, mark=diamond*] coordinates {(1,4.226) (2,4.988) (3,5.395) (4,5.696) (5,5.802) (6,6.036) (7,5.970)};
\addlegendentry{D-PACE}
\addplot[color=cDomino, mark=pentagon*] coordinates {(1,4.313) (2,4.902) (3,5.636) (4,5.864) (5,6.051) (6,6.104) (7,6.194)};
\addlegendentry{Domino}
\addplot[color=cDSpark, mark=x] coordinates {(1,4.318) (2,5.156) (3,5.634) (4,5.888) (5,6.033) (6,6.366) (7,6.207)};
\addlegendentry{DSpark-Markov}
\addplot[color=cDCore, mark=o] coordinates {(1,4.452) (2,5.131) (3,5.614) (4,5.713) (5,5.914) (6,5.929) (7,6.178)};
\addlegendentry{DCore}

\nextgroupplot[title={GSM8K},
    ymin=5.03,
    ymax=7.91,
    ytick distance=0.5]
\addplot[color=cEagle, mark=asterisk] coordinates {(1,5.510) (2,5.773) (3,5.955) (4,5.964) (5,5.959) (6,5.949) (7,5.914)};
\addplot[color=cDFlash, mark=square*] coordinates {(1,5.208) (2,5.907) (3,6.211) (4,6.483) (5,6.597) (6,6.666) (7,6.741)};
\addplot[color=cDFlare, mark=triangle*] coordinates {(1,5.405) (2,6.183) (3,6.428) (4,6.723) (5,6.911) (6,6.970) (7,6.966)};
\addplot[color=cDPace, mark=diamond*] coordinates {(1,5.784) (2,6.383) (3,6.886) (4,7.004) (5,7.200) (6,7.221) (7,7.362)};
\addplot[color=cDomino, mark=pentagon*] coordinates {(1,5.952) (2,6.783) (3,7.159) (4,7.471) (5,7.530) (6,7.738) (7,7.655)};
\addplot[color=cDSpark, mark=x] coordinates {(1,6.000) (2,6.771) (3,7.114) (4,7.356) (5,7.578) (6,7.663) (7,7.612)};
\addplot[color=cDCore, mark=o] coordinates {(1,5.887) (2,6.568) (3,6.917) (4,7.139) (5,7.294) (6,7.319) (7,7.380)};

\nextgroupplot[title={HumanEval},
    ymin=3.56,
    ymax=6.25,
    ytick distance=1]
\addplot[color=cEagle, mark=asterisk] coordinates {(1,4.547) (2,4.873) (3,5.054) (4,5.197) (5,5.228) (6,5.192) (7,5.243)};
\addplot[color=cDFlash, mark=square*] coordinates {(1,3.727) (2,4.340) (3,4.677) (4,4.868) (5,5.016) (6,5.080) (7,5.064)};
\addplot[color=cDFlare, mark=triangle*] coordinates {(1,3.956) (2,4.427) (3,4.932) (4,4.977) (5,5.232) (6,5.389) (7,5.400)};
\addplot[color=cDPace, mark=diamond*] coordinates {(1,4.090) (2,4.588) (3,5.107) (4,5.192) (5,5.367) (6,5.535) (7,5.572)};
\addplot[color=cDomino, mark=pentagon*] coordinates {(1,4.198) (2,4.791) (3,5.353) (4,5.681) (5,5.958) (6,6.089) (7,5.917)};
\addplot[color=cDSpark, mark=x] coordinates {(1,4.227) (2,5.004) (3,5.473) (4,5.655) (5,5.791) (6,5.933) (7,5.977)};
\addplot[color=cDCore, mark=o] coordinates {(1,4.323) (2,4.819) (3,5.159) (4,5.300) (5,5.469) (6,5.567) (7,5.519)};

\nextgroupplot[title={Alpaca},
    ymin=2.15,
    ymax=3.67,
    ytick distance=0.25]
\addplot[color=cEagle, mark=asterisk] coordinates {(1,3.057) (2,3.284) (3,3.472) (4,3.550) (5,3.580) (6,3.566) (7,3.547)};
\addplot[color=cDFlash, mark=square*] coordinates {(1,2.240) (2,2.489) (3,2.683) (4,2.771) (5,2.846) (6,2.878) (7,2.893)};
\addplot[color=cDFlare, mark=triangle*] coordinates {(1,2.341) (2,2.600) (3,2.756) (4,2.875) (5,2.954) (6,2.988) (7,2.984)};
\addplot[color=cDPace, mark=diamond*] coordinates {(1,2.351) (2,2.587) (3,2.783) (4,2.861) (5,2.913) (6,2.979) (7,2.997)};
\addplot[color=cDomino, mark=pentagon*] coordinates {(1,2.429) (2,2.763) (3,2.988) (4,3.143) (5,3.317) (6,3.318) (7,3.386)};
\addplot[color=cDSpark, mark=x] coordinates {(1,2.472) (2,2.799) (3,3.038) (4,3.174) (5,3.227) (6,3.307) (7,3.326)};
\addplot[color=cDCore, mark=o] coordinates {(1,2.428) (2,2.653) (3,2.817) (4,2.918) (5,2.989) (6,3.022) (7,3.030)};

\nextgroupplot[title={MT-Bench},
    ymin=2.26,
    ymax=3.70,
    ytick distance=0.5,
    xlabel={Epochs},
    xlabel style={font=\small}]
\addplot[color=cEagle, mark=asterisk] coordinates {(1,3.123) (2,3.384) (3,3.502) (4,3.583) (5,3.590) (6,3.610) (7,3.533)};
\addplot[color=cDFlash, mark=square*] coordinates {(1,2.350) (2,2.669) (3,2.751) (4,2.979) (5,2.936) (6,3.031) (7,2.995)};
\addplot[color=cDFlare, mark=triangle*] coordinates {(1,2.487) (2,2.682) (3,2.918) (4,2.946) (5,3.078) (6,3.070) (7,3.140)};
\addplot[color=cDPace, mark=diamond*] coordinates {(1,2.454) (2,2.742) (3,2.901) (4,2.887) (5,3.044) (6,3.051) (7,3.099)};
\addplot[color=cDomino, mark=pentagon*] coordinates {(1,2.516) (2,2.869) (3,3.105) (4,3.293) (5,3.349) (6,3.412) (7,3.440)};
\addplot[color=cDSpark, mark=x] coordinates {(1,2.588) (2,2.892) (3,3.117) (4,3.351) (5,3.308) (6,3.424) (7,3.370)};
\addplot[color=cDCore, mark=o] coordinates {(1,2.575) (2,2.832) (3,3.007) (4,3.053) (5,3.055) (6,3.151) (7,3.165)};

\nextgroupplot[title={SWE-Bench},
    ymin=2.23,
    ymax=3.24,
    ytick distance=0.25]
\addplot[color=cEagle, mark=asterisk] coordinates {(1,2.857) (2,3.066) (3,3.142) (4,3.142) (5,3.168) (6,3.169) (7,3.157)};
\addplot[color=cDFlash, mark=square*] coordinates {(1,2.298) (2,2.511) (3,2.628) (4,2.724) (5,2.760) (6,2.765) (7,2.794)};
\addplot[color=cDFlare, mark=triangle*] coordinates {(1,2.372) (2,2.588) (3,2.724) (4,2.806) (5,2.860) (6,2.883) (7,2.893)};
\addplot[color=cDPace, mark=diamond*] coordinates {(1,2.438) (2,2.657) (3,2.756) (4,2.850) (5,2.916) (6,2.925) (7,2.931)};
\addplot[color=cDomino, mark=pentagon*] coordinates {(1,2.445) (2,2.693) (3,2.854) (4,2.997) (5,3.041) (6,3.080) (7,3.073)};
\addplot[color=cDSpark, mark=x] coordinates {(1,2.490) (2,2.747) (3,2.887) (4,2.988) (5,3.078) (6,3.088) (7,3.084)};
\addplot[color=cDCore, mark=o] coordinates {(1,2.488) (2,2.694) (3,2.810) (4,2.913) (5,2.950) (6,2.970) (7,2.977)};

\end{groupplot}
\node[rotate=-90, anchor=center, font=\scriptsize\scshape]
    at ([xshift=0.35cm]tc c3r1.east) {Low-Entropy};
\node[rotate=-90, anchor=center, font=\scriptsize\scshape]
    at ([xshift=0.35cm]tc c3r2.east) {High-Entropy};
\end{tikzpicture}
\caption{\textbf{Acceptance during training.} Checkpoints from $0.2$ to
$3.0$ epochs on a fixed Open-PerfectBlend corpus, Qwen3-8B, $B=16$,
$1|0$, $C=1$, along one training trajectory per recipe.}
\label{fig:training-curves-subset}
\end{figure*}

%% file: section/Appendix-accuracy.tex
\section{Answer-Accuracy Sanity Check}
\label{sec:appendix-accuracy}

Tab.~\ref{tab:accuracy_qwen3_8b} reports answer accuracy at all three
temperature pairs. With a greedy target ($0|0$), every drafter stays within
$0.5$ points of the greedy target-only reference on GSM8K and $0.4$ on
MATH-500. With a sampled target ($1|0$ and $1|1$), accuracy remains close
to the shared target-only reference
without a consistent directional shift, providing a sanity check of the
evaluation pipeline. Losslessness follows from the verification rule in
App.~\ref{sec:appendix-protocol}.

%% file: main.bbl
\begin{thebibliography}{52}
\providecommand{\natexlab}[1]{#1}

\bibitem[{Abramovich et~al.(2026)Abramovich, Ashkenazi, Putterman, Chislett,
  Mitra, Rouhani, Zilberstein, and Geifman}]{abramovich2026speedbench}
Talor Abramovich, Maor Ashkenazi, Izzy Putterman, Benjamin Chislett, Tiyasa
  Mitra, Bita~Darvish Rouhani, Ran Zilberstein, and Yonatan Geifman. 2026.
\newblock \href {https://arxiv.org/abs/2604.09557} {{SPEED-Bench}: A unified
  and diverse benchmark for speculative decoding}.
\newblock \emph{arXiv preprint arXiv:2604.09557}.
\newblock ICML 2026.

\bibitem[{Austin et~al.(2021)Austin, Odena, Nye, Bosma, Michalewski, Dohan,
  Jiang, Cai, Terry, Le, and Sutton}]{austin2021mbpp}
Jacob Austin, Augustus Odena, Maxwell Nye, Maarten Bosma, Henryk Michalewski,
  David Dohan, Ellen Jiang, Carrie Cai, Michael Terry, Quoc Le, and Charles
  Sutton. 2021.
\newblock \href {https://arxiv.org/abs/2108.07732} {Program synthesis with
  large language models}.
\newblock \emph{arXiv preprint arXiv:2108.07732}.

\bibitem[{Chen et~al.(2023)Chen, Borgeaud, Irving, Lespiau, Sifre, and
  Jumper}]{chen2023accelerating}
Charlie Chen, Sebastian Borgeaud, Geoffrey Irving, Jean-Baptiste Lespiau,
  Laurent Sifre, and John Jumper. 2023.
\newblock \href {https://arxiv.org/abs/2302.01318} {Accelerating large language
  model decoding with speculative sampling}.
\newblock \emph{arXiv preprint arXiv:2302.01318}.

\bibitem[{Chen et~al.(2026)Chen, Liang, and Liu}]{chen2026dflash}
Jian Chen, Yesheng Liang, and Zhijian Liu. 2026.
\newblock \href {https://arxiv.org/abs/2602.06036} {{DFlash: Block Diffusion
  for Flash Speculative Decoding}}.
\newblock \emph{arXiv preprint arXiv:2602.06036}.

\bibitem[{Chen et~al.(2021)Chen, Tworek, Jun, Yuan, Pinto, Kaplan, Edwards,
  Burda, Joseph, Brockman et~al.}]{chen2021humaneval}
Mark Chen, Jerry Tworek, Heewoo Jun, Qiming Yuan, Henrique Ponde de~Oliveira
  Pinto, Jared Kaplan, Harri Edwards, Yuri Burda, Nicholas Joseph, Greg
  Brockman, et~al. 2021.
\newblock \href {https://arxiv.org/abs/2107.03374} {Evaluating large language
  models trained on code}.
\newblock \emph{arXiv preprint arXiv:2107.03374}.

\bibitem[{Cheng et~al.(2026{\natexlab{a}})Cheng, Wang, Zhu, Shao, Chen, and
  Hu}]{cheng-etal-2026-edsd}
Longkai Cheng, Ximing Wang, Jiangcai Zhu, Kailai Shao, Chao Chen, and Haixiang
  Hu. 2026{\natexlab{a}}.
\newblock \href {https://doi.org/10.18653/v1/2026.acl-long.2145} {{EDSD}:
  Entropy-driven design for faster speculative decoding}.
\newblock In \emph{Proceedings of the 64th Annual Meeting of the Association
  for Computational Linguistics (Volume 1: Long Papers)}, pages 46243--46260,
  San Diego, California, United States. Association for Computational
  Linguistics.

\bibitem[{Cheng et~al.(2026{\natexlab{b}})Cheng, Yu, Shao, Li, Xiong, Qian,
  Zhu, Ma, Zhang, Ye, Chen, Deng, Yu, Dai, Zhang, Wei, Tan, Yang, Xu, Wu, Xu,
  Wang, Chen, Tian, Bi, Hao, Chen, Cao, Zhang, Xu, Zhang, Zhao, and
  Liang}]{cheng2026dspark}
Xin Cheng, Xingkai Yu, Chenze Shao, Jiashi Li, Yunfan Xiong, Yi~Qian, Jiaqi
  Zhu, Shirong Ma, Xiaokang Zhang, Jiasheng Ye, Qinyu Chen, Chengqi Deng,
  Jiping Yu, Damai Dai, Zhengyan Zhang, Yixuan Wei, Yixuan Tan, Wenkai Yang,
  Runxin Xu, and 14 others. 2026{\natexlab{b}}.
\newblock \href {https://arxiv.org/abs/2607.05147} {{DSpark}:
  Confidence-scheduled speculative decoding with semi-autoregressive
  generation}.
\newblock \emph{arXiv preprint arXiv:2607.05147}.

\bibitem[{Cobbe et~al.(2021)Cobbe, Kosaraju, Bavarian, Chen, Jun, Kaiser,
  Plappert, Tworek, Hilton, Nakano, Hesse, and Schulman}]{cobbe2021gsm8k}
Karl Cobbe, Vineet Kosaraju, Mohammad Bavarian, Mark Chen, Heewoo Jun, Lukasz
  Kaiser, Matthias Plappert, Jerry Tworek, Jacob Hilton, Reiichiro Nakano,
  Christopher Hesse, and John Schulman. 2021.
\newblock \href {https://arxiv.org/abs/2110.14168} {Training verifiers to solve
  math word problems}.
\newblock \emph{arXiv preprint arXiv:2110.14168}.

\bibitem[{{DeepSeek-AI}(2024)}]{deepseekai2024deepseekv3}
{DeepSeek-AI}. 2024.
\newblock \href {https://arxiv.org/abs/2412.19437} {{DeepSeek-V3} technical
  report}.
\newblock \emph{arXiv preprint arXiv:2412.19437}.

\bibitem[{{DeepSeek-AI}(2026)}]{deepseekai2026v4prodspark}
{DeepSeek-AI}. 2026.
\newblock \href {https://huggingface.co/deepseek-ai/DeepSeek-V4-Pro-DSpark}
  {{DeepSeek-V4-Pro-DSpark} model card}.
\newblock Hugging Face.

\bibitem[{Fu et~al.(2024)Fu, Bailis, Stoica, and Zhang}]{fu2024lookahead}
Yichao Fu, Peter Bailis, Ion Stoica, and Hao Zhang. 2024.
\newblock \href {https://proceedings.mlr.press/v235/fu24a.html} {Break the
  sequential dependency of {LLM} inference using lookahead decoding}.
\newblock In \emph{International Conference on Machine Learning}.

\bibitem[{He et~al.(2024)He, Zhong, Cai, Lee, and He}]{he2024rest}
Zhenyu He, Zexuan Zhong, Tianle Cai, Jason~D Lee, and Di~He. 2024.
\newblock \href {https://aclanthology.org/2024.naacl-long.88/} {{REST}:
  Retrieval-based speculative decoding}.
\newblock In \emph{Proceedings of the 2024 Conference of the North American
  Chapter of the Association for Computational Linguistics: Human Language
  Technologies (Volume 1: Long Papers)}.

\bibitem[{Hendrycks et~al.(2021)Hendrycks, Burns, Kadavath, Arora, Basart,
  Tang, Song, and Steinhardt}]{hendrycks2021math}
Dan Hendrycks, Collin Burns, Saurav Kadavath, Akul Arora, Steven Basart, Eric
  Tang, Dawn Song, and Jacob Steinhardt. 2021.
\newblock \href {https://openreview.net/pdf?id=7Bywt2mQsCe} {Measuring
  mathematical problem solving with the math dataset}.
\newblock In \emph{Advances in Neural Information Processing Systems Datasets
  and Benchmarks Track}.

\bibitem[{Huang et~al.(2026)Huang, Zhang, Zhang, Lin, Xu, and
  Zhang}]{huang2026dominodecouplingcausalmodeling}
Jianuo Huang, Yaojie Zhang, Qituan Zhang, Hao Lin, Hanlin Xu, and Linfeng
  Zhang. 2026.
\newblock \href {https://arxiv.org/abs/2605.29707} {Domino: Decoupling causal
  modeling from autoregressive drafting in speculative decoding}.
\newblock \emph{Preprint}, arXiv:2605.29707.

\bibitem[{Hui et~al.(2026)Hui, Huang, Campos~Salas, Sun, Pemberton, Song,
  Khetan, and Karypis}]{hui2026peagle}
Mude Hui, Xin Huang, Jaime Campos~Salas, Yue Sun, Nathan Pemberton, Xiang Song,
  Ashish Khetan, and George Karypis. 2026.
\newblock \href {https://arxiv.org/abs/2602.01469} {{P-EAGLE}:
  Parallel-drafting {EAGLE} with scalable training}.
\newblock \emph{arXiv preprint arXiv:2602.01469}.

\bibitem[{Jain et~al.(2025)Jain, Han, Gu, Li, Yan, Zhang, Wang, Solar-Lezama,
  Sen, and Stoica}]{jain2025livecodebench}
Naman Jain, King Han, Alex Gu, Wen-Ding Li, Fanjia Yan, Tianjun Zhang, Sida~I.
  Wang, Armando Solar-Lezama, Koushik Sen, and Ion Stoica. 2025.
\newblock \href
  {https://proceedings.iclr.cc/paper_files/paper/2025/file/94074dd5a072d28ff75a76dabed43767-Paper-Conference.pdf}
  {Livecodebench: Holistic and contamination free evaluation of large language
  models for code}.
\newblock In \emph{International Conference on Learning Representations}.

\bibitem[{Jimenez et~al.(2024)Jimenez, Yang, Wettig, Yao, Pei, Press, and
  Narasimhan}]{jimenez2024swebench}
Carlos~E. Jimenez, John Yang, Alexander Wettig, Shunyu Yao, Kexin Pei, Ofir
  Press, and Karthik Narasimhan. 2024.
\newblock \href
  {https://proceedings.iclr.cc/paper_files/paper/2024/hash/edac78c3e300629acfe6cbe9ca88fb84-Abstract-Conference.html}
  {{SWE-bench}: Can language models resolve real-world github issues?}
\newblock In \emph{International Conference on Learning Representations}.

\bibitem[{Karimi et~al.(2026)Karimi, Gao, and Hassanpour}]{karimi2026dblast}
Amirmohammad Karimi, Chao Gao, and Negar Hassanpour. 2026.
\newblock \href {https://arxiv.org/abs/2608.05448} {{DBLast}: Dependent block
  drafting for stochastic speculative decoding}.
\newblock \emph{arXiv preprint arXiv:2608.05448}.

\bibitem[{Kim et~al.(2026)Kim, Kim, Cho, and Seo}]{kim2026runtime}
Sungkyun Kim, Jaemin Kim, Yeongpil Cho, and Jiwon Seo. 2026.
\newblock \href {https://doi.org/10.3390/electronics15143179}
  {Runtime-controlled evaluation of speculative decoding methods}.
\newblock \emph{Electronics}, 15(14):3179.

\bibitem[{Labonne(2024)}]{labonne2024openperfectblend}
Maxime Labonne. 2024.
\newblock \href {https://huggingface.co/datasets/mlabonne/open-perfectblend}
  {Open-{PerfectBlend}}.
\newblock \url{https://huggingface.co/datasets/mlabonne/open-perfectblend}.

\bibitem[{Leviathan et~al.(2023)Leviathan, Kalman, and
  Matias}]{leviathan2023fast}
Yaniv Leviathan, Matan Kalman, and Yossi Matias. 2023.
\newblock \href {https://proceedings.mlr.press/v202/leviathan23a.html} {Fast
  inference from transformers via speculative decoding}.
\newblock In \emph{International Conference on Machine Learning}.

\bibitem[{Li et~al.(2024{\natexlab{a}})Li, Chiang, Frick, Dunlap, Wu, Zhu,
  Gonzalez, and Stoica}]{li2024arenahard}
Tianle Li, Wei-Lin Chiang, Evan Frick, Lisa Dunlap, Tianhao Wu, Banghua Zhu,
  Joseph~E. Gonzalez, and Ion Stoica. 2024{\natexlab{a}}.
\newblock \href {https://arxiv.org/abs/2406.11939} {From crowdsourced data to
  high-quality benchmarks: Arena-hard and benchbuilder pipeline}.
\newblock \emph{arXiv preprint arXiv:2406.11939}.

\bibitem[{Li et~al.(2023)Li, Zhang, Dubois, Taori, Gulrajani, Guestrin, Liang,
  and Hashimoto}]{li2023alpacaeval}
Xuechen Li, Tianyi Zhang, Yann Dubois, Rohan Taori, Ishaan Gulrajani, Carlos
  Guestrin, Percy Liang, and Tatsunori~B. Hashimoto. 2023.
\newblock \href {https://github.com/tatsu-lab/alpaca_eval} {{AlpacaEval}: An
  automatic evaluator of instruction-following models}.
\newblock \url{https://github.com/tatsu-lab/alpaca_eval}.

\bibitem[{Li et~al.(2024{\natexlab{b}})Li, Wei, Zhang, and
  Zhang}]{li2024eagle2}
Yuhui Li, Fangyun Wei, Chao Zhang, and Hongyang Zhang. 2024{\natexlab{b}}.
\newblock \href {https://aclanthology.org/2024.emnlp-main.422/} {{EAGLE-2}:
  Faster inference of language models with dynamic draft trees}.
\newblock In \emph{Conference on Empirical Methods in Natural Language
  Processing}.

\bibitem[{Li et~al.(2024{\natexlab{c}})Li, Wei, Zhang, and Zhang}]{li2024eagle}
Yuhui Li, Fangyun Wei, Chao Zhang, and Hongyang Zhang. 2024{\natexlab{c}}.
\newblock \href {https://proceedings.mlr.press/v235/li24bt.html} {{EAGLE}:
  Speculative sampling requires rethinking feature uncertainty}.
\newblock In \emph{International Conference on Machine Learning}.

\bibitem[{Li et~al.(2025)Li, Wei, Zhang, and Zhang}]{NEURIPS2025_c7b5a35e}
Yuhui Li, Fangyun Wei, Chao Zhang, and Hongyang Zhang. 2025.
\newblock \href {https://doi.org/10.52202/085713-4562} {Eagle-3: Scaling up
  inference acceleration of large language models via training-time test}.
\newblock In \emph{Advances in Neural Information Processing Systems},
  volume~38, pages 136737--136756. Curran Associates, Inc.

\bibitem[{Lightman et~al.(2024)Lightman, Kosaraju, Burda, Edwards, Baker, Lee,
  Leike, Schulman, Sutskever, and Cobbe}]{lightman2024verify}
Hunter Lightman, Vineet Kosaraju, Yura Burda, Harri Edwards, Bowen Baker, Teddy
  Lee, Jan Leike, John Schulman, Ilya Sutskever, and Karl Cobbe. 2024.
\newblock \href
  {https://proceedings.iclr.cc/paper_files/paper/2024/file/aca97732e30bcf1303bc22ac3924fd16-Paper-Conference.pdf}
  {Let's verify step by step}.
\newblock In \emph{International Conference on Learning Representations}.

\bibitem[{Liu et~al.(2026{\natexlab{a}})Liu, Cen, Shi, Qin, Zhang, Liu, Fan,
  Zhao, Xie, Zhang, Liu, Yu, and Zhu}]{liu2026angelspec}
Hong Liu, Rui Cen, Junhan Shi, Guangshuo Qin, Jiebin Zhang, Tianyu Liu, Runzhi
  Fan, Guoliang Zhao, Ruobing Xie, Kai Zhang, Song Liu, Guanghua Yu, and
  Jianchen Zhu. 2026{\natexlab{a}}.
\newblock \href {https://arxiv.org/abs/2607.25852} {{AngelSpec}: Towards
  real-world high performance inference with speculative decoding}.
\newblock \emph{arXiv preprint arXiv:2607.25852}.

\bibitem[{Liu et~al.(2026{\natexlab{b}})Liu, Yu, Park, Stoica, and
  Cheung}]{liu2026illusion}
Xiaoxuan Liu, Jiaxiang Yu, Jongseok Park, Ion Stoica, and Alvin Cheung.
  2026{\natexlab{b}}.
\newblock \href {https://arxiv.org/abs/2601.11580v2} {Speculative decoding:
  Performance or illusion?}
\newblock \emph{arXiv preprint arXiv:2601.11580}.

\bibitem[{Mainardi et~al.(2026)Mainardi, Sandikci, and
  Vanschoren}]{mainardi-etal-2026-empirical}
Luca Mainardi, Selcuk Sandikci, and Joaquin Vanschoren. 2026.
\newblock \href {https://doi.org/10.18653/v1/2026.eacl-long.255} {An empirical
  study of speculative decoding for small language models}.
\newblock In \emph{Proceedings of the 19th Conference of the European Chapter
  of the Association for Computational Linguistics (Volume 1: Long Papers)},
  pages 5483--5497, Rabat, Morocco. Association for Computational Linguistics.

\bibitem[{{Mathematical Association of America}(2025)}]{maa2025aime}
{Mathematical Association of America}. 2025.
\newblock \href {https://maa.org/maa-invitational-competitions/} {American
  invitational mathematics examination (aime) 2025}.
\newblock \url{https://maa.org/maa-invitational-competitions/}.

\bibitem[{Miao et~al.(2024)Miao, Oliaro, Zhang, Cheng, Wang, Zhang, Wong, Zhu,
  Yang, Shi, Shi, Chen, Arfeen, Abhyankar, and Jia}]{miao2024specinfer}
Xupeng Miao, Gabriele Oliaro, Zhihao Zhang, Xinhao Cheng, Zeyu Wang, Zhengxin
  Zhang, Rae Ying~Yee Wong, Alan Zhu, Lijie Yang, Xiaoxiang Shi, Chunan Shi,
  Zhuoming Chen, Daiyaan Arfeen, Reyna Abhyankar, and Zhihao Jia. 2024.
\newblock \href {https://arxiv.org/abs/2305.09781} {{SpecInfer}: Accelerating
  large language model serving with tree-based speculative inference and
  verification}.
\newblock In \emph{Proceedings of the 29th ACM International Conference on
  Architectural Support for Programming Languages and Operating Systems}, pages
  932--949.

\bibitem[{{NVIDIA}(2026{\natexlab{a}})}]{nvidia2026nemotron35lightning}
{NVIDIA}. 2026{\natexlab{a}}.
\newblock \href
  {https://huggingface.co/nvidia/NVIDIA-Nemotron-3.5-Lightning-30B-A3B-BF16/blob/main/README.md}
  {{NVIDIA-Nemotron-3.5-Lightning-30B-A3B-BF16} model card}.
\newblock Hugging Face.

\bibitem[{{NVIDIA}(2026{\natexlab{b}})}]{nvidia2026nemotron35dflash}
{NVIDIA}. 2026{\natexlab{b}}.
\newblock \href
  {https://huggingface.co/nvidia/NVIDIA-Nemotron-3.5-Lightning-30B-A3B-NVFP4-DFlash}
  {{NVIDIA-Nemotron-3.5-Lightning-30B-A3B-NVFP4-DFlash} model card}.
\newblock Hugging Face.

\bibitem[{Oda et~al.(2026)Oda, Mathieu, Knyazhitskiy, and
  Chakhvadze}]{oda2026treesfrommarginals}
Yuma Oda, Ryan Mathieu, Roman Knyazhitskiy, and Artur Chakhvadze. 2026.
\newblock \href {https://arxiv.org/abs/2607.06763} {Trees from marginals:
  Autoregressive drafting with factorized priors}.
\newblock \emph{Preprint}, arXiv:2607.06763.

\bibitem[{Pope et~al.(2023)Pope, Douglas, Chowdhery, Devlin, Bradbury,
  Levskaya, Heek, Xiao, Agrawal, and Dean}]{pope2023efficient}
Reiner Pope, Sholto Douglas, Aakanksha Chowdhery, Jacob Devlin, James Bradbury,
  Anselm Levskaya, Jonathan Heek, Kefan Xiao, Shivani Agrawal, and Jeff Dean.
  2023.
\newblock \href
  {https://proceedings.mlsys.org/paper_files/paper/2023/file/c4be71ab8d24cdfb45e3d06dbfca2780-Paper-mlsys2023.pdf}
  {Efficiently scaling transformer inference}.
\newblock In \emph{Proceedings of Machine Learning and Systems}.

\bibitem[{{Qwen Team}(2026)}]{qwen2026qwen38mtp}
{Qwen Team}. 2026.
\newblock \href {https://huggingface.co/Qwen/Qwen3.8-27B} {{Qwen3.8-27B} model
  card}.
\newblock Hugging Face.

\bibitem[{Rusanovsky et~al.(2026)Rusanovsky, Miron, Uziel, Belhasin, Guo,
  Zilberstein, Ashkenazi, and Elad}]{rusanovsky2026lilicorr}
Matan Rusanovsky, Yoav Miron, Roy Uziel, Omer Belhasin, Hao Guo, Ran
  Zilberstein, Maor Ashkenazi, and Michael Elad. 2026.
\newblock \href {https://arxiv.org/abs/2608.20530} {{LiLiCorr}: Lightweight
  likelihood correlation of parallel drafts for speculative decoding}.
\newblock \emph{arXiv preprint arXiv:2608.20530}.

\bibitem[{Samarin et~al.(2026)Samarin, Krutikov, Shevtsov, Skvortsov, Fisin,
  and Golubev}]{samarin2026lk}
Alexander Samarin, Sergei Krutikov, Anton Shevtsov, Sergei Skvortsov, Filipp
  Fisin, and Alexander Golubev. 2026.
\newblock \href {https://arxiv.org/abs/2602.23881} {{LK} losses: Direct
  acceptance rate optimization for speculative decoding}.
\newblock \emph{arXiv preprint arXiv:2602.23881}.
\newblock ICML 2026.

\bibitem[{Shazeer(2019)}]{shazeer2019fast}
Noam Shazeer. 2019.
\newblock \href {https://arxiv.org/abs/1911.02150} {Fast transformer decoding:
  One write-head is all you need}.
\newblock \emph{arXiv preprint arXiv:1911.02150}.

\bibitem[{Stern et~al.(2018)Stern, Shazeer, and Uszkoreit}]{stern2018blockwise}
Mitchell Stern, Noam Shazeer, and Jakob Uszkoreit. 2018.
\newblock \href
  {https://proceedings.neurips.cc/paper/2018/hash/c4127b9194fe8562c64dc0f5bf2c93bc-Abstract.html}
  {Blockwise parallel decoding for deep autoregressive models}.
\newblock In \emph{Advances in Neural Information Processing Systems}.

\bibitem[{Sun et~al.(2025)Sun, Li, Li, Lian, Lin, Zhen, Yang, Chen, Yu, Yuan,
  and Ma}]{sun2025benchmark}
Shengyin Sun, Yiming Li, Xing Li, Yingzhao Lian, Weizhe Lin, Hui-Ling Zhen,
  Zhiyuan Yang, Chen Chen, Xianzhi Yu, Mingxuan Yuan, and Chen Ma. 2025.
\newblock \href {https://arxiv.org/abs/2509.04474} {Scaling up, speeding up: A
  benchmark of speculative decoding for efficient {LLM} test-time scaling}.
\newblock \emph{arXiv preprint arXiv:2509.04474}.

\bibitem[{{Tencent Hy Team}(2026)}]{tencent2026hy4preview}
{Tencent Hy Team}. 2026.
\newblock \href {https://github.com/Tencent-Hunyuan/Hy4-preview} {{Hy4
  preview}}.
\newblock Official model repository.

\bibitem[{Wu et~al.(2026{\natexlab{a}})Wu, Wang, Geuter, Yao, Jiang, Zhang, Li,
  and Yu}]{wu2026dpal}
Tianyu Wu, Zhuohan Wang, Jonathan Geuter, Yu~Yao, Weifan Jiang, Chengxi Zhang,
  Ju~Li, and Minlan Yu. 2026{\natexlab{a}}.
\newblock \href {https://arxiv.org/abs/2610.28355} {Beyond the point mass:
  Acceptance-derived weights and distributional losses for speculative
  decoding}.
\newblock \emph{arXiv preprint arXiv:2610.28355}.

\bibitem[{Wu et~al.(2026{\natexlab{b}})Wu, Yao, Qi, Zheng, Wang, Ma, Liao,
  Lakkaraju, Li, and Du}]{wu2026dpacedynamicpositionawarecrossentropy}
Tianyu Wu, Yu~Yao, Zhenting Qi, Han Zheng, Zhuohan Wang, Haoran Ma, Lawrence
  Liao, Himabindu Lakkaraju, Ju~Li, and Yilun Du. 2026{\natexlab{b}}.
\newblock \href {https://arxiv.org/abs/2605.18810} {{D-PACE}: Dynamic
  position-aware cross-entropy for parallel speculative drafting}.
\newblock \emph{Preprint}, arXiv:2605.18810.

\bibitem[{Xia et~al.(2024)Xia, Yang, Dong, Wang, Li, Ge, Liu, Li, and
  Sui}]{xia-etal-2024-unlocking}
Heming Xia, Zhe Yang, Qingxiu Dong, Peiyi Wang, Yongqi Li, Tao Ge, Tianyu Liu,
  Wenjie Li, and Zhifang Sui. 2024.
\newblock \href {https://aclanthology.org/2024.findings-acl.456/} {Unlocking
  efficiency in large language model inference: A comprehensive survey of
  speculative decoding}.
\newblock In \emph{Findings of the Association for Computational Linguistics:
  ACL 2024}, pages 7655--7671.

\bibitem[{{Xiaomi MiMo}(2026)}]{xiaomi2026mimoultraspeed}
{Xiaomi MiMo}. 2026.
\newblock \href {https://mimo.mi.com/docs/en-US/news/latest/1000tps} {{Xiaomi
  MiMo} partners with {TileRT}: {1T} model breaks 1000 tokens/s output speed
  for the first time}.
\newblock Official technical announcement.

\bibitem[{Yan et~al.(2025)Yan, Agarwal, and
  Venkataraman}]{yan-etal-2025-decoding}
Minghao Yan, Saurabh Agarwal, and Shivaram Venkataraman. 2025.
\newblock \href {https://aclanthology.org/2025.naacl-long.328/} {Decoding
  speculative decoding}.
\newblock In \emph{Proceedings of the 2025 Conference of the Nations of the
  Americas Chapter of the Association for Computational Linguistics: Human
  Language Technologies}, pages 6460--6473.

\bibitem[{Yang et~al.(2025)Yang, Li, Yang, Zhang, Hui, Zheng, Yu, Gao, Huang,
  Lv et~al.}]{yang2025qwen3}
An~Yang, Anfeng Li, Baosong Yang, Beichen Zhang, Binyuan Hui, Bo~Zheng, Bowen
  Yu, Chang Gao, Chengen Huang, Chenxu Lv, et~al. 2025.
\newblock \href {https://arxiv.org/abs/2505.09388} {{Qwen3} technical report}.
\newblock \emph{arXiv preprint arXiv:2505.09388}.

\bibitem[{Zhang et~al.(2026)Zhang, Yu, Liu, Yu, Li, Zhu, Duo, Xiong, Song, Yu,
  Zhu, and Li}]{zhang2026dflare}
Jiebin Zhang, Zhenghan Yu, Song Liu, Eugene~J Yu, Zheng Li, Dawei Zhu,
  Jiangshan Duo, Weimin Xiong, Yifan Song, Guanghua Yu, Jianchen Zhu, and
  Sujian Li. 2026.
\newblock \href {https://arxiv.org/abs/2606.02091} {{DFlare}: Scaling up draft
  capacity for block diffusion speculative decoding}.
\newblock \emph{arXiv preprint arXiv:2606.02091}.

\bibitem[{Zheng et~al.(2023)Zheng, Chiang, Sheng, Zhuang, Wu, Zhuang, Lin, Li,
  Li, Xing, Zhang, Gonzalez, and Stoica}]{zheng2023mtbench}
Lianmin Zheng, Wei-Lin Chiang, Ying Sheng, Siyuan Zhuang, Zhanghao Wu, Yonghao
  Zhuang, Zi~Lin, Zhuohan Li, Dacheng Li, Eric~P. Xing, Hao Zhang, Joseph~E.
  Gonzalez, and Ion Stoica. 2023.
\newblock \href
  {https://proceedings.neurips.cc/paper_files/paper/2023/hash/91f18a1287b398d378ef22505bf41832-Abstract-Datasets_and_Benchmarks.html}
  {Judging llm-as-a-judge with mt-bench and chatbot arena}.
\newblock In \emph{Advances in Neural Information Processing Systems Datasets
  and Benchmarks Track}.

\bibitem[{Zheng et~al.(2024)Zheng, Yin, Xie, Sun, Huang, Yu, Cao, Kozyrakis,
  Stoica, Gonzalez, Barrett, and Sheng}]{zheng2024sglang}
Lianmin Zheng, Liangsheng Yin, Zhiqiang Xie, Chuyue Sun, Jeff Huang, Cody~Hao
  Yu, Shiyi Cao, Christos Kozyrakis, Ion Stoica, Joseph~E. Gonzalez, Clark
  Barrett, and Ying Sheng. 2024.
\newblock \href
  {https://proceedings.neurips.cc/paper_files/paper/2024/hash/724be4472168f31ba1c9ac630f15dec8-Abstract-Conference.html}
  {{SGLang}: Efficient execution of structured language model programs}.
\newblock In \emph{Advances in Neural Information Processing Systems}.

\end{thebibliography}
